\documentclass{article}

\usepackage[main,final]{neurips_2026}

\usepackage[utf8]{inputenc}
\usepackage[T1]{fontenc}
\DeclareMathSizes{7.5}{7}{5}{5}
\usepackage{iftex}
\usepackage{url}
\usepackage{microtype}
\usepackage{graphicx}
\usepackage{subcaption}
\usepackage{booktabs}
\usepackage{nicefrac}
\usepackage{algorithm}
\usepackage{algorithmic}
\usepackage{hyperref}
\usepackage{pifont}
\usepackage{amssymb}
\usepackage{xcolor}
\usepackage{multirow}

\ifPDFTeX
  \DeclareUnicodeCharacter{2013}{--}
  \DeclareUnicodeCharacter{2014}{---}
  \DeclareUnicodeCharacter{2019}{'}
\else
  \catcode`–=\active
  \protected\def–{--}
  \catcode`—=\active
  \protected\def—{---}
  \catcode`’=\active
  \protected\def’{'}
\fi

\usepackage{amsmath}
\usepackage{amssymb}
\usepackage{mathtools}
\usepackage{amsthm}
\usepackage{mathrsfs}
\usepackage{bm}
\usepackage{enumitem}
\usepackage{graphicx}
\usepackage{placeins}

\usepackage{tikz}
\usetikzlibrary{matrix, positioning, decorations.pathreplacing, arrows.meta, patterns, shapes.geometric, calc, fit, shapes, backgrounds, shadows}

\theoremstyle{plain}
\newtheorem{theorem}{Theorem}[section]
\newtheorem{proposition}[theorem]{Proposition}
\newtheorem{lemma}[theorem]{Lemma}

\theoremstyle{definition}
\newtheorem{definition}[theorem]{Definition}
\newtheorem{assumption}[theorem]{Assumption}
\newtheorem{axiom}[theorem]{Axiom}
\theoremstyle{remark}
\newtheorem{remark}[theorem]{Remark}
\usepackage[textsize=tiny]{todonotes}

\usepackage{listings}
\usepackage{xcolor}

\lstdefinelanguage{Lean}{
  keywords={theorem, def, by, have, calc, obtain, use, intro, exact, import, section, end, variable, let},
  sensitive=true,
  morecomment=[l]{--},
  morecomment=[s]{/-}{-/},
  morestring=[b]",
  literate=
    {α}{$\alpha$}1
    {β}{$\beta$}1
    {γ}{$\gamma$}1
    {δ}{$\delta$}1
    {ε}{$\varepsilon$}1
    {θ}{$\theta$}1
    {λ}{$\lambda$}1
    {≤}{$\le$}1
    {≥}{$\ge$}1
    {→}{$\to$}1
    {∀}{$\forall$}1
    {∃}{$\exists$}1
    {∈}{$\in$}1        
    {∧}{$\land$}1      
    {∨}{$\lor$}1
    {¬}{$\neg$}1
    {≠}{$\ne$}1
    {ℝ}{$\mathbb{R}$}1
    {ℕ}{$\mathbb{N}$}1
    {ℤ}{$\mathbb{Z}$}1
    {⊆}{$\subseteq$}1
    {⟨}{$\langle$}1
    {⟩}{$\rangle$}1
    {₀}{$_0$}1         
    {₁}{$_1$}1
    {ᵢ}{$_i$}1
    {⁻¹}{$^{-1}$}1
    {^2}{{$^2$}}1      
}

\definecolor{nat_blue}{RGB}{46, 102, 164}
\definecolor{nat_red}{RGB}{180, 50, 45}
\definecolor{nat_green}{RGB}{85, 140, 90}
\definecolor{nat_orange}{RGB}{220, 130, 40}
\definecolor{nat_grey}{RGB}{100, 100, 100}
\definecolor{bg_grey}{RGB}{248, 249, 250}
\definecolor{line_grey}{RGB}{60, 60, 60}

\definecolor{nat_blue}{RGB}{0, 114, 178}     
\definecolor{nat_red}{RGB}{213, 94, 0}       
\definecolor{nat_orange}{RGB}{230, 159, 0}   
\definecolor{nat_green}{RGB}{0, 158, 115}    
\definecolor{nat_grey}{RGB}{80, 80, 80}
\definecolor{line_grey}{RGB}{100, 100, 100}
\definecolor{bg_grey}{RGB}{248, 248, 248}

\definecolor{stage1}{RGB}{110, 40, 140}     
\definecolor{stage2}{RGB}{213, 94, 0}     
\definecolor{stage3}{RGB}{0, 158, 115}    
\definecolor{neural}{RGB}{0, 80, 160}       
\definecolor{physics}{RGB}{180, 20, 80}     
\definecolor{grey}{RGB}{80, 80, 80}
\definecolor{lightgrey}{RGB}{230, 230, 230}

\tikzset{
    >={Stealth[inset=0pt, length=5pt, width=4pt]},
    base/.style={font=\sffamily\fontsize{7.5}{8.5}\selectfont, align=center, line width=0.6pt},
    tensor/.style={base, rectangle, rounded corners=2pt, draw=#1!80!black, fill=#1!12, minimum height=1.8em, minimum width=4.2em},
    process/.style={base, rectangle, rounded corners=1pt, draw=line_grey!50, fill=white, inner sep=4pt, minimum width=5.2em},
    operator/.style={base, circle, draw=line_grey!50, fill=white, inner sep=1pt, minimum size=1.5em},
    label_panel/.style={font=\sffamily\large\bfseries, anchor=north west},
    flow/.style={draw=line_grey!80, line width=0.7pt, ->},
    feedback/.style={draw=nat_orange, line width=0.8pt, dashed, ->, align=center},
    annotation/.style={font=\sffamily\fontsize{7}{8}\selectfont\itshape, color=line_grey!80, align=left}
}

\DeclareMathOperator*{\argmin}{arg\,min}
\newcommand{\eqdef}{\ensuremath{\stackrel{\mbox{\upshape\tiny def.}}{=}}}

\newcommand{\bx}{\bm{x}}

\newcommand{\x}{\bm{\mathrm{x}}}

\newcommand{\mach}{\mathrm{Ma}}

\newcommand{\velocity}{\bm{\mathrm{v}}}
\newcommand{\Velocity}{\bm{\mathrm{u}}}

\newcommand{\latticevelocity}{\bm{\mathrm{c}}}

\newcommand{\F}{\boldsymbol{f}}

\newcommand{\Feqlattice}{\boldsymbol{f}^{\mathrm{eq}}}

\newcommand{\W}{\mathrm{W}}

\newcommand{\kh}[1]{{\color{blue}\textsf{\small KH: #1}}}

\usepackage{todonotes}
\tikzset{notestyleraw/.append style={sharp corners}}
\newif\ifreviewcomments
\reviewcommentstrue
\newif\ifshowresolvedcomments
\showresolvedcommentsfalse
\newcounter{reviewcomment}

\usepackage[capitalize,noabbrev]{cleveref}
\DeclareRobustCommand{\bluerev}[1]{\texorpdfstring{#1}{#1}}

\title{Learning the Entropic Relaxation: Stable Coarse-Grid Emulation with Neural Discrete Equilibrium}

\author{%
  Jose Antonio Lara Benitez \thanks{Equal contribution.}\\
  Rice University and LBNL\\
  \texttt{jlarabenitez@lbl.gov}
  \And
  Kareem Hegazy$^*$\\
  ICSI and University of California at Berkeley
  \AND
  Anastasis Kratsios\\
  McMaster University and the Vector Institute
  \And
  Ivan Dokmani\'{c}\\
  University of Basel
  \And
  Maarten V. de Hoop\\
  Rice University
  \And
  Michael W. Mahoney\\
  ICSI, LBNL, and University of California at Berkeley
}

\begin{document}

\maketitle

\begin{abstract}

The ability to emulate complicated and computationally-expensive physical systems in an efficient and stable manner is a popular motivation for scientific machine learning (SciML). 
Stable numerical solvers tend to be computationally expensive, as they attempt to resolve small (spatial and temporal) scales that receive and dissipate spectral energy. 
Learned coarse-grid forecasters remove the dissipative channel, allowing unresolved frequencies to re-emerge as grid-scale noise (aliasing). 
In this paper, we use the Neural Discrete Equilibrium (\emph{NeurDE}) method to explore stable emulation at extremely coarse grids, demonstrating physically-principled control over entropy dynamics that reinforces the physical system's dissipative mechanism.
\emph{Entropically-stabilized NeurDE} produces stable long-time rollouts for thousands of forecasting steps on large Reynolds number turbulent flows ($\mathrm{Re}=100{,}000$) with very coarse grids (down to $\mathbb{R}^{16\times16}$).
At these very coarse grid scales, \emph{entropically-stabilized NeurDE} can be more stable than even our stabilized data-generation numerical~method.
\end{abstract}

\section{Introduction}

\begingroup
\newcommand{\entropyintrorevision}[1]{{\color{red}#1}}

\begin{figure*}[t]
    \centering
    \resizebox{0.9\linewidth}{!}{\input{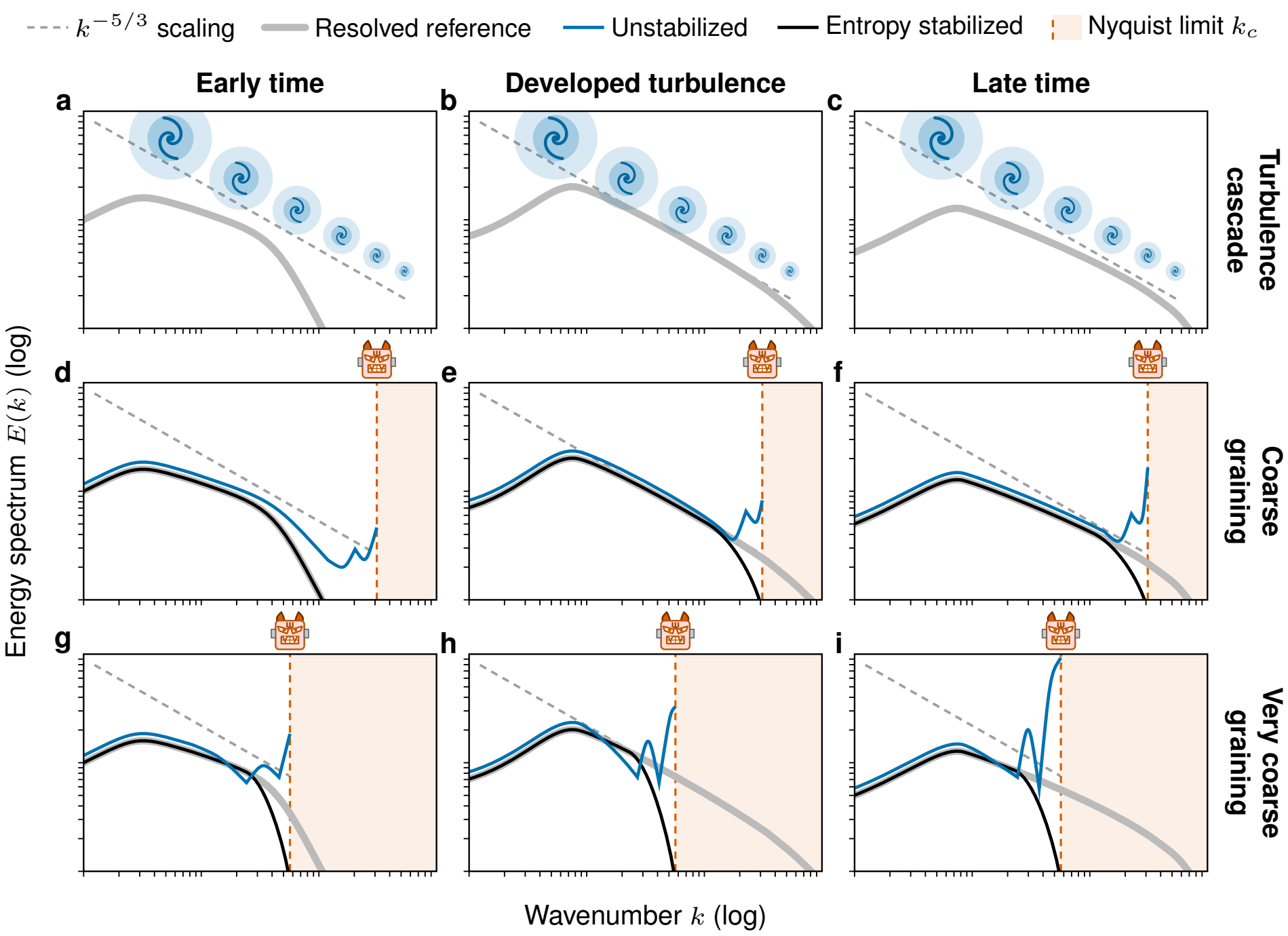}}
    \caption{\textbf{Coarse discretization interrupts scale-to-scale energy transfer.}
    Solid curves show the resolved spectral power at successive stages of a
    rollout; dashed lines indicate an idealized inertial-range decay. As
    turbulence develops, a coarse representation loses the high-wavenumber
    dissipative tail and can accumulate grid-scale content, unless the unresolved
    transfer is modeled. The learned entropic relaxation supplies a local
    dissipative channel for this missing transfer.\vspace{-2mm}}
    \label{fig:discretization}
\end{figure*}

Stable coarse-grid simulation can significantly reduce compute time, but it is challenging because discretizing the physical system disrupts the system's ability to dissipate energy~\citep{pope2000turbulent,sagaut2006large,meneveau2000scale}. Figure~\ref{fig:discretization} summarizes this failure mode.
In fluid dynamics, small-scale (high-frequency) structures such as eddies eventually dissipate via viscosity, converting kinetic energy into thermal energy at a characteristic dissipation scale $L$~\citep{pope2000turbulent,frisch1995turbulence}.
When the grid resolution $(\delta x)$ is fine enough to resolve this scale ($\delta x<L$), the simulation exhibits stability. Coarse grids, however, have $\delta x \gg L$: structures can no longer shrink toward $L$~\citep{ishihara2009study,sagaut2006large}.
The grid cutoff acts as a nonphysical ``boundary'' in Fourier space: truncating the physical cascade, aliasing, and forcing this closure error to feed unresolved content back into the represented state~\citep{canuto2006spectral,patterson1971spectral,gao2025discretization}.
Kinetic energy then accumulates in the highest resolved wavenumbers and re-enters the system as spurious noise~\citep{frisch1995turbulence,sagaut2006large} (bottom rows of Fig.~\ref{fig:discretization}).
This effect is strongest when the cutoff coincides with dynamically active scales.  In two-dimensional turbulence, for example, the inverse-energy and direct-enstrophy cascades make the direction of transfer scale-dependent~\citep{kraichnan1967inertial,batchelor1969computation,boffetta2012two,bohr1998dynamical,frisch1995turbulence}, so truncating these exchanges without a compensating dissipative mechanism lets the represented state gain spurious order or energy that the unresolved continuum system would have dissipated. 
Entropy stability constrains the discrete dynamics, including in shock capturing and turbulence, entropic stabilization also provides an implicit subgrid closure.
We call the associated error an \emph{entropy-stability error}.

A robust scientific machine learning (SciML) emulator must learn a mechanism that absorbs, redistributes, or dampens the energy that is not supported by the coarse grid~\citet{smagorinsky1963general,meneveau2000scale,sagaut2006large}.
Otherwise, SciML methods that learn simple representations or primarily interpolate the represented state during training will inherit this problem, as in Fig.~\ref {fig:discretization}.
This can be challenging in a purely data-driven setting, where the input data are often already coarsened, with the subgrid dynamics  removed~\citet{sakarvadia2025false,gao2025discretization}.

Physical principles can help supply what the data lack.
Conservation of mass and momentum is not enough, because a closure can conserve both exactly while injecting energy into the resolved scales, as a negative eddy viscosity does~\citet{smagorinsky1963general,sagaut2006large}.
The missing ingredient is the direction of dissipation.
In a fluid, viscosity only removes kinetic energy, and in kinetic theory this irreversibility is Boltzmann's H-theorem, the form of the second law stating that collisions never decrease entropy~\citet{kruger2017lattice,succi2001lattice}.
An entropy-stability error violates this direction, because energy that viscosity should have removed remains in the represented state.
Entropy-stable numerical methods avoid this error by enforcing a discrete entropy inequality~\citet{tadmor1987numerical,tadmor2003entropy}.

For learned closures, entropy offers a guide that does not depend on the data.
The entropy inequality involves only the model's own states, so it can be checked along any rollout and on any grid.
It does not replace data, however, because a closure that overdampens can remove critical flow dynamics.
Data, therefore, determine how much the closure should dissipate, and entropy determines which updates are admissible.
Physics-informed neural networks typically add such conditions as penalties or constraints on generic architectures~\citet{raissi2019physics,cai2021physics,patel2022thermodynamically,beucler2021enforcing}, rather than acting on the dissipative mechanism itself; but this popular ``physics-informed'' approach has well-known failure modes~\citet{krishnapriyan2021characterizing}.

In this work, we use the Neural Discrete Equilibrium (\emph{NeurDE}) framework~\citet{benitez2025neural} (a novel method that, among other things, can evolve a flow through lattice Boltzmann populations) to develop a method that reinforces dissipation, and we use our solution to explore stable emulation at extremely coarse grid scales.
Each lattice Boltzmann step streams these populations exactly between grid points, and then it relaxes them toward a local equilibrium in a collision step, which is where dissipation happens.
The standard Bhatnagar--Gross--Krook (BGK) collision~\citet{bhatnagar1954model} conserves mass and momentum but does not guarantee the H-theorem, and so it becomes unstable on underresolved grids at low viscosity~\citet{karlin1999perfect,boghosian2001entropic}.
Entropic lattice Boltzmann methods (ELBM) restore the H-theorem with a single relaxation parameter at each grid point; and this can act as a locally adaptive viscosity~\citet{boghosian2001entropic,karlin2003entropic,hosseini2023entropic}.
However, ELBM computes this parameter from a local entropy balance (Section~\ref{sec:entropic_stability}), which keeps each collision admissible but is not designed to reproduce the high-resolution dynamics on a coarse grid.
\citet{fischer2025optimal} learn $\alpha$ via multi-agent reinforcement learning, while assuming knowledge of the target's energy spectrum, thus limiting their method's applicability to general systems when reference spectral statistics are unavailable. Our entropy constraint does not require these statistics.
We instead learn the relaxation parameter with a Deep Equilibrium Model (DEQ)~\citet{bai2019deep}, whose fixed-point iterations mirror this entropic solve.
The model is trained to match filtered high-resolution data and to stay consistent with the H-theorem.
We call the resulting model \emph{entropically-stabilized NeurDE}, code available at \cite{hegazyCode}

We test \emph{entropically-stabilized NeurDE} on turbulent flow: a workhorse tool of modern science and engineering, which is generally limited in accuracy by the numerical resolution~\citet{ferziger2019computational,pope2000turbulent,ishihara2009study}.
A major challenge of simulating turbulence is not the governing equations, but instead that turbulence
generates a large range of scales and is scale-invariant~\citet{frisch1995turbulence,pope2000turbulent}.
Direct numerical simulation (DNS) must resolve the small structures that transfer and dissipate energy or enstrophy,
and the required grid size and stable time step grow rapidly with Reynolds number~\citet{ishihara2009study,tran2009number,pope2000turbulent}. SciML offers a complementary route to DNS~\citet{kochkov2021machine,azizzadenesheli2024neural,wu2022learning,mccabe2025walrus}.
Turbulence, therefore, is an ideal probe to evaluate how well \emph{entropically-stabilized NeurDE} and other ML and numerical benchmarks handle sub-grid dynamics.

Our main contributions are:
\begin{itemize}
    \item \textbf{An entropy-stabilized physics-based model (\emph{NeurDE}).} To achieve stable emulations at extremely coarse grids, we extend \emph{NeurDE} with an entropic closure that supplies the missing dissipative channel
    through a physically principled optimization task: enforcing the second law of thermodynamics~\citet{callen1985thermodynamics}.
    \item \textbf{Explicit fixed-point guarantees.} We prove that
    the DEQ~\citet{bai2019deep} update admits a fixed point which, if the update is a contraction, is unique and approached geometrically with DEQ depth, as is its entropy residual. Training enforces the second law there.
    \item \textbf{Challenging stress tests and benchmarks.} We demonstrate \emph{entropically-stabilized NeurDE's} stable emulation capabilities on turbulent systems with $\mathrm{Re}=100{,}000$, for $\mathcal{O}(1000)$ autoregressive steps, and on decreasing numerical grids: $\mathbb{R}^{128\times128}$ to $\mathbb{R}^{16\times16}$. We compare \emph{entropically-stabilized NeurDE} to the spectral Fourier Neural Operator (FNO)~\citet{li2021fourier} and the pretrained SciML foundation model Walrus~\citet{mccabe2025walrus} (models which are $\mathcal{O}(10^3)$ and $\mathcal{O}(10^7)$ times larger than \emph{entropically-stabilized NeurDE}). As a more challenging comparison, we benchmark against ELBM (the data generation~simulator).

\end{itemize}
\endgroup

\section{Related Work}
\label{sec:related_works}

\begingroup
\paragraph{The cost of resolved simulation.}
DNS resolves the active modes of a flow, but its cost
grows rapidly with Reynolds number and with the stable
time step required by the smallest resolved
scales~\citet{trebotich2024exascale,fischer2022nekrs}. Standard estimates put
the active degrees of freedom at
$O(\mathrm{Re}(1+\log \mathrm{Re})^{1/3})$ in two
dimensions~\citet{tran2009number,batchelor1969computation} and
$O(\mathrm{Re}^{9/4})$ in three dimensions~\citet{kolmogorov1991local,pope2001turbulent}.
Large-eddy simulation reduces this cost by evolving filtered variables and
closing the effect of unresolved motion through subgrid
models~\citet{meneveau2000scale,sagaut2006large}. The same lens applies to
learned coarse-grid forecasting: what is missing on the coarse grid is not
just resolution, but the effect of unresolved transfer on the represented
state.

\paragraph{Pure ML surrogates.}
A first family of methods replaces the solver entirely with a neural network map
between discretized function spaces. Neural operators such as
FNO~\citet{li2021fourier,kovachki2024operator} are flexible and fast, but
Fourier truncation and learned high-frequency transfer can limit robustness
in underresolved regimes~\citet{sakarvadia2025false}. 
\citet{benitez2024out} derive out-of-distribution risk bounds
for neural operators applied to the Helmholtz equation; and
\citet{kratsios2024mixtures} analyze active expert complexity in mixtures of
neural operators.
Physics-informed neural
networks encode continuous PDE residuals~\citet{raissi2019physics,cai2021physics},
yet their optimization landscape is fragile for nontrivial, e.g., stiff or multiscale
dynamics~\citet{krishnapriyan2021characterizing}. Foundation models for PDEs
such as Poseidon~\citet{herde2024poseidon} and
Walrus~\citet{mccabe2025walrus} broaden coverage by training on large
collections of downsampled simulations, at the cost of substantial data and
compute. The common thread is that solver-free surrogates have no built-in
mechanism to capture missing information from the grid, and stability over long
horizons typically has to be enforced through training objectives or
regularization, rather than by the integrator~itself.

\paragraph{Hybrid solvers with learned components.}
A second family of models keeps a numerical solver and inserts learning into specific
operators~\citet{brunton2020machine,kochkov2021machine}. The learned component
inherits locality, conservation, and discretization structure from the host
method, which improves generalization. 
Most prominently, \emph{NeurDE}~\citep{benitez2025neural} learns local equilibria
within a lattice Boltzmann solver.
Stability is the central
remaining difficulty: short-horizon objectives can accumulate error during
long rollouts, and one-step predictions that look plausible can still violate
invariants or dissipative structure. Existing remedies include
physics-informed penalties~\citet{beucler2021enforcing,raissi2019physics},
long-horizon closure optimization~\citet{fischer2025optimal}, and statistical
stabilization such as the Thermalizer~\citet{pedersen2025thermalizer}. These
add stability as an ad hoc training-time correction or post-hoc filter, rather than as
a property of the integrator~itself.

\endgroup

\section{Entropy-Stabilized Kinetic Closure}

\begingroup
\subsection{Lattice Boltzmann Background}
\label{sec:neurde}

The kinetic backbone of \emph{NeurDE} is the lattice Boltzmann method (LBM), 
which discretizes the Boltzmann equation by representing the one-particle distribution $f(t,\bx,\velocity)$ through populations
$\F_i(t,\bx)$ at discrete single-particle velocities $\velocity_i$~\citet{bellotti2025fourth}.
The Bhatnagar--Gross--Krook (BGK) collision operator~\citet{bhatnagar1954model} relaxes these populations toward equilibrium $\Feqlattice_i$:
\begin{equation}
    \F_i(t+1,\bx+\velocity_i)
    =\F_i(t,\bx)+\frac{1}{\tau}
    \bigl[\Feqlattice_i(\boldsymbol{U}(t,\bx))-\F_i(t,\bx)\bigr],
    \label{eq:main_BGK}
\end{equation}
where $\tau$ is the relaxation time and $\boldsymbol{U}$ contains the conserved macroscopic variables. 
Each step combines a local, nonlinear collision with exact streaming: the post-collision population $\F_i^{\mathrm{coll}}(t,\bx)$ moves to $\bx+\velocity_i$.

Within this model, macroscopic observables are recovered as moments of the populations:
\begin{equation}
    \boldsymbol{U}=(\rho,\rho\Velocity,\rho E)^\top
    =\sum_i(1,\velocity_i,\tfrac12|\velocity_i|^2)^\top\F_i,
    \label{eq:main_macroscopic}
\end{equation}
where $\rho$, $\Velocity$, and $E$ denote density, macroscopic velocity, and specific total energy; the energy component applies to energy-conserving lattices.
\emph{NeurDE}~\citet{benitez2025neural} learns the equilibrium map $\boldsymbol{U}\mapsto\Feqlattice(\boldsymbol{U})$, 
while retaining the prescribed streaming step. Collision preserves the conserved moments, provided the predicted equilibrium satisfies the same moment constraints
as the incoming populations. 
The full operator splitting and original formulation are retained in Appendix~\ref{appx:lbm_background}.
\endgroup
\begingroup
\subsection{Entropic Stability}
\label{sec:entropic_stability}

Conservation alone does not ensure nonnegative entropy production. 
For positive pre-collision populations $\F$ and lattice weights $w_i>0$, the discrete H-function and collision stability condition are
\begin{align}
    H(\F) &= \sum_i \F_i\log\left(\frac{\F_i}{w_i}\right),
    \label{eq:H_fxn}\\
    H(\F^{\mathrm{coll}}) &\le H(\F).
    \label{eq:H_condition}
\end{align}
The inequality is the discrete H-theorem: physical entropy increases as $H$ decreases. 
By enforcing Eq.~\ref{eq:H_condition}, one also enforces the second law of thermodyanics; see Appendix~\ref{appx:entropic_background} for the thermodynamic interpretation. 
Classical entropic LBM (ELBM)~\citet{karlin1999perfect} uses an equilibrium $\Feqlattice$ minimizing $H$ at fixed conserved moments and seeks $\alpha>0$ satisfying
\begin{equation}
    \F^{\mathrm{mirr}}(\alpha)
    =\F+\alpha(\Feqlattice-\F),
    \qquad H(\F^{\mathrm{mirr}}(\alpha))=H(\F).
    \label{eq:f_mirror}
\end{equation}
When such a positive mirror state exists, the collision
\begin{equation}
    \F^{\mathrm{coll}}
    =(1-\beta)\F+\beta\F^{\mathrm{mirr}}(\alpha)
    =\F+\alpha\beta(\Feqlattice-\F)
    \label{eq:ELBM_collision}
\end{equation}
satisfies Eq.~\ref{eq:H_condition}, by the convexity of $H$, with $\beta=1/(2\tau)\in(0,1]$ set by the nominal viscosity (Appendix~\ref{appx:mirror}). 
Near entropic equilibrium, $\alpha\approx2$ recovers BGK. At fixed $\beta$, $\alpha<2$ increases the effective viscosity relative to BGK, 
while $\alpha>2$ decreases it~\citet{hosseini2023entropic}.
The former promotes local smoothing and damping of high-frequency components, while the latter reduces damping and may permit backscatter~\citep{hosseini2023entropic,buzzicotti2021inertial}.
Therefore, the $\alpha$ parameter provides a physically-principled mechanism to modulate resolved high-frequency components; and thus it opens the possibility of modeling energy transfer across the grid cutoff (the ``Maxwell's demon'' in Fig.~\ref{fig:discretization}), without changing the Shannon-Nyquist limit.

This entropy-balance equation defines a local nonlinear solve at each site and time step. 
We replace this expensive solve with a Deep Equilibrium Model (DEQ; Section~\ref{sec:method}), using entropy consistency as a training target and rollout diagnostic. 
A learned fixed point does not automatically satisfy Eq.~\ref{eq:H_condition}.
\endgroup

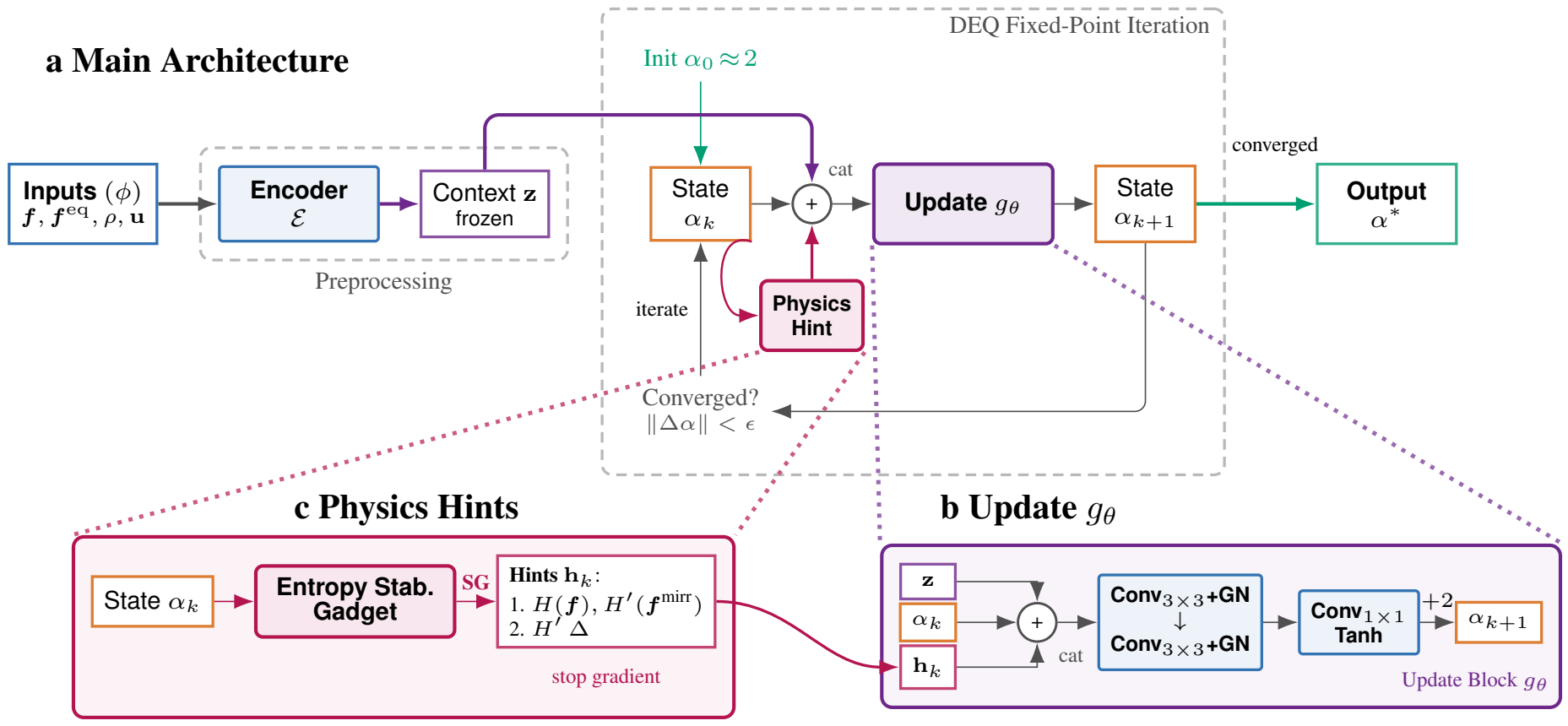
\begin{figure*}[t]
\centering
\resizebox{0.9\linewidth}{!}{\begin{tikzpicture}[
    x=1cm, y=1cm,
    node distance=0.6cm,
    >={Latex[length=2mm, width=1.5mm]},
    font=\sffamily\footnotesize,
    base_line/.style={line width=0.8pt},
    thick_line/.style={line width=1.0pt},
    distinct_dash/.style={dash pattern=on 3pt off 2pt},
    tensor/.style={
        rectangle,
        draw=#1!80,
        fill=white,
        text=black,
        base_line,
        minimum width=1.0cm,
        minimum height=0.6cm,
        font=\sffamily\scriptsize,
        align=center
    },
    block/.style={
        rectangle,
        draw=#1!80,
        fill=#1!7,
        base_line,
        minimum width=1.6cm,
        minimum height=0.75cm,
        align=center,
        font=\sffamily\scriptsize\bfseries,
        rounded corners=1pt
    },
    update_block/.style={
        rectangle,
        draw=stage1,
        fill=stage1!10,
        line width=1.0pt,
        minimum width=1.8cm,
        minimum height=0.8cm,
        align=center,
        font=\sffamily\scriptsize\bfseries,
        rounded corners=2pt
    },
    entropy_gadget/.style={
        rectangle,
        draw=physics,
        fill=physics!10,
        line width=1.0pt,
        minimum width=2.0cm,
        minimum height=0.7cm,
        align=center,
        font=\sffamily\scriptsize\bfseries,
        rounded corners=2pt
    },
    op/.style={
        circle,
        draw=grey,
        fill=white,
        line width=0.7pt,
        inner sep=0pt,
        minimum size=1.1em,
        font=\tiny
    },
    groupbox/.style={
        rectangle,
        draw=grey!40,
        distinct_dash, 
        inner sep=6pt,
        rounded corners=3pt,
        line width=0.8pt
    },
    panel_label/.style={
        font=\normalsize\bfseries,
        text=black,
        anchor=north west
    }
]

\node[panel_label] (label_a) at (-4.5, 0.5) {a Main Architecture};

\node[tensor=neural, minimum width=1.4cm, minimum height=0.8cm] (inputs) at (-4.0, -1.2) {\textbf{Inputs} $(\phi)$\\[-1pt]{\tiny $\F, \Feqlattice, \rho,\Velocity$}};

\node[block=neural, right=0.6cm of inputs] (encoder) {Encoder\\$\mathcal{E}$};
\node[tensor=stage1, right=0.4cm of encoder] (z) {Context $\mathbf{z}$\\[-1pt]{\tiny frozen}};

\draw[->, draw=grey, thick_line] (inputs) -- (encoder);
\draw[->, draw=stage1, thick_line] (encoder) -- (z);

\node[tensor=stage2, right=1.0cm of z] (alpha) {State\\$\alpha_k$};
\node[above=0.85cm of alpha, font=\scriptsize, text=stage3] (init) {Init $\alpha_0\!\approx\!2$};
\draw[->, draw=stage3] (init) -- (alpha);

\node[op, right=0.4cm of alpha] (concat) {+};

\node[update_block, right=0.4cm of concat] (update) {Update $g_\theta$};

\node[entropy_gadget, below=0.55cm of concat, minimum width=.3cm, minimum height=.7cm, font=\sffamily\tiny\bfseries] (gadget_a) {Physics\\Hint};

\node[tensor=stage2, right=0.4cm of update] (next_alpha) {State\\$\alpha_{k+1}$};

\node[below=1.35cm of alpha, font=\scriptsize, text=grey, align=center] (check) {Converged?\\[-1pt]{\tiny $\|\Delta\alpha\| < \epsilon$}};

\draw[->, draw=grey] (alpha) -- (concat);
\draw[->, draw=grey] (concat) -- (update);
\draw[->, draw=grey] (update) -- (next_alpha);

\draw[->, draw=physics, line width=0.8pt] (gadget_a.north) -- (concat.south);
\node[above right=0.01cm and -0.1cm of concat, font=\tiny, text=grey] {cat};

\draw[->, draw=physics, line width=0.6pt] (alpha.south east) to[out=-200, in=180, looseness=1.5] (gadget_a.west);

\draw[->, draw=grey, rounded corners=3pt] (next_alpha.south) |- (check.east);
\draw[->, draw=grey] (check.north) -- node[pos=0.5, left, font=\tiny] {iterate} (alpha.south);

\draw[->, draw=stage1, thick_line, rounded corners=4pt] 
    (z.north) -- ++(0, 0.55) -| (concat.north);

\node[tensor=stage3, right=1.2cm of next_alpha, minimum width=1.4cm, minimum height=0.8cm] (output) {\textbf{Output}\\$\alpha^*$};
\draw[->, draw=stage3, thick_line] (next_alpha) -- node[midway, above, font=\tiny, yshift=10pt, xshift=5.5pt] {converged} (output);

\begin{scope}[on background layer]
    \node[groupbox, fit=(encoder) (z), inner sep=5pt, label={[font=\scriptsize, text=grey, yshift=-0.45cm, anchor=south]south:Preprocessing}] (g_pre) {};
    \node[groupbox, fit=(alpha) (next_alpha) (check) (concat) (init) (update) (gadget_a), inner sep=8pt, label={[font=\scriptsize, text=grey, anchor=north east, xshift=-1pt, yshift=1pt]north east:DEQ Fixed-Point Iteration}] (g_loop) {};
\end{scope}

\node[panel_label] (label_c) at (-2., -4.) {c Physics Hints};

\node[tensor=stage2, minimum height=0.5cm] (c_alpha) at (-3.3, -5.2) {State $\alpha_k$};

\node[entropy_gadget, right=0.4cm of c_alpha, minimum width=2.0cm, minimum height=0.55cm] (phys_eng) {
    Entropy Stab.\\[-1pt]Gadget
};

\node[tensor=physics, right=0.4cm of phys_eng, align=left, font=\tiny, minimum width=1.2cm, minimum height=0.55cm] (c_hints) {
    \textbf{Hints $\mathbf{h}_k$}:\\
    1. $H(\F), H'(\F^{\text{mirr}})$\\
    2. $H'$ $\Delta$
};

\draw[->, draw=physics] (c_alpha) -- (phys_eng);
\draw[->, draw=physics] (phys_eng) -- node[above, font=\tiny\bfseries, text=physics] {SG} (c_hints);

\node[below=0.08cm of c_hints, font=\tiny, text=physics, align=center] (sg_text) {stop gradient};

\begin{scope}[on background layer]
    \node[groupbox, draw=physics, solid, fill=physics!5, line width=1.0pt, fit=(c_alpha) (c_hints) (phys_eng) (sg_text), inner sep=5pt] (gc) {};
\end{scope}

\node[panel_label] (label_b) at (4.5, -4.) {b Update $g_\theta$};

\node[tensor=stage1, minimum width=0.55cm, minimum height=0.35cm, font=\sffamily\tiny] (b_z) at (4.5, -5.0) {$\mathbf{z}$};
\node[tensor=stage2, below=0.02cm of b_z, minimum width=0.55cm, minimum height=0.35cm, font=\sffamily\tiny] (b_alpha) {$\alpha_k$};
\node[tensor=physics, below=0.02cm of b_alpha, minimum width=0.55cm, minimum height=0.35cm, font=\sffamily\tiny] (b_h) {$\mathbf{h}_k$};

\node[op, right=0.6cm of b_alpha] (b_concat) {+};
\node[below right=0.01cm and -0.05cm of b_concat, font=\tiny, text=grey] {cat};

\draw[->, draw=grey] (b_z.east) -| (b_concat.north);
\draw[->, draw=grey] (b_alpha.east) -- (b_concat.west);
\draw[->, draw=grey] (b_h.east) -| (b_concat.south);

\node[block=neural, right=0.4cm of b_concat, minimum width=1.5cm, minimum height=0.5cm, align=center, font=\sffamily\tiny\bfseries] (layers) {
    Conv$_{3\times3}$+GN\\
    $\downarrow$\\
    Conv$_{3\times3}$+GN
};

\node[block=neural, right=0.35cm of layers, minimum width=0.9cm, minimum height=0.5cm, font=\sffamily\tiny\bfseries] (head) {
    Conv$_{1\times1}$\\
    Tanh
};

\node[tensor=stage2, right=0.35cm of head, minimum width=0.55cm, minimum height=0.35cm, font=\sffamily\tiny] (b_out) {$\alpha_{k+1}$};

\draw[->, draw=grey] (b_concat) -- (layers);
\draw[->, draw=grey] (layers) -- (head);
\draw[->, draw=grey] (head) -- node[above, font=\tiny] {$+2$} (b_out);

\begin{scope}[on background layer]
    \node[groupbox, draw=stage1, solid, fill=stage1!5, line width=1.0pt, fit=(b_z) (b_out) (b_h) (layers) (head), inner sep=5pt] (gb) {};
\end{scope}

\node[font=\tiny, text=stage1, anchor=north east, yshift=0.5cm] at (gb.south east) {Update Block $g_\theta$};

\draw[dotted, line width=1.2pt, stage1!70] (update.south west) -- (gb.north west);
\draw[dotted, line width=1.2pt, stage1!70] (update.south east) -- (gb.north east);

\draw[dotted, line width=1.2pt, physics!70] (gadget_a.south west) -- (gc.north west);
\draw[dotted, line width=1.2pt, physics!70] (gadget_a.south east) -- (gc.north east);

\draw[->, draw=physics, line width=0.8pt] 
    (c_hints.east) to[out=0, in=180] (b_h.west);

\path (current bounding box.north west) +(-0.2,0.2) coordinate (nw_margin);
\path (current bounding box.south east) +(0.2,-0.4) coordinate (se_margin);
\useasboundingbox (nw_margin) rectangle (se_margin);

\end{tikzpicture}}
\caption{\textbf{Architecture of \emph{entropically-stabilized NeurDE}.} \textbf{a} The pipeline decouples feature extraction from the iterative DEQ solver. Context $\mathbf{z}$ is encoded \textit{once} and injected into every iteration. \textbf{b} The neural update block processes the concatenated context, state, and physics hints through a convolutional network to refine the stability parameter $\alpha$. \textbf{c} Physics hints include: entropy residual $G = H(\alpha) - H_\mathrm{pre}$, gradients $\Delta = H'(\alpha)$, and instability indicator. The physics hints are computed outside of the computational graph, ensuring gradients flow only through $g_\theta$ (implicit differentiation).}
\label{fig:architecture}
\end{figure*}

\section{Entropically-stabilized NeurDE Method}
\label{sec:method}

\begingroup
\emph{NeurDE}~\citet{benitez2025neural} is a method that leverages the Boltzmann equation backbone to inherit conservation and isolate the learning problem to the unknown physics closure.
Specifically, \emph{NeurDE} predicts the mesoscopic equilibrium single-particle distribution $\Feqlattice_i$ based on the macroscopic variables.
\emph{NeurDE's} neural equilibrium model improves on the fixed-time entropy
extremizer~\citet{benitez2025neural,levermore1996moment}; but this does not guarantee entropy stability across temporal rollouts.
Coarse-grid rollout fails for a reason no equilibrium closure can fix: the dissipative channel is absent from the grid, rather than mispredicted, and so an arbitrarily accurate $\Feqlattice$ cannot resolve grid-scale or finer features, and this spectral energy then re-enters as grid-scale noise.
Entropic stabilization supplies that channel.
We therefore swap the BGK collision in \emph{NeurDE} for the entropic BGK collision
of Eq.~\ref{eq:ELBM_collision}. Under the collision's admissibility and positivity
conditions, our ML surrogate aims to satisfy Eq.~\ref{eq:H_condition} by predicting $\alpha$ that minimizes the difference between each side of the equation in~Eq.~\ref{eq:f_mirror}.

\subsection{Architecture}
\label{sec:arch}

ELBM solvers typically find $\alpha$ through Newton-type iterations at each grid point and time step~\citep{karlin1999perfect,hosseini2023entropic,jonnalagadda2021single,frapolli2015entropic,ansumali2002single}.
This Newton approach derives each step from a local linearization of the entropy residual.  Instead, the DEQ step is based on the Newton optimization scheme, swapping the Newton iterations for neural network depth and improving the optimization step by a nonlinear step calculated through learned multiscale features.
Its fixed point determines $\alpha$ (Fig.~\ref{fig:architecture}).

The encoder maps mesoscopic and macroscopic inputs to a latent context
$\mathbf{z}$ \emph{once} per the collision step (Fig.~\ref{fig:architecture}a).
Each DEQ iteration then updates the current estimate $\alpha_k$ from
$(\mathbf{z},\alpha_k,\mathbf{h}_k)$, where $\mathbf{h}_k$ contains detached
physics hints (entropy residual, residual derivative, and an instability
indicator; Fig.~\ref{fig:architecture}b--c). The fixed point of this update
is the entropic relaxation parameter we use in the collision. Two design
choices are worth flagging. The hints guide the iteration but live outside
the computational graph, so gradients flow through the learned update map
rather than through a hand-coded Newton solve. 
The encoder is evaluated once
and reused across iterations, so the cost per collision scales with the
number of DEQ steps to convergence rather than with repeated feature
extraction. Appendix~\ref{appx:architecture} gives the architectural details.
\endgroup



\subsection{Fixed-Point Guarantees}
\label{sec:Foundations}
\begingroup
We analyze fixed-point existence and entropy-residual convergence under explicit assumptions on the learned DEQ update.

Let $d,F\in\mathbb{N}_{>0}$, $\Omega\subset\mathbb{R}^d$ be a bounded domain, and $\mathcal{F}$ be the feature space. 
For input $\phi\in\mathcal{F}$ (here $\phi=(\F,\Feqlattice,\rho,\Velocity)$), 
the encoder produces $\mathbf{z}=\mathcal{E}(\phi)\in\mathbb{R}^F$, 
and the DEQ update induces $G_{\theta,\mathbf{z}}:H^1(\Omega)\to H^1(\Omega)$ by $G_{\theta,\mathbf{z}}(\alpha)=g_\theta(\alpha,\mathbf{z})$.
The analysis requires the operator-level self-map and Lipschitz properties in Assumption~\ref{ass:Small}, allowing pointwise or convolutional realizations.

Fix $\beta\in[0,1]$ and a mirror-state operator $\F^{\text{mirr}}:H^1(\Omega)\to L^1_+(\Omega)$.
We use the representative continuous entropy
\begin{equation}
    H(\F) \eqdef \int_{\Omega} \F(\velocity) \log(\F(\velocity)) \, d\velocity,
\end{equation}
with the convention $0\log 0=0$. The weighted lattice entropy in Eq.~\ref{eq:H_fxn} is its finite-dimensional analogue. 
Define the entropy-balance residual $h_{\beta}:H^1(\Omega)\times L^1_+(\Omega)\to\mathbb{R}\cup\{\infty\}$ by
\begin{equation}
\label{eq:EBF}
\begin{aligned}
        h_{\beta}(\alpha; \F)
    \eqdef
        H\big(
            (1 - \beta) \F + \beta  \F^{\text{mirr}}(\alpha)
        \big)
    -
        H(\F) .
\end{aligned}
\end{equation}
At $\beta=1$, the root condition $h_1(\alpha;\F)=0$ recovers the entropy-balance equation $H(\F^{\text{mirr}}(\alpha))=H(\F)$. We assume $H(\F)< \infty$ throughout.

\begin{theorem}[\bluerev{Existence of Neural Equilibria}]
\label{thrm:existence_of_FP}
Under Assumption~\ref{ass:Small}, for every $\phi\in \mathcal{F}$ with $\mathbf{z}=\mathcal{E}(\phi)$, 
the induced operator $G_{\theta,\mathbf{z}}:H^1(\Omega)\to H^1(\Omega)$ is well defined and admits at least one equilibrium $\alpha^{\star}\in\mathcal{K}\subset H^1(\Omega)$
satisfying
\begin{equation}
\label{eq:existence_of_FP}
    \alpha^{\star}
=
    G_{\theta,\mathbf{z}}(\alpha^{\star})
    = g_\theta(\alpha^{\star},\mathbf{z})
.
\end{equation}
\end{theorem}
\begin{proof}
    See Appendix~\ref{s:Proofs__ss:existence_FPs}.
\end{proof}
On a fixed grid, Brouwer's theorem applies, provided the update is a continuous self-map of a compact convex set.
The continuum $H^1$ statement remains conditional on Assumption~\ref{ass:Small}, since the clamp bounds pointwise values, not the $H^1$ seminorm.
If the $L^2$-Lipschitz constant satisfies $L_v<1$, this fixed point is unique, and iterates initialized in $\mathcal{K}$ converge geometrically to it (Appendix~\ref{s:Proofs__ss:UniqExpConv}).
$L_v<1$ remains an assumption in our DEQ formulation~\citet{bai2019deep}, whereas monDEQ~\citet{winston2020monotone} guarantees a unique equilibrium by construction.
Under the boundedness and Lipschitz assumptions below, the entropy residual inherits this convergence rate.
\begin{theorem}[\bluerev{Exponential Convergence of Entropy-Balance Residuals}]
\label{thrm:EBF_Convergence}
In the setting of Theorem~\ref{thrm:existence_of_FP} and Proposition~\ref{prop:uniqueness_convergence}, let $0\le \beta\le 1$ and $\F\in L^1_+(\Omega)$. 
Assume that Assumption~\ref{ass:entropy_bounds} holds on a set $\mathcal{A}$. 
If $L_v<1$ and if $\alpha^{\star}$ and all but finitely many iterates $(\alpha^{(i)})_{i=0}^{\infty}$ of $G_{\theta,\mathbf{z}}$ are in $\mathcal{A}$, 
then, for all sufficiently large $i$,
\[
        \big|
                h_{\beta}(\alpha^{\star};\F)
            -
                h_{\beta}(\alpha^{(i)};\F)
        \big|
    \in
        \mathcal{O}\big(
            L_v^{i}
        \big)
.
\]
\end{theorem}
\begin{proof}
    See Appendix~\ref{s:Proofs__ss:Exp_fast_Convergence}.
\end{proof}
DEQ convergence does not imply an exact ELBM entropy root. 
$\mathcal{L}_{\mathrm{root}}$ and $\mathcal{R}_{\mathrm{H}}$ (Eq.~\ref{eq:losses}) target $h_1=0$ and $h_\beta\le0$, respectively. 
The shifted $\tanh$ restricts $\alpha$ to $[1,3]$ (Table~\ref{tab:app_neurde_hyperparameters}), excluding the trivial root $\alpha=0$, which leaves the populations unchanged.
A soft penalty on training states is not a test-time guarantee; compliance is empirical (see Fig.~\ref{fig:dpsl50_entropy}).
\endgroup

\section{Empirical Results}\label{sec:experiments_main}

We evaluate \emph{entropically-stabilized NeurDE's} emulation capabilities on extremely under-resolved grids, which drastically disrupts dissipation.
We use the 2D doubly periodic shear layer (DPSL) and the 2D Taylor Green Vortex (TGV).
DPSL is a standard benchmark for vortex roll-up, secondary instability, and vortex merger in two-dimensional turbulence~\citet{brown1995performance,minion1997performance,bedrunka2021lettuce,de2020multiphysics}.
TGV is a standard benchmark for evaluating vorticity stability and testing whether \emph{entropically-stabilized NeurDE} oversmoothes and disrupts stable dynamics.
Both DPSL and TGV experiments are low-Mach ELBM simulations of the weakly compressible Navier--Stokes shear-layer
(Appendix~\ref{appx:experiments}).
For DPSL, $\mathrm{Re}=50{,}000$ and $\mathrm{Re}=100{,}000$, for TGV $\mathrm{Re}=50{,}000$.
Both systems have periodic boundary conditions.
We train four randomly initialized models on the first half of each simulated data and test on both halves.

We benchmark \emph{entropically-stabilized NeurDE} against state-of-the-art SciML models and the ELBM that simulated the datasets.
The FNO~\citet{li2021fourier}, our first SciML benchmark, is a data-driven macroscopic spectral method that learns an
autoregressive field-to-field mapping.
Walrus~\citet{mccabe2025walrus}, our second method, is a 1.3B parameter transformer foundation model pretrained on $\mathcal{O}(10)$~TB of data.
FNO and Walrus are respectively $\mathcal{O}(10^2)$ and $\mathcal{O}(10^4)$ times larger than \emph{NeurDE}.
ML methods are rarely compared against the numerical method that generated their training data, because they rarely outperform it on accuracy and stability for challenging systems.
As a challenging benchmark, we compare against the numerical ELBM designed for stable and accurate simulations.

For \emph{entropically-stabilized NeurDE} and FNO, we train autoregressively with increasing length.
FNO and Walrus are trained on the macroscopic observables for the simplest $\mathbb{R}^{256\times256}$ case.
\emph{Entropically-stabilized NeurDE} is first trained on $\alpha$ ground truth at $\mathbb{R}^{256\times256}$ and $\mathbb{R}^{512\times512}$ grids, for the $\mathrm{Re}=50{,}000$ and $\mathrm{Re}=100{,}000$ experiments, respectively.
After, we autoregressively train on the macroscopic observables, continuing to coarsen the grid by a factor of two until reaching $\mathbb{R}^{16\times16}$.
When evaluating at $\mathbb{R}^{N\times N}$, we use the checkpoints trained up to $\mathbb{R}^{N\times N}$.
Since Walrus and \emph{entropically-stabilized NeurDE} are very different architectures, we fine-tune the pretrained Walrus within its own repository.
We train Walrus for roughly twice as long.
Further training details are provided in Appendices~\ref{app:Training} (\emph{NeurDE}), \ref{app:FNO_appx} (FNO), and \ref{app:walrus_appx} (Walrus).



\begin{figure}[t]
    \centering
    \begin{minipage}[t]{0.40\textwidth}
        \centering
        \includegraphics[width=\textwidth]{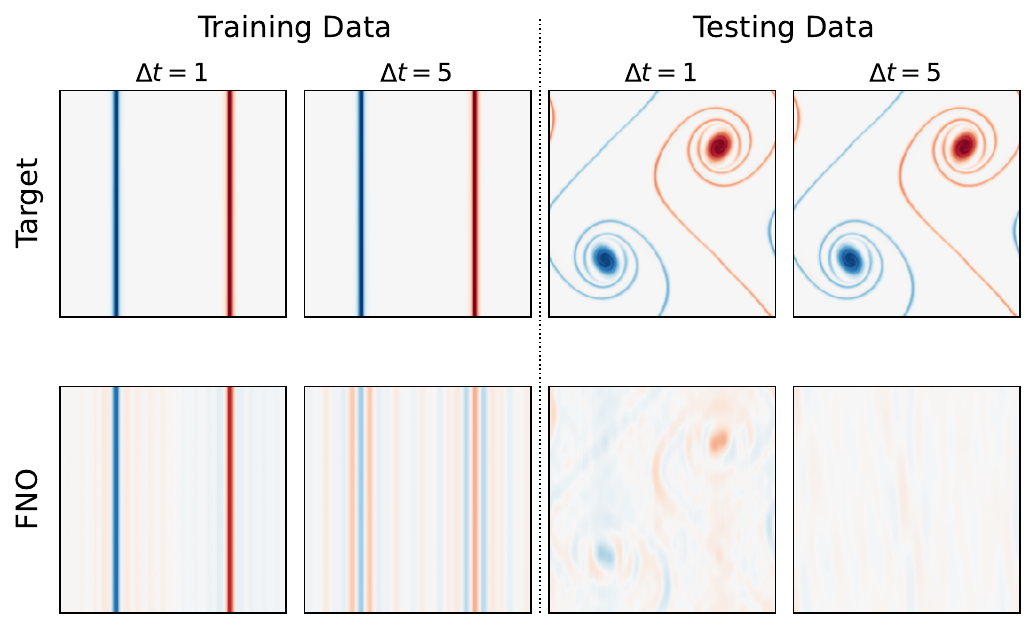}
        \caption{\textbf{FNO rollout.} Vorticity forecast on DPSL-50K at $\mathbb{R}^{256\times256}$ for 5 steps in train (left) and test (right) sets.}
        \label{fig:fno_forecast}
    \end{minipage}
    \hfill
    \begin{minipage}[t]{0.58\textwidth}
        \centering
        \includegraphics[width=\textwidth]{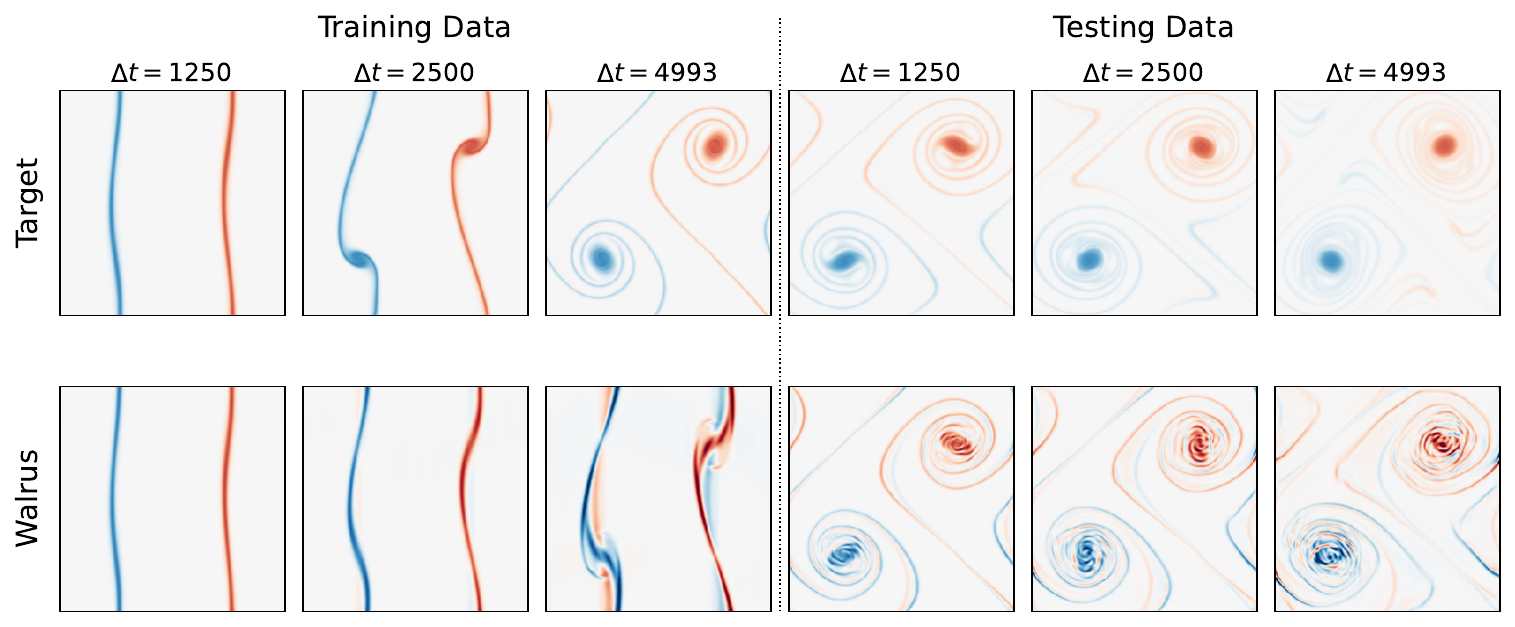}
        \caption{\textbf{Walrus rollout.} Walrus vorticity forecast on DPSL-50K at $\mathbb{R}^{256\times256}$ for the entire train (left) and test (right) datasets, roughly 5000 steps each.}
        \label{fig:walrus_forecast}
    \end{minipage}
\end{figure}

\begingroup
\begin{figure*}[t]
    \centering
    \begin{tikzpicture}
        \node[inner sep=0, anchor=south west] (img) {\includegraphics[width=0.85\linewidth]{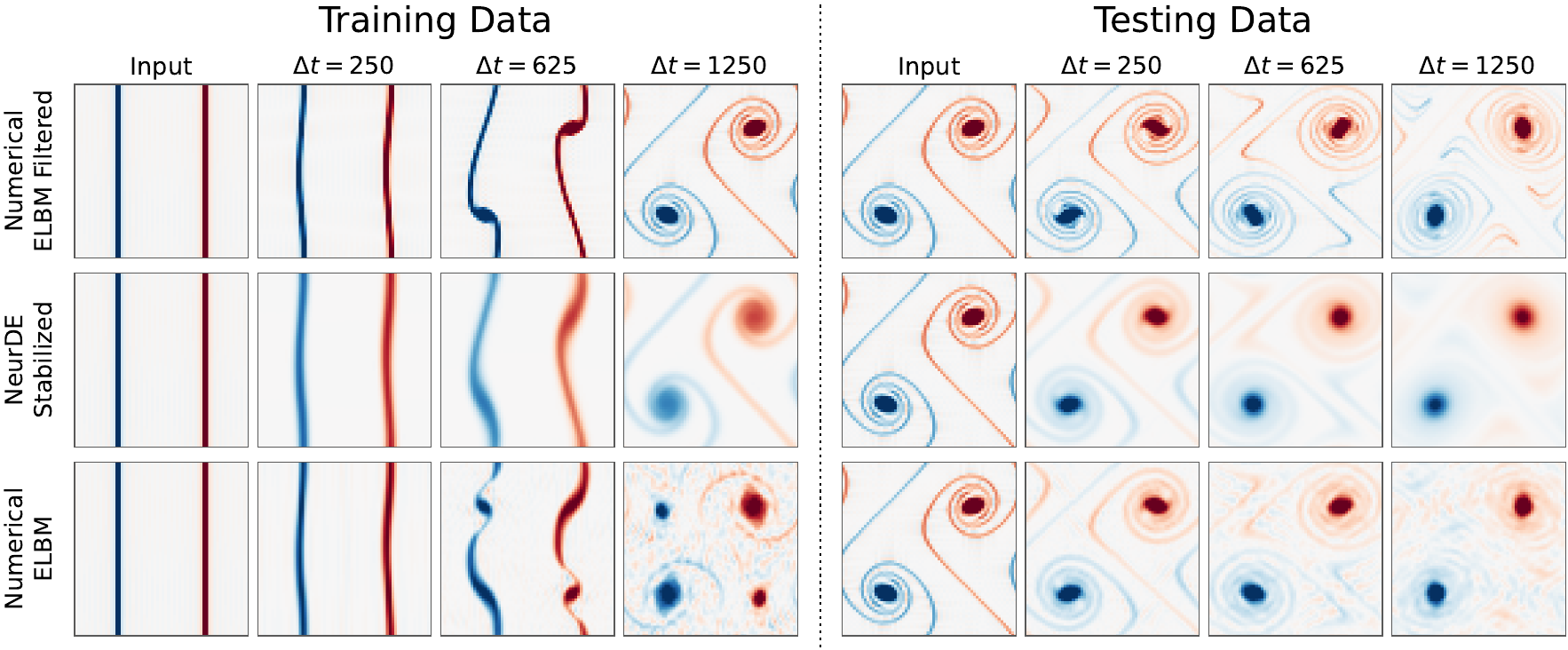}};
        \node[anchor=north west, inner sep=1pt, font=\bfseries] at (img.north west) {(a)};
    \end{tikzpicture}\par\vspace{4pt}
    \begin{tikzpicture}
        \node[inner sep=0, anchor=south west] (img) {\includegraphics[width=0.85\linewidth]{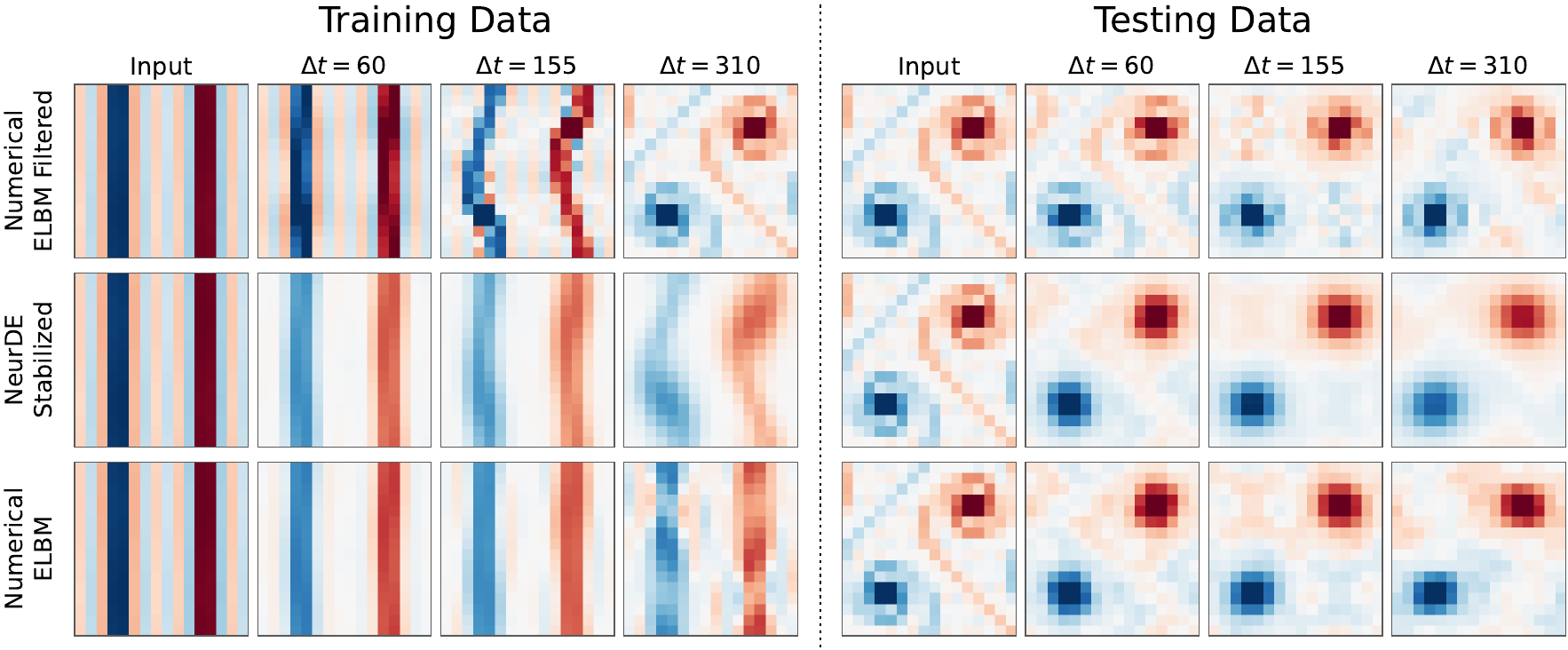}};
        \node[anchor=north west, inner sep=1pt, font=\bfseries] at (img.north west) {(b)};
    \end{tikzpicture}
    \caption{\textbf{Long-rollout comparison on the doubly periodic shear
    layer at Re=100,000.} Vorticity forecasts for the box-filtered high-resolution reference (top), \emph{entropy-stabilized NeurDE} (middle), and an underresolved
    ELBM baseline (bottom) on (a) a $64\times64$ grid and (b) a $16\times16$ grid.\vspace{-1mm}}
    \label{fig:dpsl100_forecast_64}
\end{figure*}

\paragraph{Results.}
We begin by evaluating the FNO and Walrus benchmarks, which fail to fully forecast through the simplest ($\mathbb{R}^{256\times256}$) DPSL-50K.
FNO struggles to properly take a single step and fails after 5 autoregressive steps; see Fig.~\ref{fig:fno_forecast}.
Walrus shows stable forecasts within the first $\Delta t = 1250$, but it fails to develop the vortices around $\Delta t=2500$ in the training dataset and subsequently diverges; see Fig.~\ref{fig:walrus_forecast}.
We note that incompressible turbulent dynamics were included in Walrus' pretraining data, although not this exact problem~\citet{mccabe2025walrus}.
At $\Delta t=2500$ in the test dataset, the tails no longer align and high-frequency noise enters into the center of the vortices.
In fact, test images at all times look very similar, indicating Walrus may be struggling to capture the dynamics.
Since current ML surrogates struggle to accurately emulate this high Re system, we focus on our more challenging comparison with the numerical ELBM method that generated the data.

On our most challenging dataset, DPSL-100k, \emph{entropically-stabilized NeurDE} forecasts through the entire training and test datasets for all coarsened resolutions: $\mathbb{R}^{128\times128}$, $\mathbb{R}^{64\times64}$, $\mathbb{R}^{32\times32}$, and $\mathbb{R}^{16\times16}$.
Figure~\ref{fig:dpsl100_forecast_64}a and b compare the results of \emph{entropically-stabilized NeurDE} (middle row) and the ELBM forecast (bottom row) at $\mathbb{R}^{64\times64}$ and $\mathbb{R}^{16\times16}$, respectively.
The top (ELBM filtered) row is ELBM simulations at $\mathbb{R}^{512\times512}$ and downsampled, while the bottom row is ELBM run at the coarse resolution.
ELBM at $\mathbb{R}^{64\times64}$, shows grid-scale spurious high-frequency noise as early as $\Delta t = 625$ (Fig.~\ref{fig:dpsl100_forecast_64}a).
These figures demonstrate the necessary tradeoff between stability and accuracy.
\emph{Entropically-stabilized NeurDE} stably reproduces the correct dynamics for the remaining DPSL-100k experiments and all but one DPSL-50k seed (Appendices~\ref{app:results_dp100k} and \ref{app:results_dp50k}); that seed struggles only at $\mathbb{R}^{128\times128}$, but it still outperforms the ELBM at every resolution.

\FloatBarrier
\begin{table}[t]
  \centering\small
  \caption{Doubly periodic shear layer at $\mathrm{Re}=100{,}000$: forecast metrics from the training data ($t_0=0$) at the final lead time. \emph{Entropically-stabilized NeurDE} results are the mean $\pm$ standard deviation over four seeds; ELBM results are from one deterministic run. The "Faithful Emulation" reports whether the rollout reproduces the expected structures. Field fidelity: relative $L^2$ error of velocity and vorticity $\omega$, and vorticity correlation. Spectrum diagnostics: RMSE of $\log_{10}E(k)$ over active shells, and high-$k$ energy and palinstrophy relative to the reference.}
  \label{tab:train_dp100k}
  \resizebox{\textwidth}{!}{
\setlength{\tabcolsep}{2.8pt}
    \begin{tabular}{@{}clccccccc@{}}
        \toprule
        \multirow{2}{*}{Grid} & \multicolumn{1}{c}{\multirow{2}{*}{Method}} & \multirow{2}{*}{\begin{tabular}{@{}c@{}}Faithful\\Emulation\end{tabular}} & \multicolumn{3}{c}{Field Fidelity} & \multicolumn{3}{c}{Spectrum Diagnostics} \\
        \cmidrule(lr){4-6}\cmidrule(l){7-9}
        & & & Vel. err $\downarrow$ & $\omega$ corr. $\uparrow$ & $\omega$ err $\downarrow$ & Spec. RMSE$^\dagger$ $\downarrow$ & High-$k$/ref. $\to1$ & $P$/ref. $\to1$ \\
        \midrule
        \multirow{2}{*}{$128^2$} & NeurDE & \textcolor{green!50!black}{4/4} & $\bf{0.0807 \pm 6.9{\times}10^{-5}}$ & $\bf{0.846 \pm 2.5{\times}10^{-4}}$ & $\bf{0.536 \pm 3.8{\times}10^{-4}}$ & $3.28 \pm 2.4{\times}10^{-3}$ & $0.0454 \pm 2.2{\times}10^{-4}$ & $0.104 \pm 1.4{\times}10^{-4}$ \\
         & ELBM & \textcolor{red}{0/1} & $0.281$ & $0.546$ & $0.981$ & $\bf{0.869}$ & $\bf{0.684}$ & $\bf{0.647}$ \\
        \midrule
        \multirow{2}{*}{$64^2$} & NeurDE & \textcolor{green!50!black}{4/4} & $\bf{0.154 \pm 5.7{\times}10^{-4}}$ & $\bf{0.771 \pm 3.9{\times}10^{-4}}$ & $\bf{0.641 \pm 3.7{\times}10^{-5}}$ & $3.60 \pm 9.6{\times}10^{-2}$ & $\bf{0.0283 \pm 0.0015}$ & $0.0553 \pm 0.0016$ \\
         & ELBM & \textcolor{red}{0/1} & $0.392$ & $0.542$ & $1.03$ & $\bf{0.438}$ & $1.98$ & $\bf{1.49}$ \\
        \midrule
        \multirow{2}{*}{$32^2$} & NeurDE & \textcolor{green!50!black}{4/4} & $\bf{0.304 \pm 0.045}$ & $\bf{0.684 \pm 0.032}$ & $\bf{0.730 \pm 0.032}$ & $3.72 \pm 0.70$ & $\bf{0.050 \pm 0.046}$ & $\bf{0.060 \pm 0.029}$ \\
         & ELBM & \textcolor{red}{0/1} & $0.561$ & $0.435$ & $1.16$ & $\bf{0.678}$ & $3.6$ & $2.8$ \\
        \midrule
        \multirow{2}{*}{$16^2$} & NeurDE & \textcolor{green!50!black}{4/4} & $\bf{0.561 \pm 0.034}$ & $\bf{0.600 \pm 0.017}$ & $\bf{0.800 \pm 0.013}$ & $4.2 \pm 1.3$ & $\bf{0.031 \pm 0.039}$ & $\bf{0.060 \pm 0.034}$ \\
         & ELBM & \textcolor{red}{0/1} & $0.729$ & $0.217$ & $1.29$ & $\bf{0.96}$ & $2.0$ & $2.1$ \\
        \bottomrule
    \end{tabular}
}
\end{table}
\begin{table}[!t]
  \centering\small
  \caption{Doubly periodic shear layer at $\mathrm{Re}=100{,}000$: forecast metrics from the testing data ($t_0=T/2$) at the final lead time; columns as in Table~\ref{tab:train_dp100k}.}
  \label{tab:test_dp100k}
  \resizebox{\textwidth}{!}{
\setlength{\tabcolsep}{2.8pt}
    \begin{tabular}{@{}clccccccc@{}}
        \toprule
        \multirow{2}{*}{Grid} & \multicolumn{1}{c}{\multirow{2}{*}{Method}} & \multirow{2}{*}{\begin{tabular}{@{}c@{}}Faithful\\Emulation\end{tabular}} & \multicolumn{3}{c}{Field Fidelity} & \multicolumn{3}{c}{Spectrum Diagnostics} \\
        \cmidrule(lr){4-6}\cmidrule(l){7-9}
        & & & Vel. err $\downarrow$ & $\omega$ corr. $\uparrow$ & $\omega$ err $\downarrow$ & Spec. RMSE$^\dagger$ $\downarrow$ & High-$k$/ref. $\to1$ & $P$/ref. $\to1$ \\
        \midrule
        \multirow{2}{*}{$128^2$} & NeurDE & \textcolor{green!50!black}{4/4} & $0.0519 \pm 3.3{\times}10^{-6}$ & $\bf{0.937 \pm 1.6{\times}10^{-5}}$ & $\bf{0.353 \pm 4.2{\times}10^{-5}}$ & $4.34 \pm 3.5{\times}10^{-3}$ & $0.00635 \pm 2.3{\times}10^{-5}$ & $0.0868 \pm 5.0{\times}10^{-5}$ \\
         & ELBM & \textcolor{green!50!black}{1/1} & $\bf{0.0448}$ & $0.920$ & $0.401$ & $\bf{0.257}$ & $\bf{1.08}$ & $\bf{1.02}$ \\
        \midrule
        \multirow{2}{*}{$64^2$} & NeurDE & \textcolor{green!50!black}{4/4} & $0.0956 \pm 5.9{\times}10^{-4}$ & $0.916 \pm 3.3{\times}10^{-4}$ & $0.421 \pm 8.5{\times}10^{-4}$ & $4.58 \pm 0.11$ & $0.00309 \pm 0.00047$ & $0.0780 \pm 4.0{\times}10^{-4}$ \\
         & ELBM & \textcolor{green!50!black}{1/1} & $\bf{0.0555}$ & $\bf{0.920}$ & $\bf{0.402}$ & $\bf{0.346}$ & $\bf{1.56}$ & $\bf{1.28}$ \\
        \midrule
        \multirow{2}{*}{$32^2$} & NeurDE & \textcolor{green!50!black}{4/4} & $0.175 \pm 0.019$ & $0.870 \pm 0.016$ & $0.519 \pm 0.031$ & $3.75 \pm 0.62$ & $0.0040 \pm 0.0044$ & $0.081 \pm 0.018$ \\
         & ELBM & \textcolor{green!50!black}{1/1} & $\bf{0.0785}$ & $\bf{0.931}$ & $\bf{0.368}$ & $\bf{0.430}$ & $\bf{1.2}$ & $\bf{1.1}$ \\
        \midrule
        \multirow{2}{*}{$16^2$} & NeurDE & \textcolor{green!50!black}{4/4} & $0.286 \pm 0.023$ & $0.826 \pm 0.019$ & $0.598 \pm 0.030$ & $2.85 \pm 0.45$ & $0.019 \pm 0.018$ & $0.067 \pm 0.016$ \\
         & ELBM & \textcolor{green!50!black}{1/1} & $\bf{0.224}$ & $\bf{0.884}$ & $\bf{0.469}$ & $\bf{0.331}$ & $\bf{0.82}$ & $\bf{0.69}$ \\
        \bottomrule
    \end{tabular}
}
\end{table}

Quantifying performance on chaotic systems like turbulence makes standard metrics like MSE imprecise and often requires comparisons of the energy spectrum and correlations.
Appendix~\ref{appx:turbulence_diagnostics} explains our metrics in detail: velocity error (relative error); $\omega$ corr. (vorticity correlation); $\omega$ error (relative error); spectrum RMSE; high-$k$/reference (ratio of high-$k$ energy fractions); enstrophy; palinstrophy. 
It is important to consider all of these diagnostics together. 
For example, any Fourier spectrum can be matched by a Gaussian random field lacking coherent structures, intermittency, or turbulent flux statistics \citet{pope2000turbulent,frisch1995turbulence}, so (phase-blind, second-order) metrics like $E(k)$ cannot be evaluated alone.
However, real-space metrics like MSE upweight large-valued structures, while ignoring the medium and small scales that determine the future flow.
Because low-amplitude fine structure contributes little to $L^2$ error, and high-$k$/ref measures an energy fraction rather than its distribution or phase, we also report palinstrophy (Eq.~\ref{eq:palinstrophy}), which weights the enstrophy by $k^2$.

Despite having erroneous dynamics, ELBM outperforms \emph{entropically-stabilized NeurDE} on various spectral methods, thus highlighting the challenges of quantifying emulation quality.
For all DPSL ELBM simulations on grids $\mathbb{R}^{128\times128}$ and coarser, ELBM simulations diverge and fail to recover the correct dynamics in the training set (see, e.g., Fig.~\ref {fig:dpsl100_forecast_64}, bottom row).
\emph{Entropically-stabilized NeurDE}'s strengths are at very coarse resolutions, with performance converging to ELBM as grids become finer.
Tables~\ref{tab:train_dp100k} and \ref{tab:test_dp100k} provide these measurements for DPSL-100K.
The "Faithful Emulation" column indicates whether the rollout faithfully reproduces the expected dynamics.

Stability and accuracy illustrate a tradeoff: targeted smoothing increases stability by enforcing dissipation, which causes smoothing that decreases accuracy.
Therefore, while ELBM outperforms \emph{entropically-stabilized NeurDE} on spectral measurements, this likely comes at the cost of stable simulations and ELBM's failure to faithfully simulate the training dynamics.
Specifically, by better capturing high-frequency components (Figs.~\ref{fig:app_spectra_dp100k} and \ref{fig:app_spectra_dp50k}), ELBM is incurring typical high-frequency noise that is known to cause simulation dynamics to diverge.
This is also observed in \emph{entropically-stabilized NeurDE} for the single divergent seed, as explained in Appendix~\ref{app:dpsl-50k_divergence}.

\FloatBarrier
\begin{figure}[t]
    \centering
    \begin{minipage}[t]{0.46\textwidth}
        \centering
        \includegraphics[width=\textwidth]{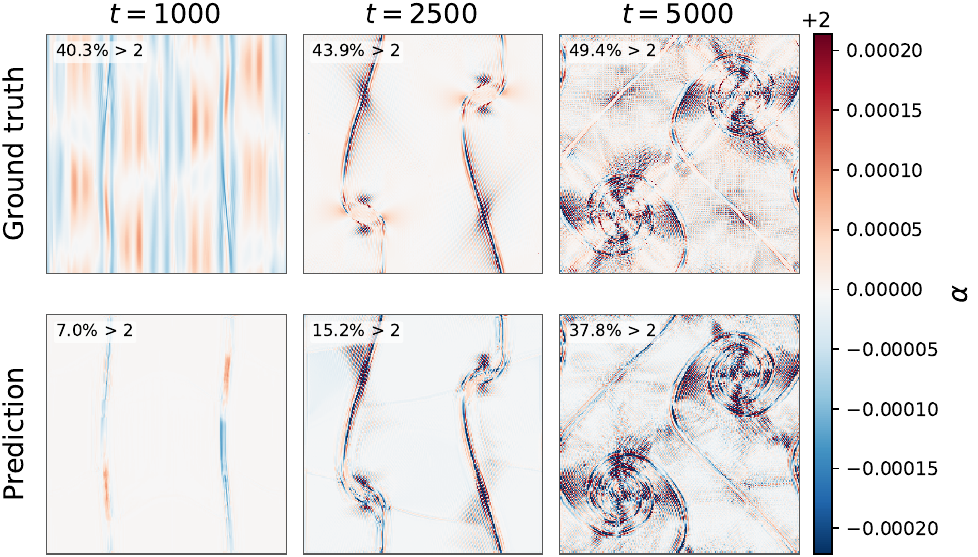}
        \caption{\textbf{$\boldsymbol{\alpha}$ prediction.} Comparison between ELBM generated and predicted $\alpha$ for DPSL-50K on a $\mathbb{R}^{256\times256}$ grid, demonstrating that  \emph{NeurDE} smooths where it needs to.}
        \label{fig:alpha_comparison}
    \end{minipage}
    \hfill
    \begin{minipage}[t]{0.52\textwidth}
        \centering
        \includegraphics[width=\textwidth]{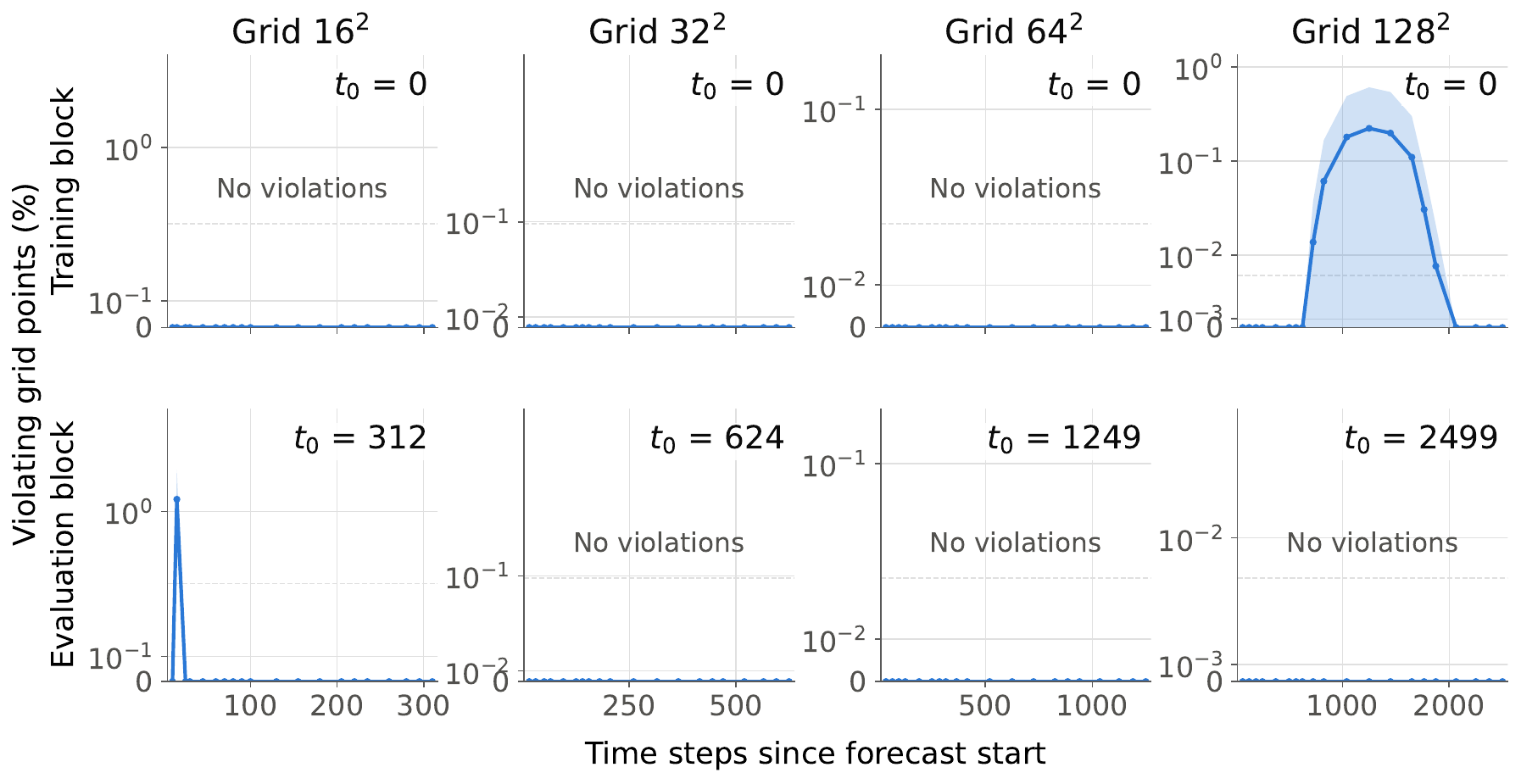}
        \caption{\textbf{Entropy Violations.} Percentage of \emph{entropically-stabilized NeurDE} rollout violating the second law of thermodynamics in DPSL-50K.}
        \label{fig:dpsl50_entropy}
    \end{minipage}
\end{figure}

\begingroup
\begin{figure*}[t]
    \centering
    \includegraphics[width=0.85\linewidth]{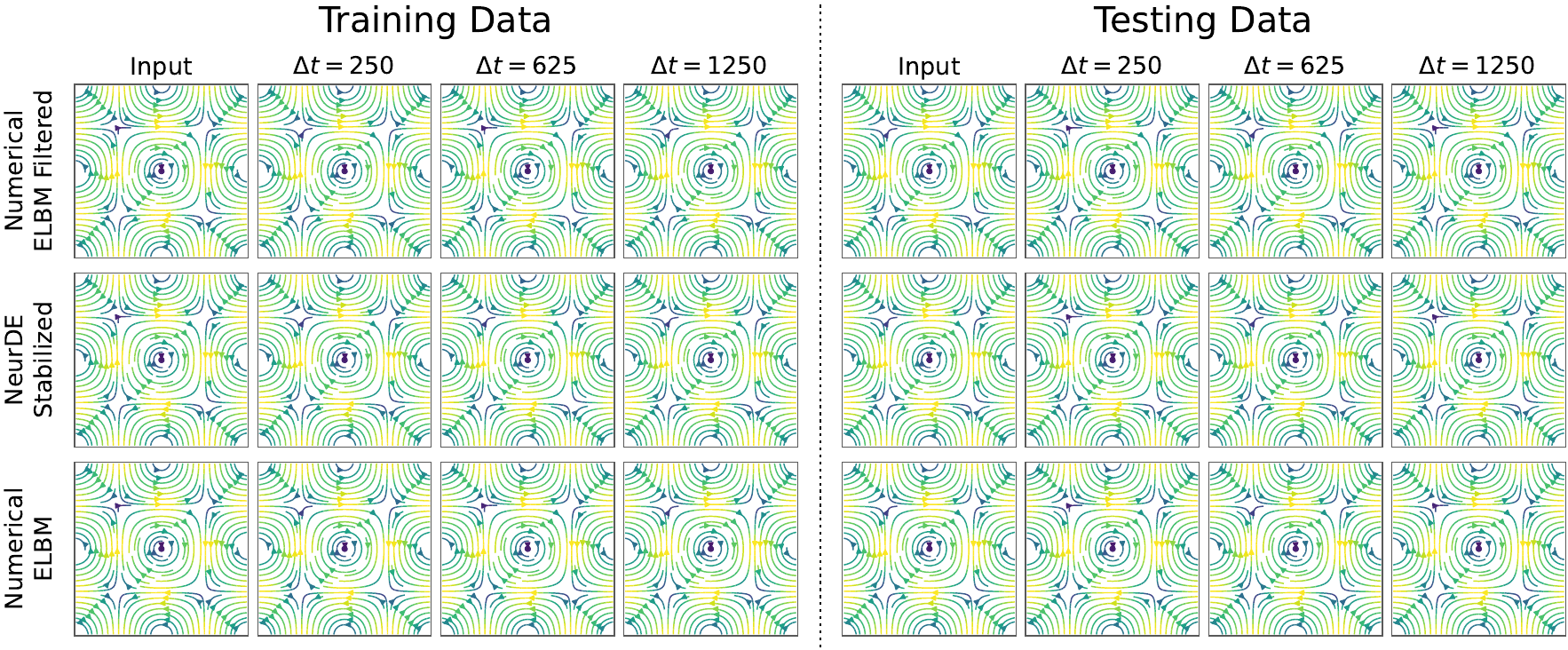}
    \caption{\textbf{Long-rollout comparison on the Taylor Green Vortex at Re=50,000.} Vorticity forecasts for the box-filtered high-resolution reference (top), \emph{entropy-stabilized NeurDE} (middle), and an underresolved
    ELBM baseline (bottom) on a $64\times64$ grid. This demonstrates that \emph{NeurDE} does not smooth where it is unnecessary.}
    \label{fig:tgv_forecast_64}
\end{figure*}

By enforcing the second law of thermodynamics and reinforcing dissipation, we provide stability through local smoothing that affects the global high-frequency components.
We investigate the DPSL-50K predicted $\alpha$ at $\mathbb{R}^{128\times128}$ to test whether \emph{entropically-stabilized NeurDE} is simply oversmoothing everywhere, and to quantify how many pixels are adding high-frequency information.
The predicted $\alpha$ (Fig.~\ref{fig:alpha_comparison}) tracks the fine ELBM-generated features, along with the large-scale structures.
Moreover, our predicted $\alpha$ affects fewer pixels than the ELBM.
Together, this indicates that our method is not simply oversmoothing, but instead it is locally smoothing where necessary and possibly more efficiently than ELBM.
The predicted $\alpha$ is less likely to increase high-frequency components compared to ELBM: the respective percentages of $\alpha > 2$ for the predicted and ELBM results are 7\% and 40.3\% at $\Delta t = 1000$; 15.2\% and 43.9\% at $\Delta t = 2500$; and 37.8\% and 49.4\% at $\Delta t = 5000$.
This likely contributes to \emph{entropically-stabilized NeurDE's} stability, since ELBM introduces sufficient spurious high-frequency noise for the dynamics to diverge at very coarse grids.

We further test \emph{entropically-stabilized NeurDE's} adherence to the second law of thermodynamics by investigating violations.
Figure~\ref{fig:dpsl50_entropy} shows the percentage of pixels that violate the second law of thermodynamics over the training and testing datasets for DPSL-50K.
We observe that \emph{entropically-stabilized NeurDE} rarely violates the second law and thus accomplishes its intended task.

For our final test of \emph{entropically-stabilized NeurDE's} ability to retain important information and not oversmooth, we train and evaluate on the 2D TGV experiment.
The TGV vortices are stable and have little spectral power in high frequencies; thus, any oversmoothing would cause them to diverge.
Figure~\ref{fig:tgv_forecast_64} shows that both \emph{entropically-stabilized NeurDE} and ELBM match the downsampled high-resolution simulations.
This indicates that \emph{entropically-stabilized NeurDE} can retain key spectral information.
For other grids, Appendix~\ref{app:results_tgv50k} provides further emulations, tables of real and spectral metrics, and plots of the spectral metrics for all grids.

\paragraph{Ablation.}

We ablate our entropic stability study with three methods on DPSL-50K: ELBM, \emph{entropically-stabilized NeurDE}, and \emph{NeurDE shell}.
\emph{NeurDE shell} improves energy spectrum alignment over \emph{entropically-stabilized NeurDE} through fine-tune training on $E(k)$ (Appendix~\ref{appx:spectral_finetuning}).
In Fig.~\ref{fig:app_forecast_dp50k}, we see that ELBM fails to be as stable as \emph{entropically-stabilized NeurDE}, indicating that the learned viscosity improves upon state-of-the-art numerical methods.
Compared to \emph{entropically-stabilized NeurDE}, \emph{NeurDE shell} significantly improves the high-frequency components on the $\mathbb{R}^{64\times64}$ grid (Appendix~\ref{appx:diagnostics_field_fidelity_64}).
Figures~\ref{fig:fields_rho_speed_vorticity_64} and \ref{fig:spectral_shell_power_spectra_64} show the real-space emulation results and spectral components, respectively, alongside more detailed comparisons.
Table~\ref{tab:fine-tune_ablation} provides the real-space and spectral metrics for all grids.
For the lower resolutions, \emph{NeurDE shell} performs similarly to \emph{entropically-stabilized NeurDE} (Appendix~\ref{appx:fine-tune_coarse}).
However, improving high-frequency alignment creates the possibility of inducing high-frequency noise that can destabilize the emulation.
For these reasons, and the small improvements outside of $\mathbb{R}^{64\times64}$, we use the more stable \emph{entropically-stabilized NeurDE}.

\newpage
\bibliographystyle{plainnat}
\bibliography{ref}

\begin{thebibliography}{95}
\providecommand{\natexlab}[1]{#1}
\providecommand{\url}[1]{\texttt{#1}}
\expandafter\ifx\csname urlstyle\endcsname\relax
  \providecommand{\doi}[1]{doi: #1}\else
  \providecommand{\doi}{doi: \begingroup \urlstyle{rm}\Url}\fi

\bibitem[Ansumali and Karlin(2000)]{ansumali2000stabilization}
Santosh Ansumali and Iliya~V. Karlin.
\newblock Stabilization of the lattice boltzmann method by the h theorem: a
  numerical test.
\newblock \emph{Physical Review E}, 62\penalty0 (6):\penalty0 7999--8003, 2000.
\newblock \doi{10.1103/PhysRevE.62.7999}.

\bibitem[Ansumali and Karlin(2002)]{ansumali2002single}
Santosh Ansumali and Iliya~V Karlin.
\newblock Single relaxation time model for entropic lattice boltzmann methods.
\newblock \emph{Physical Review E}, 65\penalty0 (5):\penalty0 056312, 2002.

\bibitem[Appell and Zabrejko(1990)]{appell1990nonlinear}
J{\"u}rgen Appell and Petr~P. Zabrejko.
\newblock \emph{Nonlinear Superposition Operators}.
\newblock Cambridge University Press, 1990.
\newblock \doi{10.1017/CBO9780511897450}.

\bibitem[Azizzadenesheli et~al.(2024)Azizzadenesheli, Kovachki, Li,
  Liu-Schiaffini, Kossaifi, and Anandkumar]{azizzadenesheli2024neural}
Kamyar Azizzadenesheli, Nikola Kovachki, Zongyi Li, Miguel Liu-Schiaffini, Jean
  Kossaifi, and Anima Anandkumar.
\newblock Neural operators for accelerating scientific simulations and design.
\newblock \emph{Nature Reviews Physics}, 6\penalty0 (5):\penalty0 320--328,
  2024.
\newblock \doi{10.1038/s42254-024-00712-5}.

\bibitem[Bai et~al.(2019)Bai, Kolter, and Koltun]{bai2019deep}
Shaojie Bai, J~Zico Kolter, and Vladlen Koltun.
\newblock Deep equilibrium models.
\newblock \emph{Advances in neural information processing systems}, 32, 2019.

\bibitem[Bai et~al.(2020)Bai, Koltun, and Kolter]{bai2020multiscale}
Shaojie Bai, Vladlen Koltun, and J~Zico Kolter.
\newblock Multiscale deep equilibrium models.
\newblock \emph{Advances in neural information processing systems},
  33:\penalty0 5238--5250, 2020.

\bibitem[Bansal et~al.(2022)Bansal, Schwarzschild, Borgnia, Emam, Huang,
  Goldblum, and Goldstein]{bansal2022end}
Arpit Bansal, Avi Schwarzschild, Eitan Borgnia, Zeyad Emam, Furong Huang, Micah
  Goldblum, and Tom Goldstein.
\newblock End-to-end algorithm synthesis with recurrent networks: Extrapolation
  without overthinking.
\newblock \emph{Advances in Neural Information Processing Systems},
  35:\penalty0 20232--20242, 2022.

\bibitem[Batchelor(1969)]{batchelor1969computation}
George~K Batchelor.
\newblock Computation of the energy spectrum in homogeneous two-dimensional
  turbulence.
\newblock \emph{The Physics of Fluids}, 12\penalty0 (12):\penalty0 II--233,
  1969.
\newblock \doi{10.1063/1.1692443}.

\bibitem[Batchelor(2000)]{batchelor2000fluid}
George~K. Batchelor.
\newblock \emph{An Introduction to Fluid Dynamics}.
\newblock Cambridge University Press, 2000.
\newblock \doi{10.1017/CBO9780511800955}.

\bibitem[Bedrunka et~al.(2021)Bedrunka, Wilde, Kliemank, Reith, Foysi, and
  Kr{\"a}mer]{bedrunka2021lettuce}
Mario~Christopher Bedrunka, Dominik Wilde, Martin Kliemank, Dirk Reith, Holger
  Foysi, and Andreas Kr{\"a}mer.
\newblock Lettuce: Pytorch-based lattice boltzmann framework.
\newblock In \emph{International Conference on High Performance Computing},
  pages 40--55. Springer, 2021.

\bibitem[Bellotti(2023)]{bellotti2023truncation}
Thomas Bellotti.
\newblock Truncation errors and modified equations for the lattice boltzmann
  method via the corresponding finite difference schemes.
\newblock \emph{ESAIM: Mathematical Modelling and Numerical Analysis},
  57\penalty0 (3):\penalty0 1225--1255, 2023.

\bibitem[Bellotti et~al.(2025)Bellotti, Helluy, and
  Navoret]{bellotti2025fourth}
Thomas Bellotti, Philippe Helluy, and Laurent Navoret.
\newblock Fourth-order entropy-stable lattice boltzmann schemes for hyperbolic
  systems.
\newblock \emph{SIAM Journal on Scientific Computing}, 47\penalty0
  (1):\penalty0 A586--A611, 2025.

\bibitem[Bengio et~al.(2009)Bengio, Louradour, Collobert, and
  Weston]{bengio2009curriculum}
Yoshua Bengio, J{\'e}r{\^o}me Louradour, Ronan Collobert, and Jason Weston.
\newblock Curriculum learning.
\newblock In \emph{Proceedings of the 26th annual international conference on
  machine learning}, pages 41--48, 2009.

\bibitem[Benitez et~al.(2025)Benitez, Guo, Hegazy, Dokmani{\'c}, Mahoney, and
  de~Hoop]{benitez2025neural}
J~Benitez, Junyi Guo, Kareem Hegazy, Ivan Dokmani{\'c}, Michael~W Mahoney, and
  Maarten~V de~Hoop.
\newblock Neural equilibria for long-term prediction of nonlinear conservation
  laws.
\newblock \emph{arXiv preprint arXiv:2501.06933}, 2025.
\newblock \doi{10.48550/arXiv.2501.06933}.

\bibitem[Benitez et~al.(2024)Benitez, Furuya, Faucher, Kratsios, Tricoche, and
  de~Hoop]{benitez2024out}
Jose Antonio~Lara Benitez, Takashi Furuya, Florian Faucher, Anastasis Kratsios,
  Xavier Tricoche, and Maarten~V de~Hoop.
\newblock Out-of-distributional risk bounds for neural operators with
  applications to the helmholtz equation.
\newblock \emph{Journal of Computational Physics}, page 113168, 2024.

\bibitem[Benzi et~al.(1993)Benzi, Ciliberto, Tripiccione, Baudet, Massaioli,
  and Succi]{benzi1993extended}
Roberto Benzi, Sergio Ciliberto, Raffaele Tripiccione, Christophe Baudet,
  F.~Massaioli, and Sauro Succi.
\newblock Extended self-similarity in turbulent flows.
\newblock \emph{Physical Review E}, 48\penalty0 (1):\penalty0 R29--R32, 1993.
\newblock \doi{10.1103/PhysRevE.48.R29}.

\bibitem[Beucler et~al.(2021)Beucler, Pritchard, Rasp, Ott, Baldi, and
  Gentine]{beucler2021enforcing}
Tom Beucler, Michael Pritchard, Stephan Rasp, Jordan Ott, Pierre Baldi, and
  Pierre Gentine.
\newblock Enforcing analytic constraints in neural networks emulating physical
  systems.
\newblock \emph{Physical review letters}, 126\penalty0 (9):\penalty0 098302,
  2021.
\newblock \doi{10.1103/PhysRevLett.126.098302}.

\bibitem[Bhatnagar et~al.(1954)Bhatnagar, Gross, and Krook]{bhatnagar1954model}
Prabhu~Lal Bhatnagar, Eugene~P Gross, and Max Krook.
\newblock A model for collision processes in gases. i. small amplitude
  processes in charged and neutral one-component systems.
\newblock \emph{Physical review}, 94\penalty0 (3):\penalty0 511, 1954.
\newblock \doi{10.1103/PhysRev.94.511}.

\bibitem[Boffetta and Ecke(2012)]{boffetta2012two}
Guido Boffetta and Robert~E Ecke.
\newblock Two-dimensional turbulence.
\newblock \emph{Annual review of fluid mechanics}, 44\penalty0 (1):\penalty0
  427--451, 2012.
\newblock \doi{10.1146/annurev-fluid-120710-101240}.

\bibitem[Boghosian et~al.(2001)Boghosian, Yepez, Coveney, and
  Wager]{boghosian2001entropic}
Bruce~M Boghosian, Jeffrey Yepez, Peter~V Coveney, and Alexander Wager.
\newblock Entropic lattice boltzmann methods.
\newblock \emph{Proceedings of the Royal Society of London. Series A:
  Mathematical, Physical and Engineering Sciences}, 457\penalty0
  (2007):\penalty0 717--766, 2001.
\newblock \doi{10.1098/rspa.2000.0689}.

\bibitem[Bohr et~al.(1998)Bohr, Jensen, Paladin, and
  Vulpiani]{bohr1998dynamical}
Tomas Bohr, Mogens~H Jensen, Giovanni Paladin, and Angelo Vulpiani.
\newblock \emph{Dynamical systems approach to turbulence}, volume~8 of
  \emph{Cambridge Nonlinear Science Series}.
\newblock Cambridge University Press, 1998.
\newblock \doi{10.1017/CBO9780511599972}.

\bibitem[Bonsall and Vedak(1962)]{bonsall1962lectures}
Frank~F Bonsall and KB~Vedak.
\newblock \emph{Lectures on some fixed point theorems of functional analysis},
  volume~26.
\newblock Tata Institute of Fundamental Research Bombay, 1962.

\bibitem[B{\"o}sch et~al.(2015)B{\"o}sch, Chikatamarla, and
  Karlin]{bosch2015entropic}
Fabian B{\"o}sch, Shyam~S Chikatamarla, and Ilya~V Karlin.
\newblock Entropic multirelaxation lattice boltzmann models for turbulent
  flows.
\newblock \emph{Physical Review E}, 92\penalty0 (4):\penalty0 043309, 2015.
\newblock \doi{10.1103/PhysRevE.92.043309}.

\bibitem[Bouchut(1999)]{bouchut1999construction}
Fran{\c{c}}ois Bouchut.
\newblock Construction of bgk models with a family of kinetic entropies for a
  given system of conservation laws.
\newblock \emph{Journal of statistical physics}, 95:\penalty0 113--170, 1999.

\bibitem[Boyd and Schwartzentruber(2017)]{boyd2017nonequilibrium}
Iain~D Boyd and Thomas~E Schwartzentruber.
\newblock \emph{Nonequilibrium gas dynamics and molecular simulation},
  volume~42.
\newblock Cambridge University Press, 2017.

\bibitem[Brown(1995)]{brown1995performance}
David~L Brown.
\newblock Performance of under-resolved two-dimensional incompressible flow
  simulations.
\newblock \emph{Journal of Computational Physics}, 122\penalty0 (1):\penalty0
  165--183, 1995.

\bibitem[Brunton et~al.(2020)Brunton, Noack, and
  Koumoutsakos]{brunton2020machine}
Steven~L Brunton, Bernd~R Noack, and Petros Koumoutsakos.
\newblock Machine learning for fluid mechanics.
\newblock \emph{Annual review of fluid mechanics}, 52\penalty0 (1):\penalty0
  477--508, 2020.

\bibitem[Buzzicotti and Tauzin(2021)]{buzzicotti2021inertial}
Michele Buzzicotti and Guillaume Tauzin.
\newblock Inertial range statistics of the entropic lattice boltzmann method in
  three-dimensional turbulence.
\newblock \emph{Physical Review E}, 104\penalty0 (1):\penalty0 015302, 2021.
\newblock \doi{10.1103/PhysRevE.104.015302}.

\bibitem[Cai et~al.(2021)Cai, Mao, Wang, Yin, and Karniadakis]{cai2021physics}
Shengze Cai, Zhiping Mao, Zhicheng Wang, Minglang Yin, and George~Em
  Karniadakis.
\newblock Physics-informed neural networks (pinns) for fluid mechanics: A
  review.
\newblock \emph{Acta Mechanica Sinica}, 37\penalty0 (12):\penalty0 1727--1738,
  2021.
\newblock \doi{10.1007/s10409-021-01148-1}.

\bibitem[Callen(1985)]{callen1985thermodynamics}
Herbert~B. Callen.
\newblock \emph{Thermodynamics and an Introduction to Thermostatistics}.
\newblock John Wiley \& Sons, New York, 2 edition, 1985.
\newblock ISBN 978-0-471-86256-7.

\bibitem[Canuto et~al.(2006)Canuto, Hussaini, Quarteroni, and
  Zang]{canuto2006spectral}
Claudio Canuto, M.~Youssuff Hussaini, Alfio Quarteroni, and Thomas~A. Zang.
\newblock \emph{Spectral Methods: Fundamentals in Single Domains}.
\newblock Scientific Computation. Springer, Berlin, Heidelberg, 2006.
\newblock \doi{10.1007/978-3-540-30726-6}.

\bibitem[{\'C}iri{\'c}(1974)]{ciric1974generalization}
Lj~B {\'C}iri{\'c}.
\newblock A generalization of banach’s contraction principle.
\newblock \emph{Proceedings of the American Mathematical society}, 45\penalty0
  (2):\penalty0 267--273, 1974.

\bibitem[Cobza{\c{s}}(2006)]{cobzas2006fixed}
S.~Cobza{\c{s}}.
\newblock Fixed point theorems in locally convex spaces---the schauder mapping
  method.
\newblock \emph{Fixed Point Theory and Applications}, 2006:\penalty0 57950,
  2006.
\newblock \doi{10.1155/FPTA/2006/57950}.

\bibitem[Davidson(2015)]{davidson2015turbulence}
Peter~A. Davidson.
\newblock \emph{Turbulence: An Introduction for Scientists and Engineers}.
\newblock Oxford University Press, 2 edition, 2015.
\newblock \doi{10.1093/acprof:oso/9780198722588.001.0001}.

\bibitem[De~Rosis and Coreixas(2020)]{de2020multiphysics}
Alessandro De~Rosis and Christophe Coreixas.
\newblock Multiphysics flow simulations using d3q19 lattice boltzmann methods
  based on central moments.
\newblock \emph{Physics of Fluids}, 32\penalty0 (11), 2020.

\bibitem[Dorschner et~al.(2016)Dorschner, B{\"o}sch, Chikatamarla, Boulouchos,
  and Karlin]{dorschner2016entropic_mrt}
Benedikt Dorschner, Fabian B{\"o}sch, Shyam~S. Chikatamarla, Konstantinos
  Boulouchos, and Ilya~V. Karlin.
\newblock Entropic multi-relaxation time lattice boltzmann model for complex
  flows.
\newblock \emph{Journal of Fluid Mechanics}, 801:\penalty0 623--651, 2016.
\newblock \doi{10.1017/jfm.2016.448}.

\bibitem[Evans(2010)]{EvansPDE}
Lawrence~C. Evans.
\newblock \emph{Partial Differential Equations}, volume~19 of \emph{Graduate
  Studies in Mathematics}.
\newblock American Mathematical Society, 2 edition, 2010.
\newblock ISBN 978-0821849743.

\bibitem[Federer(1969)]{Federer1969}
Herbert Federer.
\newblock \emph{Geometric Measure Theory}.
\newblock Die Grundlehren der mathematischen Wissenschaften. Springer, 1969.

\bibitem[Ferziger et~al.(2019)Ferziger, Peri{\'c}, and
  Street]{ferziger2019computational}
Joel~H Ferziger, Milovan Peri{\'c}, and Robert~L Street.
\newblock \emph{Computational methods for fluid dynamics}.
\newblock Springer Cham, 4 edition, 2019.
\newblock \doi{10.1007/978-3-319-99693-6}.

\bibitem[Fischer et~al.(2022)Fischer, Kerkemeier, Min, Lan, Phillips,
  Rathnayake, Merzari, Tomboulides, Karakus, Chalmers,
  et~al.]{fischer2022nekrs}
Paul Fischer, Stefan Kerkemeier, Misun Min, Yu-Hsiang Lan, Malachi Phillips,
  Thilina Rathnayake, Elia Merzari, Ananias Tomboulides, Ali Karakus, Noel
  Chalmers, et~al.
\newblock Nekrs, a gpu-accelerated spectral element navier--stokes solver.
\newblock \emph{Parallel Computing}, 114:\penalty0 102982, 2022.

\bibitem[Fischer et~al.(2025)Fischer, Kaltenbach, Litvinov, Succi, and
  Koumoutsakos]{fischer2025optimal}
Paul Fischer, Sebastian Kaltenbach, Sergey Litvinov, Sauro Succi, and Petros
  Koumoutsakos.
\newblock Optimal lattice boltzmann closures through multi-agent reinforcement
  learning.
\newblock \emph{arXiv preprint arXiv:2504.14422}, 2025.

\bibitem[Frapolli et~al.(2015)Frapolli, Chikatamarla, and
  Karlin]{frapolli2015entropic}
Nicol{\`o} Frapolli, Shyam~S Chikatamarla, and Iliya~V Karlin.
\newblock Entropic lattice boltzmann model for compressible flows.
\newblock \emph{Physical Review E}, 92\penalty0 (6):\penalty0 061301, 2015.

\bibitem[Frisch(1995)]{frisch1995turbulence}
Uriel Frisch.
\newblock \emph{Turbulence: The Legacy of A. N. Kolmogorov}.
\newblock Cambridge University Press, 1995.
\newblock \doi{10.1017/CBO9781139170666}.

\bibitem[Fung et~al.(2022)Fung, Heaton, Li, McKenzie, Osher, and
  Yin]{fung2022jfb}
Samy~Wu Fung, Howard Heaton, Qiuwei Li, Daniel McKenzie, Stanley Osher, and
  Wotao Yin.
\newblock Jfb: Jacobian-free backpropagation for implicit networks.
\newblock In \emph{Proceedings of the AAAI Conference on Artificial
  Intelligence}, volume~36, pages 6648--6656, 2022.

\bibitem[Gao et~al.(2025)Gao, Xu, Deng, and Liu]{gao2025discretization}
Wenhan Gao, Ruichen Xu, Yuefan Deng, and Yi~Liu.
\newblock Discretization-invariance? on the discretization mismatch errors in
  neural operators.
\newblock In \emph{International Conference on Learning Representations}, 2025.
\newblock URL \url{https://openreview.net/forum?id=J9FgrqOOni}.

\bibitem[Grad(1949)]{grad1949kinetic}
Harold Grad.
\newblock On the kinetic theory of rarefied gases.
\newblock \emph{Communications on pure and applied mathematics}, 2\penalty0
  (4):\penalty0 331--407, 1949.
\newblock \doi{10.1002/cpa.3160020403}.

\bibitem[Hegazy et~al.(2026)Hegazy, Benitez, and Kiefer]{hegazyCode}
Kareem Hegazy, Jose Antonio~Lara Benitez, and Alexander Kiefer.
\newblock Kinet.
\newblock \url{https://github.com/khegazy/kinet}, 2026.

\bibitem[Herde et~al.(2024)Herde, Raonic, Rohner, K{\"a}ppeli, Molinaro,
  de~B{\'e}zenac, and Mishra]{herde2024poseidon}
Maximilian Herde, Bogdan Raonic, Tobias Rohner, Roger K{\"a}ppeli, Roberto
  Molinaro, Emmanuel de~B{\'e}zenac, and Siddhartha Mishra.
\newblock Poseidon: Efficient foundation models for pdes.
\newblock \emph{Advances in Neural Information Processing Systems},
  37:\penalty0 72525--72624, 2024.

\bibitem[Hosseini et~al.(2023)Hosseini, Atif, Ansumali, and
  Karlin]{hosseini2023entropic}
Seyed~Ali Hosseini, Mohammad Atif, Santosh Ansumali, and Iliya~V Karlin.
\newblock Entropic lattice boltzmann methods: A review.
\newblock \emph{Computers \& Fluids}, 259:\penalty0 105884, 2023.
\newblock \doi{10.1016/j.compfluid.2023.105884}.

\bibitem[Ishihara et~al.(2009)Ishihara, Gotoh, and Kaneda]{ishihara2009study}
Takashi Ishihara, Toshiyuki Gotoh, and Yukio Kaneda.
\newblock Study of high--reynolds number isotropic turbulence by direct
  numerical simulation.
\newblock \emph{Annual review of fluid mechanics}, 41\penalty0 (1):\penalty0
  165--180, 2009.
\newblock \doi{10.1146/annurev.fluid.010908.165203}.

\bibitem[Jonnalagadda et~al.(2021)Jonnalagadda, Sharma, and
  Agrawal]{jonnalagadda2021single}
Anirudh Jonnalagadda, Atul Sharma, and Amit Agrawal.
\newblock Single relaxation time entropic lattice boltzmann methods: A
  developer’s perspective for stable and accurate simulations.
\newblock \emph{Computers \& Fluids}, 215:\penalty0 104792, 2021.
\newblock \doi{10.1016/j.compfluid.2020.104792}.

\bibitem[Karlin and Succi(1998)]{karlin1998equilibria}
Iliya~V Karlin and S~Succi.
\newblock Equilibria for discrete kinetic equations.
\newblock \emph{Physical Review E}, 58\penalty0 (4):\penalty0 R4053, 1998.

\bibitem[Karlin et~al.(2014)Karlin, B{\"o}sch, and
  Chikatamarla]{karlin2014entropic}
Iliya~V. Karlin, Fabian B{\"o}sch, and Shyam~S. Chikatamarla.
\newblock Gibbs' principle for the lattice-kinetic theory of fluid dynamics.
\newblock \emph{Physical Review E}, 90\penalty0 (3):\penalty0 031302, 2014.
\newblock \doi{10.1103/PhysRevE.90.031302}.

\bibitem[Karlin et~al.(1999)Karlin, Ferrante, and
  {\"O}ttinger]{karlin1999perfect}
Ilya~V Karlin, Antonio Ferrante, and Hans~Christian {\"O}ttinger.
\newblock Perfect entropy functions of the lattice boltzmann method.
\newblock \emph{Europhysics Letters}, 47\penalty0 (2):\penalty0 182--188, 1999.
\newblock \doi{10.1209/epl/i1999-00370-1}.

\bibitem[Karlin et~al.(2003)Karlin, Ansumali, De~Angelis, {\"O}ttinger, and
  Succi]{karlin2003entropic}
IV~Karlin, S~Ansumali, E~De~Angelis, HC~{\"O}ttinger, and S~Succi.
\newblock Entropic lattice boltzmann method for large scale turbulence
  simulation.
\newblock \emph{arXiv preprint cond-mat/0306003}, 2003.

\bibitem[Kochkov et~al.(2021)Kochkov, Smith, Alieva, Wang, Brenner, and
  Hoyer]{kochkov2021machine}
Dmitrii Kochkov, Jamie~A Smith, Ayya Alieva, Qing Wang, Michael~P Brenner, and
  Stephan Hoyer.
\newblock Machine learning--accelerated computational fluid dynamics.
\newblock \emph{Proceedings of the National Academy of Sciences}, 118\penalty0
  (21):\penalty0 e2101784118, 2021.

\bibitem[Kolmogorov(1991)]{kolmogorov1991local}
Andrei~Nikolaevich Kolmogorov.
\newblock The local structure of turbulence in incompressible viscous fluid for
  very large reynolds numbers.
\newblock \emph{Proceedings of the Royal Society of London. Series A:
  Mathematical and Physical Sciences}, 434\penalty0 (1890):\penalty0 9--13,
  1991.

\bibitem[Kossaifi et~al.(2025)Kossaifi, Kovachki, Li, Pitt, Liu-Schiaffini,
  Duruisseaux, George, Bonev, Azizzadenesheli, Berner, and
  Anandkumar]{kossaifi2025librarylearningneuraloperators}
Jean Kossaifi, Nikola Kovachki, Zongyi Li, David Pitt, Miguel Liu-Schiaffini,
  Valentin Duruisseaux, Robert~Joseph George, Boris Bonev, Kamyar
  Azizzadenesheli, Julius Berner, and Anima Anandkumar.
\newblock A library for learning neural operators.
\newblock \emph{arXiv preprint arXiv:2412.10354}, 2025.

\bibitem[Kovachki et~al.(2024)Kovachki, Lanthaler, and
  Stuart]{kovachki2024operator}
Nikola~B Kovachki, Samuel Lanthaler, and Andrew~M Stuart.
\newblock Operator learning: Algorithms and analysis.
\newblock \emph{arXiv preprint arXiv:2402.15715}, 2024.

\bibitem[Kraichnan(1967)]{kraichnan1967inertial}
Robert~H. Kraichnan.
\newblock Inertial ranges in two-dimensional turbulence.
\newblock \emph{The Physics of Fluids}, 10\penalty0 (7):\penalty0 1417--1423,
  1967.
\newblock \doi{10.1063/1.1762301}.

\bibitem[Kratsios et~al.(2024)Kratsios, Furuya, Lara~Benitez, Lassas, and
  de~Hoop]{kratsios2024mixtures}
Anastasis Kratsios, Takashi Furuya, Jose~Antonio Lara~Benitez, Matti Lassas,
  and Maarten de~Hoop.
\newblock Mixtures of neural operators reduce active complexity in operator
  learning.
\newblock \emph{arXiv preprint arXiv:2404.09101}, 2024.
\newblock \doi{10.48550/arXiv.2404.09101}.
\newblock URL \url{https://arxiv.org/abs/2404.09101}.

\bibitem[Krishnapriyan et~al.(2021)Krishnapriyan, Gholami, Zhe, Kirby, and
  Mahoney]{krishnapriyan2021characterizing}
Aditi Krishnapriyan, Amir Gholami, Shandian Zhe, Robert Kirby, and Michael~W
  Mahoney.
\newblock Characterizing possible failure modes in physics-informed neural
  networks.
\newblock In \emph{Advances in Neural Information Processing Systems},
  volume~34, pages 26548--26560, 2021.

\bibitem[Kr{\"u}ger et~al.(2017)Kr{\"u}ger, Kusumaatmaja, Kuzmin, Shardt,
  Silva, and Viggen]{kruger2017lattice}
Timm Kr{\"u}ger, Halim Kusumaatmaja, Alexandr Kuzmin, Orest Shardt, Goncalo
  Silva, and Erlend~Magnus Viggen.
\newblock The lattice boltzmann method.
\newblock \emph{Springer International Publishing}, 10\penalty0
  (978-3):\penalty0 4--15, 2017.

\bibitem[Latt et~al.(2020)Latt, Coreixas, Beny, and
  Parmigiani]{latt2020efficient}
Jonas Latt, Christophe Coreixas, Jo{\"e}l Beny, and Andrea Parmigiani.
\newblock Efficient supersonic flow simulations using lattice boltzmann methods
  based on numerical equilibria.
\newblock \emph{Philosophical Transactions of the Royal Society A},
  378\penalty0 (2175):\penalty0 20190559, 2020.

\bibitem[Levermore(1996)]{levermore1996moment}
C~David Levermore.
\newblock Moment closure hierarchies for kinetic theories.
\newblock \emph{Journal of statistical Physics}, 83:\penalty0 1021--1065, 1996.

\bibitem[Li et~al.(2021)Li, Kovachki, Azizzadenesheli, liu, Bhattacharya,
  Stuart, and Anandkumar]{li2021fourier}
Zongyi Li, Nikola~Borislavov Kovachki, Kamyar Azizzadenesheli, Burigede liu,
  Kaushik Bhattacharya, Andrew Stuart, and Anima Anandkumar.
\newblock Fourier neural operator for parametric partial differential
  equations.
\newblock In \emph{International Conference on Learning Representations}, 2021.
\newblock URL \url{https://openreview.net/forum?id=c8P9NQVtmnO}.

\bibitem[Loshchilov and Hutter(2019)]{loshchilov2019adamw}
Ilya Loshchilov and Frank Hutter.
\newblock Decoupled weight decay regularization.
\newblock In \emph{International Conference on Learning Representations
  (ICLR)}, 2019.
\newblock URL \url{https://arxiv.org/abs/1711.05101}.

\bibitem[McCabe et~al.(2025)McCabe, Mukhopadhyay, Marwah, Blancard, Rozet,
  Diaconu, Meyer, Wong, Sotoudeh, Bietti, et~al.]{mccabe2025walrus}
Michael McCabe, Payel Mukhopadhyay, Tanya Marwah, Bruno Regaldo-Saint Blancard,
  Francois Rozet, Cristiana Diaconu, Lucas Meyer, Kaze~WK Wong, Hadi Sotoudeh,
  Alberto Bietti, et~al.
\newblock Walrus: A cross-domain foundation model for continuum dynamics.
\newblock \emph{arXiv preprint arXiv:2511.15684}, 2025.

\bibitem[McGreivy and Hakim(2024)]{mcgreivy2024weak}
Nick McGreivy and Ammar Hakim.
\newblock Weak baselines and reporting biases lead to overoptimism in machine
  learning for fluid-related partial differential equations.
\newblock \emph{arXiv preprint arXiv:2407.07218}, 2024.

\bibitem[Meneveau and Katz(2000)]{meneveau2000scale}
Charles Meneveau and Joseph Katz.
\newblock Scale-invariance and turbulence models for large-eddy simulation.
\newblock \emph{Annual Review of Fluid Mechanics}, 32\penalty0 (1):\penalty0
  1--32, 2000.
\newblock \doi{10.1146/annurev.fluid.32.1.1}.

\bibitem[Minion and Brown(1997)]{minion1997performance}
Michael~L Minion and David~L Brown.
\newblock Performance of under-resolved two-dimensional incompressible flow
  simulations, ii.
\newblock \emph{Journal of Computational Physics}, 138\penalty0 (2):\penalty0
  734--765, 1997.

\bibitem[Pascanu et~al.(2013)Pascanu, Mikolov, and
  Bengio]{pascanu2013difficulty}
Razvan Pascanu, Tomas Mikolov, and Yoshua Bengio.
\newblock On the difficulty of training recurrent neural networks.
\newblock In \emph{International conference on machine learning}, pages
  1310--1318. Pmlr, 2013.

\bibitem[Patel et~al.(2022)Patel, Manickam, Trask, Wood, Lee, Tomas, and
  Cyr]{patel2022thermodynamically}
Ravi~G Patel, Indu Manickam, Nathaniel~A Trask, Mitchell~A Wood, Myoungkyu Lee,
  Ignacio Tomas, and Eric~C Cyr.
\newblock Thermodynamically consistent physics-informed neural networks for
  hyperbolic systems.
\newblock \emph{Journal of Computational Physics}, 449:\penalty0 110754, 2022.
\newblock \doi{10.1016/j.jcp.2021.110754}.

\bibitem[Patterson and Orszag(1971)]{patterson1971spectral}
G.~S. Patterson and Steven~A. Orszag.
\newblock Spectral calculations of isotropic turbulence: Efficient removal of
  aliasing interactions.
\newblock \emph{The Physics of Fluids}, 14\penalty0 (11):\penalty0 2538--2541,
  1971.
\newblock \doi{10.1063/1.1693365}.

\bibitem[Pedersen et~al.(2025)Pedersen, Zanna, and
  Bruna]{pedersen2025thermalizer}
Christian Pedersen, Laure Zanna, and Joan Bruna.
\newblock Thermalizer: Stable autoregressive neural emulation of spatiotemporal
  chaos.
\newblock \emph{Proceedings of the 42nd International Conference on Machine
  Learning}, 267:\penalty0 48635--48661, 2025.

\bibitem[Pope(2000)]{pope2000turbulent}
Stephen~B. Pope.
\newblock \emph{Turbulent Flows}.
\newblock Cambridge University Press, 2000.
\newblock \doi{10.1017/CBO9780511840531}.

\bibitem[Pope(2001)]{pope2001turbulent}
Stephen~B Pope.
\newblock Turbulent flows.
\newblock \emph{Measurement Science and Technology}, 12\penalty0 (11):\penalty0
  2020--2021, 2001.

\bibitem[Raissi et~al.(2019)Raissi, Perdikaris, and
  Karniadakis]{raissi2019physics}
Maziar Raissi, Paris Perdikaris, and George~E Karniadakis.
\newblock Physics-informed neural networks: A deep learning framework for
  solving forward and inverse problems involving nonlinear partial differential
  equations.
\newblock \emph{Journal of Computational physics}, 378:\penalty0 686--707,
  2019.
\newblock \doi{10.1016/j.jcp.2018.10.045}.

\bibitem[Sagaut(2006)]{sagaut2006large}
Pierre Sagaut.
\newblock \emph{Large eddy simulation for incompressible flows: an
  introduction}.
\newblock Springer, 3 edition, 2006.
\newblock \doi{10.1007/b137536}.

\bibitem[Sakarvadia et~al.(2026)Sakarvadia, Hegazy, Totounferoush, Chard, Yang,
  Foster, and Mahoney]{sakarvadia2025false}
Mansi Sakarvadia, Kareem Hegazy, Amin Totounferoush, Kyle Chard, Yaoqing Yang,
  Ian Foster, and Michael~W Mahoney.
\newblock The false promise of zero-shot super-resolution in machine-learned
  operators.
\newblock In \emph{International Conference on Learning Representations}, 2026.
\newblock \doi{10.48550/arXiv.2510.06646}.
\newblock URL \url{https://openreview.net/forum?id=hkF7ZM7fEp}.

\bibitem[Schornbaum and Rude(2016)]{schornbaum2016massively}
Florian Schornbaum and Ulrich Rude.
\newblock Massively parallel algorithms for the lattice boltzmann method on
  nonuniform grids.
\newblock \emph{SIAM Journal on Scientific Computing}, 38\penalty0
  (2):\penalty0 C96--C126, 2016.

\bibitem[Shan and He(1998)]{shan1998discretization}
Xiaowen Shan and Xiaoyi He.
\newblock Discretization of the velocity space in the solution of the boltzmann
  equation.
\newblock \emph{Physical Review Letters}, 80\penalty0 (1):\penalty0 65, 1998.
\newblock \doi{10.1103/PhysRevLett.80.65}.

\bibitem[Shan et~al.(2006)Shan, Yuan, and Chen]{shan2006kinetic}
Xiaowen Shan, Xue-Feng Yuan, and Hudong Chen.
\newblock Kinetic theory representation of hydrodynamics: a way beyond the
  navier--stokes equation.
\newblock \emph{Journal of Fluid Mechanics}, 550:\penalty0 413--441, 2006.

\bibitem[Smagorinsky(1963)]{smagorinsky1963general}
Joseph Smagorinsky.
\newblock General circulation experiments with the primitive equations: I. the
  basic experiment.
\newblock \emph{Monthly weather review}, 91\penalty0 (3):\penalty0 99--164,
  1963.
\newblock \doi{10.1175/1520-0493(1963)091<0099:GCEWTP>2.3.CO;2}.

\bibitem[Smith and Topin(2019)]{smith2019superconvergence}
Leslie~N. Smith and Nicholay Topin.
\newblock Super-convergence: Very fast training of neural networks using large
  learning rates.
\newblock In \emph{Artificial Intelligence and Machine Learning for
  Multi-Domain Operations Applications}, volume 11006 of \emph{Proc. SPIE},
  2019.
\newblock URL \url{https://arxiv.org/abs/1708.07120}.

\bibitem[Succi(2001)]{succi2001lattice}
Sauro Succi.
\newblock \emph{The lattice Boltzmann equation: for fluid dynamics and beyond}.
\newblock Oxford university press, 2001.

\bibitem[Succi and Succi(2018)]{succi2018lattice}
Sauro Succi and S~Succi.
\newblock \emph{The lattice Boltzmann equation: for complex states of flowing
  matter}.
\newblock Oxford university press, 2018.

\bibitem[Tadmor(1987)]{tadmor1987numerical}
Eitan Tadmor.
\newblock The numerical viscosity of entropy stable schemes for systems of
  conservation laws. i.
\newblock \emph{Mathematics of Computation}, 49\penalty0 (179):\penalty0
  91--103, 1987.
\newblock \doi{10.1090/S0025-5718-1987-0890255-3}.

\bibitem[Tadmor(2003)]{tadmor2003entropy}
Eitan Tadmor.
\newblock Entropy stability theory for difference approximations of nonlinear
  conservation laws and related time-dependent problems.
\newblock \emph{Acta Numerica}, 12:\penalty0 451--512, 2003.
\newblock \doi{10.1017/S0962492902000156}.

\bibitem[Taylor and Green(1937)]{taylor1937mechanism}
G.~I. Taylor and A.~E. Green.
\newblock Mechanism of the production of small eddies from large ones.
\newblock \emph{Proceedings of the Royal Society of London. Series A},
  158\penalty0 (895):\penalty0 499--521, 1937.
\newblock \doi{10.1098/rspa.1937.0036}.

\bibitem[Tran and Blackbourn(2009)]{tran2009number}
Chuong~V Tran and Luke Blackbourn.
\newblock Number of degrees of freedom of two-dimensional turbulence.
\newblock \emph{Physical Review E—Statistical, Nonlinear, and Soft Matter
  Physics}, 79\penalty0 (5):\penalty0 056308, 2009.
\newblock \doi{10.1103/PhysRevE.79.056308}.

\bibitem[Trebotich(2024)]{trebotich2024exascale}
David Trebotich.
\newblock Exascale computational fluid dynamics in heterogeneous systems.
\newblock \emph{Journal of Fluids Engineering}, 146\penalty0 (4):\penalty0
  041104, 2024.

\bibitem[Williams and Peng(1990)]{williams1990efficient}
Ronald~J Williams and Jing Peng.
\newblock An efficient gradient-based algorithm for on-line training of
  recurrent network trajectories.
\newblock \emph{Neural computation}, 2\penalty0 (4):\penalty0 490--501, 1990.

\bibitem[Winston and Kolter(2020)]{winston2020monotone}
Ezra Winston and J.~Zico Kolter.
\newblock Monotone operator equilibrium networks.
\newblock In \emph{Advances in Neural Information Processing Systems},
  volume~33, 2020.

\bibitem[Wu et~al.(2022)Wu, Maruyama, and Leskovec]{wu2022learning}
Tailin Wu, Takashi Maruyama, and Jure Leskovec.
\newblock Learning to accelerate partial differential equations via latent
  global evolution.
\newblock \emph{Advances in Neural Information Processing Systems},
  35:\penalty0 2240--2253, 2022.
\newblock \doi{2206.07681}.

\end{thebibliography}

\newpage
\appendix
\section{Kinetic Theory Background}
\label{appx:kinetic}
This appendix reviews the kinetic-theory background used in the main text. We
first recall the continuous Boltzmann description and its thermodynamic
H-theorem, then describe the Hermite--Gauss quadrature that produces a finite
velocity lattice, and finally derive the discrete entropy functional used by
entropic lattice Boltzmann methods. The goal is to make explicit where the
discrete H-function comes from and why polynomial lattice equilibria can lose
the entropy structure that the learned entropic closure is designed to restore.

\subsection{Continuous Kinetic Theory and Thermodynamic Structure}

\subsubsection{Phase Space Formulation}

Consider a dilute monatomic gas in the physical domain $\Omega \subset \mathbb{R}^d$ with $d \in \{2,3\}$. Under the molecular chaos hypothesis (\emph{Stosszahlansatz}), the state of the gas is completely characterized by the one-particle distribution function $f: [0,T] \times \Omega \times \mathbb{R}^d \to \mathbb{R}_{\geq 0}$, where $f(t, \bx, \velocity)$ represents the probability density of finding a particle at position $\bx$ with microscopic velocity $\velocity$ at time $t$.
We note that the microscopic position $\bx$ and velocity $\velocity$ are independent of each other.

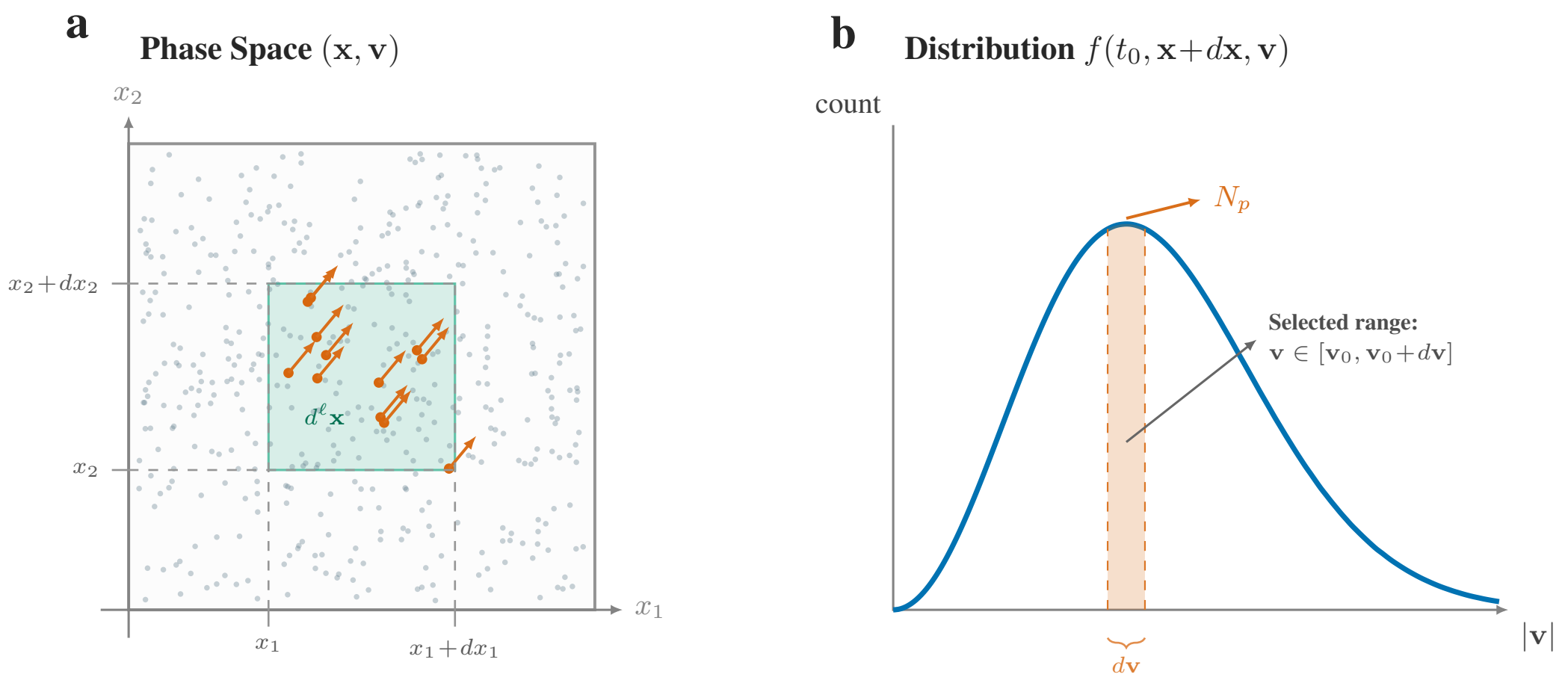
\begin{figure}[ht!]
\resizebox{\linewidth}{!}{%
\begin{tikzpicture}[
    scale=1,
    font=\sffamily\small,
    >={Latex[length=1.3mm, width=1.1mm]},
    axis/.style={line width=0.7pt, color=line_grey!80},
    tick/.style={line width=0.6pt, color=line_grey!70},
    data/.style={line width=1.6pt, color=nat_blue},
    guide/.style={dashed, line width=0.6pt, color=nat_grey!60},
    highlight/.style={color=nat_red},
    panel_label/.style={font=\bfseries\Large, text=black!85},
    panel_title/.style={font=\bfseries\normalsize, text=black!85}
]

    \pgfmathsetseed{100}
    \def\numparticles{450}

    \begin{scope}[local bounding box=physical]
        \node[panel_label, anchor=north west] at (-0.8, 6.5) {a};
        \node[panel_title, anchor=north west] at (0, 6.3) {Phase Space $(\mathbf{x}, \mathbf{v})$};

        \draw[line width=0.9pt, color=line_grey!70, fill=bg_grey!40] (0,0) rectangle (5,5);

        \def\xstart{1.5}
        \def\xend{3.5}
        \def\ystart{1.5}
        \def\yend{3.5}

        \fill[nat_green!80, opacity=0.18] (\xstart, \ystart) rectangle (\xend, \yend);
        \draw[nat_green!85, line width=0.7pt, opacity=0.7] (\xstart, \ystart) rectangle (\xend, \yend);

        \node[font=\scriptsize\bfseries, text=nat_green!70!black, align=center]
            at (2.1, 2.1) {$d^\ell\mathbf{x}$};

        \foreach \i in {1,...,\numparticles} {
            \pgfmathsetmacro{\x}{4.8*rnd + 0.1}
            \pgfmathsetmacro{\y}{4.8*rnd + 0.1}
            \fill[nat_blue!25!gray, opacity=0.35] (\x,\y) circle (0.03);
        }

        \foreach \k in {1,...,12} {
             \pgfmathsetmacro{\px}{2.0*rnd + 1.5}
             \pgfmathsetmacro{\py}{2.0*rnd + 1.5}

             \fill[nat_red!95] (\px,\py) circle (0.055);

             \draw[-{Latex[length=1.6mm, width=1.1mm]}, nat_red!90, line width=0.9pt]
                  (\px,\py) -- ++(0.3, 0.36);
        }

        \draw[->, axis] (-0.3, 0) -- (5.3, 0) node[right, font=\small] {$x_1$};
        \draw[->, axis] (0, -0.3) -- (0, 5.3) node[above, font=\small] {$x_2$};

        \draw[guide] (\xstart, 0) -- (\xstart, \yend);
        \draw[guide] (\xend, 0) -- (\xend, \yend);

        \draw[tick] (\xstart, 0) -- (\xstart, -0.18)
            node[below, font=\scriptsize, text=black!65] {$x_1$};
        \draw[tick] (\xend, 0) -- (\xend, -0.18)
            node[below, font=\scriptsize, text=black!65] {$x_1\!+\!dx_1$};

        \draw[guide] (0, \ystart) -- (\xend, \ystart);
        \draw[guide] (0, \yend) -- (\xend, \yend);

        \draw[tick] (0, \ystart) -- (-0.18, \ystart)
            node[left, font=\scriptsize, text=black!65] {$x_2$};
        \draw[tick] (0, \yend) -- (-0.18, \yend)
            node[left, font=\scriptsize, text=black!65] {$x_2\!+\!dx_2$};

    \end{scope}

    \begin{scope}[xshift=8.2cm, local bounding box=graph]
        \node[panel_label, anchor=north west] at (-0.8, 6.5) {b};
        \node[panel_title, anchor=north west] at (0, 6.3) {Distribution $f(t_0, \mathbf{x}\!+\!d\mathbf{x}, \mathbf{v})$};

        \draw[->, axis] (0,5.2) node[above left, text=black!75, font=\small] {count}
            -- (0,0) -- (6.6,0) node[below right, text=black!75, font=\small] {$|\mathbf{v}|$};

        \draw[data, smooth, domain=0:6.5, samples=120]
            plot (\x, {1.8 * \x^2 * exp(-0.16 * \x^2)});

        \def\va{2.3}
        \def\vb{2.7}

        \path[fill=nat_red!90, opacity=0.22]
            (\va,0) --
            plot[domain=\va:\vb, samples=40] (\x, {1.8 * \x^2 * exp(-0.16 * \x^2)}) --
            (\vb,0) -- cycle;

        \draw[dashed, color=nat_red!85, line width=0.5pt]
            (\va,0) -- (\va, {1.8 * \va^2 * exp(-0.16 * \va^2)});
        \draw[dashed, color=nat_red!85, line width=0.5pt]
            (\vb,0) -- (\vb, {1.8 * \vb^2 * exp(-0.16 * \vb^2)});

        \draw[->, line width=0.9pt, color=nat_red!90]
            (2.5, 4.2) -- (3.3, 4.4)
            node[right, font=\bfseries\small, text=nat_red!90] {$N_p$};

        \draw[->, line width=0.7pt, color=black!60]
            (2.5, 1.8) -- (3.9, 2.9)
            node[right, align=left, font=\scriptsize, text=black!70] {
                \textbf{Selected range:}\\[1pt]
                $\mathbf{v} \in [\mathbf{v}_0, \mathbf{v}_0\!+\!d\mathbf{v}]$
            };

        \draw[decoration={brace, amplitude=3pt, mirror}, decorate, line width=0.6pt, color=nat_red!70]
            (\va, -0.3) -- (\vb, -0.3)
            node[midway, below=2pt, font=\scriptsize, text=nat_red!80] {$d\mathbf{v}$};

    \end{scope}

\end{tikzpicture}
}
\caption{\textbf{Phase-space representation of a kinetic distribution.}
\textbf{a} Phase space $(\mathbf{x},\mathbf{v})$ at $t=t_0$. The green region
marks a spatial volume element containing particles; red particles indicate
those whose velocities lie in a selected interval.
\textbf{b} The corresponding velocity distribution function $f$. The red area
represents the particle count $N_p$ in the selected velocity interval,
connecting microscopic particle states to the macroscopic distribution.
Inspired by Boyd \& Schunk~\citet{boyd2017nonequilibrium}.}
\label{fig:phase_space}
\end{figure}

\begin{definition}[Macroscopic Observables]
For any phase space function $\psi(\velocity)$, the corresponding macroscopic field is defined by velocity averaging:
\begin{equation}
    \langle \psi \rangle(t, \bx) \eqdef \int_{\mathbb{R}^d} f(t, \bx, \velocity) \psi(\velocity) \, d\velocity.
\end{equation}
\end{definition}

The physically relevant macroscopic quantities are the conserved moments corresponding to mass, momentum, and energy:
\begin{equation}
    \boldsymbol{\psi}_{\mathrm{inv}}(\velocity) = \begin{pmatrix} 1 \\ \velocity \\ |\velocity|^2/2 \end{pmatrix}, \quad\boldsymbol{U} = \begin{pmatrix} \rho \\ \rho \Velocity \\ \rho E \end{pmatrix} = \int_{\mathbb{R}^d} f \boldsymbol{\psi}_{\mathrm{inv}} \, d\velocity,
\end{equation}
where $\rho$ is mass density, $\Velocity$ is macroscopic (bulk) velocity, and $E$ is specific total energy.

\subsubsection{The Boltzmann Equation and Collision Invariants}

The evolution of $f$ is governed by the Boltzmann equation:
\begin{equation}
    \label{eq:boltzmann}
    \partial_t f + \velocity \cdot \nabla_{\bx} f = \Omega(f),
\end{equation}
where $\Omega(f)$ is the collision operator. The physical admissibility of collision processes requires:

\begin{axiom}[Conservation Laws]
\label{ax:conservation}
The collision operator preserves mass, momentum, and energy:
\begin{equation}
    \int_{\mathbb{R}^d} \Omega(f) \boldsymbol{\psi}_{\mathrm{inv}}(\velocity) \, d\velocity = \mathbf{0}.
\end{equation}
\end{axiom}

\begin{axiom}[Boltzmann's H-Theorem]
\label{ax:h_theorem}
For collision operators satisfying Boltzmann's H-theorem and for distribution
functions $f > 0$ with $\int f \log f \, d\velocity < \infty$, the
H-functional production is non-positive:
\begin{equation}
    \frac{d}{dt} H(f)(t, \bx) = \int_{\mathbb{R}^d} \Omega(f) \log f \, d\velocity \leq 0,
\end{equation}
where the continuous H-functional (Boltzmann $H$-function) is defined as:
\begin{equation}
    \label{eq:continuous_entropy}
    H(f)(t, \bx) \eqdef \int_{\mathbb{R}^d} f(t, \bx, \velocity) \log f(t, \bx, \velocity) \, d\velocity.
\end{equation}
Equality holds if and only if $\Omega(f) = 0$ almost everywhere.
\end{axiom}

\begin{remark}
For kinetic distributions, the physical entropy has the form
$S=-k_B H$ up to additive constants fixed by the reference measure. Thus
$\dot H\leq 0$ is the $H$-theorem form of nonnegative entropy production for
the modeled collision dynamics.
\end{remark}

\subsubsection{Equilibrium as Entropy Minimization}\label{subsection:equilibrium_entropy_minimization}

The equilibrium distribution is characterized as the entropy minimizer subject to moment constraints. Let $\mathbf{U} = (\rho, \rho\Velocity, \rho E)^\top$ be a given macroscopic state with $\rho > 0$ and $E > |\Velocity|^2/2$. The unique solution to the constrained minimization problem
\begin{equation}
    f^{\mathrm{eq}} = \argmin_{g \geq 0} \left\{ H[g] \;\bigg|\; \int_{\mathbb{R}^d} g \boldsymbol{\psi}_{\mathrm{inv}} \, d\velocity =\boldsymbol{U} \right\}
\end{equation}
is the Maxwell-Boltzmann (Maxwellian) distribution:
\begin{equation}
    \label{eq:maxwellian}
    f^{\mathrm{MB}}(\velocity; \rho, \Velocity, \theta) \eqdef \frac{\rho}{(2\pi\theta)^{d/2}} \exp\left( -\frac{|\velocity - \Velocity|^2}{2\theta} \right),
\end{equation}
where the kinetic temperature is $\theta = \frac{2}{d}(E - |\Velocity|^2/2)$. This result follows from introducing Lagrange multipliers $\boldsymbol{\lambda} \in \mathbb{R}^{d+2}$ for the moment constraints; the first-order optimality condition $\delta H(f) = \boldsymbol{\lambda} \cdot \delta \int f \boldsymbol{\psi}_{\mathrm{inv}} d\velocity$ yields $1 + \log f^{\mathrm{eq}} = \boldsymbol{\lambda} \cdot \boldsymbol{\psi}_{\mathrm{inv}}$, from which \cref{eq:maxwellian} follows by identifying $\boldsymbol{\lambda}$ through the moment constraints. Convexity of $H$ ensures uniqueness.

\subsubsection{BGK Collision Model}

For computational tractability, we adopt the Bhatnagar-Gross-Krook (BGK) single-relaxation-time approximation \citet{bhatnagar1954model}
\begin{equation}
    \label{eq:bgk}
    \Omega(f) \eqdef -\frac{1}{\tau}(f - f^{\mathrm{MB}}),
\end{equation}
where $\tau > 0$ is the relaxation time related to viscosity in hydrodynamics. This model satisfies Axioms \ref{ax:conservation} and \ref{ax:h_theorem}, and in the hydrodynamic limit $\tau \to 0$ (Chapman-Enskog expansion) recovers the compressible Navier-Stokes equations with kinematic viscosity $\nu = \theta \tau$.

\subsection{Spectral Discretization of Velocity Space}

We now construct a finite-dimensional approximation of the Boltzmann equation that preserves the spectral structure of the continuous theory.

\subsubsection{Hermite Polynomial Basis}\label{subsection:Hermite}

\begin{definition}[Probabilist Hermite Polynomials]
The multidimensional Hermite polynomials $\{\mathcal{H}^{(\mathbf{n})}(\velocity)\}$ indexed by multi-indices $\mathbf{n} \in \mathbb{N}_0^d$ are the tensor products of univariate polynomials satisfying the orthogonality relation:
\begin{equation*}
    \int_{\mathbb{R}^d} \mathcal{H}^{(\mathbf{n})}(\velocity) \mathcal{H}^{(\mathbf{m})}(\velocity) \omega(\velocity) \, d\velocity = \delta_{\mathbf{n}\mathbf{m}} \, \mathbf{n}! \, \mathbb{I}^{(|\mathbf{n}|)},
\end{equation*}
where
\begin{equation}\label{eq:weight_function}
\omega(\velocity) \eqdef (2\pi)^{-d/2} e^{-|\velocity|^2/2}
\end{equation}

is the standard Gaussian weight function, $\mathbf{n}! = n_1! \cdots n_d!$, $|\mathbf{n}| = n_1 + \cdots + n_d$, and $\mathbb{I}^{(k)}$ denotes the rank-$k$ identity tensor \citet{shan1998discretization,kruger2017lattice,succi2018lattice}.
\end{definition}

\begin{definition}[Polynomial Space]\label{def:polynomial_space}
We denote by $\mathbb{P}_N(\mathbb{R}^d)$ the space of all real-valued polynomials of total degree at most $N$:
\begin{equation}
    \mathbb{P}_N(\mathbb{R}^d) = \left\{ P(\velocity) = \sum_{|\mathbf{n}| \leq N} c_{\mathbf{n}} \velocity^{\mathbf{n}} \;\bigg|\; c_{\mathbf{n}} \in \mathbb{R} \right\},
\end{equation}
where $\velocity^{\mathbf{n}} =\velocity_1^{n_1} \cdots\velocity_d^{n_d}$ denotes the monomial corresponding to multi-index $\mathbf{n}$.
\end{definition}

\begin{definition}[Grad Expansion]
For fixed truncation order $N \in \mathbb{N}$, the $N$-th order Grad approximation of $f$ is~\citet{grad1949kinetic}:
\begin{equation}
    \label{eq:grad_expansion}
    f^{(N)}(\velocity) \eqdef \omega(\velocity) \sum_{|\mathbf{n}|=0}^N \frac{1}{\mathbf{n}!} \mathbf{a}^{(\mathbf{n})} : \mathcal{H}^{(\mathbf{n})}(\velocity),
\end{equation}
where $\mathbf{a}^{(\mathbf{n})}$ are symmetric moment tensors and $:$ denotes full tensor contraction. The moment tensors are determined by:
\begin{equation}
    \mathbf{a}^{(\mathbf{n})} = \int_{\mathbb{R}^d} f(\velocity) \mathcal{H}^{(\mathbf{n})}(\velocity) \, d\velocity.
\end{equation}
\end{definition}

\begin{remark}
By construction, the truncated Grad expansion satisfies $f^{(N)}/\omega \in \mathbb{P}_N(\mathbb{R}^d)$. For the BGK model with Maxwellian equilibrium, the low-Mach Hermite expansion of $f^{\mathrm{MB}}/\omega$ truncated at $N=2$ yields the standard polynomial lattice equilibrium; the exact Maxwellian ratio is not itself a second-order polynomial except in this truncated approximation.
\end{remark}

\subsubsection{Gauss-Hermite Quadrature}

To evaluate moments without continuous integration, we employ Gaussian quadrature adapted to the Hermite weight \citet{shan1998discretization,shan2006kinetic,kruger2017lattice,succi2001lattice}.

\begin{definition}[Gauss-Hermite Quadrature Rule]
\label{def:quadrature}
A Gauss-Hermite quadrature rule of degree $M$ consists of a finite set of velocity nodes $\mathcal{V} = \{\velocity_i\}_{i=1}^Q \subset \mathbb{R}^d$ and positive weights $\{w_i\}_{i=1}^Q$ such that:
\begin{equation}
    \label{eq:quadrature_exactness}
    \int_{\mathbb{R}^d} P(\velocity) \omega(\velocity) \, d\velocity = \sum_{i=1}^Q w_i P(\velocity_i),
\end{equation}
for all $P \in \mathbb{P}_M(\mathbb{R}^d)$ ; see Definition~\ref{def:polynomial_space}.
\end{definition}

\begin{assumption}[Quadrature Order]
\label{ass:quadrature_order}
We assume $M \geq 2N$, ensuring exact integration of all moments up to order $N$ in the Grad expansion and exact projection of the BGK collision operator.
\end{assumption}

\begin{remark}
Standard lattices satisfy this requirement: D2Q9 has $M=5$ (sufficient for $N=2$), D3Q27 has $M=5$, and higher-order lattices like D3Q121 achieve $M=9$.
\end{remark}

\subsubsection{Discrete Populations}

The key to preserving thermodynamic structure is the correct definition of discrete populations.

\begin{definition}[Lattice Populations]
\label{def:populations}
The discrete particle populations $\{\F_i\}_{i=1}^Q$ are defined by:
\begin{equation}
    \label{eq:population_definition}
    \F_i(t, \bx) \eqdef w_i \frac{f^{(N)}(t, \bx, \velocity_i)}{\omega(\velocity_i)}.
\end{equation}
\end{definition}

Under Definition~\ref{def:populations} and Assumption~\ref{ass:quadrature_order}, the discrete moments exactly reproduce the continuous moments of $f^{(N)}$. Specifically, for any smooth function $\boldsymbol{\psi}$ in $\velocity$ with $\deg( \partial_{\velocity}^k {\boldsymbol{\psi}}) \leq N$, we have:

\begin{equation}
    \label{eq:moment_preservation}
    \sum_{i=1}^Q \F_i \boldsymbol{\psi}(\velocity_i) = \int_{\mathbb{R}^d} f^{(N)}(\velocity) \boldsymbol{\psi}(\velocity) \, d\velocity \approx \int_{\mathbb{R}^d} f(\velocity) \boldsymbol{\psi}(\velocity) \, d\velocity.
\end{equation}
In particular, the conserved moments are preserved exactly: $\sum_{i=1}^Q \F_i \boldsymbol{\psi}_{\mathrm{inv}}(\velocity_i) =\boldsymbol{U}$ as the function $\boldsymbol{\psi}$ is a polynomial and the quadrature is exact.

To verify this, note that by the Grad expansion \cref{eq:grad_expansion}, we have $f^{(N)}/\omega \in \mathbb{P}_N$. For any $\boldsymbol{\psi}$ with $\deg(\boldsymbol{\psi}) \leq N$, the product $\boldsymbol{\psi} \cdot (f^{(N)}/\omega) \in \mathbb{P}_{2N} \subseteq \mathbb{P}_M$ by Assumption \ref{ass:quadrature_order}. Applying the quadrature rule \cref{eq:quadrature_exactness}:
\begin{align*}
    \int_{\mathbb{R}^d} f^{(N)} \boldsymbol{\psi} \, d\velocity &= \int_{\mathbb{R}^d} \frac{f^{(N)}(\velocity)}{\omega(\velocity)} \boldsymbol{\psi}(\velocity) \omega(\velocity) \, d\velocity \\
    &= \sum_{i=1}^Q w_i \frac{f^{(N)}(\velocity_i)}{\omega(\velocity_i)} \boldsymbol{\psi}(\velocity_i) = \sum_{i=1}^Q \F_i \boldsymbol{\psi}(\velocity_i),
\end{align*}
where we used Definition \ref{def:populations} in the last equality. Furthermore, because $f$ and $f^{(N)}$ coincide on the first moments; the quadrature rule is exact for the relevant polynomials, then
\[
\int_{\mathbb{R}^d} f\boldsymbol{\psi} \, d\velocity = \int_{\mathbb{R}^d} f^{(N)} \boldsymbol{\psi} \, d\velocity =\sum_{i=1}^Q \F_i \boldsymbol{\psi}(\velocity_i)
\]
\begingroup
\renewcommand{\kh}[1]{{\textsf{\small KH: #1}}}

\subsection{Lattice Boltzmann Background}
\label{appx:lbm_background}


\begingroup
The kinetic backbone of \emph{NeurDE} is based on the Boltzmann equation (BE) and the evolution of the discretized single-particle distribution $f(t,\bx,\velocity)$.
Here, $f(t,\bx,\velocity)$ describes the mesoscopic particle state at position $\bx$ and particle velocity $\velocity$.
The distribution is updated via the Bhatnagar–Gross–Krook (BGK)~\citet{bhatnagar1954model}  particle collision operator, which relaxes $f$ towards an equilibrium distribution $f^{\text{eq}}$.
The lattice Boltzmann
method realizes this view on a finite velocity
lattice~\citet{bellotti2025fourth} by evolving populations $\F_i(t,\bx)$ at
positions $\bx$ and discrete velocities $\velocity_i$:
\begin{align}
    \F_i(t+1, \bx+\velocity_i) &= \F_i(t, \bx)
    + \Omega(\F_i(t, \bx), \boldsymbol{U}(t, \bx)) \label{eq:Boltzmann}\\
    \Omega_\text{BGK} &= \tfrac{1}{\tau}\left(\Feqlattice_i (\boldsymbol{U}(t, \bx)) - \F_i(t, \bx) \right)
    \label{eq:BGK}
\end{align}
with relaxation time $\tau$ and conserved macroscopic variable $\boldsymbol{U}(t, \bx)$ (density, momentum, and, for
energy-conserving lattices, energy).

This formulation is well suited to hybrid ML for two reasons. First, the BGK
update separates cleanly into a local collision and a global streaming step:
\begin{subequations}\label{eq:LBM_algorithm_dimensionless}
\begin{align}
\F_i(t+1,\bx) &\approx \Phi^{\mathcal{S}}\Phi^{\mathcal{C}}_\text{BGK} \F_i(t,\bx)\\
\F_i^{\text{coll}}(t,\bx) &= \Phi^{\mathcal{C}}_\text{BGK} \F_i(t, \bx) \nonumber \\
= &\F_i(t,\bx) + \frac{1}{\tau}\bigl[\Feqlattice_i(\boldsymbol{U}(t,\bx)) - \F_i(t,\bx)\bigr]
\label{eq:LBM_collision}\\
\F_i(t+1,\bx) &= \Phi^{\mathcal{S}} \F_i^{\text{coll}}(t,\bx) = \F_i^{\text{coll}}(t,\bx-\latticevelocity_i)
\label{eq:LBM_streaming}
\end{align}
\end{subequations}
The collision $\Phi^{\mathcal{C}}_\text{BGK}$ is local but nonlinear, which
makes it both highly parallelizable and the natural place to insert a learned
component~\citet{schornbaum2016massively,benitez2025neural}. The streaming step
$\Phi^{\mathcal{S}}$ is linear and is implemented exactly as a shift on the
lattice. Second, conserved macroscopic observables are recovered by simple
moments of the populations, which preserves conservation laws by construction:
\begin{equation}
\boldsymbol{U}(t, \bx) = (\rho, \rho\Velocity, \rho E)^\top = \sum_i ( 1, \velocity_i, \tfrac{1}{2} \velocity_i^2)^\top \F_i.
\label{eq:macroscopic}
\end{equation}

\emph{NeurDE} inherits the BE mass and momentum conservation and only needs to specify the remaining system-specific physics governed by $\Feqlattice_i(\boldsymbol{U}(t,\bx))$.
To do this, \emph{NeurDE} learns the system's closure to predict the mesoscopic equilibrium from the macroscopic observables $\boldsymbol{U}$.
By learning this macro-to-mesoscopic mapping, \emph{NeurDE} and (Eq.~\ref{eq:macroscopic}) naturally couple the mesoscopic particle (molecular) and macroscopic (observable) scales.
\endgroup

\subsection{Discrete Entropy Functional}\label{section:DiscreteEntropyFunction}

Standard lattice Boltzmann implementations employ polynomial equilibria (e.g., second-order Hermite expansions) that approximate the continuous entropy minimizer $H(f)$ but do not generally minimize the discrete entropy functional. This discrepancy can manifest as spurious H-function growth, $\Delta H_{\text{discrete}} > 0$, equivalently negative physical entropy production, which accumulates into numerical instability. The next subsection develops a quadrature-consistent discrete entropy functional on the discrete velocity set.

\subsubsection{Projection of the Continuous Entropy}

We begin by projecting the continuous entropy functional onto the discrete velocity space. Starting from the continuous entropy and introducing the weight function:
\begin{equation*}
    H(f) = \int_{\mathbb{R}^d} f \log f \, d\velocity = \int_{\mathbb{R}^d} \frac{f(\velocity)}{\omega(\velocity)} \log f(\velocity) \, \omega(\velocity) \, d\velocity.
\end{equation*}

For the Grad approximation $f^{(N)}$, we have $f^{(N)}/\omega \in \mathbb{P}_N$ by construction. The integrand can be decomposed as:
\begin{equation*}
    \frac{f^{(N)}(\velocity)}{\omega(\velocity)} \log f^{(N)}(\velocity) = \frac{f^{(N)}(\velocity)}{\omega(\velocity)} \left[ \log\left(\frac{f^{(N)}(\velocity)}{\omega(\velocity)}\right) + \log \omega(\velocity) \right].
\end{equation*}

The first term involves $(f^{(N)}/\omega) \log(f^{(N)}/\omega)$. Since $f^{(N)}/\omega \in \mathbb{P}_N$, we can write:
\begin{equation*}
    \frac{f^{(N)}(\velocity)}{\omega(\velocity)} = \sum_{|\mathbf{n}| \leq N} c_{\mathbf{n}} \mathcal{H}^{(\mathbf{n})}(\velocity),
\end{equation*}
for some coefficients $c_{\mathbf{n}} \in \mathbb{R}$ determined by the moment tensors $\mathbf{a}^{(\mathbf{n})}$ through the Hermite expansion. The logarithm introduces nonpolynomial terms; however, for populations close to equilibrium (the relevant regime for LBM), we assume $f^{(N)}/\omega$ is strictly positive on the support of the quadrature nodes $\{\velocity_i\}_{i=1}^Q$. Under this condition and requiring quadrature order $M \geq 2N + 2$, the quadrature rule captures the leading-order entropy contributions with controllable error.

Applying the quadrature rule to approximate the continuous integral:
\begin{align*}
    H(f^{(N)}) &\approx \sum_{i=1}^Q w_i \frac{f^{(N)}(\velocity_i)}{\omega(\velocity_i)} \log f^{(N)}(\velocity_i) \\
    &= \sum_{i=1}^Q w_i \frac{f^{(N)}(\velocity_i)}{\omega(\velocity_i)} \left[ \log\left(\frac{f^{(N)}(\velocity_i)}{\omega(\velocity_i)}\right) + \log w_i - \log w_i + \log \omega(\velocity_i) \right] \\
    &= \sum_{i=1}^Q w_i \frac{f^{(N)}(\velocity_i)}{\omega(\velocity_i)} \left[ \log\left(\frac{f^{(N)}(\velocity_i)}{w_i \omega(\velocity_i)}\right) + \log \omega(\velocity_i) \right].
\end{align*}

Using Definition \ref{def:populations}, we have $f^{(N)}(\velocity_i) = \F_i \omega(\velocity_i) / w_i$, which yields:
\begin{equation*}
    H(f^{(N)}) \approx \sum_{i=1}^Q \F_i \left[ \log\left(\frac{\F_i}{w_i}\right) + \log \omega(\velocity_i) \right].
\end{equation*}

We now analyze the second term. Since $\log \omega(\velocity) = -\frac{d}{2}\log(2\pi) - \frac{1}{2}|\velocity|^2$, we have:
\begin{equation*}
    \mathcal{I}(\F) \eqdef \sum_{i=1}^Q \F_i \log \omega(\velocity_i) = -\frac{d}{2}\log(2\pi) \sum_{i=1}^Q \F_i - \frac{1}{2}\sum_{i=1}^Q \F_i |\velocity_i|^2.
\end{equation*}

Recognizing the discrete moments for mass $\rho = \sum_{i=1}^Q \F_i$ and total energy $\rho E = \frac{1}{2}\sum_{i=1}^Q \F_i |\velocity_i|^2$ (by \cref{eq:moment_preservation} with $\boldsymbol{\psi} = |\velocity|^2/2$), we obtain:
\begin{equation*}
    \mathcal{I}(\F) = -\frac{d}{2}\log(2\pi)\rho - \rho E.
\end{equation*}

Since mass $\rho$ and total energy $\rho E$ are collision invariants (conserved by Axiom \ref{ax:conservation}), we have $\frac{d}{dt}\mathcal{I}(\F) = 0$ identically along Boltzmann dynamics. Thus $H(f^{(N)})$ and the first term have identical dynamical content.

\begin{definition}[Discrete Entropy Functional]
\label{def:discrete_entropy}
The discrete entropy functional (discrete $H$-function) is defined as:
\begin{equation}
    \label{eq:discrete_entropy}
    H_{\mathcal{L}}(\F)\eqdef \sum_{i=1}^Q \F_i \log\left(\frac{\F_i}{w_i}\right),
\end{equation}
which is a discrete relative-entropy functional with respect to the quadrature weights $\mathbf{w} = (w_1, \ldots, w_Q)^\top$. When $\sum_i \F_i=1$, it is the usual Kullback--Leibler divergence $D_{\mathrm{KL}}(\F\,\|\,\mathbf{w})$; for fixed non-unit mass it differs from the normalized KL form only by mass-dependent affine terms.
\end{definition}

At the quadrature level, the projected entropy is approximated by the decomposition
\begin{equation*}
    H(f^{(N)}) \approx H_{\mathcal{L}}(\F)+ \mathcal{I}(\F),
\end{equation*}
where $\mathcal{I}(\F)$ is a conserved quantity. Thus, within the discrete quadrature approximation, minimizing the discrete entropy $H_{\mathcal{L}}[\F]$ subject to moment constraints $\sum_{i=1}^Q \F_i \boldsymbol{\psi}_{\mathrm{inv}}(\velocity_i) =\boldsymbol{U}$ is the entropy-consistent discrete counterpart of the continuous constrained minimization problem~\citet{karlin1999perfect,hosseini2023entropic}.

\begin{remark}[Notation]
Throughout the main text, for simplicity of notation, we denote the discrete entropy $H_{\mathcal{L}}$ simply as $H$, as the context implies the discrete functional unless otherwise stated.
\end{remark}

\subsubsection[Reference measure w]{Reference measure $\mathbf{w}$}
\label{subsection:MB_as_W}
The definition of the discrete entropy in \cref{eq:discrete_entropy} is the canonical entropic-LBM choice for preserving the thermodynamic structure of the continuous Boltzmann equation upon velocity discretization~\citet{karlin1999perfect,hosseini2023entropic}. It is easy to see that the Maxwellian \cref{eq:maxwellian} and the weight function of the quadrature \cref{eq:weight_function} are related as,
\[
f^{\mathrm{MB}}(\velocity; \rho, \Velocity, \theta)= \frac{\rho}{\theta^{d/2}} \omega \left( \frac{\velocity- \Velocity}{\sqrt{\theta}}\right),
\]
In particular, if we consider the absolute equilibrium state of the fluid at rest with unit density and temperature ($\rho=1, \Velocity=\mathbf{0}, \theta=1$).
In this state, the continuous Maxwell-Boltzmann distribution function coincides exactly with the Gaussian weight function $\omega(\velocity)$ used for the Hermite spectral projection:
\begin{equation*}\label{eq:omega_as_distr}
    f^{\mathrm{rest}}(\velocity;1,\mathbf{0},1) = \frac{1}{(2\pi)^{d/2}} e^{-|\velocity|^2/2} \equiv \omega(\velocity).
\end{equation*}
Substituting this physical state into the definition of the lattice populations (\cref{def:populations}) yields the discrete equilibrium populations at rest:
\begin{equation}\label{eq:MX_as_weight}
    \F_i^{\mathrm{rest}} = w_i \frac{f^{\mathrm{rest}}(\velocity_i)}{\omega(\velocity_i)} = w_i \frac{\omega(\velocity_i)}{\omega(\velocity_i)} = w_i.
\end{equation}
This identity shows that the vector of quadrature weights $\mathbf{w}$ is the
discrete population vector corresponding to the unit-density rest state.
Moreover, since the quadrature is exact for the zeroth moment, by \cref{eq:quadrature_exactness} we have
\[
\sum_{i=1}^Q w_i = \int \omega(\velocity) d\velocity = 1.
\]
Thus, $\mathbf{w}$ is a valid \emph{probability mass function} representing
the normalized rest equilibrium.

Consequently, the functional $H_{\mathcal{L}}(\F)$ measures the relative
entropic cost of the current dynamic state $\F$ with respect to the rest
reference $\mathbf{w}$. Minimizing this divergence subject to nonzero momentum
constraints $\mathbf{U}$ gives the discrete analogue of the continuous
constrained-entropy problem that recovers the moving Maxwell--Boltzmann
distribution.

\subsubsection{Entropic Stabilization Motivation}

Standard lattice Boltzmann methods employ polynomial equilibria derived by truncating the Hermite expansion of the Maxwellian. The Maxwellian minimizes the continuous entropy $H(f)$, but the discretization error in the Gauss-Hermite quadrature means that polynomial approximations do not generally minimize the discrete entropy $H_{\mathcal{L}}$. The discrete entropy $H_{\mathcal{L}}$ is strictly convex in $\F$, as its Hessian matrix $\frac{\partial^2 H_{\mathcal{L}}}{\partial \F_i \partial \F_j} = \frac{\delta_{ij}}{\F_i}$ is positive definite for $\F > \mathbf{0}$. Therefore, relaxation toward polynomial equilibria that do not minimize $H_{\mathcal{L}}$ can violate the discrete $H$-theorem, producing entropy growth that accumulates into numerical instability over long time integration.
\begingroup
\renewcommand{\kh}[1]{{\textsf{\small KH: #1}}}

\subsection{Entropic Stability}
\label{appx:entropic_background}

\begingroup
To avoid adding spurious noise (erroneous information) from our coarse discretization, we enforce the second law of thermodynamics by making sure entropy increases or stays constant at every autoregressive step.
Naturally, as an isolated system evolves, it reaches equilibrium where disorder and entropy are maximized.
Thus, a decreasing entropy indicates that energy (erroneous information) was added to the system.

The BE backbone provides a simple calculation of entropy from the H-function and $\F_i$.
Defined up to reference-measure constants, the H-function and entropy $S(\F)$ are
\begin{align}
    H(\F) &= \sum_i \F_i\log \left( \frac{\F_i}{w_i} \right)\label{appx:eq:H_fxn}\\
    S(\F) &= - k_B H(\F), \label{eq:entropy}
\end{align}
where $k_B$ is the Boltzmann constant and $w_i$ are the quadrature weights of the discrete velocity lattice~\citet{kruger2017lattice,succi2001lattice}. When
$\sum_i \F_i=\sum_i w_i=1$, Eq.~\ref{appx:eq:H_fxn} is the Kullback--Leibler
divergence $D_{\mathrm{KL}}(\F\Vert\mathbf{w})$; for nonunit mass it is the
corresponding unnormalized relative entropy. Therefore, $H(\F)$ measures how far the
population is from the lattice representation of the Maxwell--Boltzmann rest
state (Appendix~\ref{section:DiscreteEntropyFunction}).

To enforce the second law of thermodynamics, we must satisfy the discrete H-theorem: the H function must decrease or remain constant after the collision operator.
A collision that satisfies the H-theorem
\begin{equation}
    H(\F^{\text{coll}}) \le H(\F^\text{pre}), \label{appx:eq:H_condition}
\end{equation}
dissipates nonequilibrium structure rather than creating spurious grid-scale order.
Entropic LBM (ELBM)
enforces this condition by a clever construction~\citet{karlin1999perfect}: it
introduces a local parameter $\alpha(t,\bx)$ that stretches the relaxation
toward an \emph{entropic mirror state},
\begin{equation}
    \F^{\text{mirr}}(\alpha) = \F^\text{pre} + \alpha (\Feqlattice - \F^\text{pre}). \label{appx:eq:f_mirror}
\end{equation}
Classical ELBM chooses $\alpha$ as an admissible positive root of
$H(\F^{\text{mirr}}(\alpha)) = H(\F^\text{pre})$, which implies
Eq.~\ref{appx:eq:H_condition} under the positivity conditions in
Appendix~\ref{appx:mirror}.
At each time step and each spatial grid point, this root is generally found via expensive iterative second-order optimizers. The entropic parameter $\alpha$ then
turns the BGK collision into
\begin{align}
    \Omega_\text{EBGK} &= \alpha(t,\bx) \beta \left( \Feqlattice_i(\boldsymbol{U}(t, \bx)) - \F_i(t, \bx) \right) \label{eq:collision_EBGK} \\
    \F^\text{coll}_i &= \Phi^{\mathcal{C}}_\text{EBGK} \F_i \nonumber \\
    &= \F_i + \alpha(t,\bx) \beta \left( \Feqlattice_i(\boldsymbol{U}(t, \bx)) - \F_i(t, \bx) \right), \label{appx:eq:ELBM_collision}
\end{align}
where $\beta=1/(2\tau)$ is fixed by the target viscosity~\citet{kruger2017lattice}
and $\Phi^{\mathcal{C}}_\text{EBGK}$ is the entropic BGK collision operator.

Intuitively, $\alpha$ is a flow-dependent thermodynamic dial controlling an artificial viscosity. Near
equilibrium, $\alpha\approx 2$ and the entropic collision reduces to standard BGK relaxation. 
In underresolved or strongly nonequilibrium regions, the entropy condition alters $\alpha$ to modify the effective relaxation $\alpha\beta$ by increasing an artificial viscosity.
The larger viscosity increases local dissipation
of high-frequency structures \emph{before} they reach the grid resolution.
Figure~\ref{fig:discretization} demonstrates this in Fourier space, where the artificial viscosity dampens the energy spectrum near the grid cutoff, prohibiting energy and information from being lost to the unresolved subgrid frequencies.
The artificial viscosity trades unresolved high-frequency fidelity for entropy stability on the
represented grid.

Learning the $\alpha$ parameter, which controls the artificial viscosity and has no closed-form solution, presents a robust learning problem.
Similar to $\Feqlattice_i$, $\alpha$ is local and calculated by an unknown nonlinear function, making it well-suited to neural networks and highly parallelizable.
From the described thermodynamic constraints, we create physics-informed losses to enrich this model with physics principles (Appendix~\ref{appx:entropic_stabilization}).
These principles are resolution invariant and provide supplemental information to help the model make predictions on otherwise missing sub-grid information.
\endgroup

\endgroup

\subsection{Entropic Stabilization via Involution}
\label{appx:entropic_stabilization}

To enforce the discrete $H$-theorem, the Entropic Lattice Boltzmann Method
(ELBM) constructs collision updates for which an admissible positive
entropy-balance root implies $\Delta H_{\mathcal{L}} \leq 0$ by
design~\citet{karlin1999perfect,ansumali2000stabilization,hosseini2023entropic}.

\subsubsection{Entropic Involution and Mirror State}\label{appx:mirror}

\begin{definition}[Entropic Involution]
\label{def:entropic_involution}
Given pre-collision populations $\F = (\F_1, \ldots, \F_Q)^\top$ and a target equilibrium $\F^{\mathrm{eq}} = (\F_1^{\mathrm{eq}}, \ldots, \F_Q^{\mathrm{eq}})^\top$, the entropic mirror state parameterized by $\alpha$ is defined as
\begin{equation}
    \label{eq:mirror_state}
    \F^{\mathrm{mirr}}(\alpha) \eqdef \F + \alpha(\F^{\mathrm{eq}} - \F) = (1-\alpha)\F + \alpha \F^{\mathrm{eq}}.
\end{equation}
\end{definition}

The parameter $\alpha$ is determined by requiring that the mirror state has exactly the same entropy as the pre-collision state.
\begin{definition}[Entropy Balance Equation]
\label{def:entropy_balance}
The entropic parameter $\alpha^*\ge 1$ is the nontrivial positive root of
\begin{equation}
    \label{eq:entropy_balance}
    h_{\beta=1}(\alpha) \eqdef H_{\mathcal{L}}(\F^{\mathrm{mirr}}(\alpha)) - H_{\mathcal{L}}(\F)= 0,
\end{equation}
 with the special value \(\alpha=2\) corresponding to the algebraic reflection \(\F^{\mathrm{mirr}}_{(2)} = 2\Feqlattice - \F\) \citet{karlin1999perfect,ansumali2000stabilization,karlin2014entropic}.
\end{definition}

The post-collision state is obtained by convex mixing,
\begin{equation}
\label{appx_eq:elbm-update-rev}
    \F^{\mathrm{coll}} \;=\; (1-\beta)\F \;+\; \beta\,\F^{\mathrm{mirr}}{(\alpha)}.
\end{equation}
Here, \(\beta \in [0,1]\) sets the nominal viscosity in smooth regions, while \(\alpha\) is chosen locally via \eqref{eq:entropy_balance} to enforce entropy balance and preserve positivity. Because \(H\) is convex and \(\F^{\mathrm{coll}}\) is a convex combination of \(\F\) and \(\F^{\mathrm{mirr}}{(\alpha)}\), with \(H[\F^{\mathrm{mirr}}{(\alpha)}] = H(\F)\) by construction, Jensen's inequality implies
\begin{equation}
\label{appx_eq:jensen-rev}
    H(\F^{\mathrm{coll}}) = H\!\bigl((1-\beta)\F + \beta\,\F^{\mathrm{mirr}}{(\alpha)}\bigr) \le H(\F),
\end{equation}
establishing the discrete H-theorem for the collision step.

\newpage
\section{Proofs of Theoretical Guarantees}
\label{s:Proofs}
\begingroup

\subsection{Preliminaries}\label{s:preliminaries_proof}
We recall from Section~\ref{sec:Foundations} the following: let $d,F$ be positive integers, let $\Omega\subset \mathbb{R}^d$ be a bounded domain, let $\mathcal{F}$ be a feature space, and fix a nonempty collection of encoders $\mathcal{E}:\mathcal{F}\to\mathbb{R}^F$. For each encoded context $\mathbf{z}=\mathcal{E}(\phi)$, the DEQ update induces an operator $G_{\theta,\mathbf{z}}:H^1(\Omega)\to H^1(\Omega)$.

\begin{definition}[Sobolev space]\label{def:Sobolev}
We write $H^1(\Omega)$ for the (Hilbertian) Sobolev space on $\Omega$; whose norm we denote by $\|\cdot\|_{H^1(\Omega)}$ and is given for any $u\in L^2(\Omega;\mathbb{R}^d)$ by
\begin{equation}
\label{eq:sobolevnorm}
        \|u\|_{H^1(\Omega)}
    \eqdef
        \Bigg(
        \int_{\velocity\in \Omega}\,
            \|u(\velocity)\|_2^2 + \|\nabla u(\velocity)\|_2^2
        \,
            d\velocity
        \Bigg)^{1/2}
\end{equation}
where $\nabla$ denotes the weak derivative of $u$ (provided that it exists).

\end{definition}

\begin{definition}[$L^1_+(\Omega)$ space]\label{def:space_L}
We write $L^1_+(\Omega)$ to denote the set of nonnegative finite-mass densities or population fields on $\Omega$; that is, the collection of every $\F\in L^1(\Omega)$ satisfying $\F\ge 0$ Lebesgue-a.e.\ and $0<\int_\Omega \F(\velocity)\,d\velocity<\infty$. The probability-normalized case has unit mass, while the lattice-Boltzmann populations used in the experiments may carry nonunit local mass.

\end{definition}

\begin{assumption}[Uniform bounds of $\F$ and $\F^{\text{mirr}}$]\label{ass:entropy_bounds}
Let $\mathcal{A}\subset H^1(\Omega)$ be a set on which $\F^{\text{mirr}}:\mathcal{A}\to L^1_+(\Omega)$ is Lipschitz with respect to the $L^2(\Omega)$ norm on $\mathcal{A}$ and the $L^1(\Omega)$ norm on its image, and on which there exist constants $0<m<M<\infty$ such that
\begin{equation}
\label{eq:hb__uniform_bounds}
        m
    \le
        (1-\beta)\F(\velocity) + \beta\, \F^{\text{mirr}}(\alpha)(\velocity)
    \le
        M
\end{equation}
for Lebesgue-a.e. $\velocity\in\Omega$ and for all $\alpha\in\mathcal{A}$.
\end{assumption}

\paragraph{From a pointwise symbol to an operator-level update.}
The architecture in this paper realizes the DEQ update by a finite-depth
convolutional network whose spatial coupling is not pointwise. Accordingly,
the fixed-point analysis below is phrased at the level of the operator
induced on $H^1(\Omega)$ rather than as a Nemytskii (superposition) lift of
a pointwise kernel. For each $\phi\in\mathcal{F}$ with encoded context
$\mathbf{z}=\mathcal{E}(\phi)\in\mathbb{R}^F$, write
\[
    G_{\theta,\mathbf{z}}: H^1(\Omega)\to H^1(\Omega),
    \qquad
    \alpha\mapsto g_\theta(\alpha,\mathbf{z})
\]
for the operator that takes the current iterate to the next.
The Nemytskii lift of a pointwise $C^1$ symbol (Lemma~\ref{lem:welldefined_contraction})
is one realization of $G_{\theta,\mathbf{z}}$; convolutional updates with
bounded layer norms and Lipschitz activations are another. The arguments
below depend only on operator-level Lipschitz and self-map properties of
$G_{\theta,\mathbf{z}}$ on a suitable admissible set, and are agnostic to
whether the spatial coupling is pointwise or convolutional.

\begin{assumption}[Operator-level regularity of the DEQ update]
\label{ass:Small}

The domain $\Omega\subset\mathbb{R}^d$ is bounded with Lipschitz boundary.
There exists a nonempty closed bounded convex set
$\mathcal{K}\subseteq H^1(\Omega)$ -- the \emph{admissible set of stability
parameters} -- such that, for every input feature $\phi\in\mathcal{F}$ with
encoded context $\mathbf{z}=\mathcal{E}(\phi)$, the induced operator
$G_{\theta,\mathbf{z}}:\alpha\mapsto g_\theta(\alpha,\mathbf{z})$ satisfies
the following:
\begin{enumerate}[label=(A\arabic*),leftmargin=*]
    \item \label{ass:Small:i}
    \emph{(Well-definedness.)}
    $G_{\theta,\mathbf{z}}: H^1(\Omega)\to H^1(\Omega)$ is a well-defined map.
    \item \label{ass:Small:ii}
    \emph{(Invariance of $\mathcal{K}$.)}
    $G_{\theta,\mathbf{z}}(\mathcal{K})\subseteq \mathcal{K}$.
    \item \label{ass:Small:iii}
    \emph{($L^2$-Lipschitzness on $\mathcal{K}$.)}
    There exists a constant $L_v\ge 0$ such that
    \begin{equation}
    \label{eq:growth_condition}
            \|G_{\theta,\mathbf{z}}(\alpha_0)-G_{\theta,\mathbf{z}}(\alpha_1)\|_{L^2(\Omega)}
        \le
            L_v\,\|\alpha_0-\alpha_1\|_{L^2(\Omega)},
        \qquad
        \alpha_0,\alpha_1\in \mathcal{K}.
    \end{equation}
\end{enumerate}

\end{assumption}

\begin{remark}[Pointwise (Nemytskii) realization and the role of $B_1$]
\label{rem:nemytskii_special}
A common way to realize Assumption~\ref{ass:Small} is the pointwise case in
which $g_\theta:\mathbb{R}^d\times\mathbb{R}^F\to\Omega$ is continuously
differentiable with bounded Jacobian
$\sup_{\velocity\in\mathbb{R}^d}\|Jg_\theta(\velocity,v)\|_2\le L_v$.
Lemma~\ref{lem:welldefined_contraction} shows that, with $\Omega$ bounded
and $0\in\Omega$, the Nemytskii lift
$T_{g_\theta:\phi}:H^1(\Omega)\to H^1(\Omega)$ is then well-defined and
$L_v$-Lipschitz in $L^2$. With the additional structural conditions
$g_\theta(0,v)=0$ for every $v\in\mathbb{R}^F$ and $L_v\le 1$, the lift also
stabilizes the Sobolev unit ball
$B_1\eqdef\{u\in H^1(\Omega):\|u\|_{H^1(\Omega)}\le 1\}$, providing a
concrete pointwise realization of Assumption~\ref{ass:Small} with
$\mathcal{K}=B_1$. In that special case, however, the strict-contraction
condition $L_v<1$ forces the unique fixed point to be the zero function,
since $T_{g_\theta:\phi}0=0$. The general admissible-set formulation here
avoids that artifact: dropping the zero-anchoring condition allows
$\mathcal{K}$ to be a translate of a Sobolev ball centered near the entropic
value $\alpha\approx 2$, which matches the $\tanh$-bounded shifted output of
the architecture used in this paper.
\end{remark}

\begin{remark}[Convolutional realization]
\label{rem:convolutional_realization}
The architecture used in this paper realizes $G_{\theta,\mathbf{z}}$ as a
finite composition of convolutional layers with bounded operator norms and
Lipschitz pointwise activations, followed by a $\tanh$-bounded shifted output
near the entropic value $\alpha\approx 2$. The shifted output keeps each
iterate in a fixed pointwise interval, which gives rise to a closed bounded
convex admissible set $\mathcal{K}$ that is a translate of a Sobolev ball
rather than the zero-centered $B_1$. The Lipschitz constant on $\mathcal{K}$
is bounded by the product of $L^2$ operator norms of the convolutional
layers and the activation Lipschitz constants; we do not unfold these
architecture-dependent constants here, because the analysis below depends
only on the operator-level conditions in Assumption~\ref{ass:Small}.
\end{remark}

\subsection{Existence of Fixed Points}
\label{s:Proofs__ss:existence_FPs}
We first establish Theorem~\ref{thrm:existence_of_FP}.
\begin{lemma}[Pointwise (Nemytskii) realization of Assumption~\ref{ass:Small}]
\label{lem:welldefined_contraction}
Let $\Omega\subset\mathbb{R}^d$ be a bounded domain with Lipschitz
boundary. Suppose $g_\theta:\mathbb{R}^d\times\mathbb{R}^F\to\Omega$ is
continuously differentiable, and that for every $v\in\mathbb{R}^F$ the
Jacobian satisfies
\[
    L_v\eqdef
    \sup_{\velocity\in\mathbb{R}^d}\,\|Jg_\theta(\velocity,v)\|_2
    < \infty.
\]
Then for every $\phi\in\mathcal{F}$ the Nemytskii-type composition operator
\begin{equation}
\label{eq:composition}
\begin{aligned}
T_{g_\theta:\phi}:
(H^1(\Omega),\|\cdot\|_{H^1(\Omega)})
& \rightarrow
(H^1(\Omega),\|\cdot\|_{L^2(\Omega)})
\\
u & \mapsto g_\theta(u,\mathcal{E}(\phi))
\end{aligned}
\end{equation}
is well-defined and $L_v$-Lipschitz as a map into $L^2(\Omega)$. This is the
standard superposition-operator viewpoint for Nemytskii maps on Sobolev
spaces~\citet{appell1990nonlinear}. If, in addition,
$0\in\Omega$, $g_\theta(0,v)=0$ for every $v\in\mathbb{R}^F$, and $L_v\le 1$,
then $T_{g_\theta:\phi}$ also maps the Sobolev unit ball
$B_1\eqdef\{u\in H^1(\Omega):\|u\|_{H^1(\Omega)}\le 1\}$ to itself, providing
a concrete realization of Assumption~\ref{ass:Small} with $\mathcal{K}=B_1$.
\end{lemma}
\begin{proof}[{Proof of Lemma~\ref{lem:welldefined_contraction}}]
Fix $\phi\in\mathcal{F}$ and abbreviate $
    v   \eqdef   \mathcal{E}(\phi)\in\mathbb{R}^F
$ and $\Phi_v(\cdot)  \eqdef  g_\theta(\cdot,v):\mathbb{R}^d\to\Omega$.
We interpret~\eqref{eq:composition} in the standard Nemytskii sense, i.e.\ for every $\velocity\in \Omega$ let $\big(T_{g_\theta:\phi}u\big)(\velocity)
  \eqdef
    g_\theta\big(u(\velocity),v\big)$.
We first show that $\Phi_v$ is $L_v$-Lipschitz on $\mathbb{R}^d$.  To see this first fix $x,y\in\mathbb{R}^d$.  Since $g_\theta(\cdot,v)$ is $C^1$ on $\mathbb{R}^d$, the Fundamental Theorem of Calculus along the line segment $t\mapsto y+t(x-y)$ yields
\begin{equation}
\label{eq:FTOC}
    \Phi_v(x)-\Phi_v(y)
=
    \int_0^1
        Jg_\theta\big(y+t(x-y),v\big)\,(x-y)
    \,dt.
\end{equation}
The $L_v$-Lipschitzness of $\Phi_v$ on $\mathbb{R}^d$ now follows upon taking Euclidean norms across~\eqref{eq:FTOC} and using the Jacobian bound in the lemma's hypothesis
\[
    \big\|\Phi_v(x)-\Phi_v(y)\big\|_2
\le
    \int_0^1
        \big\|Jg_\theta\big(y+t(x-y),v\big)\big\|_2\,\|x-y\|_2
    \,dt
\le
    L_v\,\|x-y\|_2.
\]
Next, we show that the composition operator is indeed well-defined on $H^1(\Omega)$.  To this end, the $L_v$-Lipschitzness of $\Phi_v$ implies Lebesgue-a.e.\ differentiability by Rademacher's Theorem, cf.~\citep[Theorem 3.1.6]{Federer1969}.
We then directly observe that $ \|D\Phi_v(\cdot)\|_2 \le L_v$ a.e.\ on $\mathbb{R}^d$.
Now, for any $u\in H^1(\Omega)$ we compute the $H^1(\Omega)$-Sobolev norm of the composition $\Phi_v\circ u$ via the chain rule; yielding
\[
        \Phi_v\circ u \in H^1(\Omega)
    \qquad\text{and}\qquad
    \nabla(\Phi_v\circ u)(\velocity)
    =
        Jg_\theta\big(u(\velocity),v\big)\,\nabla u(\velocity)
\]
for Lebesgue-a.e.\ $\velocity\in\Omega$.
In particular, we have that $
    \|\nabla(\Phi_v\circ u)\|_{L^2(\Omega)}
\le
    L_v\,\|\nabla u\|_{L^2(\Omega)}
$.
Now, as $\Phi_v(\mathbb{R}^d)\subseteq\Omega$ and since $\Omega$ is bounded, then the quantity
$
M_\Omega\eqdef \sup_{x\in\Omega}\|x\|_2
$
must be finite.  Consequently, we have that $\|\Phi_v(u(\cdot))\|_2\le M_\Omega<\infty$ Lebesgue-almost everywhere; whence
\[
    \|\Phi_v\circ u\|_{L^2(\Omega)}
\le
    |\Omega|^{1/2}\,M_\Omega
.
\]
Therefore $\Phi_v\circ u\in H^1(\Omega)$, so $T_{g_\theta:\phi}$ is well-defined as a map into the \emph{set} $H^1(\Omega)$.

It remains to verify the Lipschitz estimate.  Let $u,w\in H^1(\Omega)$.  By the first step, for Lebesgue-a.e.\ $\velocity\in\Omega$, we have $
    \big\|g_\theta\big(u(\velocity),v\big)-g_\theta\big(w(\velocity),v\big)\big\|_2
\le
    L_v\,\|u(\velocity)-w(\velocity)\|_2$.
Squaring and integrating yields
\[
    \|T_{g_\theta:\phi}u - T_{g_\theta:\phi}w\|_{L^2(\Omega)}^2
=
    \int_{\Omega}
        \big\|g_\theta\big(u(\velocity),v\big)-g_\theta\big(w(\velocity),v\big)\big\|_2^2
    \,d\velocity
\le
    L_v^2\int_{\Omega}\|u(\velocity)-w(\velocity)\|_2^2\,d\velocity
,
\]
so
$
    \|T_{g_\theta:\phi}u - T_{g_\theta:\phi}w\|_{L^2(\Omega)}
\le
    L_v\,\|u-w\|_{L^2(\Omega)}
\le
    L_v\,\|u-w\|_{H^1(\Omega)}
$.
This proves that $T_{g_\theta:\phi}$ is $L_v$-Lipschitz as a map into $L^2(\Omega)$.

\paragraph{Conditional self-map on $B_1$.}
Suppose now the additional hypotheses $0\in\Omega$, $g_\theta(0,v)=0$ for every $v$, and $L_v\le 1$.  The preceding estimate then gives the non-expansive bound
\begin{equation}
\label{eq:non_expansive}
    \|T_{g_\theta:\phi}u - T_{g_\theta:\phi}w\|_{L^2(\Omega)}
\le
    \|u-w\|_{L^2(\Omega)}.
\end{equation}
Since $g_\theta(0,v)=0$, the constant function $0\in H^1(\Omega)$ satisfies
$T_{g_\theta:\phi}0=0$, and~\eqref{eq:non_expansive} implies
\[
    \|T_{g_\theta:\phi}u\|_{L^2(\Omega)}
=
    \|T_{g_\theta:\phi}u - T_{g_\theta:\phi}0\|_{L^2(\Omega)}
\le
    L_v\,\|u\|_{L^2(\Omega)}
\le
    L_v\,\|u\|_{H^1(\Omega)}.
\]
Combining this with $\|\nabla(T_{g_\theta:\phi}u)\|_{L^2(\Omega)}\le L_v\,\|\nabla u\|_{L^2(\Omega)}$ yields
\[
    \|T_{g_\theta:\phi}u\|_{H^1(\Omega)}^2
=
    \|T_{g_\theta:\phi}u\|_{L^2(\Omega)}^2
    +
    \|\nabla(T_{g_\theta:\phi}u)\|_{L^2(\Omega)}^2
\le
    L_v^2\,\|u\|_{H^1(\Omega)}^2.
\]
In particular, if $\|u\|_{H^1(\Omega)}\le 1$ then $\|T_{g_\theta:\phi}u\|_{H^1(\Omega)}\le 1$, i.e.\ $T_{g_\theta:\phi}(B_1)\subseteq B_1$. This realizes Assumption~\ref{ass:Small} with $\mathcal{K}=B_1$, but as noted in Remark~\ref{rem:nemytskii_special} this special case forces $0$ to be a fixed point of $T_{g_\theta:\phi}$, which is not the regime of interest for the architecture used in this paper.
\end{proof}
We are now in a position to prove Theorem~\ref{thrm:existence_of_FP}.
\begin{proof}[{Proof of Theorem~\ref{thrm:existence_of_FP}}]
Fix $\phi\in\mathcal{F}$ and write $\mathbf{z}=\mathcal{E}(\phi)$.
By Assumption~\ref{ass:Small}\ref{ass:Small:i}--\ref{ass:Small:ii}, the
operator $G_{\theta,\mathbf{z}}: H^1(\Omega)\to H^1(\Omega)$ is well-defined
and leaves the admissible set $\mathcal{K}$ invariant, and
by~\ref{ass:Small:iii} its restriction to $\mathcal{K}$ is $L_v$-Lipschitz
in the $L^2$ metric, hence continuous in $L^2$.

Since $\mathcal{K}$ is bounded in $H^1(\Omega)$ and $\Omega$ is a bounded
domain with Lipschitz boundary, the Rellich--Kondrachov
theorem~\citep[Theorem 5.1]{EvansPDE} implies that $\mathcal{K}$ is
relatively compact in $L^2(\Omega)$. To see that $\mathcal{K}$ is closed in
$L^2(\Omega)$, let $u_n\in\mathcal{K}$ with $u_n\to u$ in $L^2(\Omega)$. The
sequence is bounded in the reflexive space $H^1(\Omega)$, so a subsequence
converges weakly in $H^1(\Omega)$ to some $\tilde u\in H^1(\Omega)$; the same
subsequence converges strongly in $L^2(\Omega)$ to $u$, hence $\tilde u=u$
a.e. By Mazur's theorem the closed convex set $\mathcal{K}\subseteq H^1(\Omega)$
is weakly closed, so $\tilde u\in\mathcal{K}$ and therefore $u\in\mathcal{K}$.
Hence $\mathcal{K}$ is compact in $L^2(\Omega)$.

Since $G_{\theta,\mathbf{z}}|_{\mathcal{K}}$ is continuous in the $L^2$ metric
and maps $\mathcal{K}$ to itself, the Schauder--Tychonoff fixed-point
theorem applies; see~\citep[page 136]{bonsall1962lectures} and the discussion
in~\citet{cobzas2006fixed}. Therefore $G_{\theta,\mathbf{z}}$ admits a fixed
point $\alpha^{\star}\in\mathcal{K}$ satisfying~\eqref{eq:existence_of_FP}.
\end{proof}

\subsection{Uniqueness and Exponential Convergence}\label{s:Proofs__ss:UniqExpConv}
If, moreover, the Lipschitz constant is strictly below one, then the preceding result strengthens substantially: the iterates generated by our model converge to the equilibrium $\alpha^{\star}$ at an \emph{exponential} rate.  In particular, under this condition, the equilibrium becomes \emph{unique}.

\begin{proposition}[Exponential Convergence]
\label{prop:uniqueness_convergence}
In the setting of Theorem~\ref{thrm:existence_of_FP}, suppose additionally
that $L_v<1$ in~\eqref{eq:growth_condition}.
Then the equilibrium $\alpha^{\star}\in\mathcal{K}$ is unique,
and for any $\alpha^{(0)}\in\mathcal{K}$ the sequence of iterates
$(\alpha^{(i)})_{i=0}^{\infty}$ defined for every non-negative integer $i$ by
$\alpha^{(i+1)}\eqdef G_{\theta,\mathbf{z}}(\alpha^{(i)})=g_\theta(\alpha^{(i)},\mathcal{E}(\phi))$
satisfies the exponential convergence bound
\begin{equation}
\label{eq:exponential_convergence}
    \|\alpha^{\star}-\alpha^{(i)}\|_{L^2(\Omega)}
\le
    \|\alpha^{(0)}-\alpha^{\star}\|_{L^2(\Omega)}
    \,
    L_v^i
.
\end{equation}
\end{proposition}

\begin{proof}[{Proof of Proposition~\ref{prop:uniqueness_convergence}}]
Under Assumption~\ref{ass:Small} with $L_v<1$, the operator
$G_{\theta,\mathbf{z}}$ is a strict contraction on $\mathcal{K}$ in the $L^2$
metric and maps $\mathcal{K}$ into itself. The proof of
Theorem~\ref{thrm:existence_of_FP} shows that $\mathcal{K}$ is closed in
$L^2(\Omega)$, hence complete in this metric. The Banach contraction
principle (also a special case of the generalizations
in~\citet{ciric1974generalization}) then yields (i) a unique fixed point of
$G_{\theta,\mathbf{z}}$ in $\mathcal{K}$, and (ii) the exponential
estimate~\eqref{eq:exponential_convergence}. By
Theorem~\ref{thrm:existence_of_FP} and (i), this unique fixed point is
$\alpha^{\star}$. The argument does \emph{not} require
$G_{\theta,\mathbf{z}}(0)=0$; in particular, $\alpha^{\star}$ need not be the
zero function and, for the architecture used in this paper, lies near the
entropic value $\alpha\approx 2$ rather than at the origin.
\end{proof}

\subsection{Convergence of the Entropy Balance Functions}
\label{s:Proofs__ss:Exp_fast_Convergence}
To obtain convergence of the associated entropy-balance functions, we require
a stability lemma. Although Proposition~\ref{prop:uniqueness_convergence}
gives convergence of the architecture iterates under a strict-contraction
assumption, we still need to control how the entropy-balance map varies with
respect to its input density. Concretely, we show that the entropy-balance map
is Lipschitz on any class of densities satisfying the stated positivity and
boundedness assumptions.
\begin{lemma}[Lipschitzness of the Entropy Balance Function]
\label{lemma:Lip}
Fix $0\le \beta \le 1$ and $\F\in L^1_+(\Omega)$, and assume that Assumption~\ref{ass:entropy_bounds} holds on $\mathcal{A}$.
There exists a constant $L\ge 0$ such that the entropy balance function $h_{\beta}(\cdot;\F)|_{\mathcal{A}}$ is $L$-Lipschitz on $\mathcal{A}$ with respect to the $L^2(\Omega)$ metric.  Moreover, we have
\begin{equation}
\label{eq:hb__Lip_estimate}
    L
\le
        \beta\,
        \Big(
            \sup_{t\in[m,M]}\,\big|1+\log(t)\big|
        \Big)\,
        \operatorname{Lip}( \F^{\text{mirr}};\mathcal{A})
\end{equation}
where $\operatorname{Lip}( \F^{\text{mirr}};\mathcal{A})$ denotes any Lipschitz constant of $ \F^{\text{mirr}}$ on $\mathcal{A}$ with respect to the $L^2(\Omega)$ norm on $\mathcal{A}$ and the $L^1(\Omega)$ norm on its image.
\end{lemma}
\begin{proof}[{Proof of Lemma~\ref{lemma:Lip}}]
Define $\phi:[0,\infty)\to \mathbb{R}$ by $\phi(x)\eqdef x\log(x)$ with $\phi(0)\eqdef 0$.
Then $\phi\in C^1((0,\infty))$ and $\phi'(x)=1+\log(x)$ for all $x>0$.
In particular, for any $x,y\in[m,M]$, the mean value theorem yields
\[
        \big|\phi(x)-\phi(y)\big|
    \le
        \Big(
            \sup_{t\in[m,M]}\,\big|\phi'(t)\big|
        \Big)\,
        |x-y|
    =
        \Big(
            \sup_{t\in[m,M]}\,\big|1+\log(t)\big|
        \Big)\,
        |x-y|
.
\]
Now fix $\alpha\in\mathcal{A}$ and set $
        \F^{\text{coll}}_{\alpha}
    \eqdef
        (1-\beta)\F + \beta\, \F^{\text{mirr}}(\alpha)
    \in
        L^1_+(\Omega)
$. By~\eqref{eq:hb__uniform_bounds} we have $\F^{\text{coll}}_{\alpha}(\velocity)\in[m,M]$ for Lebesgue-a.e.\ $\velocity\in\Omega$ and all
$\alpha\in\mathcal{A}$.
Consequently, for any $\alpha_0,\alpha_1\in\mathcal{A}$,
\[
\begin{aligned}
\big|
        h_{\beta}(\alpha_0;\F)-h_{\beta}(\alpha_1;\F)
\big|
&=
\big|
        H(\F^{\text{coll}}_{\alpha_0})-H(\F^{\text{coll}}_{\alpha_1})
\big|
\\
&=
\bigg|
        \int_{\Omega}\,
            \phi\!\big(\F^{\text{coll}}_{\alpha_0}(\velocity)\big)
        -
            \phi\!\big(\F^{\text{coll}}_{\alpha_1}(\velocity)\big)
        \,d\velocity
\bigg|
\\
&\le
        \int_{\Omega}\,
            \big|
                \phi\!\big(\F^{\text{coll}}_{\alpha_0}(\velocity)\big)
            -
                \phi\!\big(\F^{\text{coll}}_{\alpha_1}(\velocity)\big)
            \big|
        \,d\velocity
\\
&\le
        \Big(
            \sup_{t\in[m,M]}\,\big|1+\log(t)\big|
        \Big)\,
        \int_{\Omega}\,
            \big|
                \F^{\text{coll}}_{\alpha_0}(\velocity)-\F^{\text{coll}}_{\alpha_1}(\velocity)
            \big|
        \,d\velocity
\\
&=
        \Big(
            \sup_{t\in[m,M]}\,\big|1+\log(t)\big|
        \Big)\,
        \|\F^{\text{coll}}_{\alpha_0}-\F^{\text{coll}}_{\alpha_1}\|_{L^1(\Omega)}
\\
&=
        \beta\,
        \Big(
            \sup_{t\in[m,M]}\,\big|1+\log(t)\big|
        \Big)\,
        \| \F^{\text{mirr}}(\alpha_0)- \F^{\text{mirr}}(\alpha_1)\|_{L^1(\Omega)}
\\
&\le
        \beta\,
        \Big(
            \sup_{t\in[m,M]}\,\big|1+\log(t)\big|
        \Big)\,
        \operatorname{Lip}( \F^{\text{mirr}};\mathcal{A})\,
        \|\alpha_0-\alpha_1\|_{L^2(\Omega)}.
\end{aligned}
\]
This is precisely~\eqref{eq:hb__Lip_estimate}.
\end{proof}
We can now prove convergence of the entropy-balance residual under the
hypotheses of Proposition~\ref{prop:uniqueness_convergence} and
Assumption~\ref{ass:entropy_bounds}.
\begin{proof}[{Proof of Theorem~\ref{thrm:EBF_Convergence}}]
Under the hypotheses of Proposition~\ref{prop:uniqueness_convergence}, the exponential convergence estimate
in~\eqref{eq:exponential_convergence} holds.
By assumption, $\alpha^{\star}\in\mathcal{A}$ and there exists $I\in\mathbb{N}$ such that
$\alpha^{(i)}\in\mathcal{A}$ for all $i\in\mathbb{N}_+$ with $i\ge I$.
Define the (tail) set
\[
        \mathcal{A}^{\star}
    \eqdef
        \{\alpha^{\star}\}
        \cup
        \{\alpha^{(i)}\}_{i\in\mathbb{N}_+,\, i\ge I}
    \subseteq
        \mathcal{A}
    \,.
\]
Consequently, Lemma~\ref{lemma:Lip} applies on $\mathcal{A}^{\star}$, and therefore there exists $L\ge 0$
(cf.~\eqref{eq:hb__Lip_estimate}) such that the restriction $h_{\beta}|_{\mathcal{A}^{\star}}$ is $L$-Lipschitz.

Hence, for every $i\in\mathbb{N}_+$ with $i\ge I$ we obtain
\[
\begin{aligned}
        \big|
                h_{\beta}(\alpha^{\star};\F)
            -
                h_{\beta}(\alpha^{(i)};\F)
        \big|
&\le
        L\,\|\alpha^{\star}-\alpha^{(i)}\|_{L^2(\Omega)}
\\
&\le
        L\,\|\alpha^{(0)}-\alpha^{\star}\|_{L^2(\Omega)}\,L_v^{i},
\end{aligned}
\]
where the final inequality follows from~\eqref{eq:exponential_convergence}.
This concludes the proof upon setting
\[
        \tilde{L}
    \eqdef
        L\,\|\alpha^{(0)}-\alpha^{\star}\|_{L^2(\Omega)}
\]
and relabeling $\tilde{L}$ as $L$.
\end{proof}

\endgroup
\newpage

\section{Model Architecture}\label{appx:architecture}

\begingroup
Fig.~\ref{fig:architecture} illustrates the DEQ framework used to approximate
Eq.~\ref{eq:EBF}. Since $\beta$ is determined by the target Reynolds
number~\citet{karlin1998equilibria}, we solve the entropy-balance equation at
$\beta=1$. For an exact positive entropy-balance root, the post-collision
convex combination satisfies the discrete H-theorem for any
$\beta \in [0,1]$ by Jensen's inequality (Eq.~\ref{appx_eq:jensen-rev}). The
adaptive relaxation parameter $\alpha^{\star}$ is then represented as the fixed
point of a learnable transformation. Following a decoupled conditioning
strategy~\citet{bai2019deep,bai2020multiscale,fung2022jfb}, the architecture
separates high-dimensional feature extraction from the recurrent dynamics that
govern the fixed-point iterations.

\subsection{Connection to Entropic Root Finding}

The DEQ model~\citet{bai2019deep} is motivated by the fact that the entropic
stabilizer is already a root-finding problem. For fixed populations $\F$, let
\[
    r(\alpha;\F) \eqdef h_{\beta=1}(\alpha;\F).
\]
The classical ELBM solve applies a damped Newton map
\[
    \mathcal{N}_{\lambda}(\alpha;\F)
    \eqdef
    \alpha - \lambda\,
    \frac{r(\alpha;\F)}{\partial_{\alpha} r(\alpha;\F)},
    \qquad 0 < \lambda \le 1.
\]
Thus every admissible entropic root $r(\alpha^\star;\F)=0$ is equivalently a
fixed point $\alpha^\star=\mathcal{N}_{\lambda}(\alpha^\star;\F)$, provided
$\partial_{\alpha}r(\alpha^\star;\F)\neq 0$. We use the DEQ as a learned
Newton-style fixed-point solver: instead of running Newton iterations independently
at every lattice site and timestep, the learnable map $g_\theta$ approximates
this fixed-point iteration while conditioning on mesoscopic, macroscopic, and
physics-hint features. The guarantees in Section~\ref{sec:Foundations}
abstract the conditions under which this learned implicit iteration has a
well-defined attractor. When the learned map approximates the Newton root map,
the attractor approximates an entropy-balance root; the fixed-point theorem
alone guarantees existence and convergence of the learned iteration, not zero
entropy residual by itself.
\endgroup

The computational pipeline begins with an encoder that processes microscopic
nonequilibrium features $(\F_{\mathrm{neq}} = \F - \Feqlattice)$ together with
macroscopic variables $(\rho,\mathbf{u})$ (Fig.~\ref{fig:architecture}a).
Because NeurDE operates at the mesoscopic level, these inputs can be augmented
with higher-order moments, including momentum flux and stress
tensors~\citet{bosch2015entropic}. All features are normalized and mapped into a latent
context vector $\mathbf{z}$, which is computed once per time step and reused in
each DEQ iteration. At every iteration, the DEQ update function $g_\theta$
(Fig.~\ref{fig:architecture}b) receives three concatenated inputs: the frozen
context $\mathbf{z}$, the current state $\alpha_k$, and the physics-based hint
$\mathbf{h}_k$. These are processed by a convolutional network of
\texttt{Conv+GroupNorm} layers followed by a $\tanh$ activation, and the output
is shifted by $+2$ to keep the predicted $\alpha_{k+1}$ near the physically
admissible relaxation range. To guide the solver, a physics engine computes
auxiliary hints $\mathbf{h}_k$ from the current iterate $\alpha_k$, including
the entropy residual, its derivative, and additional instability indicators
when available~\citet{latt2020efficient}.
By embedding these physically meaningful signals into the update rule, the
recurrent map is trained toward low entropy residuals and stable rollouts
rather than treated as an unconstrained black-box predictor.


\begin{algorithm}[ht!]
\caption{Entropy Root DEQ: Complete Forward Pass}
\label{alg:main}
\begin{algorithmic}[1]
\STATE \textbf{Input:} Distribution functions $F, F_{\text{eq}} \in \mathbb{R}^{B \times Q \times H \times W}$
\STATE \hspace{1.5em} Macroscopic variables $\{\rho, \mathbf{u}, T, \ldots\}$
\STATE \hspace{1.5em} Previous state $\alpha_{\text{prev}}$ (optional warm-start)
\STATE \textbf{Output:} Stability parameter $\alpha \in \mathbb{R}^{B \times 1 \times H \times W}$
\STATE \textcolor{blue}{// Feature Encoding (Injection Pattern - Runs ONCE)}
\STATE Concatenate input features: $\phi \gets [\F, \Feqlattice, \rho, \mathbf{u}, \ldots]$
\STATE Encode features: $\mathbf{z} \gets \text{FeatureEncoder}(\phi)$ \hfill $\triangleright$ Eq.~\ref{eq:encoder}
\STATE \textcolor{blue}{// Physics Setup (Pre-computation)}
\STATE Compute lattice weights: $w_i \gets w_i(T)$ for $i = 1, \ldots, Q$
\STATE Compute reference entropy
\STATE Compute physical bound
\STATE \textcolor{blue}{// Initialize Alpha (Warm-start if available)}
\IF{$\alpha_{\text{prev}}$ available and batch sizes match}
    \STATE $\alpha^{(0)} \gets \text{clamp}(\alpha_{\text{prev}}, \alpha_{\min}, \alpha_{\max})$
\ELSE
    \STATE $\alpha^{(0)} \gets \alpha_{\text{init}} \cdot \mathbf{1}_{B \times 1 \times H \times W}$ \hfill $\triangleright$ Typically $\alpha_{\text{init}} = 2.0$
\ENDIF
\STATE \textcolor{blue}{// DEQ Fixed-Point Iteration}
\STATE Define update function $g_\theta(\alpha; \mathbf{z}, F, F_{\text{eq}}, H_{\text{pre}}, \alpha_{\max})$ \hfill 
\STATE Solve fixed point: $\alpha^* \gets \text{DEQSolver}(g_\theta, \alpha^{(0)})$ \hfill $\triangleright$ 
\STATE \textcolor{blue}{// Final Processing}
\STATE $\alpha \gets \text{clamp}(\alpha^*, \alpha_{\min}, \alpha_{\max})$
\STATE \textbf{Return} $\alpha$
\end{algorithmic}
\end{algorithm}

\subsection{Feature Encoder Architecture}
\label{subsec:encoder_arch}

The feature encoder computes the injection vector $\mathbf{z}$ that is reused across all DEQ iterations:

\begin{equation}
\label{eq:encoder}
\mathbf{z} = \text{FeatureEncoder}(\phi) = \phi_2(\text{GN}(\phi_1(\text{GN}(\phi))))
\end{equation}

where $\phi_1, \phi_2$ are $3 \times 3$ convolutional layers with group normalization (GN) and activation functions (ReLU or GELU). The encoder maps from input dimension $C_{\text{in}}$ to hidden dimension $d_h$.

\subsection[Update gtheta]{Update $g_\theta$}
\label{sec:network_architectures}

\subsubsection{Layer Configuration}

\begin{table}[ht]
\centering
\caption{Update Block layer specifications}
\label{tab:update_block_layers}
\begin{tabular}{lcccc}
\toprule
Layer & Input Dim & Output Dim & Kernel & Activation \\
\midrule
Input GN & $d_h + 4$ & $d_h + 4$ & -- & -- \\
Conv1 & $d_h + 4$ & $d_h$ & $3 \times 3$ & -- \\
GN1 & $d_h$ & $d_h$ & -- & -- \\
Act1 & $d_h$ & $d_h$ & -- & GELU/ReLU \\
\midrule
Conv2 & $d_h$ & $d_h$ & $3 \times 3$ & -- \\
GN2 & $d_h$ & $d_h$ & -- & -- \\
Act2 & $d_h$ & $d_h$ & -- & GELU/ReLU \\
$\vdots$ & $\vdots$ & $\vdots$ & $\vdots$ & $\vdots$ \\
\midrule
Conv$_{N-2}$ & $d_h$ & $d_h/2$ & $3 \times 3$ & -- \\
GN$_{N-2}$ & $d_h/2$ & $d_h/2$ & -- & -- \\
Act$_{N-2}$ & $d_h/2$ & $d_h/2$ & -- & GELU/ReLU \\
Conv$_{N-1}$ & $d_h/2$ & $1$ & $1 \times 1$ & -- \\
Tanh & $1$ & $1$ & -- & $\tanh$ \\
\bottomrule
\end{tabular}
\end{table}

All convolutional layers use padding=1 (except the final $1 \times 1$ conv) to maintain spatial dimensions.

\subsection{Detached Physics Hints}

Physics hints are computed with \texttt{torch.no\_grad()}. This design choice means:

\begin{enumerate}
    \item Gradients flow only through the neural network $\text{UpdateBlock}$
    \item Physics computations provide guidance but are not differentiable
    \item This prevents instabilities from entropy gradient flow
    \item Follows the DEQFlow pattern for stability
\end{enumerate}

\subsection{Model Hyperparameters}

\begin{table}[!htbp]
  \centering\small
  \caption{NeurDE architecture hyperparameters, shared by every case, grid, and seed. The equilibrium is the analytic entropic one; the network predicts the ELBM relaxation parameter $\alpha$ at each lattice site as the fixed point $y^\star$ of a deep equilibrium (DEQ) model. All convolutions use zero padding of width one.}
  \label{tab:app_neurde_hyperparameters}
  \begin{tabular}{@{}llp{0.45\textwidth}@{}}
    \toprule
    Component & Hyperparameter & Value \\
    \midrule
    Inputs     & Channels & 14: $f_i$ (9, D2Q9), $\lVert f^{\mathrm{neq}}\rVert_2$ (1), $\rho$ (1), $\mathbf{u}$ (2), $\omega$ (1) \\
               & Preprocessing & $\operatorname{sign}(x)\,\log(1+10^{5}\lvert x\rvert)$ \\
    \midrule
    Encoder    & Layers & 2 conv $3\times3$: $14\to64\to64$, ReLU \\
               & Evaluations & once per collision step \\
    \midrule
    Update block & Layers & 5 conv: $3\times3$ $68\to64$; $2\times$ $3\times3$ $64\to64$; $3\times3$ $64\to32$; $1\times1$ $32\to1$; ReLU between \\
               & Input  & 64 encoder + 1 $\tanh(y_k)$ + 3 physics hints \\
               & Damping $\lambda$ & $0.1$: $y_{k+1}=(1-\lambda)\,y_k+\lambda\,f_\theta$ \\
    \midrule
    Physics hints & Entropy residual $G$ & $\operatorname{sign}(G)\,\log(1+10^{10}\lvert G\rvert)$ \\
    (no gradient) & Newton direction & $\tanh(10^{3}\,\delta)$, $\delta$ = Newton step clipped to $[-1,1]$ \\
               & Instability & $\min\{\log(1+1/(\lvert H'\rvert+10^{-4})),\,5\}$ \\
    \midrule
    Output     & $\alpha$ map & $\alpha_0+w\tanh(y^\star)$, $w=\alpha_0-\alpha_{\min}$ if $y^\star<0$, else $\alpha_{\max}-\alpha_0$ \\
               & $(\alpha_{\min},\alpha_0,\alpha_{\max})$ & $(1,\,2,\,3)$ \\
    \midrule
    DEQ solver & Forward / backward & Broyden / Broyden, implicit differentiation \\
               & Max.\ iterations / tolerance & 18 / $10^{-3}$ (both passes) \\
               & Initial state $y_0$ & previous step's $\alpha$ if available, else $\alpha_0$ \\
    \midrule
    Initialization & Conv weights / biases & $\mathcal{N}(0,\,0.01^2)$ / $0$ \\
    \midrule
    Precision  & Stability pretraining & float32; bf16 autocast in encoder only \\
               & Solver training & float64, no autocast \\
    \midrule
    Size       & Trainable parameters & 176{,}641 (encoder 45{,}056; update block 131{,}585) \\
    \bottomrule
  \end{tabular}
\end{table}

\section{Experiments and Coarse Graining}\label{appx:experiments}
This appendix describes the implementation of the spatio-temporal
coarse-graining pipeline used to process the high-resolution shear-layer
reference described in Section~\ref{sec:experiments_main}.

\subsection{Problem Setup}
\label{app:problem_setup}
The two-dimensional incompressible Navier--Stokes equations are represented as
\begin{equation}
\label{eq:NS}
\begin{aligned}
\partial_t \mathbf{u} + (\mathbf{u}\cdot\nabla)\mathbf{u} &= -\nabla p + \nu \nabla^2 \mathbf{u}, \\
\nabla \cdot \mathbf{u} &= 0,
\end{aligned}
\end{equation}
where $\mathbf{u}=(u_x,u_y)$ is the velocity, $p$ is the pressure, and $\nu$ is the kinematic viscosity.

The LB method provides a mesoscopic discretization of these dynamics by
evolving particle distributions on a lattice. Although weakly compressible, LB
recovers the incompressible limit for low Mach numbers ($\mach\lesssim 0.1$).
Its local update rules eliminate the need for a global Poisson solver, and the
scheme is inherently parallelizable, making it a widely used tool for
high-performance fluid simulation and multiphysics
coupling~\citet{bouchut1999construction,bellotti2023truncation}. These features
make LB a natural platform for evaluating machine-learning closures in
challenging underresolved flows.

\paragraph{Numerical framework.}
We perform simulations on a two-dimensional lattice with a D2Q9 discrete
velocity set, $\velocity_i\in \{0, \pm 1\}^{\otimes 2}$ for
$i=1,\ldots,Q$, in the isothermal, weakly compressible regime. The discrete
entropic equilibrium $\Feqlattice$ admits a closed-form solution
\citet{hosseini2023entropic}, allowing us to isolate the stability parameter
$\alpha$ as the sole quantity requiring closure.

\begin{figure}[ht!]
\centering
\begin{tikzpicture}[scale=5]

  \definecolor{shearblue}{RGB}{70,130,180}
  \definecolor{pertred}{RGB}{220,20,60}
  \definecolor{velocitygreen}{RGB}{34,139,34}

  \draw[very thick, black] (0,0) rectangle (1,1);

  \draw[->, thick] (-0.08,0) -- (1.08,0) node[anchor=west] {$x$};
  \draw[->, thick] (0,-0.08) -- (0,1.08) node[anchor=south] {$y$};

  \foreach \x/\xlabel in {0/0, 0.25/0.25, 0.5/0.5, 0.75/0.75, 1/1} {
    \draw (\x,-0.01) -- (\x,-0.03);
    \node[anchor=north, font=\footnotesize] at (\x,-0.035) {$\xlabel$};
  }
  \foreach \y/\ylabel in {0/0, 0.25/0.25, 0.5/0.5, 0.75/0.75, 1/1} {
    \draw (-0.01,\y) -- (-0.03,\y);
    \node[anchor=east, font=\footnotesize] at (-0.035,\y) {$\ylabel$};
  }

  \fill[shearblue!15] (0,0.15) rectangle (1,0.35);
  \fill[shearblue!15] (0,0.65) rectangle (1,0.85);

  \draw[shearblue, very thick] (0,0.25) -- (1,0.25);
  \draw[shearblue, very thick] (0,0.75) -- (1,0.75);

  \foreach \x in {0.1, 0.3, 0.5, 0.7, 0.9} {
    \draw[->, velocitygreen, thick] (\x, 0.28) -- ++(0.06, 0);
    \draw[->, velocitygreen, thick] (\x, 0.22) -- ++(-0.06, 0);
    \draw[->, velocitygreen, thick] (\x, 0.78) -- ++(-0.06, 0);
    \draw[->, velocitygreen, thick] (\x, 0.72) -- ++(0.06, 0);
  }

  \draw[pertred, very thick, domain=0:1, samples=100, smooth]
    plot (\x, {0.25 + 0.03*sin(720*(\x+0.25))});  
  \draw[pertred, very thick, domain=0:1, samples=100, smooth]
    plot (\x, {0.75 + 0.03*sin(720*(\x+0.25))});

  \draw[->, thick, dashed, gray] (-0.05, 0.5) arc (180:270:0.05);
  \draw[->, thick, dashed, gray] (1.05, 0.5) arc (0:-90:0.05);
  \node[gray, font=\small] at (0.5, -0.16) {periodic in $x$};

  \draw[->, thick, dashed, gray] (0.5, -0.05) arc (270:360:0.05);
  \draw[->, thick, dashed, gray] (0.5, 1.05) arc (90:180:0.05);
  \node[gray, font=\small, rotate=90] at (-0.2, 0.5) {periodic in $y$};

  \node[anchor=west, align=left, font=\footnotesize,
        draw=black!30, fill=white, rounded corners=2pt,
        inner sep=4pt] at (1.05, 0.85) {
    $\Delta x = 1/512$ \\
    $\mach = 0.1$ \\
    $\lambda = 80$ \\
    $\delta = 0.05$ \\
    $\mathrm{Re} = 50{,}000$
  };

  \node[anchor=west, align=left, font=\footnotesize] at (1.05, 0.4) {
    \textcolor{shearblue}{\rule{0.3cm}{0.8pt}} Shear layers \\
    \textcolor{pertred}{\rule{0.3cm}{0.8pt}} $u_y$ perturbation \\
    \textcolor{velocitygreen}{$\rightarrow$} $u_x$ velocity
  };

  \node[anchor=center, font=\scriptsize, align=center] at (0.5, 0.125) {
    $u_x = V_0 \tanh[\lambda(y-0.25)]$
  };
  \node[anchor=center, font=\scriptsize, align=center] at (0.5, 0.875) {
    $u_x = V_0 \tanh[\lambda(0.75-y)]$
  };
  \node[anchor=center, font=\scriptsize, pertred] at (0.5, 0.55) {
    $u_y = \delta V_0 \sin[2\pi(x+0.25)]$
  };

\end{tikzpicture}
\caption{Computational setup for Kelvin--Helmholtz instability simulation. The domain $[0,1]^2$ contains two counter-rotating shear layers at $y = 0.25$ and $y = 0.75$ with hyperbolic tangent velocity profiles (blue regions). A sinusoidal perturbation in the cross-stream velocity $u_y$ (red curves) seeds the instability. Green arrows indicate the streamwise velocity direction in each layer. Periodic boundary conditions are applied in both directions (gray dashed arrows).}
\label{fig:doubly_periodic}
\end{figure}
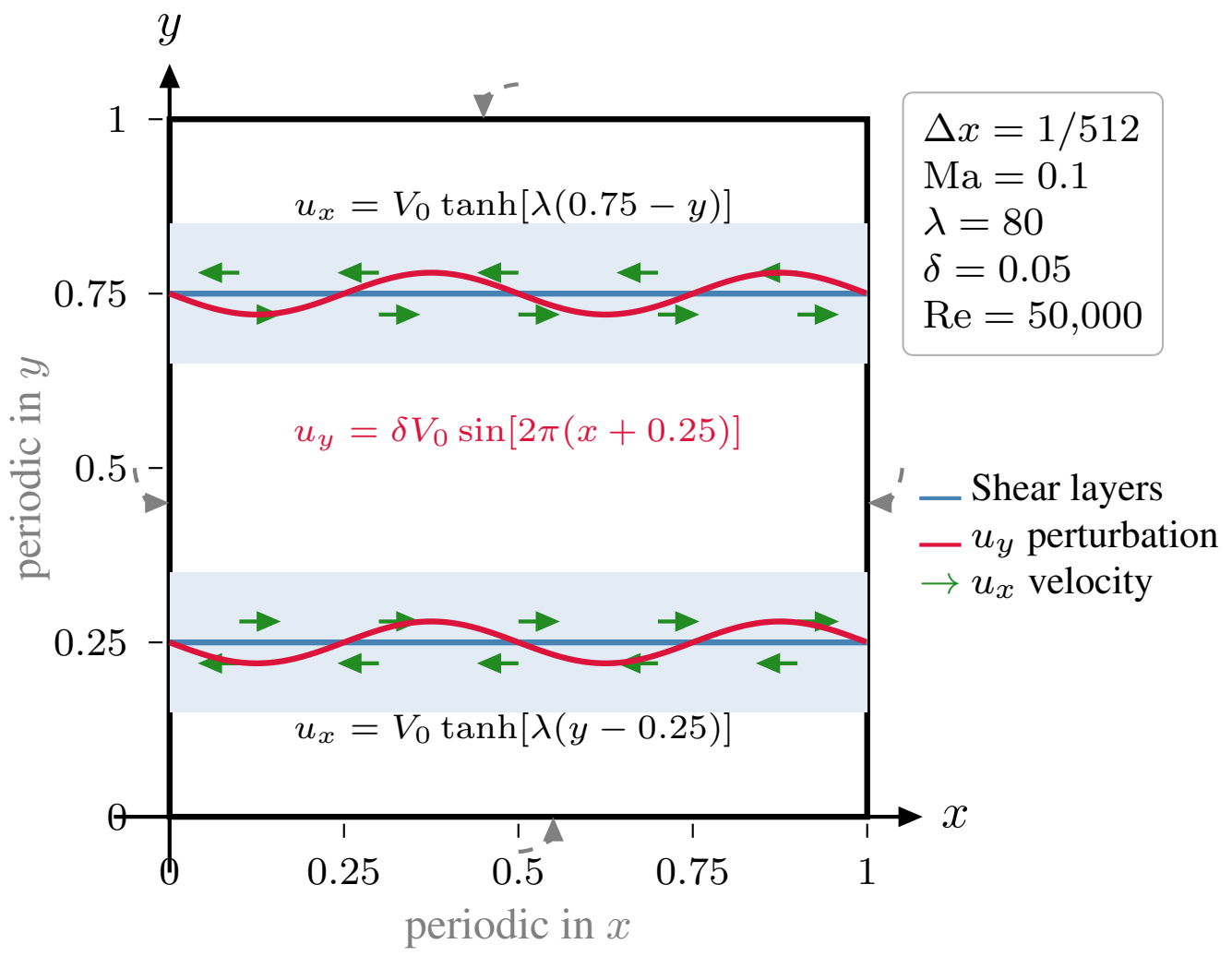

\paragraph{Computational setup and initialization.}
The domain is $(x,y)\in[0,1]\times[0,1]$, discretized with $\Delta x = 1/512$ and periodic boundary conditions in both directions. The streamwise velocity is prescribed piecewise to form two shear layers centered at $y=0.25$ and $y=0.75$:
\begin{equation}
u_x(x,y) =
\begin{cases}
V_0 \tanh\!\bigl(\lambda(y-0.25)\bigr), & y < 0.5, \\[6pt]
V_0 \tanh\!\bigl(\lambda(0.75-y)\bigr), & y \geq 0.5,
\end{cases}
\end{equation}
while the cross-stream velocity introduces a small sinusoidal perturbation,
\begin{equation}
u_y(x,y) = \delta V_0 \sin\!\bigl(2\pi(x+0.25)\bigr),
\end{equation}
which seeds the Kelvin--Helmholtz instability. The parameters are fixed as $\rho_0 = 1$ (reference density), $V_0 = \mach\,c_s$ with $\mach=0.1$ and $c_s=1/\sqrt{3}$ (reference velocity), $\lambda = 80$ (shear-layer sharpness), and $\delta = 0.05$ (perturbation amplitude). The Reynolds number is set to $\mathrm{Re}=50{,}000$, and the LB populations are initialized from the standard equilibrium distribution $\Feqlattice(\rho,\mathbf{u})$.

\paragraph{Dataset.}
The reference trajectory is generated on a $512 \times 512$ grid with
nondimensional time step
\begin{equation}
\Delta t_{\text{ref}} = \frac{c_s \, \mach}{n_x} \approx 1 \times 10^{-4}.
\label{eq:ref_timestep}
\end{equation}

We refer to this high-resolution trajectory as the reference data in the
diagnostics below. It is the resolved LB reference used for filtering and
comparison, not an exact spectral solution of the continuum equations.

\subsection{Coarse-Graining for Stability Analysis}
\label{app:coarse_graining_method}

To study stability under resolution loss, we coarse-grain the reference data to
target grids where closure effects dominate. Operating at coarser resolutions
(e.g., $128 \times 128$ or $64 \times 64$) serves three purposes:

\begin{enumerate}
    \item analyzing stability in the underresolved regime where closure effects
    dominate,
    \item enabling efficient neural-network training with larger batch sizes,
    \item reducing temporal redundancy while preserving essential advective
    dynamics.
\end{enumerate}

\begingroup
The coarse-graining process applies a box filter in both space and time before decimation, so the coarse fields are local space-time averages of the resolved trajectory rather than point samples.

\subsection{Space-Time Box Filter for Coarse-Graining}
\label{app:box_filter}

To generate data at coarser resolutions suitable for stability analysis, we
apply a space-time box filter to the reference fields and then decimate. For a field
$u_i(\mathbf{x},t)$, the filtered velocity field
$\bar{\mathbf{u}}$ at spatial resolution $\Delta x^{\text{out}}$ and temporal
spacing $\Delta t^{\text{out}}$ is obtained using a product filter

\begin{equation}
\bar{u}_i(\mathbf{x}, t) =
\int_{\mathbb{R}}
\int_{\Omega}
G_x(\mathbf{x} - \mathbf{x}'; \Delta_f)
G_t(t-t'; \Delta T_f)
u_i(\mathbf{x}', t') \, d\mathbf{x}' \, dt' .
\label{eq:spatial_filter}
\end{equation}

Here $G_x$ is the spatial box kernel and $G_t$ is the temporal box kernel. The
implementation uses the discrete version of this product filter, described
below.

\subsubsection{Mathematical Definition}

Our implementation employs the discrete box filter (also known as a top-hat filter or uniform filter) as the filtering operation in Eq.~\eqref{eq:spatial_filter}.

The continuous spatial box filter in $d$ dimensions is defined as:

\begin{equation}
G(\mathbf{x}; \Delta_f) = \prod_{i=1}^{d} G_i(x_i; \Delta_f) = \prod_{i=1}^{d} \begin{cases}
\frac{1}{\Delta_f}, & |x_i| \leq \Delta_f/2, \\
0, & \text{otherwise},
\end{cases}
\label{eq:box_filter_kernel}
\end{equation}

where $\Delta_f$ is the filter width. The temporal box filter
has the same top-hat form on an interval of width
$\Delta T_f=q\,\Delta t^{\text{in}}$. For a scalar field $u(\mathbf{x}, t)$
discretized on a uniform grid with spacing $\Delta x^{\text{in}}$ and time step
$\Delta t^{\text{in}}$, the filtered field at coarse resolution
$\Delta x^{\text{out}} = r \Delta x^{\text{in}}$ and
$\Delta t^{\text{out}} = q \Delta t^{\text{in}}$ is:

\begin{equation}
\bar{u}(\mathbf{X}, t_n)
=
\frac{1}{q\,r^d}
\sum_{m=0}^{q-1}
\sum_{\mathbf{x} \in \mathcal{N}(\mathbf{X})}
u(\mathbf{x}, t_{nq+m}) .
\label{eq:box_filter_discrete}
\end{equation}

where $\mathcal{N}(\mathbf{X})$ denotes the $r \times r \times \cdots$ neighborhood of fine-grid cells that correspond to coarse-grid cell $\mathbf{X}$. The spatial filter width is $\Delta_f = r \Delta x^{\text{in}}$, which coincides with the coarse grid spacing $\Delta x^{\text{out}}$. The temporal factor $q$ is chosen consistently with the spatial downsampling factor. Thus, the $512^2$, 20,000-frame reference gives $256^2$, $128^2$, and $64^2$ descendants with 10,000, 5,000, and 2,500 frames. The $64^2$ comparison stage used in Appendix~\ref{appx:diagnostics_field_fidelity_64} is the $256^2$, 10,000-frame space-time box-filtered reference filtered again with $r=q=4$.

\subsubsection{Implementation via Average Pooling}

In our pipeline, the discrete box filter is implemented using average pooling operations. For each spatial axis $i$, the kernel size is computed as:

\begin{equation}
k_i = \left\lceil \frac{s_i^{\text{in}}}{s_i^{\text{out}}} \right\rceil
\label{eq:kernel_size}
\end{equation}

where $s_i^{\text{in}}$ and $s_i^{\text{out}}$ are the input and output resolutions. For each output frame, the code first averages over a contiguous temporal window of length $q$ and then applies the spatial block average. The filtering operation is then:

\begin{equation}
\mathbf{X}_{\text{filtered}}^{(n)}
=
\operatorname{AvgPool}_{\mathbf{x}}
\left(
\frac{1}{q}\sum_{m=0}^{q-1}
\mathbf{X}^{(nq+m)};
k_1,k_2,\ldots
\right)
\label{eq:avgpool_operation}
\end{equation}

This operation is performed on GPU for efficiency and is applied independently to each channel (velocity component, vorticity, etc.). One-dimensional time-like arrays such as $t$ and $t_{\mathrm{index}}$ are reduced with the same temporal mean.
\endgroup

The coarse-graining pipeline consists of four main stages executed sequentially:

\begin{enumerate}
    \item \textbf{Metadata Extraction}: Preliminary scan of input HDF5 datasets to classify data types and extract dimensional information.
    \item \textbf{Preprocessing}: Dataset consolidation and validation, creating temporary merged datasets suitable for parallel processing.
    \item \textbf{Parallel Processing}: Distributed spatio-temporal downsampling across multiple workers using configurable backends (SLURM or multiprocessing).
    \item \textbf{Virtual Dataset Assembly}: Efficient merging of processed shards using HDF5 virtual datasets to avoid redundant data copies.
\end{enumerate}

\section{Spectral Finetuning}\label{appx:spectral_finetuning}
%
%
%
%

\emph{NeurDE shell}, the third method in the ablation of Section~\ref{sec:experiments_main}, is an \emph{entropically-stabilized NeurDE} that has been fine-tuned with an additional spectral objective.
We do not use this fine-tuning for our final results, because it can be challenging to apply.
Improving the agreement with the reference at high wavenumbers can also bring in high-frequency noise and aliasing, the kind of grid-scale content that destabilizes the underresolved ELBM rollouts in Section~\ref{sec:experiments_main}.
A spectral objective must therefore sharpen small-scale features without destroying the large-scale features and the stability that the entropic closure provides.
Ongoing research is looking into metrics and loss functions that strike this balance.
Still, \emph{NeurDE shell} is shown to improve upon \emph{entropically-stabilized NeurDE} for DPSL-50K while maintaining long-time emulation stability.
This appendix describes the fine-tuning procedure and reports the ablation against \emph{entropically-stabilized NeurDE} and ELBM on DPSL-50K for the $64^2$, $32^2$, and $16^2$ grids.

\subsection{Fine-tuning Procedure}
\label{app:fine-tune_procedure}

The rollout loss of Eq.~\eqref{eq:losses} compares fields in real space.
Such losses are dominated by the large, energetic structures and are insensitive to the high-wavenumber tail of the spectrum, which is where coarse rollouts lose small-scale energy or inject grid-scale artifacts.
Fine-tuning adds a term that compares the highest-wavenumber shells of the rollout's kinetic-energy spectrum to those of the box-filtered reference, so that the closure is also trained to carry the correct amount of energy at the smallest resolved scales.

\paragraph{Starting point.}
Fine-tuning starts from the trained \emph{entropically-stabilized NeurDE} checkpoint of the same grid (Appendix~\ref{app:Training}).
The architecture, the lattice Boltzmann state variables, and the rollout integrator are unchanged; only the training objective changes.

\paragraph{Spectral-shell loss.}
At each saved rollout state, we Fourier transform the velocity field and bin the kinetic energy into integer shells,
\begin{equation}
    E_n(t)
    =
    \sum_{\operatorname{round}(|\mathbf{m}|)=n}
    \frac{1}{2V^2}
    \left|\widehat{u}_i(\mathbf{m},t)\right|^2 ,
\end{equation}
where $V$ is the number of grid points, following the convention of Appendix~\ref{appx:diagnostics_spectrum}.
The spectral-shell loss penalizes the mismatch between the rollout and the box-filtered reference on the top quarter of the shells,
\begin{equation}\label{eq:spec_loss}
    \mathcal{L}_{\mathrm{spec}}(t)
    =
    \frac{1}{|\mathcal{H}|}
    \sum_{n\in\mathcal{H}}
    \left(\ln E_n(t)-\ln E_n^{\mathrm{ref}}(t)\right)^2 ,
    \qquad
    \mathcal{H}
    =
    \{n : 0.75\,n_{\max} \le n \le n_{\max}\},
\end{equation}
where $n_{\max}=N/2$ is the Nyquist shell of the $N\times N$ grid, so the loss acts on the shells $3N/8\le n\le N/2$.
This lower bound is higher than the $N/4$ cutoff of the high-$k$ energy diagnostic (Appendix~\ref{appx:turbulence_diagnostics}), so the loss acts on a narrower band than the diagnostic measures.
Because the shell sums discard the Fourier phases, this term only constrains how much energy each high-wavenumber shell carries; the real-space terms of Eq.~\eqref{eq:losses} keep the fields aligned with the reference.
The fine-tuning objective is
\begin{equation}\label{eq:fine-tune_loss}
    \mathcal{L}_{\mathrm{ft}}(\theta)
    =
    \mathcal{L}_{\mathrm{rollout}}(\theta)
    + c_{\mathrm{spec}} \sum_{t=1}^{T} \mathcal{L}_{\mathrm{spec}}(t) .
\end{equation}
We say ``spectral-shell'' rather than ``spectral norm'' because the penalty acts on shell-averaged flow spectra, not on the operator norm of a network layer.

\paragraph{Training.}
Fine-tuning uses the same rollout training, horizon curriculum, and IPTA gradient subset as the base model, with Eq.~\eqref{eq:fine-tune_loss} in place of Eq.~\eqref{eq:losses} in Algorithm~\ref{Alg:training}.
The regularizer is used only during training.
At test time there is no spectral filter, projection, or correction; \emph{NeurDE shell} is advanced with the same local update as \emph{entropically-stabilized NeurDE}.

\paragraph{Metrics.}
The field-fidelity and spectrum metrics reported below are those of Appendix~\ref{appx:turbulence_diagnostics}: the relative velocity error, the pointwise vorticity correlation, the relative vorticity error, the active-shell log-spectrum RMSE, and the ratio of high-$k$ energy fractions.
All metrics are evaluated at the same saved rollout time as the corresponding filtered reference.
In generated figure labels and axes, ``DNS'' denotes the box-filtered high-resolution reference rather than an exact spectral solution of the continuum equations.
Table~\ref{tab:fine-tune_ablation} collects the final metrics on all three grids.

\begin{table}[t]
    \centering
    \small
    \caption{\textbf{Final metrics of the fine-tuning ablation.}
    Metrics at the final saved horizon of the $64^2$ ($t_{\mathrm{index}}=2500$), $32^2$ ($t_{\mathrm{index}}=1249$), and $16^2$ ($t_{\mathrm{index}}=624$) rollouts, against the box-filtered reference.
    The first three columns measure field fidelity; the last two diagnose spectral behavior.
    Bold marks the best value within each grid block. For High-$k$/ref., best means closest to one, not smallest.}
    \label{tab:fine-tune_ablation}
    \setlength{\tabcolsep}{4pt}
    \begin{tabular}{@{}llccccc@{}}
        \toprule
        & & \multicolumn{3}{c}{Field fidelity} & \multicolumn{2}{c}{Spectrum diagnostics} \\
        \cmidrule(lr){3-5}\cmidrule(l){6-7}
        Grid & Method & Vel. err $\downarrow$ & $\omega$ corr. $\uparrow$ &
        $\omega$ err $\downarrow$ & Spec. RMSE $\downarrow$ & High-$k$/ref. $\to1$ \\
        \midrule
        \multirow{3}{*}{$64^2$}
               & NeurDE       & $0.0806$          & $0.9387$          & $2.383$          & $3.794$          & $0.0077$ \\
               & NeurDE shell & $\mathbf{0.0462}$ & $\mathbf{0.9614}$ & $\mathbf{0.275}$ & $0.540$          & $0.422$ \\
               & ELBM         & $0.0544$          & $0.9374$          & $0.355$          & $\mathbf{0.302}$ & $\mathbf{1.415}$ \\
        \midrule
        \multirow{3}{*}{$32^2$}
               & NeurDE       & $0.0686$          & $\mathbf{0.9551}$ & $\mathbf{0.2977}$ & $\mathbf{0.2593}$ & $0.6226$ \\
               & NeurDE shell & $\mathbf{0.0676}$ & $0.9534$          & $0.3021$          & $0.2613$          & $0.7478$ \\
               & ELBM         & $0.0694$          & $0.9456$          & $0.3268$          & $0.3915$          & $\mathbf{1.1259}$ \\
        \midrule
        \multirow{3}{*}{$16^2$}
               & NeurDE       & $\mathbf{0.205}$  & $\mathbf{0.910}$  & $0.425$          & $0.631$          & $0.301$ \\
               & NeurDE shell & $0.207$           & $0.909$           & $\mathbf{0.422}$ & $0.583$          & $0.365$ \\
               & ELBM         & $0.221$           & $0.897$           & $0.442$          & $\mathbf{0.519}$ & $\mathbf{0.718}$ \\
        \bottomrule
    \end{tabular}
\end{table}

\subsection[Ablation on the 64x64 Rollout]{Ablation on the $64\times64$ Rollout}
\label{appx:diagnostics_field_fidelity_64}

This ablation isolates the effect of fine-tuning at $64^2$.
We compare \emph{entropically-stabilized NeurDE}, \emph{NeurDE shell}, and under-resolved ELBM against the box-filtered high-resolution reference.
The reference target is the $256^2$, 10,000-frame stage after space-time box filtering to the $64^2$, 2,500-frame comparison grid.

\begin{figure*}[t]
    \centering
    \includegraphics[width=\textwidth]{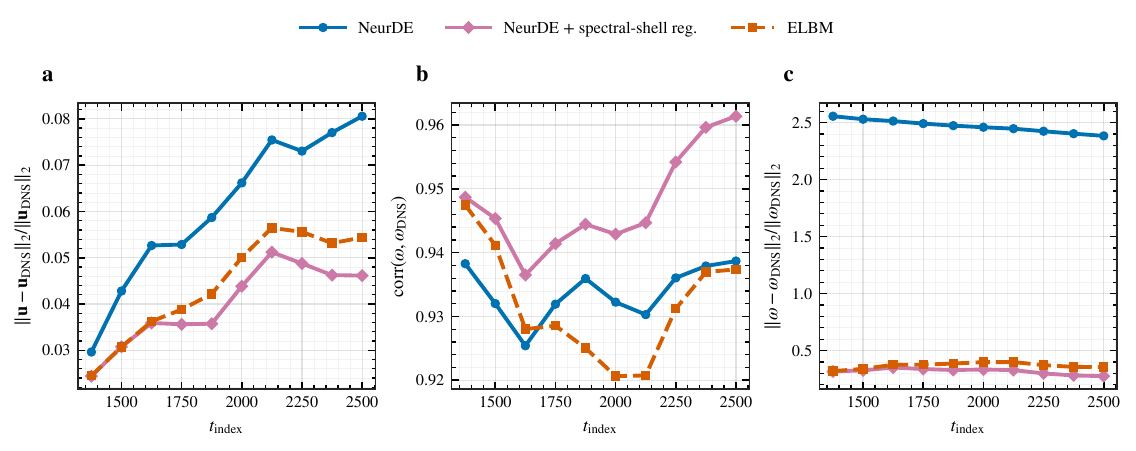}
    \caption{\textbf{Core field-level fidelity for $64^2$ forecasts.}
    \emph{Entropically-stabilized NeurDE}, \emph{NeurDE shell}, and ELBM are
    compared against the box-filtered high-resolution reference. (a) Relative velocity
    error, (b) vorticity correlation, and (c) relative vorticity error over
    rollout time. Fine-tuning substantially reduces the
    velocity and vorticity errors of the base model and increases
    vorticity correlation over the later rollout.}
    \label{fig:field_fidelity_64}
\end{figure*}

Figure~\ref{fig:field_fidelity_64} shows the three field-fidelity metrics over the rollout.
Fine-tuning gives the best field-level agreement in this comparison (Table~\ref{tab:fine-tune_ablation}), whereas ELBM gives the lowest final spectrum log-RMSE while carrying excess high-wavenumber energy relative to the reference.
The pointwise vorticity correlation measures phase and coherent-structure fidelity rather than only global statistics.
After the early transient, \emph{NeurDE shell} remains better aligned with the box-filtered reference, while the base model and ELBM decorrelate.

\begin{figure*}[t]
    \centering
    \includegraphics[width=\textwidth]{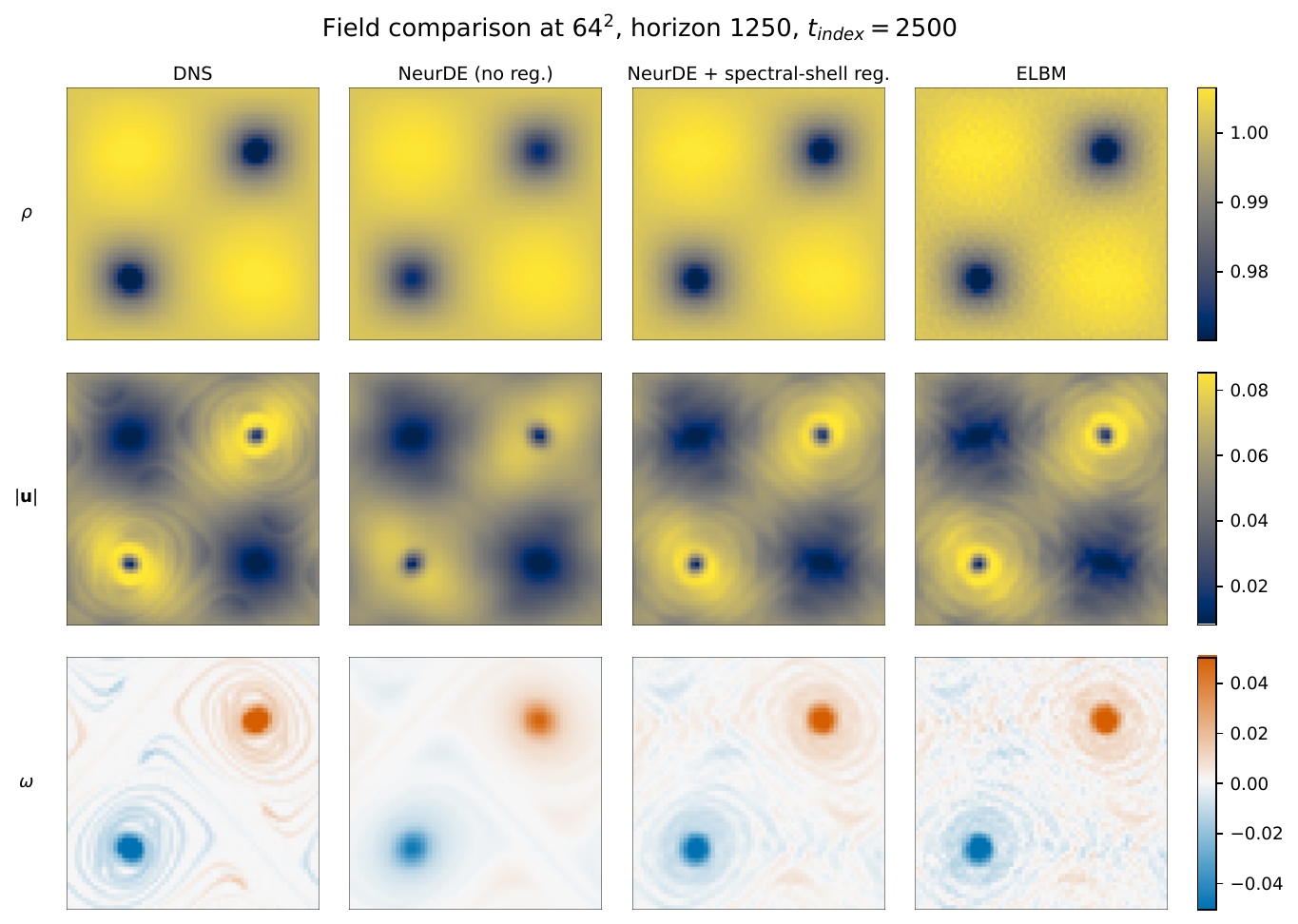}
    \caption{\textbf{Field snapshots for the $64^2$ rollout at
    $t_{\mathrm{index}}=2500$.} Columns compare the box-filtered reference,
    \emph{entropically-stabilized NeurDE}, \emph{NeurDE shell}, and ELBM;
    rows show density $\rho$, speed $|\mathbf{u}|$, and vorticity $\omega$.
    The base \emph{entropically-stabilized NeurDE} forecast is visibly over-smoothed in vorticity,
    while fine-tuning restores sharper vortex structure and
    better phase alignment with the reference. ELBM preserves the main vortices but shows
    excess small-scale texture in the fields.}
    \label{fig:fields_rho_speed_vorticity_64}
\end{figure*}

The field snapshots in Fig.~\ref{fig:fields_rho_speed_vorticity_64} explain the correlation improvement visually.
\emph{NeurDE shell} better preserves the signed vortex cores and surrounding filaments at the final saved horizon, whereas the base model over-smooths the vorticity field.

\begin{figure*}[t]
    \centering
    \includegraphics[width=\textwidth]{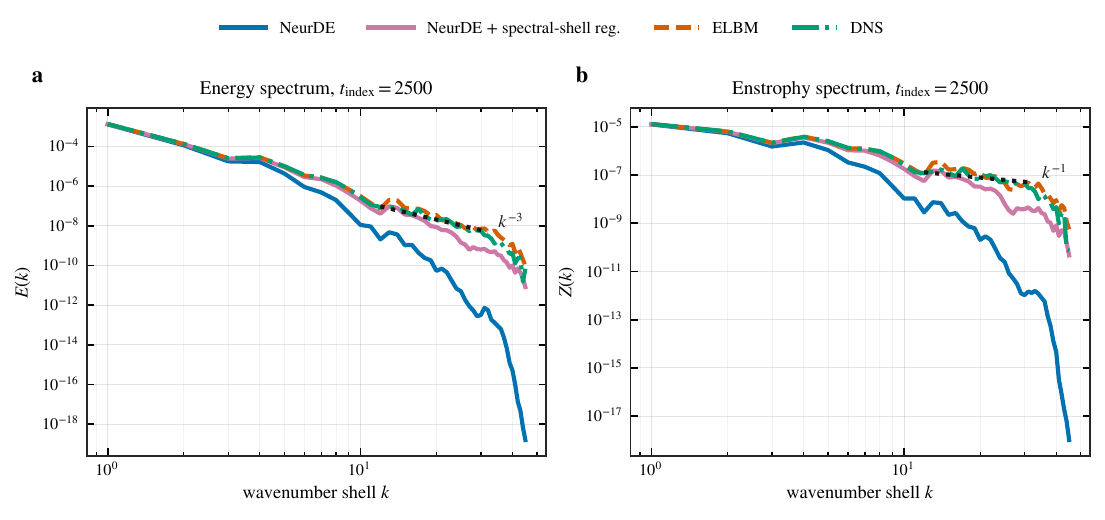}
    \caption{\textbf{Energy and enstrophy spectra at
    $t_{\mathrm{index}}=2500$.} The base \emph{entropically-stabilized NeurDE} forecast loses
    high-wavenumber energy and enstrophy, while \emph{NeurDE shell}
    restores much of the missing spectral content. ELBM remains strongest
    on the final energy-spectrum log-RMSE, so the improvement should be read
    as improved field-level fidelity and reduced spectral collapse rather than
    better spectral accuracy on every metric.}
    \label{fig:spectral_shell_power_spectra_64}
\end{figure*}

Figure~\ref{fig:spectral_shell_power_spectra_64} shows the shell-averaged energy spectrum of Appendix~\ref{appx:diagnostics_spectrum} together with the analogous enstrophy spectrum,
\begin{equation}
    Z_n
    =
    \sum_{\operatorname{round}(|\mathbf{m}|)=n}
    \frac{1}{2V^2}
    \left|\widehat{\omega}(\mathbf{m})\right|^2 .
\end{equation}
At $t_{\mathrm{index}}=2500$, the energy-spectrum log-RMSE is $3.79$ for the base model, $0.540$ for \emph{NeurDE shell}, and $0.302$ for ELBM.
Fine-tuning therefore greatly reduces the spectral collapse of the base model, while ELBM remains closest to the reference by this final spectral RMSE.

Two-point correlations measure spatial structure. With
$\mathbf{u}'=\mathbf{u}-\langle\mathbf{u}\rangle$ and
$\omega'=\omega-\langle\omega\rangle$, we use
\begin{equation}
    C_{uu}(r)
    =
    \frac{
        \left\langle
            \mathbf{u}'(\mathbf{x})\cdot
            \mathbf{u}'(\mathbf{x}+\mathbf{r})
        \right\rangle
    }{
        \left\langle |\mathbf{u}'(\mathbf{x})|^2 \right\rangle
    },
    \qquad
    C_{\omega\omega}(r)
    =
    \frac{
        \left\langle
            \omega'(\mathbf{x})
            \omega'(\mathbf{x}+\mathbf{r})
        \right\rangle
    }{
        \left\langle |\omega'(\mathbf{x})|^2 \right\rangle
    }.
\end{equation}

\begin{figure*}[t]
    \centering
    \includegraphics[width=\textwidth]{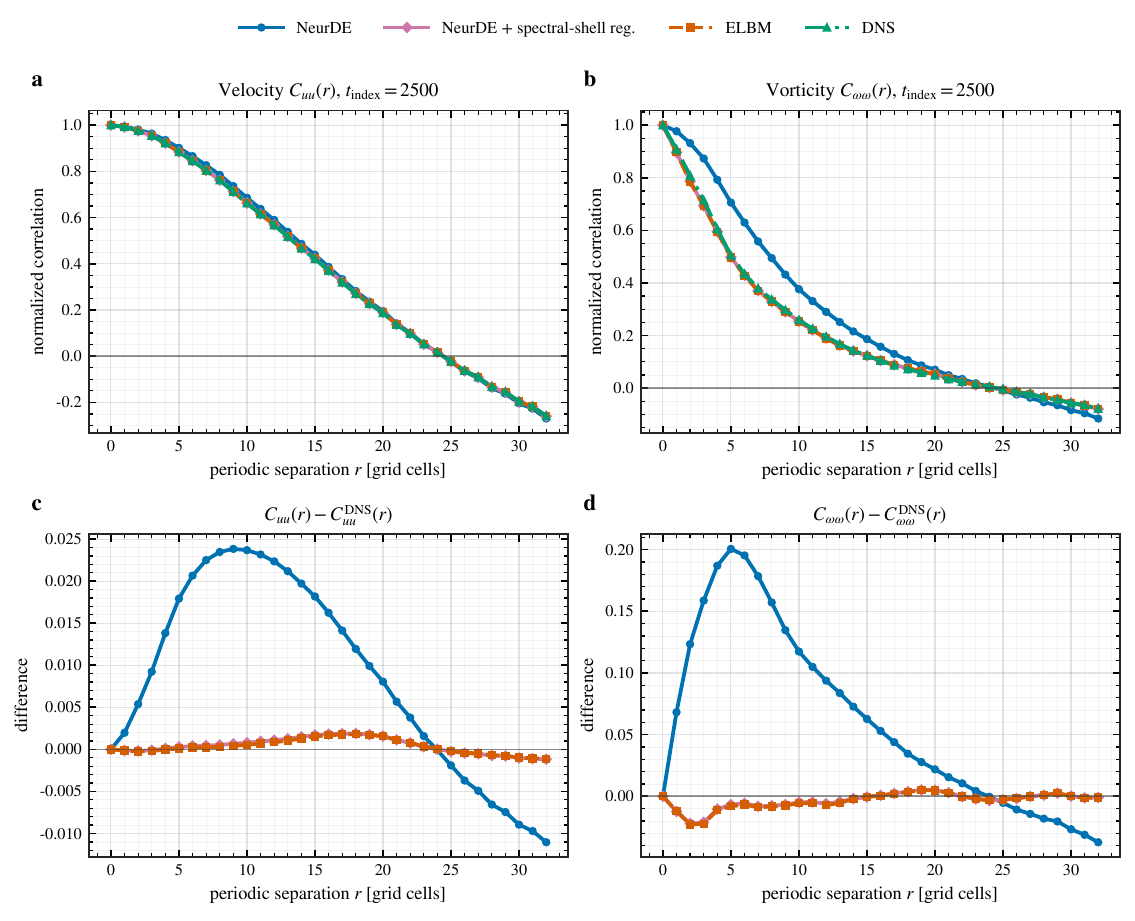}
    \caption{\textbf{Two-point correlations at $t_{\mathrm{index}}=2500$.}
    Velocity correlations are similar across methods, but vorticity
    correlations expose the main failure of the base model: it
    preserves overly smooth long-range structure and does not reproduce the
    reference-like small-scale decorrelation. \emph{NeurDE shell} and
    ELBM track the reference vorticity correlation much more closely.}
    \label{fig:spectral_shell_two_point_correlations_64}
\end{figure*}

The remaining panels (Figs.~\ref{fig:spectral_shell_all_fidelity_64}--\ref{fig:spectral_shell_two_point_error_evolution_64}) give the time-resolved fidelity metrics, flow statistics, artifact diagnostics, and two-point error evolution for the same rollouts.

\begin{figure*}[p]
    \centering
    \includegraphics[width=\textwidth]{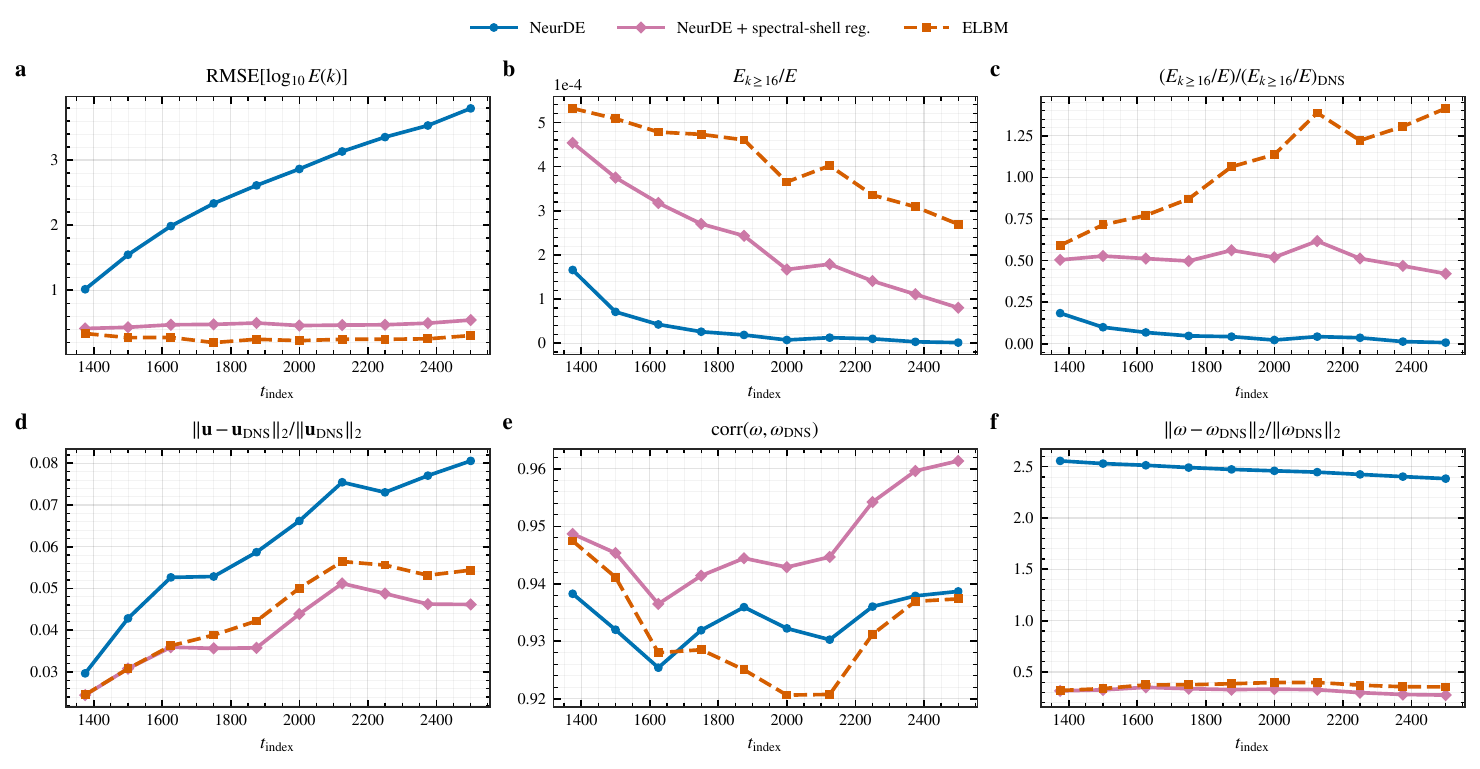}
    \caption{\textbf{Supplementary fidelity metrics for the $64^2$
    fine-tuning ablation.} This figure reports the full collection of
    time-resolved fidelity metrics used to compare \emph{entropically-stabilized NeurDE},
    \emph{NeurDE shell}, and ELBM against the filtered reference.}
    \label{fig:spectral_shell_all_fidelity_64}
\end{figure*}

\begin{figure*}[p]
    \centering
    \includegraphics[width=\textwidth]{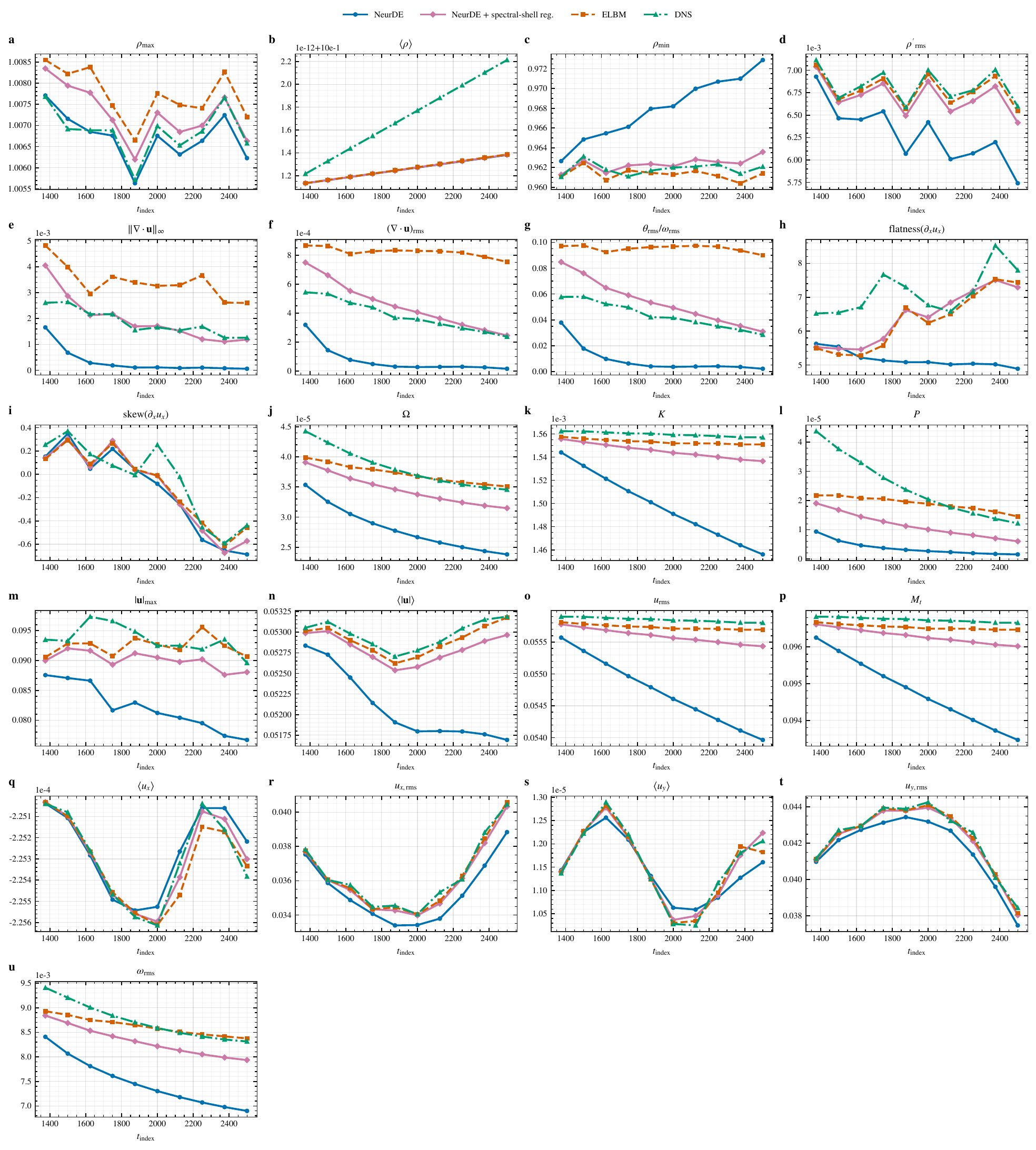}
    \caption{\textbf{Supplementary flow statistics for the $64^2$
    fine-tuning ablation.} The statistics summarize density, kinetic,
    vorticity, divergence, and derivative-shape diagnostics for the reference and each
    rollout.}
    \label{fig:spectral_shell_all_statistics_64}
\end{figure*}

\begin{figure*}[p]
    \centering
    \includegraphics[width=\textwidth]{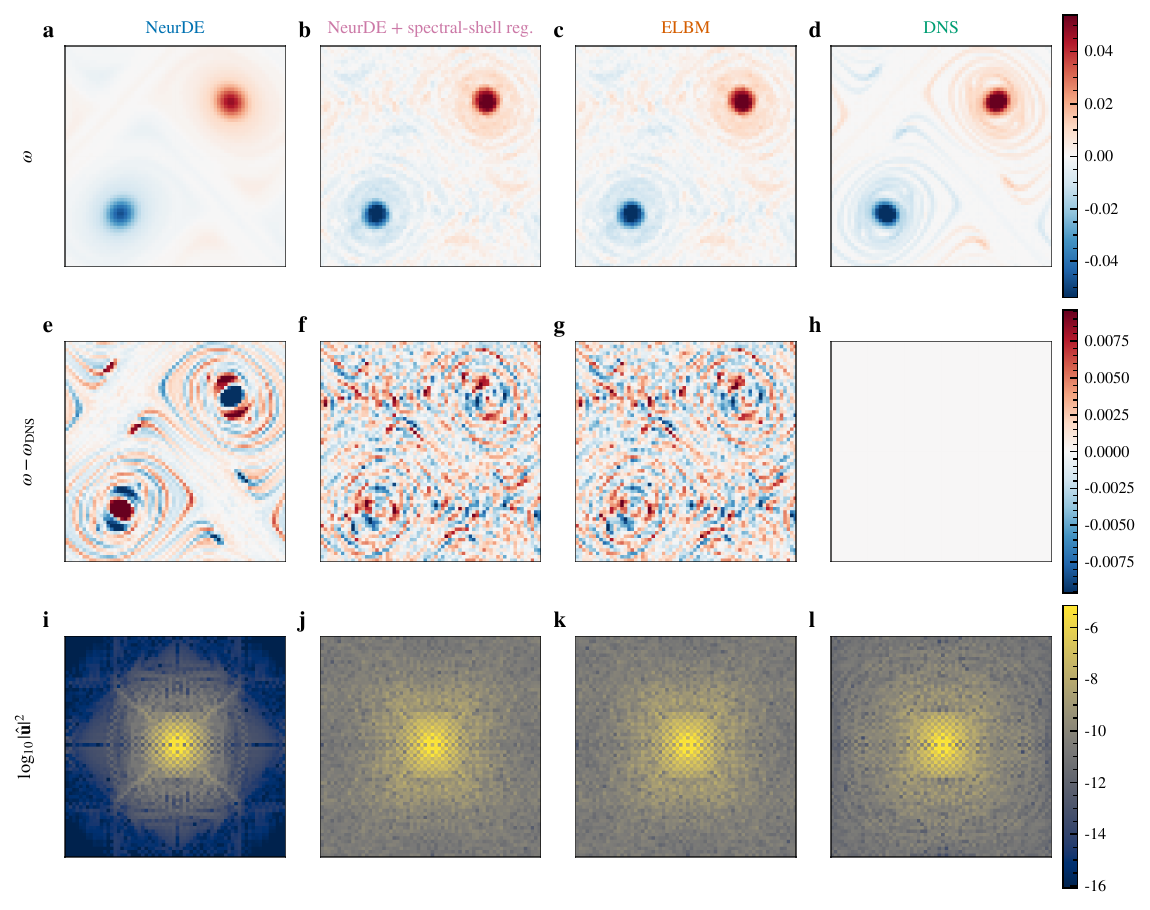}
    \caption{\textbf{Supplementary artifact diagnostics for the $64^2$
    fine-tuning ablation.} Fourier-space and artifact-oriented diagnostics
    expose how the base forecast loses high-wavenumber content and how
    fine-tuning recovers a larger fraction of the resolved
    reference spectral structure.}
    \label{fig:spectral_shell_artifact_diagnostics_64}
\end{figure*}

\begin{figure*}[p]
    \centering
    \includegraphics[width=\textwidth]{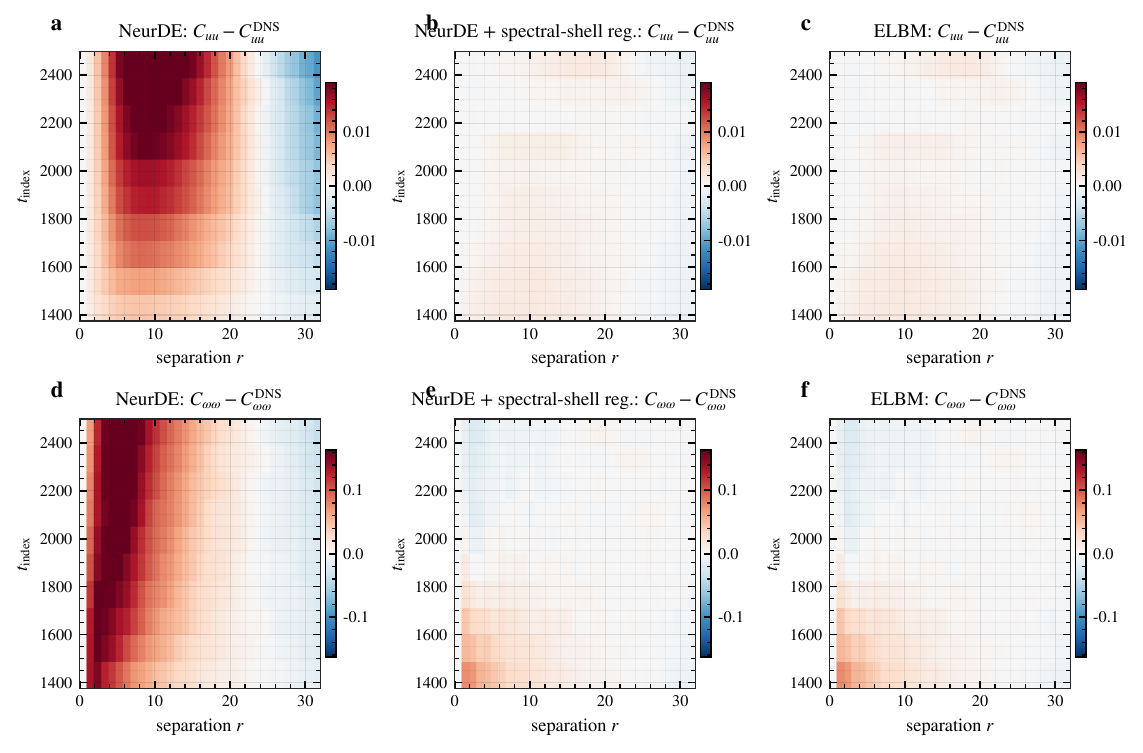}
    \caption{\textbf{Supplementary two-point error evolution for the $64^2$
    fine-tuning ablation.} The curves track correlation-error evolution over
    rollout time and separate velocity-correlation errors from the more
    discriminating vorticity-correlation errors.}
    \label{fig:spectral_shell_two_point_error_evolution_64}
\end{figure*}

\FloatBarrier

\subsection[Ablation on the 32x32 and 16x16 Rollouts]{Ablation on the $32\times32$ and $16\times16$ Rollouts}
\label{appx:fine-tune_coarse}

On the coarser grids, fine-tuning has a much smaller effect than at $64^2$ (Table~\ref{tab:fine-tune_ablation}).
The base model does not show the spectral collapse seen at $64^2$: its spectrum log-RMSE is $0.259$ at $32^2$ and $0.631$ at $16^2$, compared with $3.794$ at $64^2$.
The spectral-shell loss therefore has less missing high-wavenumber energy to restore.
Fine-tuning raises the high-$k$ energy fraction from $0.623$ to $0.748$ of the reference value at $32^2$ and from $0.301$ to $0.365$ at $16^2$, and it lowers the spectrum log-RMSE at $16^2$ from $0.631$ to $0.583$.
The field metrics of the two \emph{NeurDE} variants differ by at most $1.5\%$ on both grids, and neither variant is better on all of them.
Both learned closures have lower velocity error, lower vorticity error, and higher vorticity correlation than ELBM on both grids.
ELBM has the lowest spectrum log-RMSE at $16^2$ and carries excess high-$k$ energy at $32^2$; at $16^2$ all three methods stay below the reference high-$k$ energy.

\subsection{Summary}
\label{app:fine-tune_summary}

Across the three grids, fine-tuning changes the field metrics little except at $64^2$, where the base model collapses spectrally and fine-tuning removes the collapse: the vorticity error drops from $2.383$ to $0.275$ and the high-$k$ energy fraction rises from $0.0077$ to $0.422$ of the reference value.
At $32^2$ and $16^2$, the base model does not collapse spectrally, and fine-tuning mainly raises the high-$k$ energy fraction while every field metric stays within $1.5\%$ of the base model (Appendix~\ref{appx:fine-tune_coarse}).
ELBM keeps the lowest final log-spectrum RMSE at $64^2$ and $16^2$ while carrying excess high-$k$ energy at $64^2$ and $32^2$.
Fine-tuning therefore helps where \emph{entropically-stabilized NeurDE} over-smooths, and otherwise leaves its field fidelity essentially unchanged, which is the behavior reported in Section~\ref{sec:experiments_main}.

\section{\emph{NeurDE} Training Procedure}\label{app:Training}
The surrogate's effectiveness depends on the training protocol as well as the
architecture. Because the entropic parameter $\alpha$ directly controls
long-time stability in underresolved LB simulations, we train it in the same
setting in which it will be used: embedded in the rollout. The protocol
combines three elements: rollout-based training, curriculum learning, and the
Incremental Progress Training Algorithm (IPTA)~\citet{bansal2022end}.

\subsection{Stability Model Pretraining}

Training the $\alpha$ stability model begins with a typical supervised training routine from predicted ELBM-generated $\alpha$.
As the ELBM grids become finer, $\alpha \rightarrow 2$.
Therefore, we train against the $\alpha$ values from the coarsest grid where ELBM can reproduce the expected dynamics.
We train the stability model by comparing its output directly to the ELBM $\alpha$ through an MSE loss.

\subsection{Rollout-Based Training}
Rather than optimizing only one-step predictions, the surrogate is embedded
directly into the ELBM update and trained through multi-step rollouts. This
recursive formulation exposes the model to compounding errors and encourages
updates that remain stable over trajectories. The rollout loss balances three
objectives:
\begin{equation}\label{eq:losses}
\mathcal{L}_{\mathrm{rollout}}(\theta) = \sum_{t=1}^T  
    c_{\mathrm{root}} \, \mathcal{L}_{\mathrm{root}}(t) 
    + c_\omega \, \mathcal{L}_\omega(t) 
    + c_{\mathrm{H}} \, \mathcal{R}_{\mathrm{H}}(t),
\end{equation}
where $T$ is the rollout horizon. In practice, $T$ is limited by GPU memory,
and the loss is a weighted combination of approximate recovery of the entropic
root, preservation of vortical structures, and compliance with the discrete
$H$-theorem.

\subsection{Curriculum Learning}
Applying rollout training with a fixed, long horizon from the start is unstable
and memory intensive. We therefore use a curriculum over the rollout horizon
$T$~\citet{bengio2009curriculum,bansal2022end}. The model is first trained on
short trajectories and is then exposed to longer horizons as training
progresses. The effective horizon at epoch $e$ is determined by a monotonically
increasing schedule $g(e)$:
\begin{equation} 
T_\mathrm{curr}(e) = \min \big(T_\mathrm{max}, \lfloor g(e) \rfloor \big), 
\end{equation} 
where $T_\mathrm{max}$ is the maximum horizon feasible within hardware memory
constraints.\footnote{The floor function, denoted as $\lfloor x \rfloor$, gives
the greatest integer that is less than or equal to $x$.} This schedule lets the
model learn stable short-term updates before it is required to maintain
dynamical consistency over longer intervals.

\subsection{Incremental Progress Training Algorithm (IPTA)}
Curriculum learning controls the forward rollout length, but gradient
computation remains the limiting factor. On short horizons, consecutive states
are often nearly identical, giving weak gradients. Backpropagating through the
full long-horizon trajectory is typically infeasible due to memory limitations.
Truncated Backpropagation Through Time (TBPTT) reduces memory by
backpropagating through a fixed recent window of timesteps
\citet{williams1990efficient}, but this introduces a recency bias and ignores
longer-range dependencies that matter for physical consistency
\citet{pascanu2013difficulty}.

We use IPTA~\citet{bansal2022end} to decouple the rollout horizon from the
backpropagation path. Instead of using a contiguous block of steps, IPTA
computes the gradient over a sparse subset
$\mathcal{S}_k(e) \subseteq \{1, \dots, T_\mathrm{curr}(e)\}$:
\begin{equation}
\nabla_\theta \mathcal{L}_\mathrm{IPTA}(e) = 
\sum_{t \in \mathcal{S}_k(e)} \nabla_\theta \ell^{(t)}.
\end{equation}
This selective summation makes the memory cost proportional to the subset size
$|\mathcal{S}_k(e)|$, not the full horizon $T_\mathrm{curr}(e)$. This is what
allows the curriculum to use effective horizons longer than those feasible with
standard backpropagation.

We use two strategies to construct $\mathcal{S}_k(e)$. The first is a
\textit{deterministic} schedule that selects timesteps sparsely near the
beginning of a trajectory and more densely toward the end, where rollout errors
have had more time to compound. The second is a \textit{stochastic} schedule
that samples timesteps from a distribution biased toward distant states. The
deterministic version gives a lower-variance learning signal; the stochastic
version gives a more exploratory estimate of long-range temporal effects.

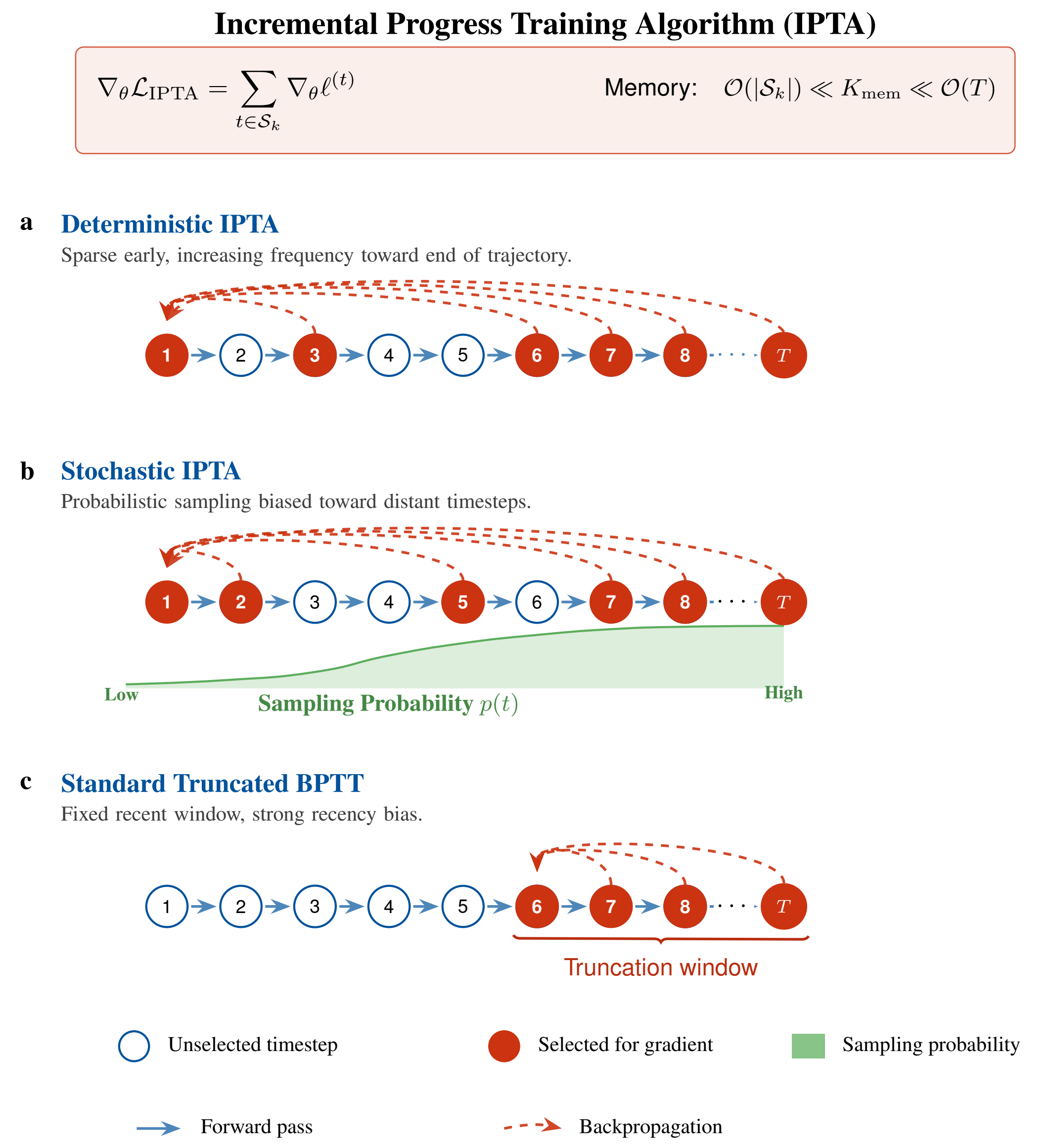
\begin{figure}[ht!]
    \centering

\begin{tikzpicture}[
    scale=1.1, 
    >=Stealth,
    timestep/.style={circle, draw=natureblue, fill=white, line width=0.8pt, minimum size=5mm, font=\scriptsize\sffamily},
    selected/.style={timestep, draw=naturered, fill=naturered, text=white, font=\scriptsize\bfseries\sffamily},
    fwd_arrow/.style={->, natureblue!70, line width=1pt, shorten >=1pt, shorten <=1pt}, 
    back_arrow/.style={->, naturered, dashed, line width=1pt, opacity=0.9},
    panel_label/.style={font=\normalsize\bfseries, anchor=west, text=black},
    panel_title/.style={font=\normalsize\bfseries, anchor=west, text=natureblue},
    panel_desc/.style={font=\footnotesize, text=naturegray, text width=8cm, anchor=west}
]

\definecolor{natureblue}{RGB}{0,82,155}
\definecolor{naturered}{RGB}{204,51,17}
\definecolor{naturegreen}{RGB}{85,170,85}
\definecolor{naturegray}{RGB}{60,60,60} 

\node[font=\large\bfseries, text=black] at (0, 9.2) {Incremental Progress Training Algorithm (IPTA)};
\node[rectangle, draw=naturered!80, fill=naturered!8, rounded corners=3pt,
      inner sep=8pt, text width=12cm, align=center, line width=0.5pt] at (0, 8.3) {
    $\displaystyle \nabla_\theta \mathcal{L}_\mathrm{IPTA} = \sum_{t \in \mathcal{S}_k} \nabla_\theta \ell^{(t)}$
    \hfill 
    {\small \textsf{Memory: } $\mathcal{O}(|\mathcal{S}_k|) \ll K_\mathrm{mem} \ll \mathcal{O}(T)$}
};

\node[panel_label] at (-6.5, 6.8) {a};
\node[panel_title] at (-6.0, 6.8) {Deterministic IPTA};
\node[panel_desc] at (-6.0, 6.4) {Sparse early, increasing frequency toward end of trajectory.};

\foreach \i in {1,...,8} {
    \coordinate (dt\i) at (-5.5+\i*0.9, 5.2);
    \ifnum\i=1 \node[selected] (n_dt\i) at (dt\i) {\i}; \else \node[timestep] (n_dt\i) at (dt\i) {\i}; \fi
    \ifnum\i>1 \pgfmathtruncatemacro{\prev}{\i-1} \draw[fwd_arrow] (n_dt\prev) -- (n_dt\i); \fi
}
\coordinate (dtT_pos) at ($(dt8)+(1.2,0)$);
\node[selected] (n_dtT) at (dtT_pos) {$T$};

\draw[fwd_arrow] (n_dt8) -- (n_dtT) node[midway, fill=white, inner sep=1pt] {$\cdots$}; 

\foreach \t in {3,6,7,8} { \node[selected] at (n_dt\t) {\t}; }

\coordinate (det_target) at ($(n_dt1.north)+(0,0.15)$);
\draw[back_arrow] (n_dt3.north) .. controls +(0,0.4) and +(0,0.4) .. (det_target);
\draw[back_arrow] (n_dt6.north) .. controls +(0,0.5) and +(0,0.5) .. (det_target);
\draw[back_arrow] (n_dt7.north) .. controls +(0,0.6) and +(0,0.6) .. (det_target);
\draw[back_arrow] (n_dt8.north) .. controls +(0,0.65) and +(0,0.65) .. (det_target);
\draw[back_arrow] (n_dtT.north) .. controls +(-0.5,0.7) and +(0.5,0.7) .. (det_target);

\node[panel_label] at (-6.5, 3.8) {b};
\node[panel_title] at (-6.0, 3.8) {Stochastic IPTA};
\node[panel_desc] at (-6.0, 3.4) {Probabilistic sampling biased toward distant timesteps.};

\coordinate (st1) at (-4.6, 2.2);
\coordinate (st8) at (-4.6+7*0.9, 2.2);
\coordinate (stT) at ($(st8)+(1.2,0)$);

\draw[naturegreen, thick, opacity=0.95]
    plot[smooth, tension=0.9] coordinates {
        ($(st1)+(-0.50,-1.0)$) ($(st1)+(0.7,-0.95)$) ($(st1)+(1.8,-0.85)$)
        ($(st1)+(2.7,-0.65)$) ($(st1)+(3.6,-0.50)$) ($(st1)+(4.5,-0.40)$)
        ($(st1)+(5.4,-0.33)$) ($(st1)+(6.3,-0.30)$) ($(stT)+( 0.00,-0.29)$)
    };
\fill[naturegreen, opacity=0.20] 
    plot[smooth, tension=0.9] coordinates {
        ($(st1)+(-0.50,-1.0)$) ($(st1)+(0.7,-0.95)$) ($(st1)+(1.8,-0.85)$)
        ($(st1)+(2.7,-0.65)$) ($(st1)+(3.6,-0.50)$) ($(st1)+(4.5,-0.40)$)
        ($(st1)+(5.4,-0.33)$) ($(st1)+(6.3,-0.30)$) ($(stT)+( 0.00,-0.29)$)
    }
    -- ($(stT)+(0,-1.05)$) -- ($(st1)+(-0.50,-1.05)$) -- cycle;

\node[font=\small\bfseries, text=naturegreen!80!black] at ($(st1)+(2.7,-1.25)$) {Sampling Probability $p(t)$};
\node[font=\scriptsize\bfseries, text=naturegreen!80!black] at ($(st1)+(-0.55,-1.12)$) {Low};
\node[font=\scriptsize\bfseries, text=naturegreen!80!black] at ($(stT)+(0.00,-1.12)$) {High};

\foreach \i in {1,...,8} {
    \coordinate (pos\i) at (-5.5+\i*0.9, 2.2);
    \ifnum\i=1 \node[selected] (n_st\i) at (pos\i) {\i}; \else \node[timestep] (n_st\i) at (pos\i) {\i}; \fi
    \ifnum\i>1 \pgfmathtruncatemacro{\prev}{\i-1} \draw[fwd_arrow] (n_st\prev) -- (n_st\i); \fi
}
\node[selected] (n_stT) at (stT) {$T$};
\draw[fwd_arrow] (n_st8) -- (n_stT) node[midway, fill=white, inner sep=1pt, text=black] {$\cdots$};

\foreach \t in {2,5,7,8} { \node[selected] at (n_st\t) {\t}; }

\coordinate (sto_target) at ($(n_st1.north)+(0,0.15)$);
\draw[back_arrow] (n_st2.north) .. controls +(0,0.3) and +(0,0.3) .. (sto_target);
\draw[back_arrow] (n_st5.north) .. controls +(0,0.5) and +(0,0.5) .. (sto_target);
\draw[back_arrow] (n_st7.north) .. controls +(0,0.6) and +(0,0.6) .. (sto_target);
\draw[back_arrow] (n_st8.north) .. controls +(0,0.65) and +(0,0.65) .. (sto_target);
\draw[back_arrow] (n_stT.north) .. controls +(-0.5,0.7) and +(0.5,0.7) .. (sto_target);

\node[panel_label] at (-6.5, 0.0) {c};
\node[panel_title] at (-6.0, 0.0) {Standard Truncated BPTT};
\node[panel_desc] at (-6.0, -0.4) {Fixed recent window, strong recency bias.};

\foreach \i in {1,...,8} {
    \coordinate (posb\i) at (-5.5+\i*0.9, -1.5);
    \node[timestep] (n_bt\i) at (posb\i) {\i};
    \ifnum\i>1 \pgfmathtruncatemacro{\prev}{\i-1} \draw[fwd_arrow] (n_bt\prev) -- (n_bt\i); \fi
}
\coordinate (btT_pos) at ($(n_bt8)+(1.2,0)$);
\node[selected] (n_btT) at (btT_pos) {$T$};
\draw[fwd_arrow] (n_bt8) -- (n_btT) node[midway, fill=white, inner sep=1pt, text=black] {$\cdots$};

\foreach \t in {6,7,8} { \node[selected] at (n_bt\t) {\t}; }

\coordinate (bptt_target) at ($(n_bt6.north)+(0,0.15)$);
\draw[back_arrow] (n_bt7.north) .. controls +(0,0.40) and +(0,0.40) .. (bptt_target);
\draw[back_arrow] (n_bt8.north) .. controls +(0,0.40) and +(0,0.40) .. (bptt_target);
\draw[back_arrow] (n_btT.north) .. controls +(0,0.50) and +(0,0.50) .. (bptt_target);

\draw[decoration={brace,mirror,raise=5pt}, decorate, line width=1.0pt, naturered!90!black]
    ($(n_bt6.south west)+(-0.1,0)$) -- ($(n_btT.south east)+(0.1,0)$)
    node[midway, below=10pt, font=\small\sffamily, text=naturered!90!black] {Truncation window};

\node[timestep, minimum size=4mm] (leg_unselected) at (-5, -3.2) {};
\node[font=\footnotesize, anchor=west] at ($(leg_unselected)+(0.3,0)$) {Unselected timestep};

\node[selected, minimum size=4mm] (leg_selected) at (-0.5, -3.2) {};
\node[font=\footnotesize, anchor=west] at ($(leg_selected)+(0.3,0)$) {Selected for gradient};

\fill[naturegreen, opacity=0.7] (3, -3.35) rectangle (3.4, -3.05);
\node[font=\footnotesize, anchor=west] at (3.5, -3.2) {Sampling probability};

\draw[fwd_arrow] (-5, -4.2) -- (-4.4, -4.2);
\node[font=\footnotesize, anchor=west] at (-4.3, -4.2) {Forward pass};

\draw[back_arrow] (-0.5, -4.2) to[bend left=12] (0.2, -4.2);
\node[font=\footnotesize, anchor=west] at (0.3, -4.2) {Backpropagation};

\end{tikzpicture}
\caption{Gradient accumulation in the Incremental Progress Training Algorithm (IPTA). 
By computing gradients over a sparse subset of timesteps, either deterministically (denser near the end) or stochastically (biased toward distant steps), IPTA removes the recency bias of truncated backpropagation and enables memory-efficient long-horizon training.}\label{fig:IPTA}
\end{figure}

\subsubsection{Training Algorithm}
The practical implementation is summarized in \Cref{Alg:training}. The
surrogate is unrolled for a curriculum-selected horizon, and the model
parameters are updated using gradients from a sparse subset of timesteps whose
size is constrained by the available hardware memory.

\vspace{0.5em}

\begin{algorithm}[ht!]
   \caption{Training the DEQ Surrogate for $\alpha$}
   \label{Alg:training}
\begin{algorithmic}[1]
   \STATE {\bfseries Input:} DEQ parameters $\theta$, rollout data $\{\F^{(t)}\}_{t=1}^T$, curriculum schedule $\{T_\mathrm{curr}(e)\}$, IPTA subset strategy, max subset size $K_\mathrm{mem}$ (hardware-dependent memory budget), vorticity reference $\boldsymbol{\omega}_{\mathrm{ref}}$, weights $c_{\mathrm{root}}, c_\omega, c_{\mathrm{H}}$
   \STATE {\bfseries Output:} Trained DEQ parameters $\theta$
   \FOR{each epoch $e$}
      \STATE Set rollout horizon $T \leftarrow T_\mathrm{curr}(e)$
      \FOR{each rollout trajectory}
         \FOR{each timestep $t=1,\dots,T$}
            \STATE Compute $\mathbf{z} \leftarrow \mathcal{E}(\F^{(t)}, \F^{\mathrm{eq}}, \rho, \Velocity)$
            \STATE Predict entropic parameter $\alpha_\theta^{(t)} = \mathrm{DEQ}(\alpha_\theta^{(t-1)},  \mathbf{z})$
            \STATE Apply ELBM update with $\alpha_\theta^{(t)}$ to obtain $\F^{(t+1)}$
            \STATE Evaluate entropy regularizer $\mathcal{R}_{\mathrm{H}}(t)$ and vorticity loss $\mathcal{L}_\omega(t)$
            \STATE Form one-step weighted loss $\ell^{(t)}$
         \ENDFOR
         \STATE Select subset $\mathcal{S}_k(e) \subseteq \{1,\dots,T\}$ s.t. $|\mathcal{S}_k(e)| \le K_\mathrm{mem}$ using the chosen IPTA strategy
         \STATE Accumulate gradients $\nabla_\theta \mathcal{L}_\mathrm{IPTA}(e) = \sum_{t \in \mathcal{S}_k(e)} \nabla_\theta \ell^{(t)}$
         \STATE Update $\theta$ using optimizer
         \STATE Save $\alpha_\theta^{(t)}$
      \ENDFOR
   \ENDFOR
\end{algorithmic}
\end{algorithm}

\vspace{0.5em}

Together, rollout training, the horizon curriculum, and IPTA address the main
practical tension in this problem: the closure must be trained on long
trajectories, but full backpropagation through those trajectories is too
expensive. Rollout training exposes the model to compounding error, the loss
keeps the update tied to physical consistency, and IPTA makes longer effective
horizons feasible under memory constraints.

\subsection{Hyperparameters}

\paragraph{Optimization.}
Both stages use AdamW with learning rate $5\times10^{-4}$ and weight decay $10^{-4}$. The
learning rate is warmed up linearly over the first 1000 steps and then held constant.
Gradients are clipped to norm 1, and steps whose gradient norm is an outlier
($z$-score above 2.5 relative to a running average) are scaled down further. The global
batch size is 16 for pretraining and 32 for rollout training, split evenly over the eight
GCDs. Pretraining runs in float32 with a bfloat16 encoder; rollout training runs in
float64. The validation loss is computed every 250 steps, and the checkpoint with the
lowest value is kept. Every configuration is trained with four seeds.

\paragraph{Data.}
Pretraining targets are the ELBM $\alpha$ fields at $128^2$ (TGV-50K), $256^2$
(DPSL-50K), and $384^2$ (DPSL-100K). Rollout training uses the $512^2$ ELBM simulation, 
coarse-grained to $N^2$. Each simulation is split along time into training, validation,
and test blocks in proportions 0.6/0.1/0.3 for pretraining and 0.5/0.2/0.3 for rollout
training. Rollout training on each grid starts from the model of the previous, finer
grid; the finest grid starts from the pretrained model.

\paragraph{Losses.}
Pretraining minimizes a mean squared error on $\alpha - 2$, weighted by
$1 + 10^{5}\lvert\alpha - 2\rvert$, plus a Jacobian penalty on the DEQ with weight 0.1.
The rollout loss of Eq.~\eqref{eq:losses} uses $c_{\mathrm{root}} = 10^{4}$,
$c_\omega = 10$, and $c_{\mathrm{H}} = 10$, where $\mathcal{L}_\omega$ is a Huber loss
($\delta = 10^{-4}$) on $\rho$, $\mathbf{u}$, and $\boldsymbol{\omega}$ with relative
weights 1, 1, and 5.

\paragraph{Curriculum and IPTA.}
The horizon $T_{\mathrm{curr}}$ steps through 1, 2, 3, 5, 7, 10, 17, 25, 37, 50, 75,
100, 250, 500, 1000, and 1500, with 2, 1, 1, 5, 1, 5, 1, 5, 1, 7, 1, 10, 10, 10, 10, and
10 epochs respectively (80 in total). Training moves to the next horizon early after 3
epochs without validation improvement, and unused epochs carry over, which allows the
final horizon $T_{\mathrm{max}} = 2000$ to be reached. Horizons are capped at 75\% of the
training split. We use the deterministic IPTA schedule with $K_{\mathrm{mem}} = 16$
log-spaced timesteps that become denser toward $T$.

\paragraph{Training length.}
Pretraining was configured for 10{,}000 epochs and rollout training for 80, but both
were also limited by job wall time. Table~\ref{tab:epochs} gives the epochs completed.

\begin{table}[ht]
\centering
\small
\caption{Epochs completed, mean over four seeds. Rollout training on a grid is complete
at 80 epochs; lower values were stopped by the wall-time limit. Epochs shrink as the
horizon grows, so counts are not comparable across stages.}
\label{tab:epochs}
\begin{tabular}{@{}lrrrrrrr@{}}
\toprule
 & Stability & $384^2$ & $256^2$ & $128^2$ & $64^2$ & $32^2$ & $16^2$ \\
\midrule
TGV-50K   & 374.3 & --  & --   & 51.3 & 80.0 & 80 & 80 \\
DPSL-50K  & 32.5  & --  & 14.5 & 39.5 & 56.1 & 80 & 80 \\
DPSL-100K & 16.2  & 8.6 & 16.5 & 44.8 & 68.7 & 80 & 80 \\
\bottomrule
\end{tabular}
\end{table}

\subsection{Compute Costs}

All models were trained on the OLCF Frontier supercomputer. A Frontier node contains four
AMD Instinct MI250X accelerators. Each MI250X consists of two graphics compute dies (GCDs),
and each GCD has 64\,GB of high-bandwidth memory and is exposed to software as a separate
GPU, so a node provides eight GPUs.\footnote{OLCF Frontier User Guide,
\url{https://docs.olcf.ornl.gov/systems/frontier_user_guide.html}.} Every training run
in Table~\ref{tab:training_budget} used one full node, that is, all eight GCDs. Training was
data-parallel: each GCD held a complete copy of the model, the training samples were split
evenly across the eight GCDs, and gradients were averaged across GCDs at every step with
PyTorch \texttt{DistributedDataParallel}. The spatial domain was not partitioned, so each GCD
processed whole grids. Each of the four random seeds was trained as an independent job with
this same configuration.

Table~\ref{tab:training_budget} reports the cost of each training stage, averaged over the
four seeds. The \emph{Stage} column lists the stability model, which is trained first, and
then the solver at each training resolution $N^2$, which is the reference simulation
coarse-grained to an $N \times N$ grid. Solver stages are trained from the finest grid to the
coarsest, and each stage is initialized from a checkpoint of the previous, finer stage. Each
row therefore gives the additional cost of that stage. The cost of obtaining a model at a
given resolution includes all finer stages above it. \emph{Wall time} is the elapsed time of
the stage in hours, minutes, and seconds, summed over all jobs used to complete it, because
long stages were split into several consecutive jobs that each resumed from the last
checkpoint. \emph{Node-h} is the wall time in hours multiplied by the one node used.
\emph{GCD-h} is the node-hours multiplied by the eight GCDs per node. To count physical
MI250X accelerators instead, divide this column by two. \emph{Steps} is the number of
training iterations completed, taken from the last saved checkpoint. For stages that were
stopped by the job time limit, this is a lower bound. The \emph{Total} rows sum all stages for
one seed, so the full cost of all four seeds is four times the total.

\begin{table}
\centering
\caption{Training budget of the base models, averaged over four seeds. All stages ran on
one Frontier node (8 MI250X GCDs) with data-parallel training.}
\label{tab:training_budget}
\begin{tabular}{@{}llrrrr@{}}
        \toprule
        Case & Stage & Wall time & Node-h & GCD-h & Steps \\
        \midrule
        \multirow{6}{*}{TGV-50K}
         & Stability      & 1:57:19 & 1.96 & 15.64 & 69{,}611 \\
         & Solver $128^2$ & 6:00:34 & 6.01 & 48.08 & 3{,}657 \\
         & Solver $64^2$  & 1:50:48 & 1.85 & 14.78 & 1{,}736 \\
         & Solver $32^2$  & 0:39:30 & 0.66 &  5.27 & 787 \\
         & Solver $16^2$  & 0:11:14 & 0.19 &  1.50 & 373 \\
         \cmidrule(l){2-6}
         & \textbf{Total} & \textbf{10:39:27} & \textbf{10.66} & \textbf{85.26} & \\
        \midrule
        \multirow{7}{*}{DPSL-50K}
         & Stability      & 1:28:29 & 1.47 & 11.80 & 12{,}123 \\
         & Solver $256^2$ & 6:00:45 & 6.01 & 48.10 & 2{,}453 \\
         & Solver $128^2$ & 6:00:48 & 6.01 & 48.11 & 2{,}948 \\
         & Solver $64^2$  & 5:48:24 & 5.81 & 46.45 & 1{,}539 \\
         & Solver $32^2$  & 0:44:44 & 0.75 &  5.97 & 775 \\
         & Solver $16^2$  & 0:12:43 & 0.21 &  1.70 & 373 \\
         \cmidrule(l){2-6}
         & \textbf{Total} & \textbf{20:15:55} & \textbf{20.27} & \textbf{162.12} & \\
        \midrule
        \multirow{8}{*}{DPSL-100K}
         & Stability                & 1:59:10 & 1.99 & 15.89 & 9{,}113 \\
         & Solver $384^2$ & 3:24:32 & 3.41 & 27.27 & 2{,}125 \\
         & Solver $256^2$           & 4:00:21 & 4.01 & 32.05 & 2{,}625 \\
         & Solver $128^2$           & 6:00:27 & 6.01 & 48.06 & 3{,}259 \\
         & Solver $64^2$            & 2:22:38 & 2.38 & 19.02 & 1{,}597 \\
         & Solver $32^2$            & 0:34:03 & 0.57 &  4.54 & 775 \\
         & Solver $16^2$            & 0:11:57 & 0.20 &  1.59 & 373 \\
         \cmidrule(l){2-6}
         & \textbf{Total} & \textbf{18:33:11} & \textbf{18.55} & \textbf{148.42} & \\
        \bottomrule
\end{tabular}
\end{table}

\section{Periodic Flow Diagnostics}\label{appx:periodic_diagnostics}
This appendix records the periodic-flow diagnostics that we use for offline
post-processing and will use for future turbulence analyses. These diagnostics
are independent of a particular solver implementation. They assume periodic
fields, constant-density low-Mach hydrodynamic conventions, and Fourier
derivatives on a uniform grid unless stated otherwise. The goal is to make the
normalizations explicit, so that spectra, vorticity diagnostics, and
time-series balance checks can be compared across saved rollouts without
ambiguity.

Throughout this appendix, $\langle \cdot \rangle$ denotes a spatial average
over the periodic grid. Repeated velocity-component indices are summed. The
definitions below are hydrodynamic: if a simulation is compressible or
variable-density, then density-weighted kinetic energy and compressible
energy budgets require additional terms that are not part of these diagnostics.

\subsection{Fourier and Energy-Spectrum Conventions}
\label{appx:diagnostics_spectrum}

Let $\mathbf{u}=(u_1,\ldots,u_D)$ denote a velocity field on a periodic
grid with $V=\prod_{j=1}^D N_j$ grid points. We use the unnormalized discrete
Fourier transform convention
\begin{equation}
    \widehat{u}_i(\mathbf{m})
    =
    \sum_{\mathbf{x}}
    u_i(\mathbf{x})\,
    \exp\!\left(-2\pi \mathrm{i}\,
    \sum_{j=1}^D \frac{m_j x_j}{N_j}\right),
\end{equation}
where $\mathbf{m}$ is the vector of integer FFT modes. With this convention, the zero mode
is the spatial mean and nonzero modes describe fluctuations around the mean
flow. The kinetic energy per unit mass is defined by
\begin{equation}
    K
    =
    \frac{1}{2}\left\langle u_i u_i \right\rangle,
\end{equation}
where $\langle\cdot\rangle$ denotes the spatial average and repeated velocity
indices are summed. Parseval's identity then gives
\begin{equation}
    K
    =
    \sum_{\mathbf{m}}
    \frac{1}{2V^2}
    \left|\widehat{u}_i(\mathbf{m})\right|^2.
\end{equation}
The integer-shell energy is obtained by binning Fourier modes according to
$n=\operatorname{round}(|\mathbf{m}|)$:
\begin{equation}
    E_n
    =
    \sum_{\operatorname{round}(|\mathbf{m}|)=n}
    \frac{1}{2V^2}
    \left|\widehat{u}_i(\mathbf{m})\right|^2.
\end{equation}
When the zero shell is included, $\sum_n E_n=K$ up to roundoff. If the zero
shell is omitted, the sum gives the turbulent kinetic energy after removing
mean-flow kinetic energy. This distinction is useful for comparing initialized
zero-mean periodic turbulence with rollouts that may develop a small numerical
mean.

On an isotropic periodic domain of side length $L$, the physical shell
wavenumber and shell spacing are
\begin{equation}
    \kappa_n = \frac{2\pi n}{L},
    \qquad
    \Delta \kappa = \frac{2\pi}{L}.
\end{equation}
The continuum energy spectrum is approximated by
\begin{equation}
    E(\kappa_n)
    \approx
    \frac{E_n}{\Delta \kappa},
    \qquad
    K
    \approx
    \sum_n E(\kappa_n)\Delta \kappa.
\end{equation}
The continuum conversion above assumes a common side length in all coordinate
directions. For anisotropic boxes one can still use spectral derivatives with
direction-dependent wavenumbers, but shell-averaged spectra should be
interpreted with care because spherical shells in integer-mode space no longer
correspond to isotropic shells in physical wavenumber space. These are the
standard homogeneous-turbulence spectrum conventions used in Pope, Frisch, and
Davidson~\citet{pope2000turbulent,frisch1995turbulence,davidson2015turbulence}.

\subsection{Spectral Derivatives, Vorticity, and Dissipation}
\label{appx:diagnostics_derivatives}

For a periodic scalar field $f$, spectral differentiation is computed as
\begin{equation}
    \partial_j f
    =
    \mathcal{F}^{-1}
    \left[
        \mathrm{i}\kappa_j\,\widehat{f}
    \right],
\end{equation}
where $\kappa_j=2\pi m_j/L_j$ is the physical wavenumber in direction $j$.
The divergence diagnostic is $\nabla\cdot \mathbf{u}=\partial_j u_j$. In
low-Mach incompressible or weakly compressible calculations, we use the ratio
\begin{equation}
    \frac{\theta_{\mathrm{rms}}}{\omega_{\mathrm{rms}}}
    =
    \frac{
        \left\langle (\nabla\cdot \mathbf{u})^2 \right\rangle^{1/2}
    }{
        \left\langle \omega_i\omega_i \right\rangle^{1/2}
    }
\end{equation}
as a compact measure of compressibility or solver-induced divergence error.
In two dimensions, the scalar vorticity is
\begin{equation}
    \omega = \partial_x u_y - \partial_y u_x,
\end{equation}
and the enstrophy is
\begin{equation}
    \Omega
    =
    \frac{1}{2}\left\langle \omega^2 \right\rangle.
\end{equation}
The palinstrophy is
\begin{equation}
    P
    =
    \frac{1}{2}
    \left\langle |\nabla \omega|^2 \right\rangle.
    \label{eq:palinstrophy}
\end{equation}
For three-dimensional flow, $\boldsymbol{\omega}=\nabla\times\mathbf{u}$
and the strain-rate tensor is
\begin{equation}
    S_{ij}
    =
    \frac{1}{2}
    \left(
        \partial_j u_i + \partial_i u_j
    \right).
\end{equation}
For incompressible periodic flow with kinematic viscosity $\nu$, the kinetic
energy dissipation rate is
\begin{equation}
    \varepsilon
    =
    2\nu
    \left\langle S_{ij}S_{ij}\right\rangle.
\end{equation}
The corresponding Kolmogorov length and velocity scales are
\begin{equation}
    \eta_K
    =
    \left(\frac{\nu^3}{\varepsilon}\right)^{1/4},
    \qquad
    u_\eta
    =
    (\nu\varepsilon)^{1/4}.
\end{equation}
These vorticity and strain diagnostics follow the usual incompressible-flow
notation~\citet{batchelor2000fluid,pope2000turbulent,davidson2015turbulence}.

\subsection{Time-Series Balance Checks}
\label{appx:diagnostics_balances}

For unforced incompressible periodic flow, kinetic energy decay is controlled
by viscous dissipation:
\begin{equation}
    \frac{dK}{dt}
    =
    -\varepsilon.
\end{equation}
In periodic incompressible flow, $\varepsilon=2\nu\Omega$ when
$\Omega=\frac{1}{2}\langle \omega_i\omega_i\rangle$. Thus, an effective
viscosity inferred from the energy time series is
\begin{equation}
    \nu_{\mathrm{eff},K}
    =
    -\frac{dK/dt}{2\Omega}.
\end{equation}
When the nominal viscosity is known, the ratio
$\nu_{\mathrm{eff},K}/\nu$ is a direct diagnostic of whether the rollout is
dissipating energy at the expected rate.

For the enstrophy balance, the three-dimensional identity is
\begin{equation}
    \frac{d\Omega}{dt}
    =
    \left\langle
        \omega_i S_{ij}\omega_j
    \right\rangle
    -
    2\nu P,
\end{equation}
where the first term is vortex stretching. In two dimensions the stretching
term vanishes, leaving purely viscous enstrophy decay. The corresponding
effective viscosity estimate is
\begin{equation}
    \nu_{\mathrm{eff},\Omega}
    =
    \frac{
        \left\langle \omega_i S_{ij}\omega_j \right\rangle
        -
        d\Omega/dt
    }{
        2P
    }.
\end{equation}
These balance identities are useful as regression checks for saved rollouts: large disagreement between nominal and effective viscosity typically indicates
forcing, boundary effects, insufficient resolution, or non-negligible
compressibility terms that are outside the assumptions of this appendix.

We also report basic scalar summaries: component means and root-mean-square
velocities, total RMS speed $\sqrt{\langle u_i u_i\rangle}$, and the
low-Mach RMS Mach diagnostic
\begin{equation}
    \mach_{\mathrm{rms}}
    =
    \frac{\sqrt{\langle u_i u_i\rangle}}{c_s}.
\end{equation}
For zero-mean homogeneous turbulence, this coincides with the usual turbulent
Mach number. If a rollout develops mean drift, the fluctuation-based turbulent
Mach number is obtained by replacing $u_i$ with
$u_i'=u_i-\langle u_i\rangle$. For lattice Boltzmann units, the default is
$c_s=1/\sqrt{3}$.

\subsection{Inertial-Range Guides}
\label{appx:diagnostics_inertial}

In two-dimensional hydrodynamic turbulence, the inviscid quadratic invariants
are kinetic energy and enstrophy. The Kraichnan--Batchelor dual-cascade picture
therefore predicts an inverse energy cascade with
\begin{equation}
    E(\kappa)
    \sim
    \varepsilon^{2/3}\kappa^{-5/3},
\end{equation}
and a direct enstrophy cascade with
\begin{equation}
    E(\kappa)
    \sim
    \eta_{\Omega}^{2/3}\kappa^{-3},
\end{equation}
where $\varepsilon$ is the upscale energy-transfer rate and $\eta_{\Omega}$ is
the downscale enstrophy-transfer rate. We use the subscripted notation
$\eta_{\Omega}$ to avoid confusing the enstrophy flux with the ELBM relaxation
parameter $\beta$. In this setting, the enstrophy spectrum
satisfies $Z(\kappa)=\kappa^2 E(\kappa)$, so the direct-enstrophy range
corresponds to $Z(\kappa)\sim \kappa^{-1}$. We use these scalings only for
two-dimensional hydrodynamic diagnostics~\citet{kraichnan1967inertial,boffetta2012two}.
For three-dimensional homogeneous isotropic turbulence, the default inertial
range guide is the Kolmogorov energy scaling
\begin{equation}
    E(\kappa)
    \sim
    \varepsilon^{2/3}\kappa^{-5/3},
\end{equation}
with the dissipation rate $\varepsilon$ defined above~\citet{pope2000turbulent,frisch1995turbulence}.

We also track longitudinal structure functions for periodic homogeneous fields.
For an axis-aligned separation $r\mathbf{e}_j$,
\begin{equation}
    S_p(r)
    =
    \left\langle
        \left[
            \big(
                \mathbf{u}(\mathbf{x}+r\mathbf{e}_j)
                -
                \mathbf{u}(\mathbf{x})
            \big)
            \cdot \mathbf{e}_j
        \right]^p
    \right\rangle,
\end{equation}
with averages taken over space and over selected coordinate directions. Local
scaling exponents are estimated by
\begin{equation}
    \xi_p(r)
    =
    \frac{d\log S_p(r)}{d\log r}.
\end{equation}
When the inertial interval is short, extended self-similarity instead compares
structure functions directly, e.g.
\begin{equation}
    \frac{d\log S_4(r)}{d\log S_2(r)}
    =
    \frac{\xi_4(r)}{\xi_2(r)}.
\end{equation}
This follows the ESS diagnostic of Benzi et al.~\citet{benzi1993extended}.

\subsection{Correlations and Small-Scale Shape Statistics}
\label{appx:diagnostics_correlations}

Two-point velocity correlations are computed from the fluctuating velocity
$u_i' = u_i-\langle u_i\rangle$:
\begin{equation}
    R_{uu}(r)
    =
    \left\langle
        u_i'(\mathbf{x})\,
        u_i'(\mathbf{x}+r\mathbf{e}_j)
    \right\rangle,
\end{equation}
with averages over space, velocity component $i$, and selected coordinate
directions $j$. The normalized correlation $R_{uu}(r)/R_{uu}(0)$ provides a
scale-space view that is complementary to the Fourier spectrum. In periodic
homogeneous turbulence, the correlation length and spectral distribution
encode the same large-scale organization, but correlations can be easier to
interpret when spectra are noisy or when only a short inertial interval is
available~\citet{pope2000turbulent,davidson2015turbulence}.

For small-scale intermittency checks, we use the longitudinal derivative
$a=\partial_x u_x$ and report
\begin{equation}
    S_a
    =
    \frac{\langle a^3\rangle}{\langle a^2\rangle^{3/2}},
    \qquad
    F_a
    =
    \frac{\langle a^4\rangle}{\langle a^2\rangle^2}.
\end{equation}
The skewness $S_a$ is sensitive to the asymmetric small-scale cascade, while
the flatness $F_a$ is a compact measure of heavy tails and intermittency.

\subsection{Entropic LBM Diagnostics}
\label{appx:diagnostics_lbm}

The same spectra and structure statistics are useful for evaluating entropic
lattice Boltzmann simulations, where stability and inertial-range fidelity are
often assessed through energy spectra, compensated spectra, and higher-order
structure functions~\citet{dorschner2016entropic_mrt,buzzicotti2021inertial}.
In the present hydrodynamic setting, the diagnostic role is threefold. First,
the identity $\sum_n E_n=K$ verifies the Fourier normalization and catches
shape/order errors in saved velocity tensors. Second, the energy and enstrophy
balances measure whether the learned collision introduces the intended
effective dissipation without unstable entropy growth. Third, inertial-range
statistics test whether the stabilized coarse-grid dynamics preserve the
expected cascade phenomenology rather than merely damping the flow.

For comparing a learned entropic closure against a baseline solver, we use the
following hydrodynamic checks:
\begin{enumerate}[leftmargin=*]
    \item \textbf{Energy consistency:} compare $K(t)$ and
    $\sum_n E_n(t)$, including the zero shell when the mean flow is relevant.
    \item \textbf{Dissipation consistency:} compare
    $\nu_{\mathrm{eff},K}$ and $\nu_{\mathrm{eff},\Omega}$ with the nominal
    viscosity or with the effective viscosity induced by the closure.
    \item \textbf{Cascade consistency:} compare compensated spectra against
    the appropriate $-5/3$ or $-3$ guides, using the two-dimensional
    enstrophy-cascade guide only for two-dimensional hydrodynamic data.
    \item \textbf{Coherent-structure consistency:} compare vorticity RMS,
    enstrophy, palinstrophy, and structure-function exponents across time.
\end{enumerate}
The diagnostics are therefore not additional training objectives by
themselves; they are post-processing tools for verifying that the stabilized
rollout remains dynamically credible.

\subsection{Turbulence Diagnostics}
\label{appx:turbulence_diagnostics}
The metrics 
are evaluated at the same saved rollout time as the corresponding filtered reference.
Let $\mathbf{u}$ denote the predicted velocity, $\mathbf{u}_{\mathrm{ref}}$ the reference velocity, and $\omega=\partial_x u_y-\partial_y u_x$ the scalar vorticity computed from the velocity field.
The velocity error is $\|\mathbf{u}-\mathbf{u}_{\mathrm{ref}}\|_2/
\|\mathbf{u}_{\mathrm{ref}}\|_2$, and the vorticity error is $\|\omega-\omega_{\mathrm{ref}}\|_2/\|\omega_{\mathrm{ref}}\|_2$,
with norms taken over all grid points and velocity components where applicable. The vorticity correlation is the pointwise Pearson correlation
\begin{equation}
    \operatorname{corr}(\omega,\omega_{\mathrm{ref}})
    =
    \frac{
    \langle(\omega-\langle\omega\rangle)
    (\omega_{\mathrm{ref}}-\langle\omega_{\mathrm{ref}}\rangle)\rangle
    }{
    \sqrt{\langle(\omega-\langle\omega\rangle)^2\rangle
    \langle(\omega_{\mathrm{ref}}-\langle\omega_{\mathrm{ref}}\rangle)^2\rangle}
    } .
\end{equation} 
For the spectral columns, $E(k)$ is the shell-summed kinetic-energy spectrum defined in Appendix~\ref{appx:diagnostics_spectrum}.
The reported spectrum RMSE is an active-shell log-spectrum error,
\begin{equation}
    \begin{aligned}
    \operatorname{RMSE}_{E}
    &=
    \left[
    \frac{1}{|\mathcal{A}|}
    \sum_{k\in\mathcal{A}}
    \left(\log_{10}E(k)-\log_{10}E_{\mathrm{ref}}(k)\right)^2
    \right]^{1/2},\\
    \mathcal{A}
    &=
    \{k:E_{\mathrm{ref}}(k)>10^{-12}E_{\max}\},
    \qquad
    E_{\max}=\max_j E_{\mathrm{ref}}(j).
    \end{aligned}
\end{equation}
The high-wavenumber entry is not total energy; it is the ratio of high-$k$ energy fractions, 
\begin{equation}
    \frac{
    \sum_{k\geq N/4}E(k)/\sum_k E(k)
    }{
    \sum_{k\geq N/4}E_{\mathrm{ref}}(k)/\sum_k E_{\mathrm{ref}}(k)
    } ,
\end{equation}
so a value of one means matching the reference high-$k$ fraction, values above one indicate excess grid-scale energy, and values below one indicate depleted high-wavenumber content.

\begingroup

Let $\mathbf{u}_m(\mathbf{x},t)$ be the model velocity field, 
and let $\mathbf{u}_{\mathrm{ref}}$ be the box-filtered
reference velocity. Vorticity is computed spectrally from velocity as

\begin{equation}
    \omega
    =
    \partial_x u_y - \partial_y u_x .
\end{equation}

The relative velocity error is
\begin{equation}
    \varepsilon_u(t)
    =
    \frac{
        \left\|\mathbf{u}_m(\cdot,t)
        - \mathbf{u}_{\mathrm{ref}}(\cdot,t)\right\|_2
    }{
        \left\|\mathbf{u}_{\mathrm{ref}}(\cdot,t)\right\|_2
    }.
\end{equation}

The vorticity correlation is

\begin{equation}
    \rho_\omega(t)
    =
    \frac{
        \sum_{\mathbf{x}}
        \left(\omega_m-\overline{\omega}_m\right)
        \left(\omega_{\mathrm{ref}}-\overline{\omega}_{\mathrm{ref}}\right)
    }{
        \left[
        \sum_{\mathbf{x}}\left(\omega_m-\overline{\omega}_m\right)^2
        \sum_{\mathbf{x}}
        \left(\omega_{\mathrm{ref}}-\overline{\omega}_{\mathrm{ref}}\right)^2
        \right]^{1/2}
    }.
\end{equation}

The relative vorticity error is 

\begin{equation}
    \varepsilon_\omega(t)
    =
    \frac{
        \left\|\omega_m(\cdot,t)
        - \omega_{\mathrm{ref}}(\cdot,t)\right\|_2
    }{
        \left\|\omega_{\mathrm{ref}}(\cdot,t)\right\|_2
    }.
\end{equation}

Here the discrete norm is the RMS norm over grid points and velocity components where applicable. 

The pointwise vorticity correlation measures phase and coherent-structure fidelity rather than only global statistics. 

The shell-averaged energy spectrum uses the Fourier-shell convention in
Appendix~\ref{appx:diagnostics_spectrum},

\begin{equation}
    E_n
    =
    \sum_{\operatorname{round}(|\mathbf{m}|)=n}
    \frac{1}{2V^2}
    \left|\widehat{u}_i(\mathbf{m})\right|^2,
\end{equation}

with an analogous enstrophy spectrum

\begin{equation}
    Z_n
    =
    \sum_{\operatorname{round}(|\mathbf{m}|)=n}
    \frac{1}{2V^2}
    \left|\widehat{\omega}(\mathbf{m})\right|^2.
\end{equation}

Two-point correlations measure spatial structure. With
$\mathbf{u}'=\mathbf{u}-\langle\mathbf{u}\rangle$ and
$\omega'=\omega-\langle\omega\rangle$, we use

\begin{equation}
    C_{uu}(r)
    =
    \frac{
        \left\langle
            \mathbf{u}'(\mathbf{x})\cdot
            \mathbf{u}'(\mathbf{x}+\mathbf{r})
        \right\rangle
    }{
        \left\langle |\mathbf{u}'(\mathbf{x})|^2 \right\rangle
    },
\end{equation}

and

\begin{equation}
    C_{\omega\omega}(r)
    =
    \frac{
        \left\langle
            \omega'(\mathbf{x})
            \omega'(\mathbf{x}+\mathbf{r})
        \right\rangle
    }{
        \left\langle |\omega'(\mathbf{x})|^2 \right\rangle
    }.
\end{equation}

\endgroup

\FloatBarrier

\section{Experimental Results}\label{app:results}

\newcommand{\appforecastpanel}[3]{%
  \begin{tikzpicture}
    \node[inner sep=0, anchor=south west] (img) {\includegraphics[width=0.86\textwidth]{#3}};
    \node[anchor=north west, inner sep=1pt, font=\small\bfseries] at (img.north west) {(#1)};
    \node[anchor=south, rotate=90, inner sep=0pt, font=\small] at ([xshift=-6pt]img.west) {#2};
  \end{tikzpicture}}
\newcommand{\appspectrapanel}[3][0.49\textwidth]{%
  \begin{tikzpicture}
    \node[inner sep=0, anchor=south west] (img) {\includegraphics[width=#1]{#3}};
    \node[anchor=north west, inner sep=1pt, font=\small\bfseries] at (img.north west) {(#2)};
  \end{tikzpicture}}

We report forecasts for the doubly periodic shear layer at $\mathrm{Re}=50{,}000$ (DPSL-50K)
and $\mathrm{Re}=100{,}000$ (DPSL-100K), and the Taylor Green vortex at
$\mathrm{Re}=50{,}000$ (TGV).
For all experiments, the reference is the $512^2$ numerical ELBM simulation downsampled to each grid following the techniques described in Appendix~\ref{app:box_filter}.
Training forecasts start at $t_0=0$; testing forecasts start at $t_0=T/2$, where $T$ is the rollout length.
Figures show seed 2025; tables report \emph{entropically-stabilized NeurDE} as mean $\pm$ standard deviation over four seeds and ELBM as one deterministic run.

\subsection{Doubly periodic shear layer at Re=50,000}
\label{app:results_dp50k}

We present the results for the doubly periodic shear layer at $\mathrm{Re}=50{,}000$.
The real-space and spectral metrics are presented in Tables~\ref{tab:app_train_dp50k} and~\ref{tab:app_test_dp50k} for the training and testing datasets, respectively.
Figure~\ref{fig:app_forecast_dp50k} shows real-space forecasts for a single seed, and Fig.~\ref{fig:app_spectra_dp50k} shows the spectral metrics.
These results are visually indistinguishable from those of the other seeds, except for a single case described below.

For the model trained with seed 576268, the emulation diverges noticeably, but still maintains the primary features.
This is the only instance where any \emph{entropically-stabilized NeurDE} emulation diverges from the expected results.
We highlight this example in Appendix~\ref{app:dpsl-50k_divergence}.

\begin{table}[!htbp]
  \centering\small
  \caption{Real-space and spectral metrics for doubly periodic shear layer at $\mathrm{Re}=50{,}000$. We compare emulation metrics from the training data ($t_0=0$) at the final lead time against the downsampled $512^2$ reference. \emph{Entropically-stabilized NeurDE} results are the mean $\pm$ standard deviation over four seeds; ELBM results are from one deterministic run. The "Faithful Emulation" reports whether the rollout reproduces the expected structures. Field fidelity: relative $L^2$ error of velocity and vorticity $\omega$, and vorticity correlation. Spectrum diagnostics: RMSE of $\log_{10}E(k)$ over active shells, and high-$k$ energy and palinstrophy relative to the reference. Bold marks the better method per grid.}
  \label{tab:app_train_dp50k}
  \resizebox{\textwidth}{!}{
\setlength{\tabcolsep}{2.8pt}
    \begin{tabular}{@{}clccccccc@{}}
        \toprule
        \multirow{2}{*}{Grid} & \multicolumn{1}{c}{\multirow{2}{*}{Method}} & \multirow{2}{*}{\begin{tabular}{@{}c@{}}Faithful\\Emulation\end{tabular}} & \multicolumn{3}{c}{Field Fidelity} & \multicolumn{3}{c}{Spectrum Diagnostics} \\
        \cmidrule(lr){4-6}\cmidrule(l){7-9}
        & & & Vel. err $\downarrow$ & $\omega$ corr. $\uparrow$ & $\omega$ err $\downarrow$ & Spec. RMSE$^\dagger$ $\downarrow$ & High-$k$/ref. $\to1$ & $P$/ref. $\to1$ \\
        \midrule
        \multirow{2}{*}{$128^2$} & NeurDE & \textcolor{green!50!black}{3.5/4} & $\bf{0.109 \pm 0.044}$ & $\bf{0.806 \pm 0.083}$ & $\bf{0.59 \pm 0.12}$ & $3.37 \pm 0.88$ & $0.051 \pm 0.062$ & $0.133 \pm 0.080$ \\
         & ELBM & \textcolor{red}{0/1} & $0.259$ & $0.600$ & $0.92$ & $\bf{0.967}$ & $\bf{0.48}$ & $\bf{0.540}$ \\
        \midrule
        \multirow{2}{*}{$64^2$} & NeurDE & \textcolor{green!50!black}{4/4} & $\bf{0.144 \pm 4.9{\times}10^{-3}}$ & $\bf{0.793 \pm 8.8{\times}10^{-3}}$ & $\bf{0.614 \pm 0.011}$ & $3.15 \pm 0.23$ & $0.041 \pm 0.012$ & $0.0716 \pm 0.0084$ \\
         & ELBM & \textcolor{red}{0/1} & $0.381$ & $0.582$ & $0.961$ & $\bf{0.479}$ & $\bf{1.5}$ & $\bf{1.20}$ \\
        \midrule
        \multirow{2}{*}{$32^2$} & NeurDE & \textcolor{green!50!black}{4/4} & $\bf{0.256 \pm 0.019}$ & $\bf{0.725 \pm 0.017}$ & $\bf{0.690 \pm 0.018}$ & $2.74 \pm 0.20$ & $\bf{0.098 \pm 0.046}$ & $\bf{0.093 \pm 0.025}$ \\
         & ELBM & \textcolor{red}{0/1} & $0.551$ & $0.456$ & $1.13$ & $\bf{0.767}$ & $3.4$ & $2.6$ \\
        \midrule
        \multirow{2}{*}{$16^2$} & NeurDE & \textcolor{green!50!black}{4/4} & $\bf{0.550 \pm 0.012}$ & $\bf{0.610 \pm 7.8{\times}10^{-3}}$ & $\bf{0.792 \pm 6.0{\times}10^{-3}}$ & $2.46 \pm 0.26$ & $0.034 \pm 0.013$ & $\bf{0.070 \pm 0.012}$ \\
         & ELBM & \textcolor{red}{0/1} & $0.720$ & $0.212$ & $1.29$ & $\bf{0.987}$ & $\bf{1.9}$ & $2.2$ \\
        \bottomrule
    \end{tabular}
}
\end{table}
\begin{table}[!htbp]
  \centering\small
  \caption{Real-space and spectral metrics for doubly periodic shear layer at $\mathrm{Re}=50{,}000$. We compare emulation metrics from the testing data ($t_0=T/2$) at the final lead time, against the downsampled $512^2$ reference. \emph{Entropically-stabilized NeurDE} results are the mean $\pm$ standard deviation over four seeds; ELBM results are from one deterministic run. The "Faithful Emulation" reports whether the rollout reproduces the expected structures. Field fidelity: relative $L^2$ error of velocity and vorticity $\omega$, and vorticity correlation. Spectrum diagnostics: RMSE of $\log_{10}E(k)$ over active shells, and high-$k$ energy and palinstrophy relative to the reference. Bold marks the better method per grid.}
  \label{tab:app_test_dp50k}
  \resizebox{\textwidth}{!}{
\setlength{\tabcolsep}{2.8pt}
    \begin{tabular}{@{}clccccccc@{}}
        \toprule
        \multirow{2}{*}{Grid} & \multicolumn{1}{c}{\multirow{2}{*}{Method}} & \multirow{2}{*}{\begin{tabular}{@{}c@{}}Faithful\\Emulation\end{tabular}} & \multicolumn{3}{c}{Field Fidelity} & \multicolumn{3}{c}{Spectrum Diagnostics} \\
        \cmidrule(lr){4-6}\cmidrule(l){7-9}
        & & & Vel. err $\downarrow$ & $\omega$ corr. $\uparrow$ & $\omega$ err $\downarrow$ & Spec. RMSE$^\dagger$ $\downarrow$ & High-$k$/ref. $\to1$ & $P$/ref. $\to1$ \\
        \midrule
        \multirow{2}{*}{$128^2$} & NeurDE & \textcolor{green!50!black}{4/4} & $0.045 \pm 0.011$ & $0.959 \pm 0.010$ & $0.288 \pm 0.039$ & $3.64 \pm 0.70$ & $0.015 \pm 0.024$ & $0.141 \pm 0.054$ \\
         & ELBM & \textcolor{green!50!black}{1/1} & $\bf{0.021}$ & $\bf{0.970}$ & $\bf{0.244}$ & $\bf{0.286}$ & $\bf{0.96}$ & $\bf{0.921}$ \\
        \midrule
        \multirow{2}{*}{$64^2$} & NeurDE & \textcolor{green!50!black}{4/4} & $0.0875 \pm 0.0015$ & $0.935 \pm 9.9{\times}10^{-4}$ & $0.375 \pm 3.0{\times}10^{-3}$ & $3.56 \pm 0.12$ & $0.0051 \pm 0.0018$ & $0.107 \pm 2.4{\times}10^{-3}$ \\
         & ELBM & \textcolor{green!50!black}{1/1} & $\bf{0.0548}$ & $\bf{0.939}$ & $\bf{0.349}$ & $\bf{0.308}$ & $\bf{1.4}$ & $\bf{1.17}$ \\
        \midrule
        \multirow{2}{*}{$32^2$} & NeurDE & \textcolor{green!50!black}{4/4} & $0.146 \pm 9.1{\times}10^{-3}$ & $0.904 \pm 6.8{\times}10^{-3}$ & $0.456 \pm 0.016$ & $2.89 \pm 0.23$ & $0.0125 \pm 0.0081$ & $0.122 \pm 0.013$ \\
         & ELBM & \textcolor{green!50!black}{1/1} & $\bf{0.0697}$ & $\bf{0.945}$ & $\bf{0.328}$ & $\bf{0.440}$ & $\bf{1.25}$ & $\bf{1.16}$ \\
        \midrule
        \multirow{2}{*}{$16^2$} & NeurDE & \textcolor{green!50!black}{4/4} & $0.267 \pm 7.6{\times}10^{-3}$ & $0.846 \pm 6.3{\times}10^{-3}$ & $0.569 \pm 9.6{\times}10^{-3}$ & $2.34 \pm 0.14$ & $0.0254 \pm 0.0064$ & $0.0812 \pm 0.0048$ \\
         & ELBM & \textcolor{green!50!black}{1/1} & $\bf{0.209}$ & $\bf{0.901}$ & $\bf{0.435}$ & $\bf{0.348}$ & $\bf{0.783}$ & $\bf{0.709}$ \\
        \bottomrule
    \end{tabular}
}
\end{table}

\begin{figure}[t]
  \centering
  \appforecastpanel{a}{$128^2$}{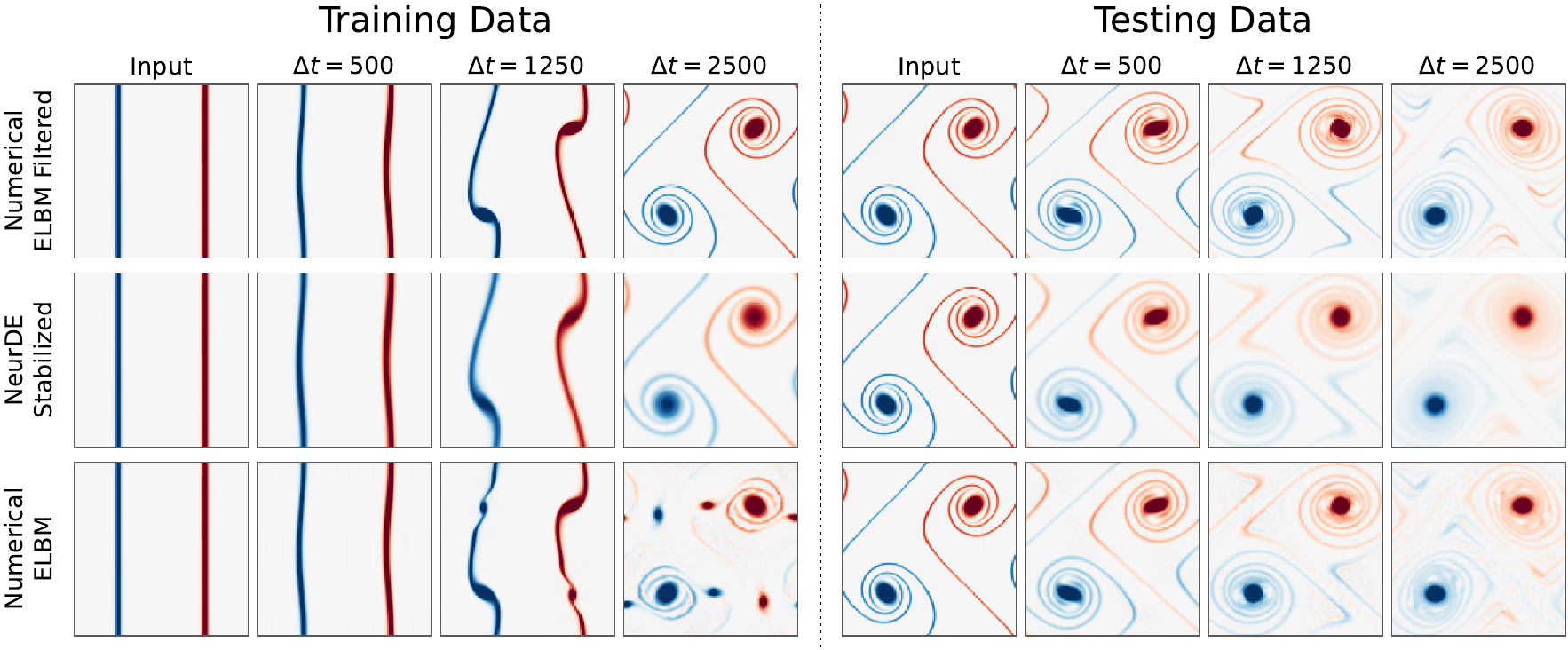}\par\vspace{2pt}
  \appforecastpanel{b}{$64^2$}{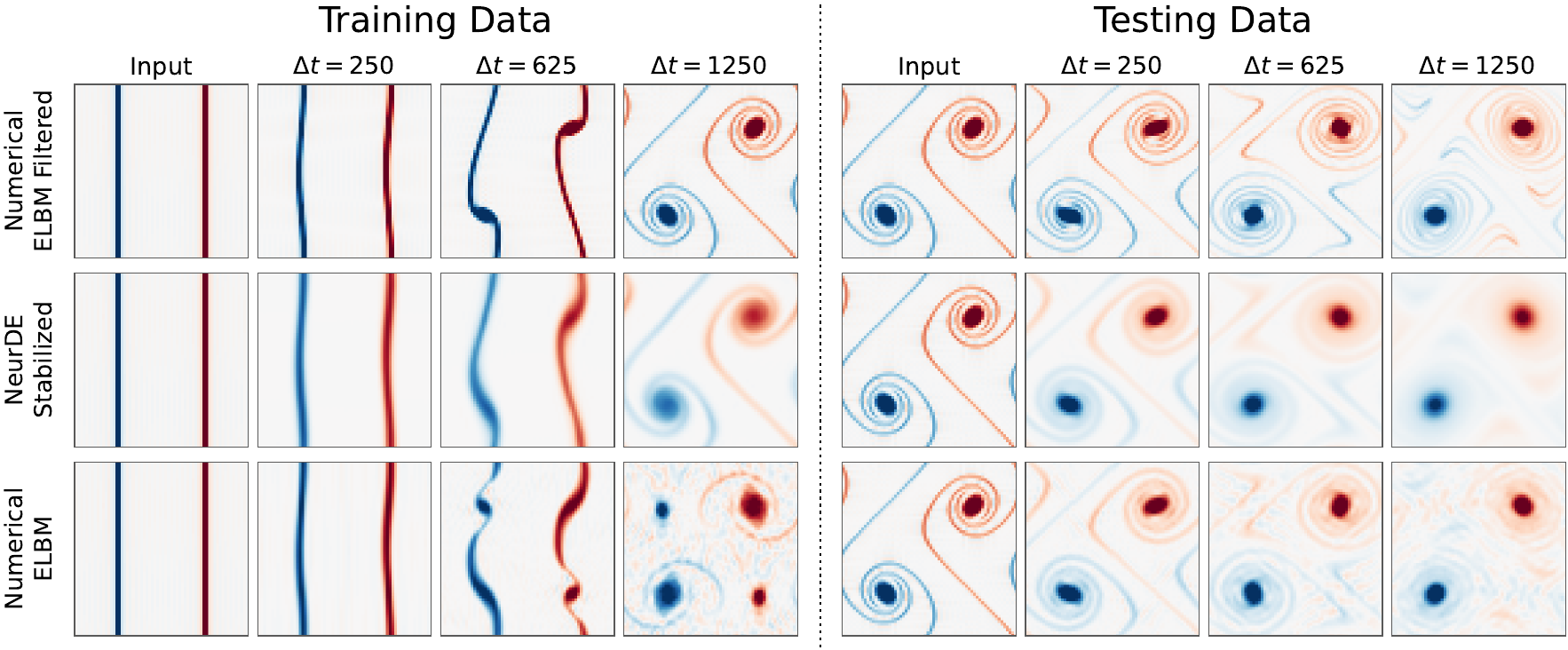}\par\vspace{2pt}
  \appforecastpanel{c}{$32^2$}{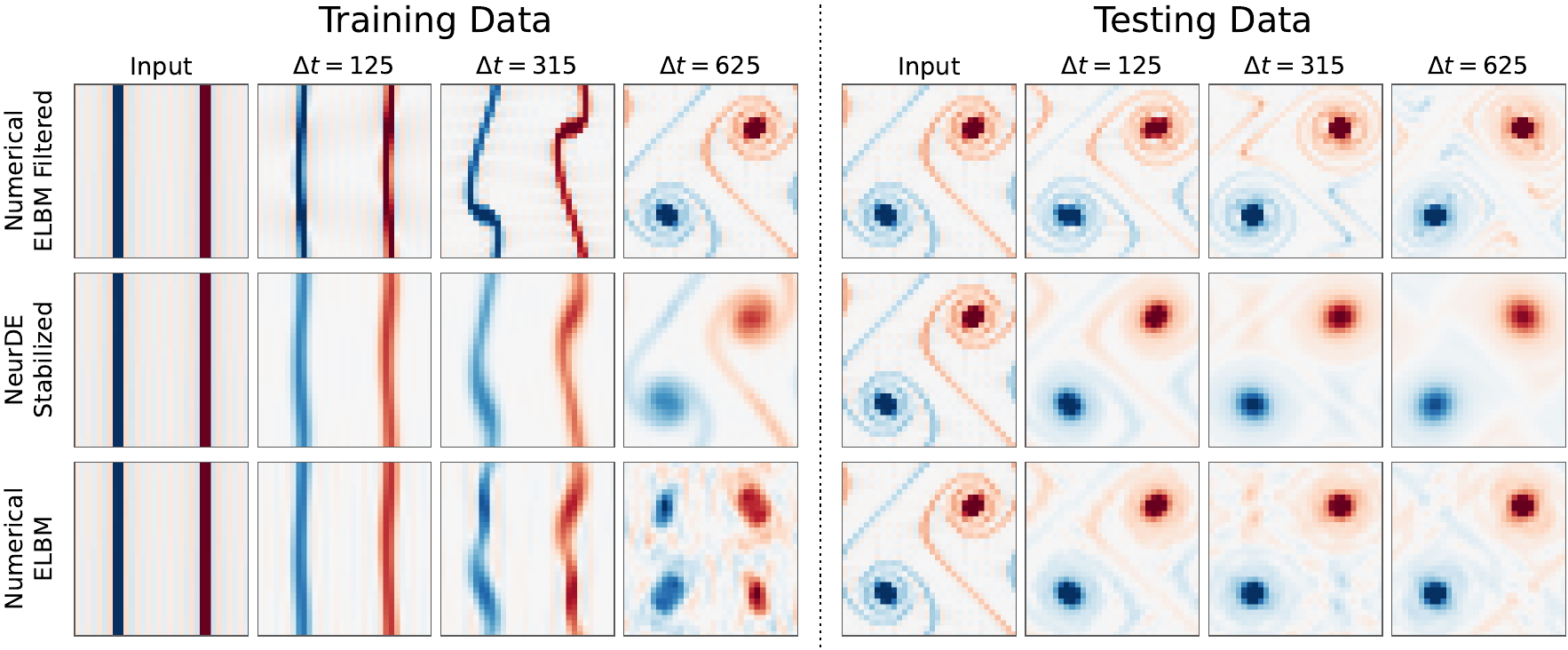}\par\vspace{2pt}
  \appforecastpanel{d}{$16^2$}{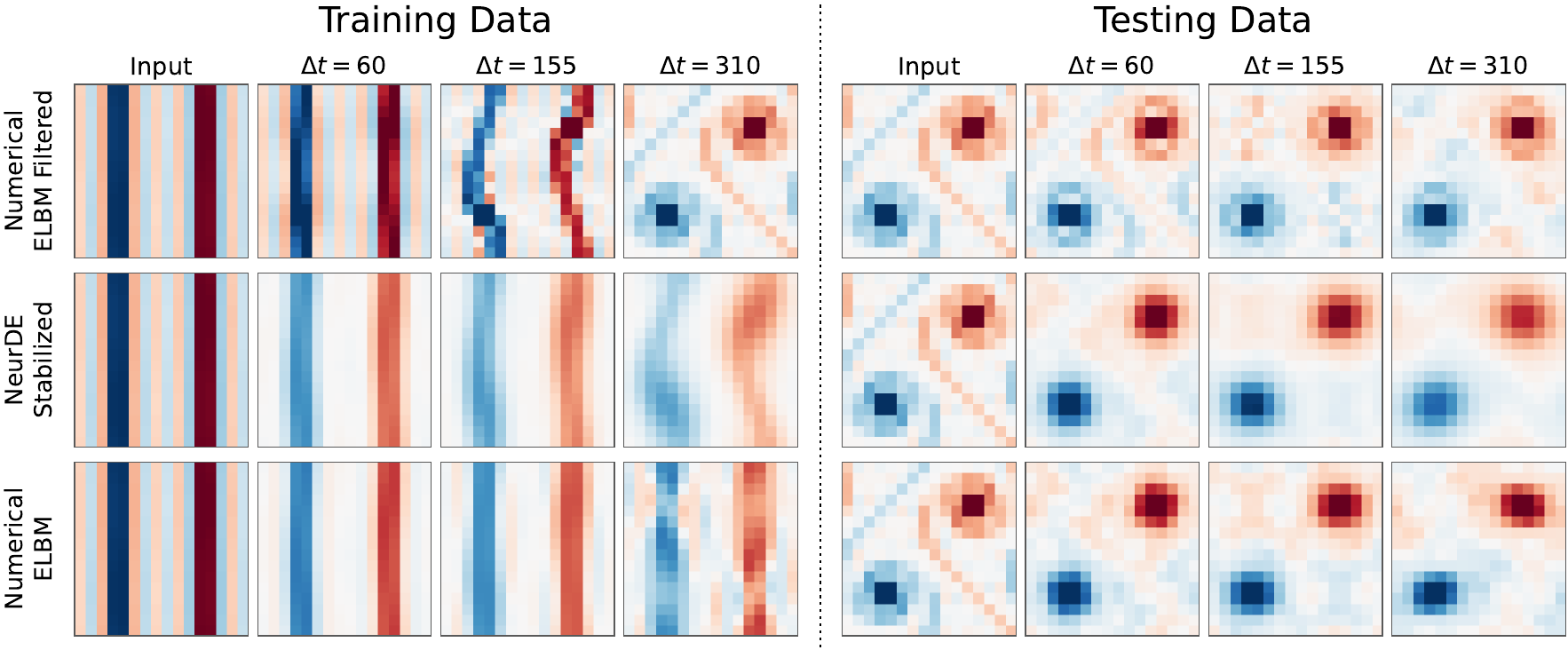}
  \caption{Doubly periodic shear layer at $\mathrm{Re}=50{,}000$: vorticity forecasts on grids (a)~$128^2$, (b)~$64^2$, (c)~$32^2$, and (d)~$16^2$; the grid size is also marked to the left of each panel. In each panel, the rows are the downsampled $512^2$ reference (top), \emph{entropically-stabilized NeurDE} (middle), and numerical ELBM run at that grid (bottom). The left block starts from the training data ($t_0=0$), the right block from the testing data ($t_0=T/2$); columns are lead times $\Delta t$ in solver steps.}
  \label{fig:app_forecast_dp50k}
\end{figure}

\begin{figure}[t]
  \centering
  \appspectrapanel{a}{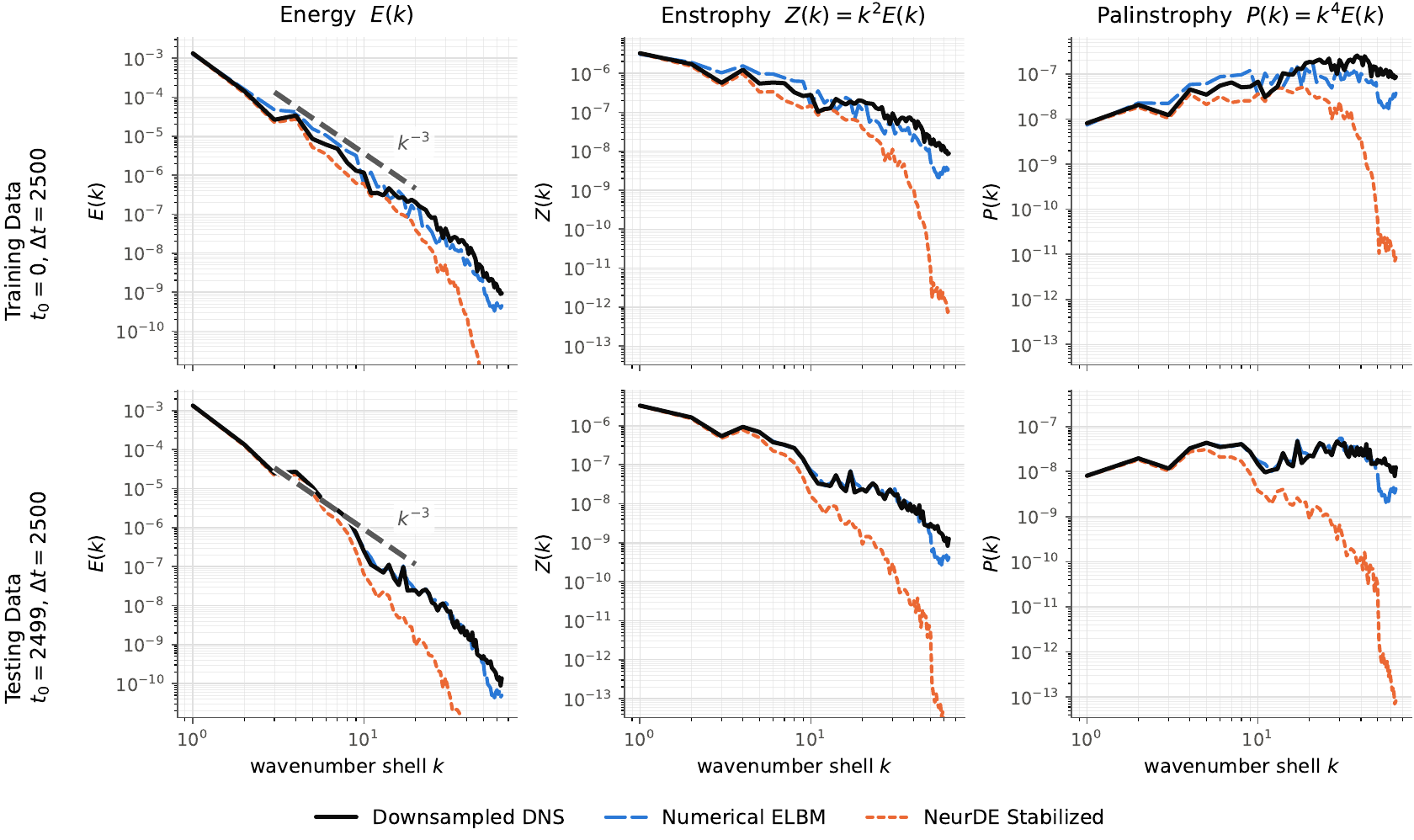}\hfill
  \appspectrapanel{b}{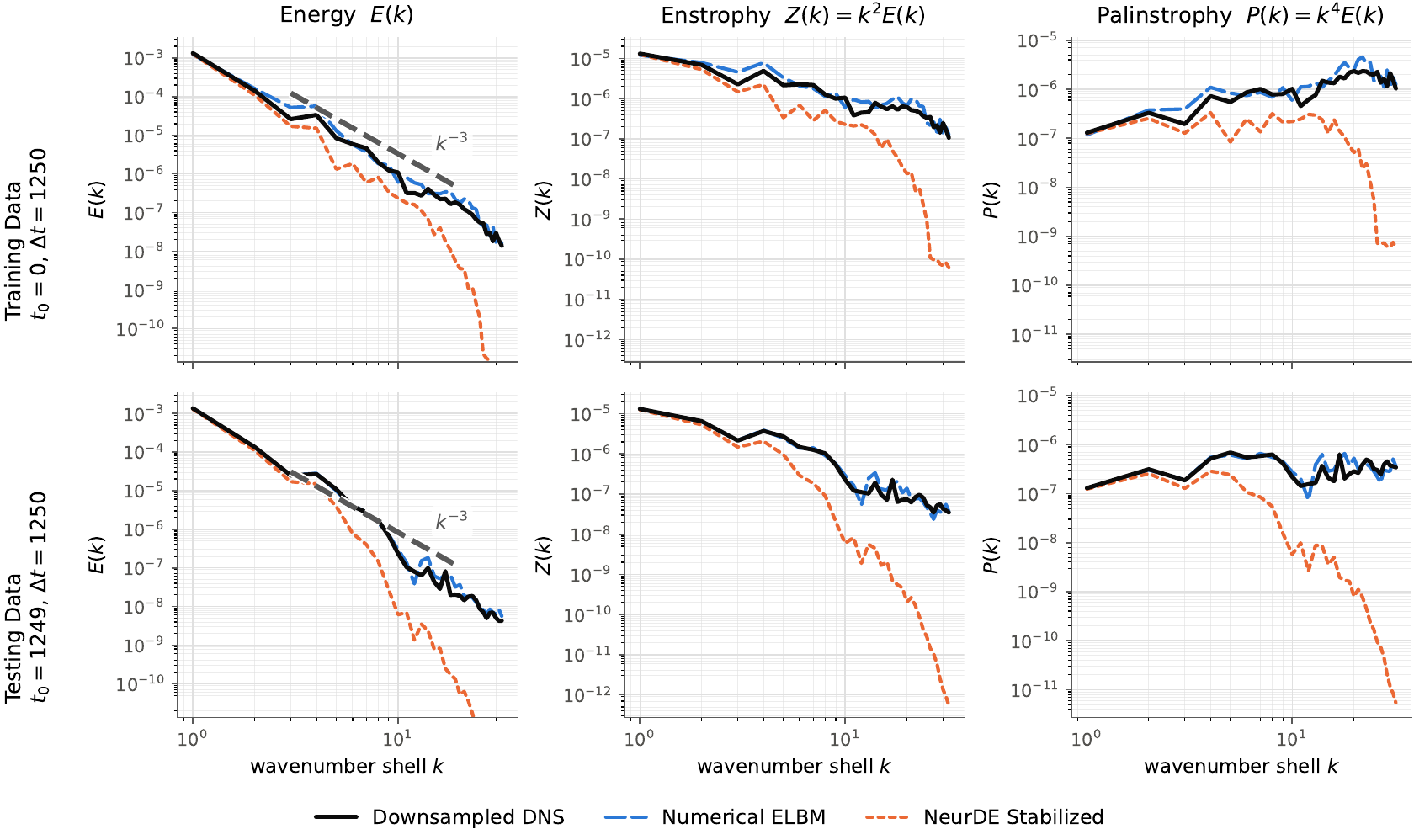}\par\smallskip
  \appspectrapanel{c}{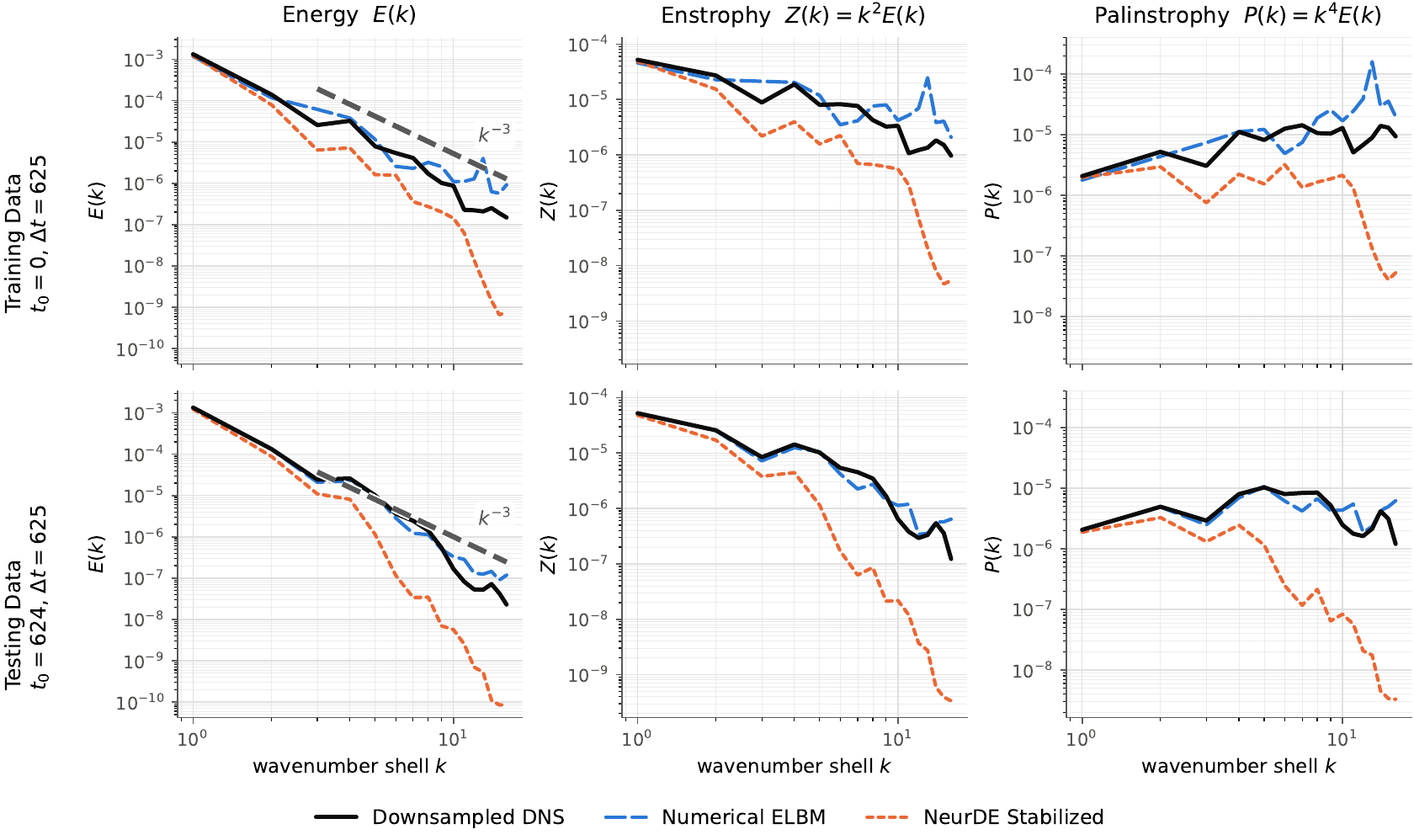}\hfill
  \appspectrapanel{d}{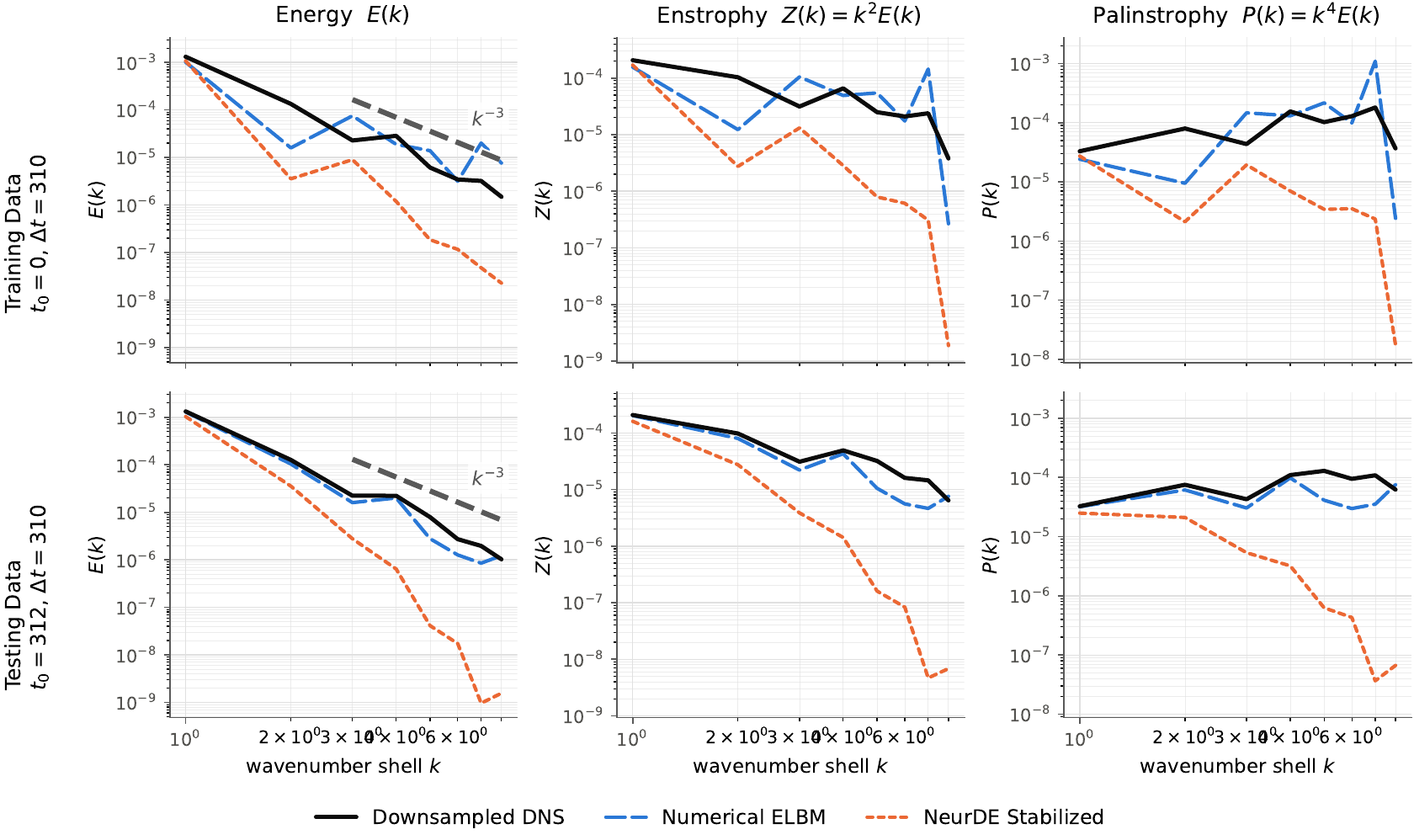}
  \caption{Doubly periodic shear layer at $\mathrm{Re}=50{,}000$: shell-averaged spectra at the final lead time on grids (a)~$128^2$, (b)~$64^2$, (c)~$32^2$, and (d)~$16^2$. Columns are the energy $E(k)$, enstrophy $Z(k)=k^2E(k)$, and palinstrophy $P(k)=k^4E(k)$ spectra, where $k$ is the wavenumber shell; rows are forecasts from the training (top) and testing (bottom) data. Curves compare the downsampled $512^2$ reference, numerical ELBM, and \emph{entropically-stabilized NeurDE}.}
  \label{fig:app_spectra_dp50k}
\end{figure}

\begin{figure}
    \centering
    \includegraphics[width=0.95\linewidth]{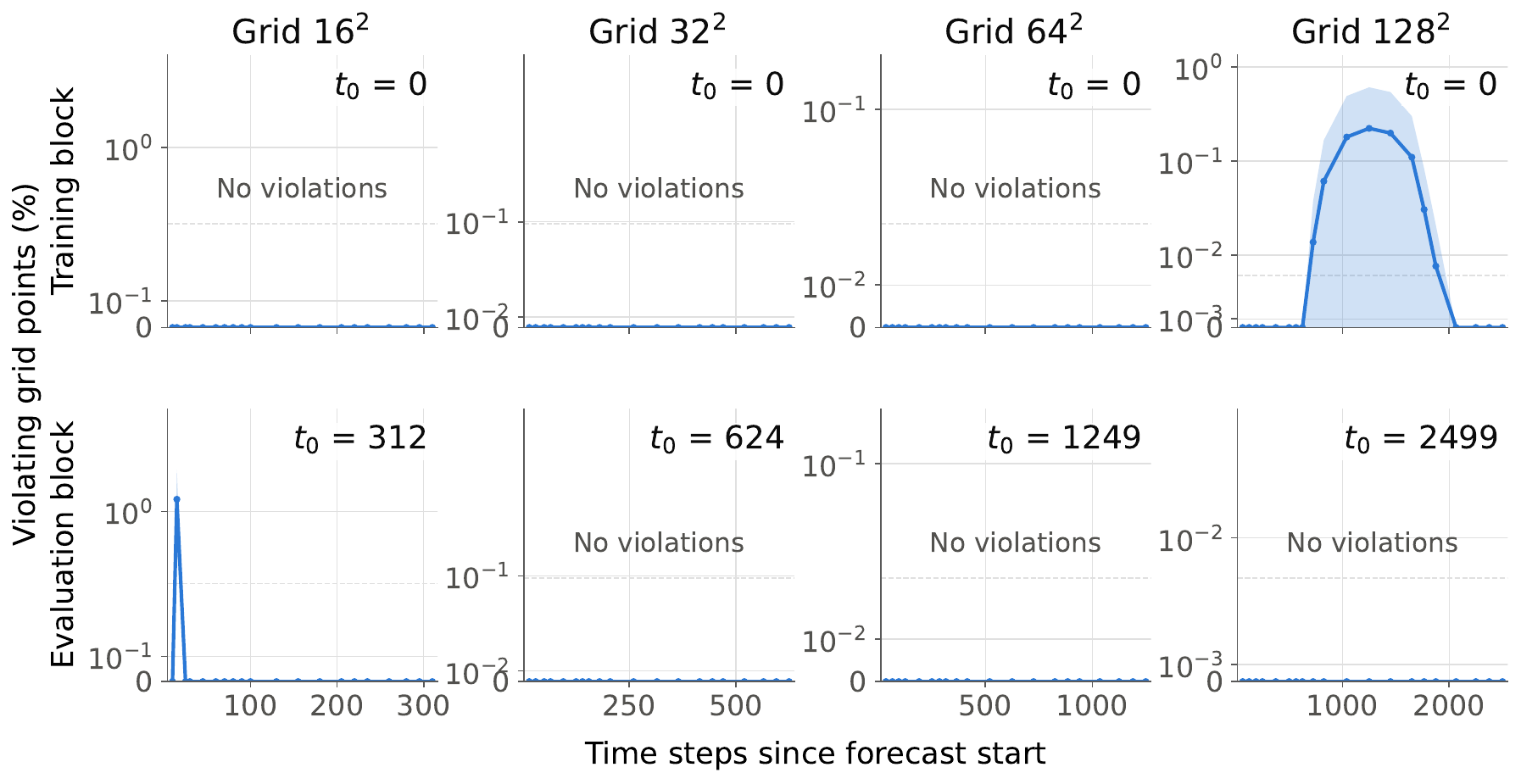}
    \caption{The percentage of pixels that violate the second law of thermodynamics over the \emph{NeurdE} emulation of the doubly periodic shear layer at $\mathrm{Re}=50{,}000$. We show the training and testing datasets for all grids. The mean and standard deviation are taken over all seeds.}
    \label{fig:app_entropy_dp50k}
\end{figure}

\FloatBarrier

\subsubsection{Single Seed Divergence\label{app:dpsl-50k_divergence}}
We highlight the emulation results of the \emph{entropically-stabilized NeurDE} model trained on DPSL-50K with seed 576268 and evaluated on the $128^2$ grid.
We note that this is the only noticeable real-space divergence amongst all the models evaluated on all the coarsened grids.
This includes this same model (seed 576268) evaluated on DPSL-50K on coarser grids.
Notably, despite this divergence, this model is still more stable than ELBM.

Figures~\ref{fig:app_dp50k_seed576268}a and b show the real-space and spectral components of the emulation.
At late times in train set emulation, \emph{entropically-stabilized NeurDE} begins to diverge from the ground truth.
While two erroneous vortices begin to emerge, it still captures the general features of the ground truth dynamics.
For ELBM, these vortices have already formed before $\Delta t = 1250$.
Therefore, while this specific \emph{entropically-stabilized NeurDE} model is less stable than all the others, it is still more stable than the ELBM comparison.

The spectral metrics of this emulation (Fig.~\ref{fig:app_dp50k_seed576268}b) perform better than the metrics for the emulation of seed 2025 (Fig.~\ref{fig:app_spectra_dp50k}a).
This further supports our claim that quantifying simulation and emulation quality of chaotic dynamics is very challenging.
Again, we see that improved spectral methods have led to significant divergence in the simulation compared to the 2025 seed.

\begin{figure}[p]
  \centering
  \appforecastpanel{a}{$128^2$}{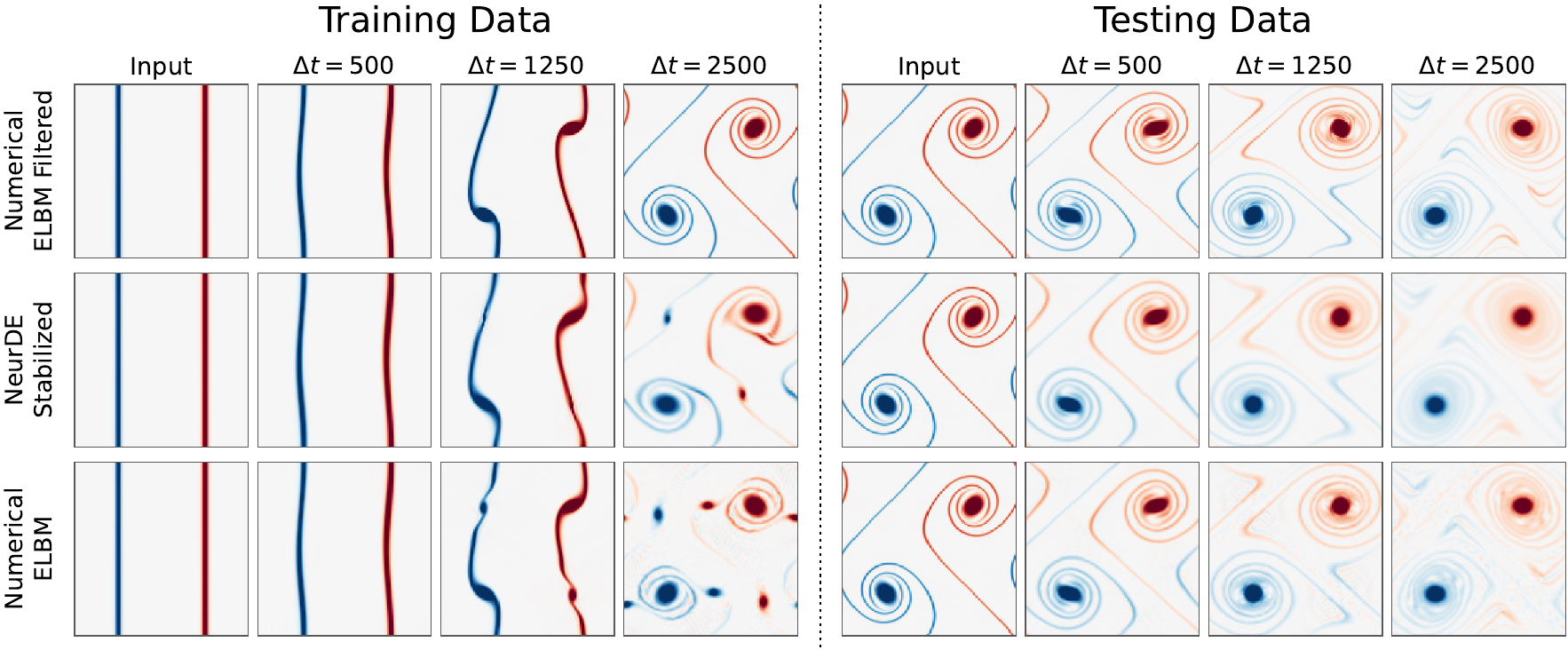}\par\vspace{4pt}
  \appspectrapanel[0.86\textwidth]{b}{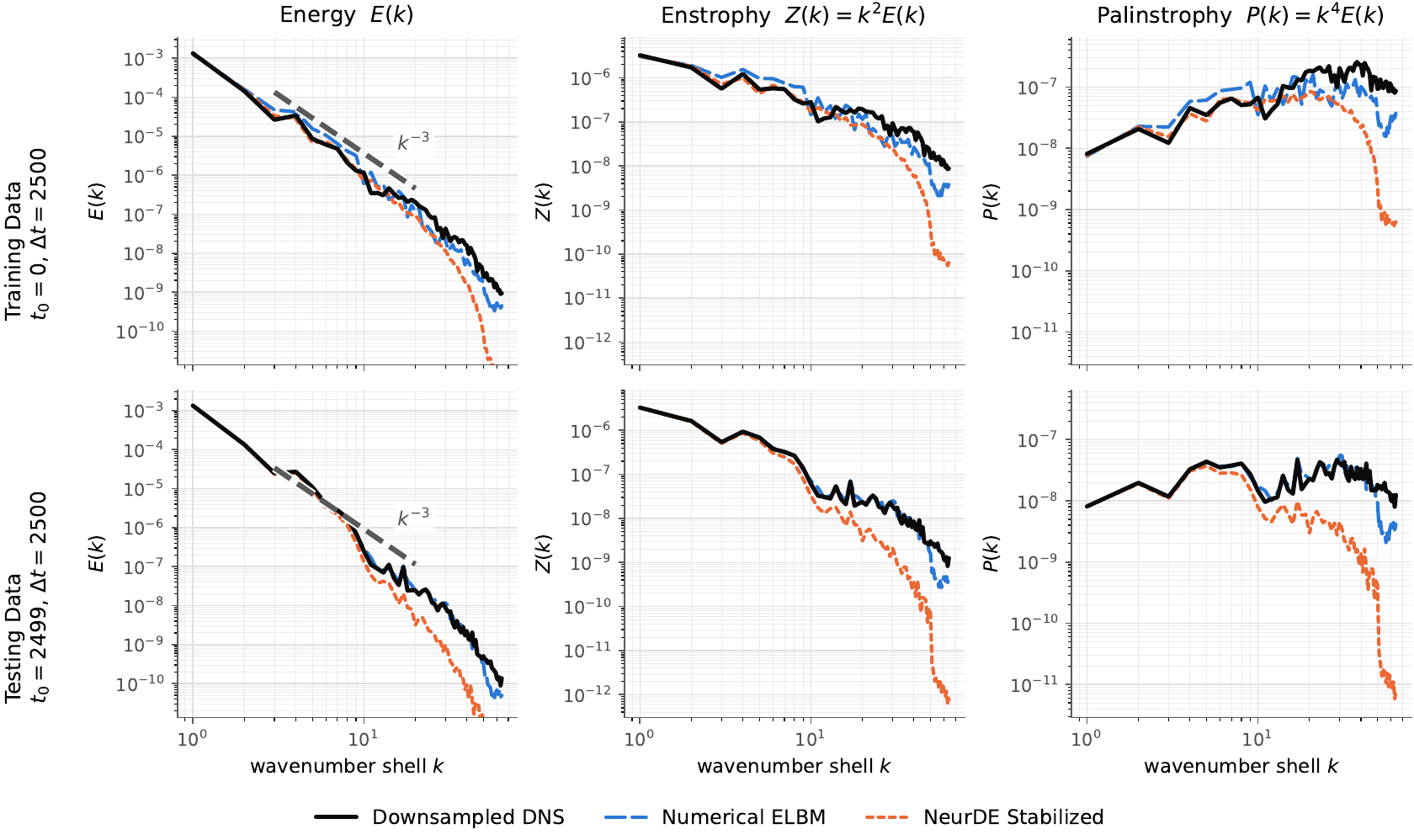}
  \caption{Doubly periodic shear layer at $\mathrm{Re}=50{,}000$, seed 576268 on the $128^2$ grid: (a) vorticity forecasts and (b) shell-averaged spectra at the final lead time, laid out as in Figs.~\ref{fig:app_forecast_dp50k} and~\ref{fig:app_spectra_dp50k}. The rollout stays finite, but its forecast from the training data departs from the other three seeds: at $\Delta t=2500$ its velocity error is $0.174$ (other seeds $0.080$--$0.096$) and its vorticity correlation is $0.683$ ($0.829$--$0.862$). It also keeps more energy at high $k$ than the other seeds (high-$k$/ref.\ $0.142$ against $0.007$--$0.034$). Its forecast from the testing data is in line with, and on the field metrics slightly better than, the other seeds.}
  \label{fig:app_dp50k_seed576268}
\end{figure}

\clearpage

\subsection{Doubly periodic shear layer at Re=100,000}
\label{app:results_dp100k}

We present the results for the doubly periodic shear layer at $\mathrm{Re}=100{,}000$.
The real-space and spectral metrics are presented in Tables~\ref{tab:app_train_dp100k} and~\ref{tab:app_test_dp100k} for the training and testing datasets, respectively.
Figure~\ref{fig:app_forecast_dp100k} shows real-space forecasts for a single seed, and Fig.~\ref{fig:app_spectra_dp100k} shows the spectral metrics.
These results are visually indistinguishable from those of the other seeds.

\begin{table}[!htbp]
  \centering\small
  \caption{Real-space and spectral metrics for doubly periodic shear layer at $\mathrm{Re}=100{,}000$. We compare emulation metrics from the training data ($t_0=0$) at the final lead time against the downsampled $512^2$ reference. \emph{Entropically-stabilized NeurDE} results are the mean $\pm$ standard deviation over four seeds; ELBM results are from one deterministic run. The "Faithful Emulation" reports whether the rollout reproduces the expected structures. Field fidelity: relative $L^2$ error of velocity and vorticity $\omega$, and vorticity correlation. Spectrum diagnostics: RMSE of $\log_{10}E(k)$ over active shells, and high-$k$ energy and palinstrophy relative to the reference. Bold marks the better method per grid.}
  \label{tab:app_train_dp100k}
  \resizebox{\textwidth}{!}{}
\end{table}
\begin{table}[!htbp]
  \centering\small
  \caption{Real-space and spectral metrics for doubly periodic shear layer at $\mathrm{Re}=100{,}000$. We compare emulation metrics from the testing data ($t_0=T/2$) at the final lead time against the downsampled $512^2$ reference. \emph{Entropically-stabilized NeurDE} results are the mean $\pm$ standard deviation over four seeds; ELBM results are from one deterministic run. The "Faithful Emulation" reports whether the rollout reproduces the expected structures. Field fidelity: relative $L^2$ error of velocity and vorticity $\omega$, and vorticity correlation. Spectrum diagnostics: RMSE of $\log_{10}E(k)$ over active shells, and high-$k$ energy and palinstrophy relative to the reference. Bold marks the better method per grid.}
  \label{tab:app_test_dp100k}
  \resizebox{\textwidth}{!}{}
\end{table}

\begin{figure}[p]
  \centering
  \appforecastpanel{a}{$128^2$}{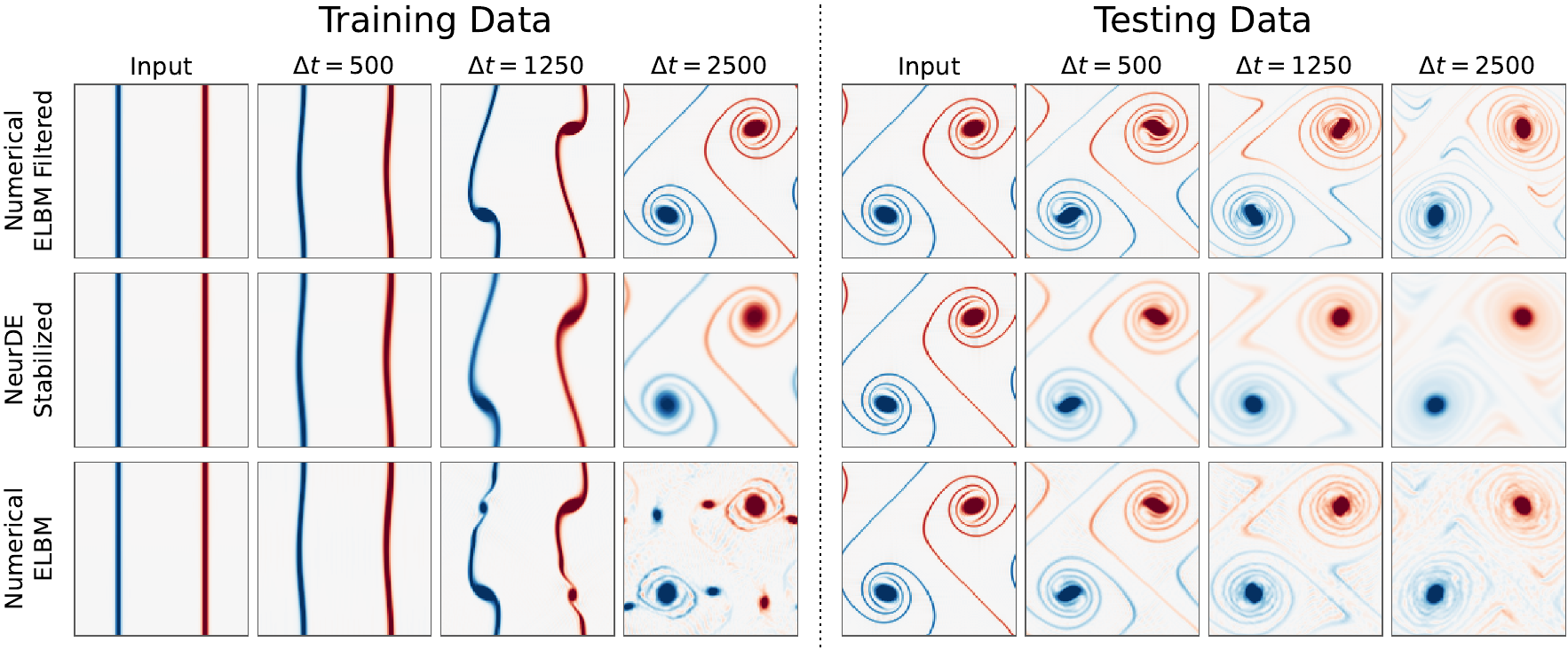}\par\vspace{2pt}
  \appforecastpanel{b}{$64^2$}{Figures/rollout_results/base/dp100k/tier-64/forecast_seed-2025.pdf}\par\vspace{2pt}
  \appforecastpanel{c}{$32^2$}{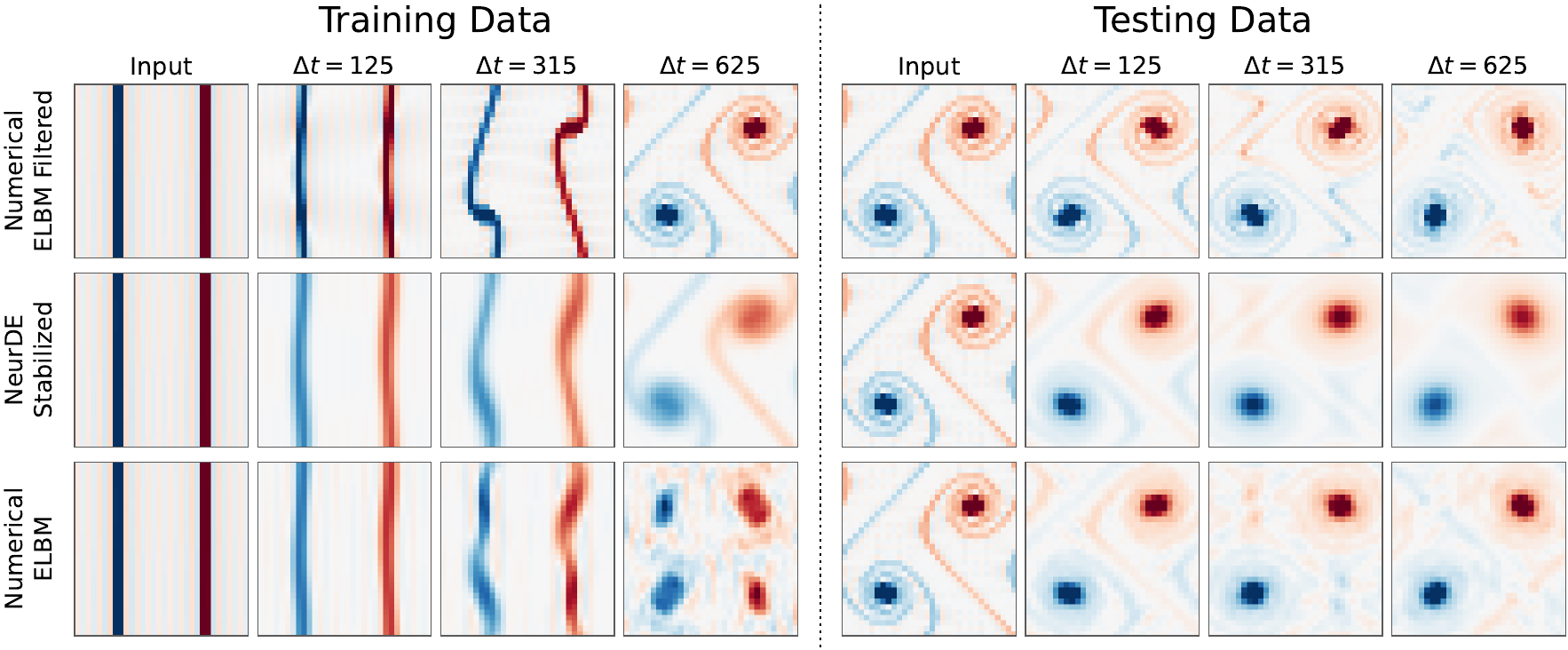}\par\vspace{2pt}
  \appforecastpanel{d}{$16^2$}{Figures/rollout_results/base/dp100k/tier-16/forecast_seed-2025.pdf}
  \caption{Doubly periodic shear layer at $\mathrm{Re}=100{,}000$: vorticity forecasts on grids (a)~$128^2$, (b)~$64^2$, (c)~$32^2$, and (d)~$16^2$; the grid size is also marked to the left of each panel. In each panel, the rows are the downsampled $512^2$ reference (top), \emph{entropically-stabilized NeurDE} (middle), and numerical ELBM run at that grid (bottom). The left block starts from the training data ($t_0=0$), the right block from the testing data ($t_0=T/2$); columns are lead times $\Delta t$ in solver steps.}
  \label{fig:app_forecast_dp100k}
\end{figure}

\begin{figure}[p]
  \centering
  \appspectrapanel{a}{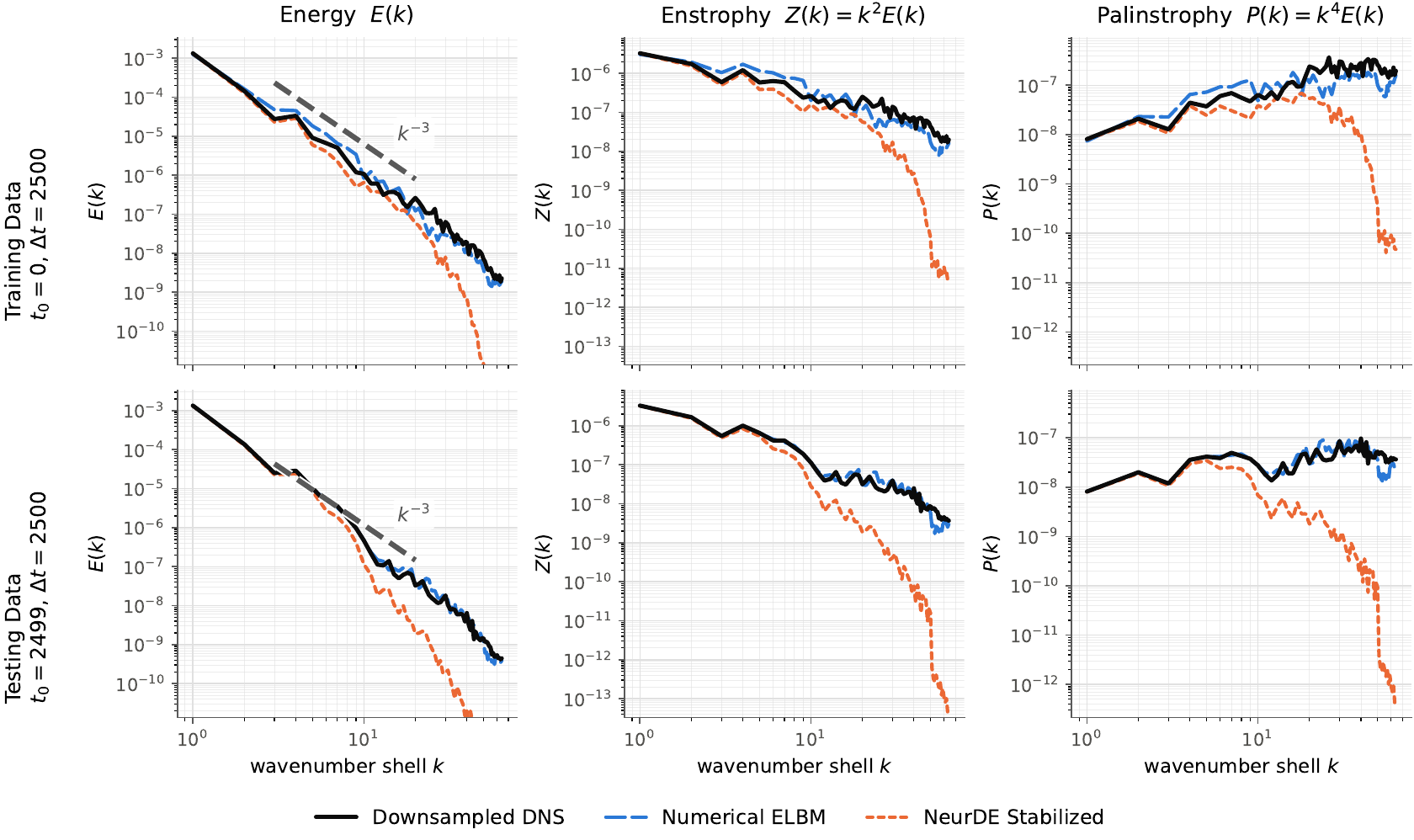}\hfill
  \appspectrapanel{b}{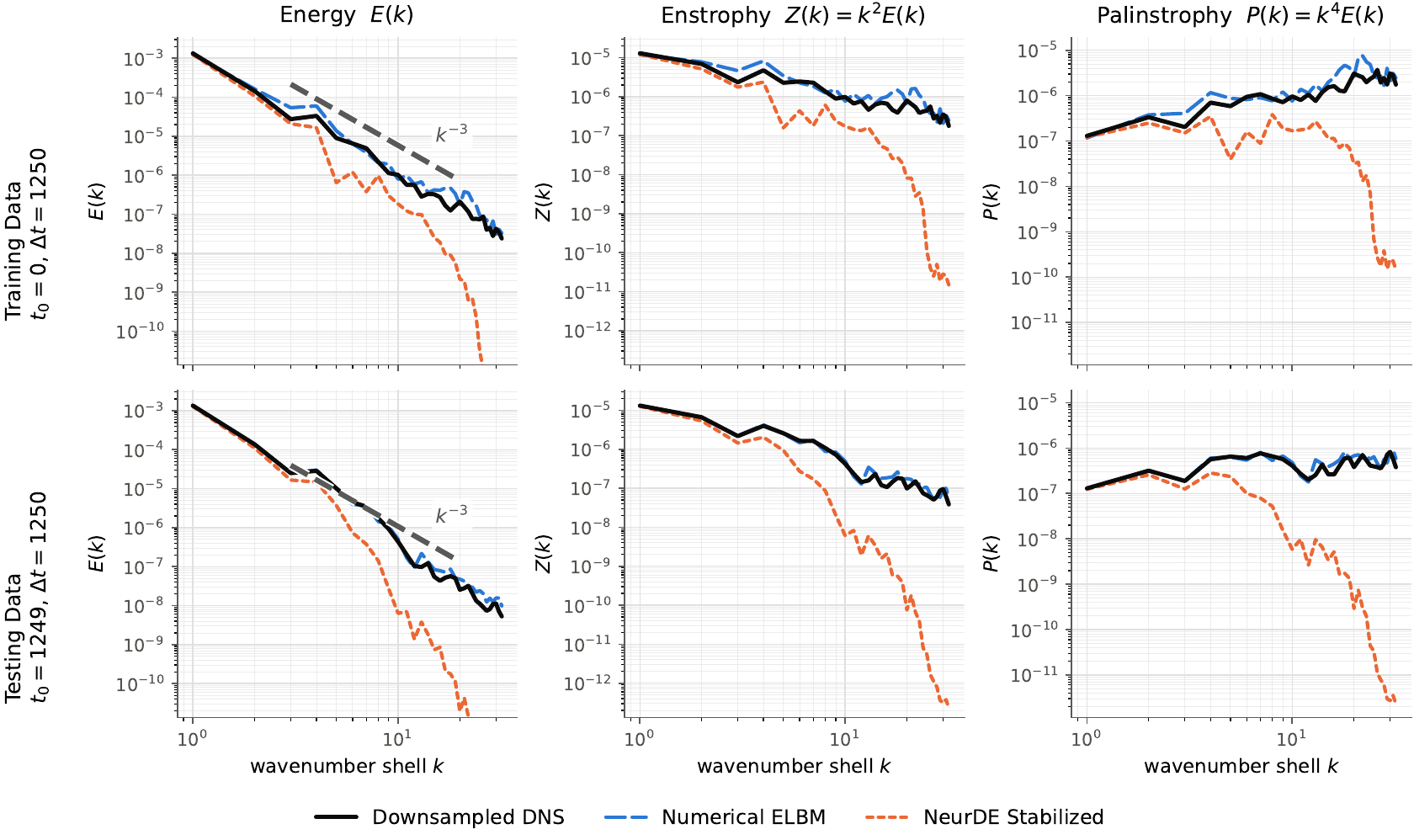}\par\smallskip
  \appspectrapanel{c}{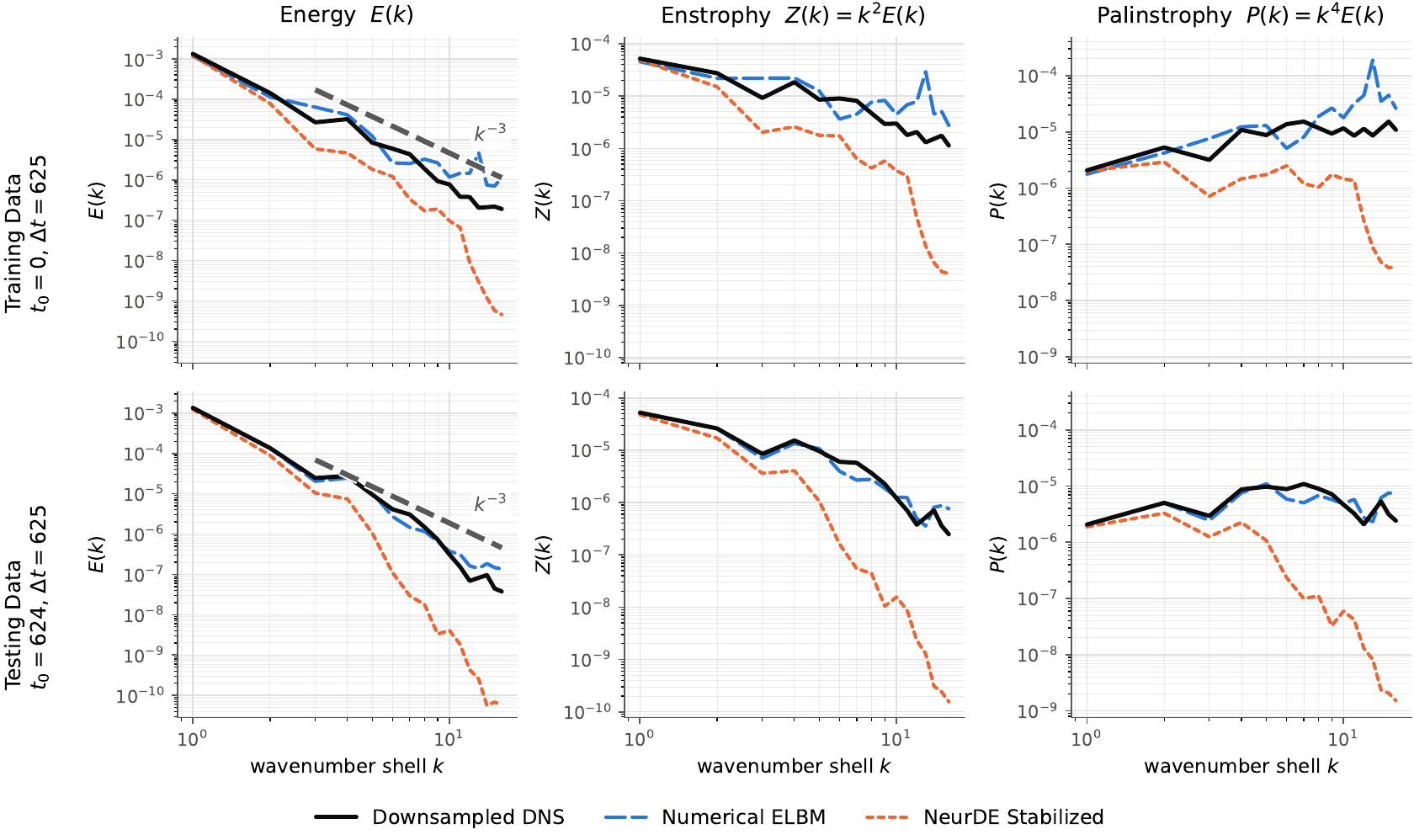}\hfill
  \appspectrapanel{d}{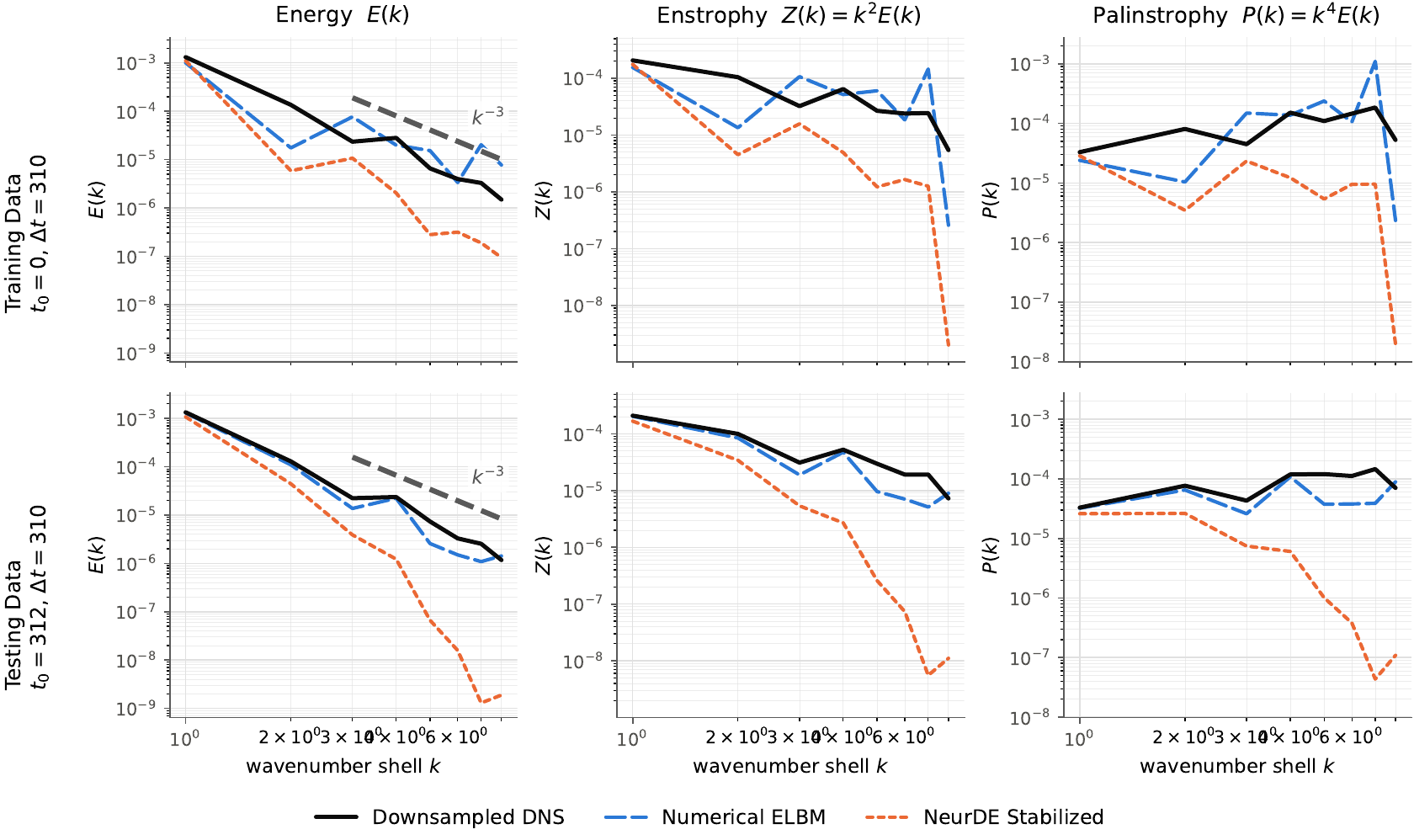}
  \caption{Doubly periodic shear layer at $\mathrm{Re}=100{,}000$: shell-averaged spectra at the final lead time on grids (a)~$128^2$, (b)~$64^2$, (c)~$32^2$, and (d)~$16^2$. Columns are the energy $E(k)$, enstrophy $Z(k)=k^2E(k)$, and palinstrophy $P(k)=k^4E(k)$ spectra, where $k$ is the wavenumber shell; rows are forecasts from the training (top) and testing (bottom) data. Curves compare the downsampled $512^2$ reference, numerical ELBM, and \emph{entropically-stabilized NeurDE}.}
  \label{fig:app_spectra_dp100k}
\end{figure}

\begin{figure}
    \centering
    \includegraphics[width=0.95\linewidth]{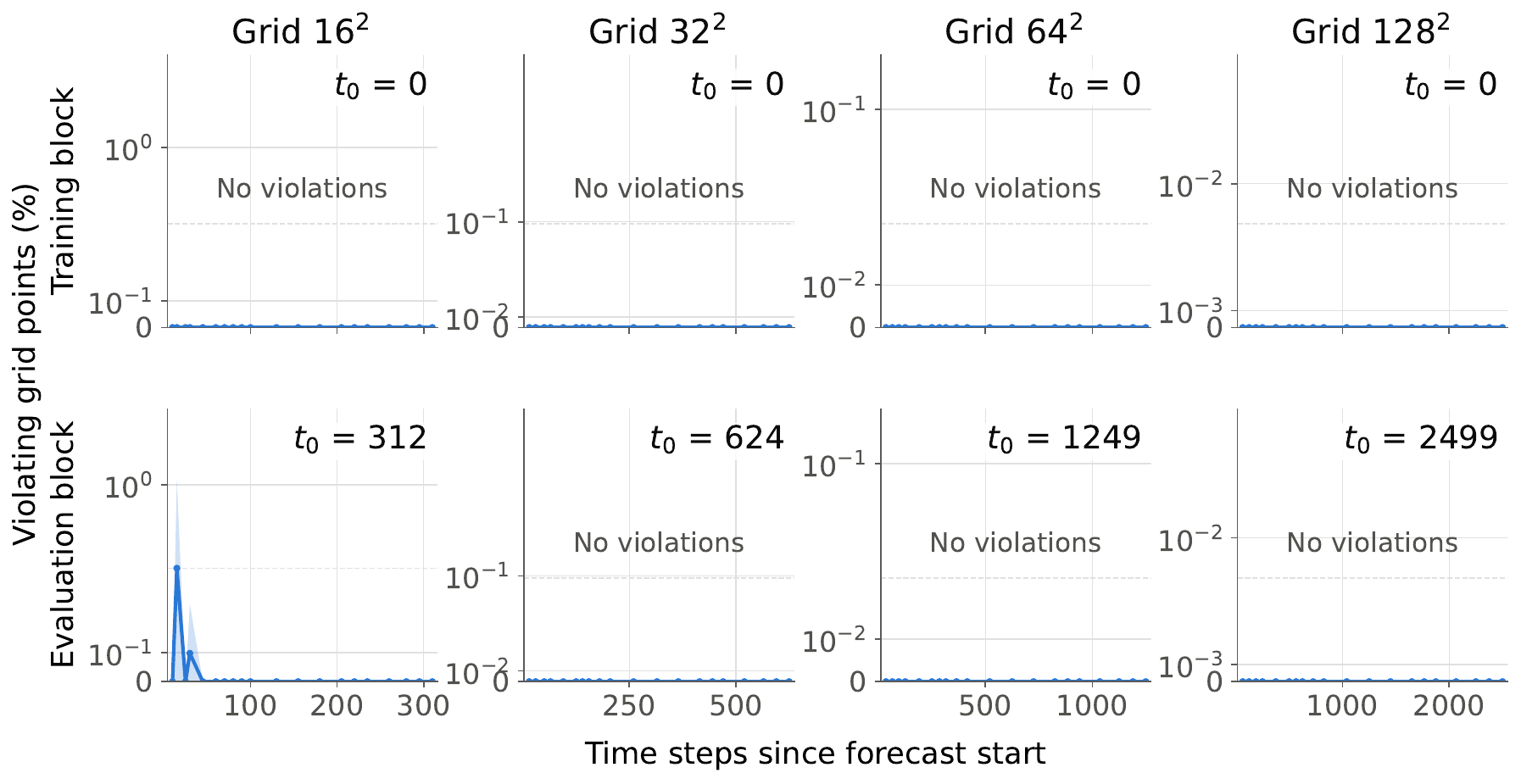}
    \caption{The percentage of pixels that violate the second law of thermodynamics over the \emph{NeurdE} emulation of the doubly periodic shear layer at $\mathrm{Re}=100{,}000$. We show the training and testing datasets for all grids. The mean and standard deviation are taken over all seeds.}
    \label{fig:app_entropy_dp100k}
\end{figure}

\clearpage

\subsection{Taylor Green Vortex at Re=50,000}
\label{app:results_tgv50k}

We present the results for the Taylor Green vortex experiment at $\mathrm{Re}=50{,}000$.
The real-space and spectral metrics are presented in Tables~\ref{tab:app_train_tgv50k} and~\ref{tab:app_test_tgv50k} for the training and testing datasets, respectively.
Figure~\ref{fig:app_forecast_tgv50k} shows real-space forecasts for a single seed, and Fig.~\ref{fig:app_spectra_tgv50k} shows the spectral metrics.
These results are visually indistinguishable from those of the other seeds.

The two spectrum diagnostics (Spec.\ RMSE and High-$k$/ref.) are left blank at every resolution.
This follows from the flow itself.
We initialize TGV with the single-mode field $u = U_0 \sin x \cos y$, $v = -U_0 \cos x \sin y$, which is an exact solution of the two-dimensional incompressible Navier--Stokes equations: the pressure gradient balances the nonlinear term, so the vortex decays in place and passes no energy to smaller scales~\citep{taylor1937mechanism}.
In our weakly compressible simulations ($\mathrm{Ma} = 0.1$), the energy therefore stays in the lowest wavenumber shells over the whole forecast window.
The shells above the high-$k$ cutoff ($N/4$ on an $N \times N$ grid) hold energy only at the level of floating-point roundoff.
Both spectrum diagnostics then reduce to comparisons of roundoff.
High-$k$/ref.\ divides the predicted high-$k$ energy fraction by a reference fraction that is effectively zero.
Spec.\ RMSE is a logarithmic error that weights every retained shell equally, so it is dominated by shells whose reference energy is numerical noise rather than flow.
The computed values confirm this.
At the final horizon, \emph{entropically-stabilized NeurDE}'s High-$k$/ref.\ for TGV ranges from 20 to $7 \times 10^{3}$ across seeds and grids, and it does not change monotonically with resolution.
We therefore omit these two columns for TGV.
The velocity error, the vorticity error and correlation, and the palinstrophy ratio remain well defined and are reported.

\begin{table}[!htbp]
  \centering\small
  \caption{Real-space and spectral metrics for TGV at $\mathrm{Re}=50{,}000$. We compare emulation metrics from the training data ($t_0=0$) at the final lead time against the downsampled $512^2$ reference. \emph{Entropically-stabilized NeurDE} results are the mean $\pm$ standard deviation over four seeds, ELBM results are from one deterministic run. The "Faithful Emulation" reports whether the rollout reproduces the expected structures. Field fidelity: relative $L^2$ error of velocity and vorticity $\omega$, and vorticity correlation. Spectrum diagnostics: RMSE of $\log_{10}E(k)$ over active shells, and high-$k$ energy and palinstrophy relative to the reference. Bold marks the better method per grid.}
  \label{tab:app_train_tgv50k}
  \resizebox{\textwidth}{!}{
\setlength{\tabcolsep}{2.8pt}
    \begin{tabular}{@{}clccccccc@{}}
        \toprule
        \multirow{2}{*}{Grid} & \multicolumn{1}{c}{\multirow{2}{*}{Method}} & \multirow{2}{*}{\begin{tabular}{@{}c@{}}Faithful\\Emulation\end{tabular}} & \multicolumn{3}{c}{Field Fidelity} & \multicolumn{3}{c}{Spectrum Diagnostics} \\
        \cmidrule(lr){4-6}\cmidrule(l){7-9}
        & & & Vel. err $\downarrow$ & $\omega$ corr. $\uparrow$ & $\omega$ err $\downarrow$ & Spec. RMSE$^\dagger$ $\downarrow$ & High-$k$/ref. $\to1$ & $P$/ref. $\to1$ \\
        \midrule
        \multirow{2}{*}{$128^2$} & NeurDE & \textcolor{green!50!black}{4/4} & $(6.14 \pm 0.81){\times}10^{-4}$ & $1.00 \pm 1.4{\times}10^{-9}$ & $(6.23 \pm 0.82){\times}10^{-4}$ & -- & -- & $1.00 \pm 1.7{\times}10^{-4}$ \\
         & ELBM & \textcolor{green!50!black}{1/1} & $\bf{2.53{\times}10^{-4}}$ & $1.00$ & $\bf{2.58{\times}10^{-4}}$ & -- & -- & $1.00$ \\
        \midrule
        \multirow{2}{*}{$64^2$} & NeurDE & \textcolor{green!50!black}{4/4} & $0.00209 \pm 0.00049$ & $1.00 \pm 3.5{\times}10^{-8}$ & $0.00213 \pm 0.00046$ & -- & -- & $0.996 \pm 1.0{\times}10^{-3}$ \\
         & ELBM & \textcolor{green!50!black}{1/1} & $\bf{0.00162}$ & $1.00$ & $\bf{0.00179}$ & -- & -- & $\bf{0.997}$ \\
        \midrule
        \multirow{2}{*}{$32^2$} & NeurDE & \textcolor{green!50!black}{4/4} & $0.00954 \pm 6.8{\times}10^{-5}$ & $1.00 \pm 3.7{\times}10^{-9}$ & $0.00951 \pm 6.7{\times}10^{-5}$ & -- & -- & $0.981 \pm 1.4{\times}10^{-4}$ \\
         & ELBM & \textcolor{green!50!black}{1/1} & $\bf{0.00317}$ & $1.00$ & $\bf{0.00355}$ & -- & -- & $\bf{0.995}$ \\
        \midrule
        \multirow{2}{*}{$16^2$} & NeurDE & \textcolor{green!50!black}{4/4} & $0.0474 \pm 2.2{\times}10^{-5}$ & $1.00 \pm 7.8{\times}10^{-9}$ & $0.0474 \pm 2.2{\times}10^{-5}$ & -- & -- & $0.908 \pm 4.2{\times}10^{-5}$ \\
         & ELBM & \textcolor{green!50!black}{1/1} & $\bf{0.0228}$ & $1.00$ & $\bf{0.0228}$ & -- & -- & $\bf{0.956}$ \\
        \bottomrule
    \end{tabular}
}
\end{table}
\begin{table}[!htbp]
  \centering\small
  \caption{Real-space and spectral metrics for TGV at $\mathrm{Re}=50{,}000$. We compare emulation metrics from the testing data ($t_0=T/2$) at the final lead time against the downsampled $512^2$ reference. \emph{Entropically-stabilized NeurDE} results are the mean $\pm$ standard deviation over four seeds, ELBM results are from one deterministic run. The "Faithful Emulation" reports whether the rollout reproduces the expected structures. Field fidelity: relative $L^2$ error of velocity and vorticity $\omega$, and vorticity correlation. Spectrum diagnostics: RMSE of $\log_{10}E(k)$ over active shells, and high-$k$ energy and palinstrophy relative to the reference. Bold marks the better method per grid.}
  \label{tab:app_test_tgv50k}
  \resizebox{\textwidth}{!}{
\setlength{\tabcolsep}{2.8pt}
    \begin{tabular}{@{}clccccccc@{}}
        \toprule
        \multirow{2}{*}{Grid} & \multicolumn{1}{c}{\multirow{2}{*}{Method}} & \multirow{2}{*}{\begin{tabular}{@{}c@{}}Faithful\\Emulation\end{tabular}} & \multicolumn{3}{c}{Field Fidelity} & \multicolumn{3}{c}{Spectrum Diagnostics} \\
        \cmidrule(lr){4-6}\cmidrule(l){7-9}
        & & & Vel. err $\downarrow$ & $\omega$ corr. $\uparrow$ & $\omega$ err $\downarrow$ & Spec. RMSE$^\dagger$ $\downarrow$ & High-$k$/ref. $\to1$ & $P$/ref. $\to1$ \\
        \midrule
        \multirow{2}{*}{$128^2$} & NeurDE & \textcolor{green!50!black}{4/4} & $(5.62 \pm 0.79){\times}10^{-4}$ & $1.00 \pm 2.5{\times}10^{-9}$ & $(5.76 \pm 0.81){\times}10^{-4}$ & -- & -- & $1.00 \pm 1.7{\times}10^{-4}$ \\
         & ELBM & \textcolor{green!50!black}{1/1} & $\bf{3.14{\times}10^{-4}}$ & $1.00$ & $\bf{3.18{\times}10^{-4}}$ & -- & -- & $\bf{0.999}$ \\
        \midrule
        \multirow{2}{*}{$64^2$} & NeurDE & \textcolor{green!50!black}{4/4} & $0.00225 \pm 0.00048$ & $1.00 \pm 3.9{\times}10^{-8}$ & $0.00227 \pm 0.00046$ & -- & -- & $0.996 \pm 1.0{\times}10^{-3}$ \\
         & ELBM & \textcolor{green!50!black}{1/1} & $\bf{0.00180}$ & $1.00$ & $\bf{0.00197}$ & -- & -- & $\bf{0.997}$ \\
        \midrule
        \multirow{2}{*}{$32^2$} & NeurDE & \textcolor{green!50!black}{4/4} & $0.00970 \pm 6.8{\times}10^{-5}$ & $1.00 \pm 3.4{\times}10^{-9}$ & $0.00971 \pm 6.8{\times}10^{-5}$ & -- & -- & $0.981 \pm 1.4{\times}10^{-4}$ \\
         & ELBM & \textcolor{green!50!black}{1/1} & $\bf{0.00319}$ & $1.00$ & $\bf{0.00365}$ & -- & -- & $\bf{0.995}$ \\
        \midrule
        \multirow{2}{*}{$16^2$} & NeurDE & \textcolor{green!50!black}{4/4} & $0.0478 \pm 2.4{\times}10^{-5}$ & $1.00 \pm 7.7{\times}10^{-9}$ & $0.0478 \pm 2.4{\times}10^{-5}$ & -- & -- & $0.907 \pm 4.5{\times}10^{-5}$ \\
         & ELBM & \textcolor{green!50!black}{1/1} & $\bf{0.0231}$ & $1.00$ & $\bf{0.0232}$ & -- & -- & $\bf{0.955}$ \\
        \bottomrule
    \end{tabular}
}
\end{table}

\begin{figure}[p]
  \centering
  \appforecastpanel{a}{$128^2$}{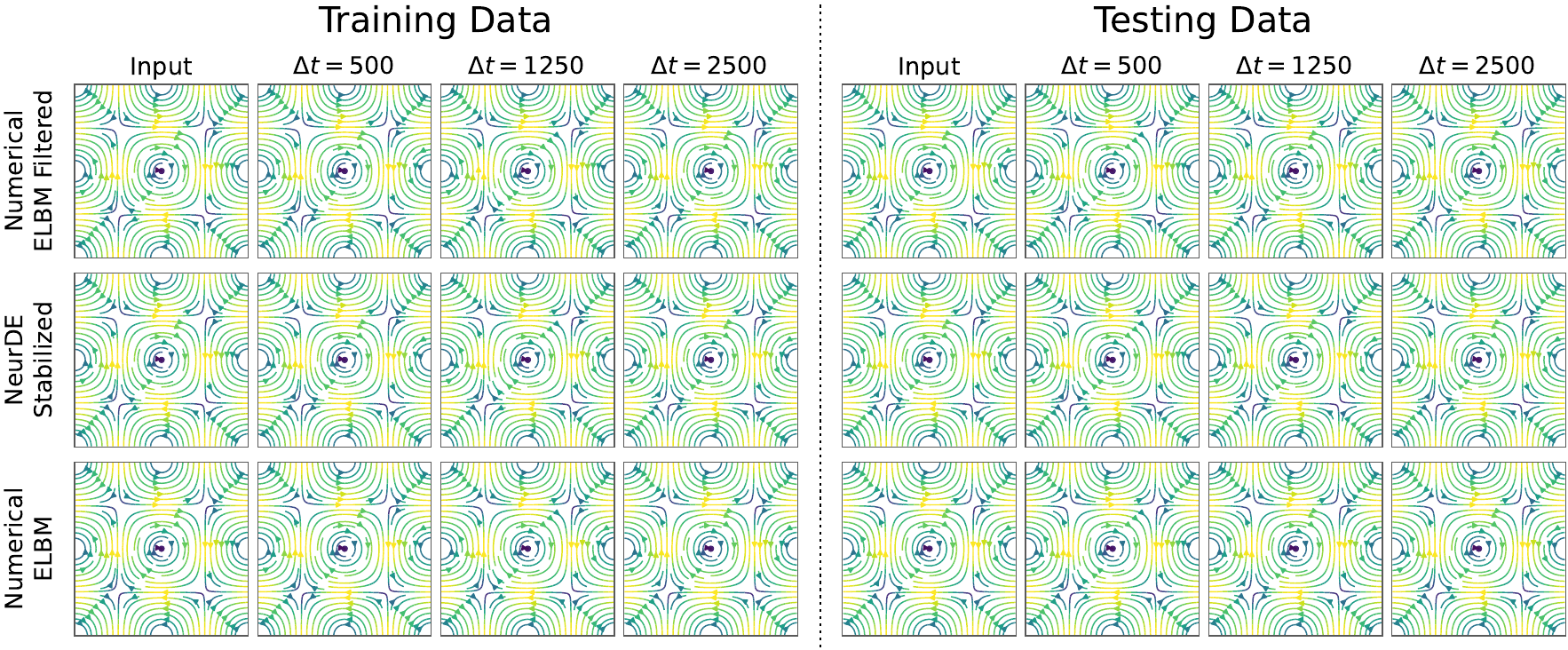}\par\vspace{2pt}
  \appforecastpanel{b}{$64^2$}{Figures/rollout_results/base/tgv50k/tier-64/forecast_seed-2025.pdf}\par\vspace{2pt}
  \appforecastpanel{c}{$32^2$}{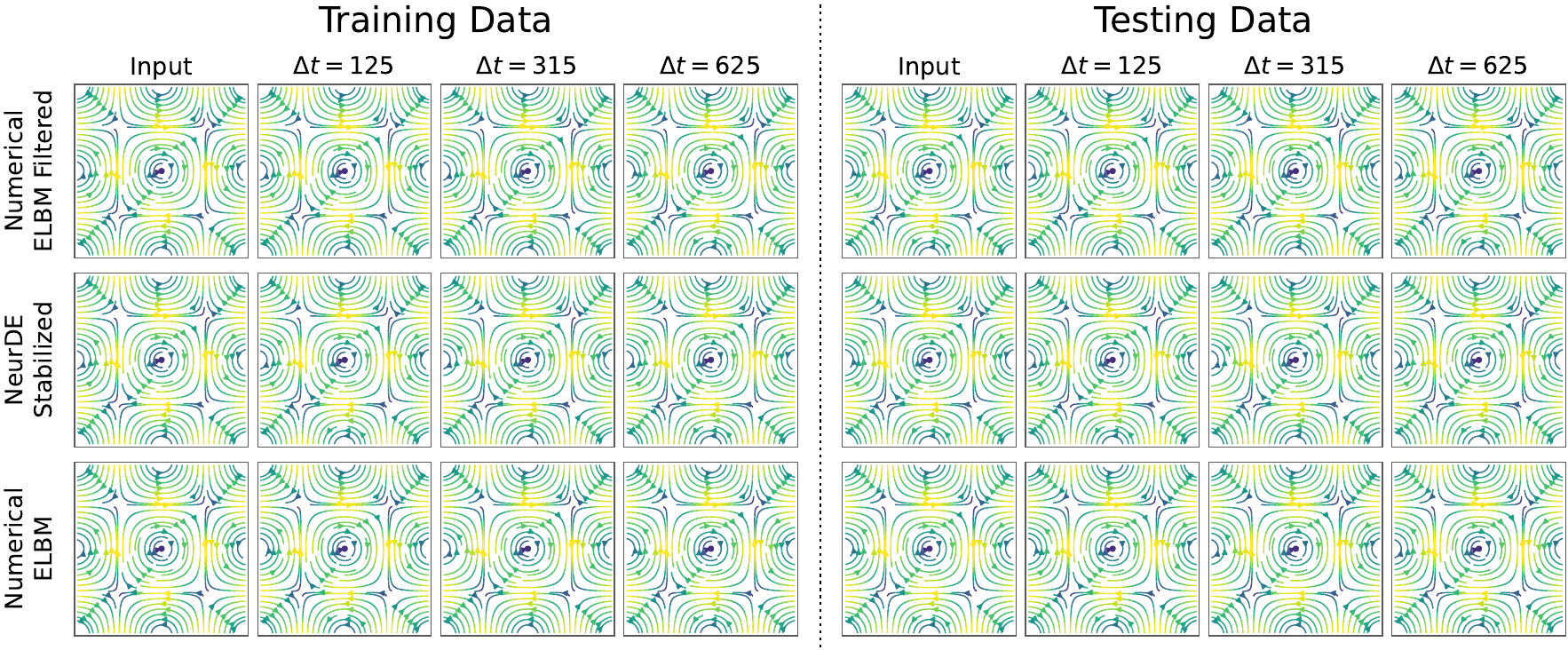}\par\vspace{2pt}
  \appforecastpanel{d}{$16^2$}{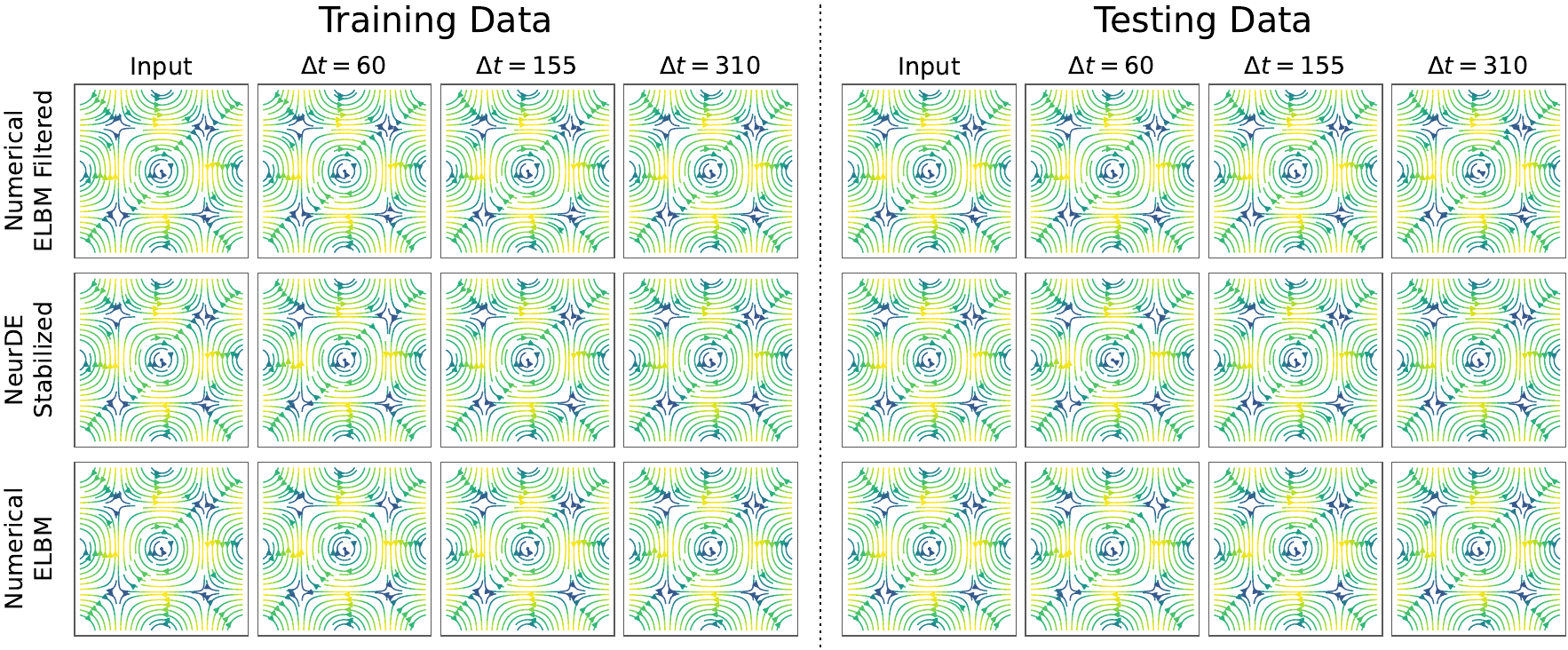}
  \caption{TGV velocity streamline emulations at $\mathrm{Re}=50{,}000$, colored by speed on grids (a)~$128^2$, (b)~$64^2$, (c)~$32^2$, and (d)~$16^2$; the grid size is also marked to the left of each panel. In each panel, the rows are the downsampled $512^2$ reference (top), \emph{entropically-stabilized NeurDE} (middle), and numerical ELBM run at that grid (bottom). The left block starts from the training data ($t_0=0$), the right block from the testing data ($t_0=T/2$); columns are lead times $\Delta t$ in solver steps.}
  \label{fig:app_forecast_tgv50k}
\end{figure}

\begin{figure}[p]
  \centering
  \appspectrapanel{a}{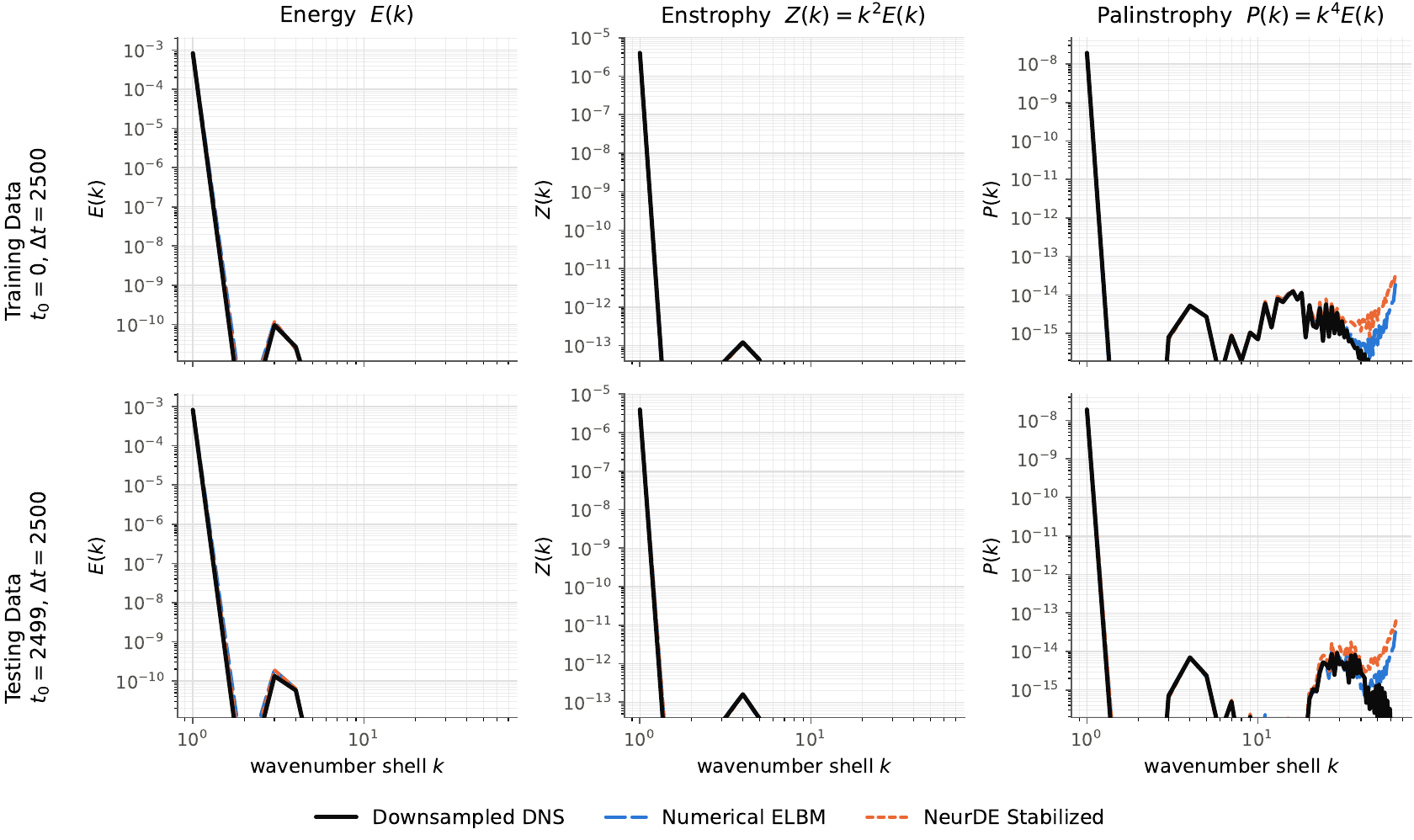}\hfill
  \appspectrapanel{b}{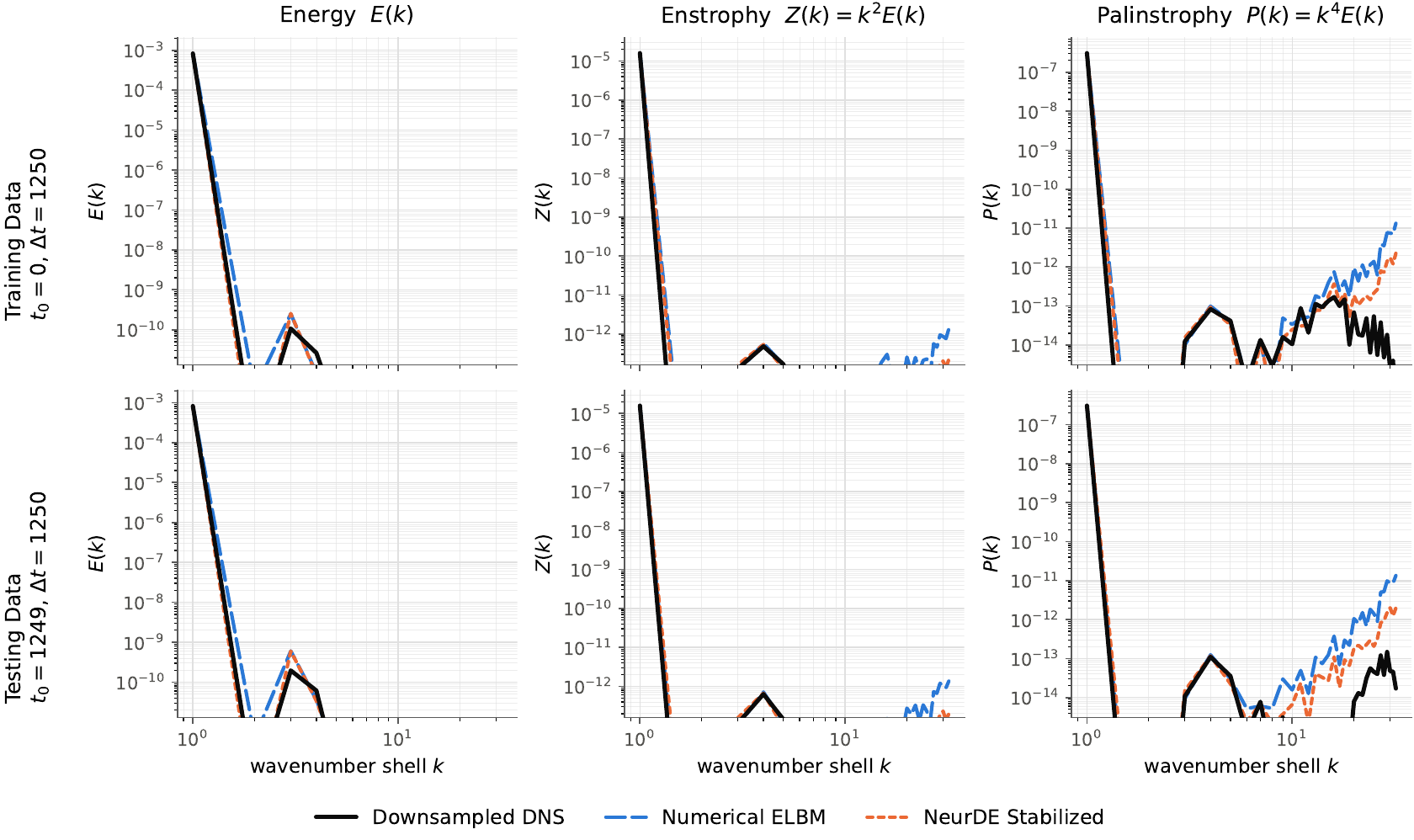}\par\smallskip
  \appspectrapanel{c}{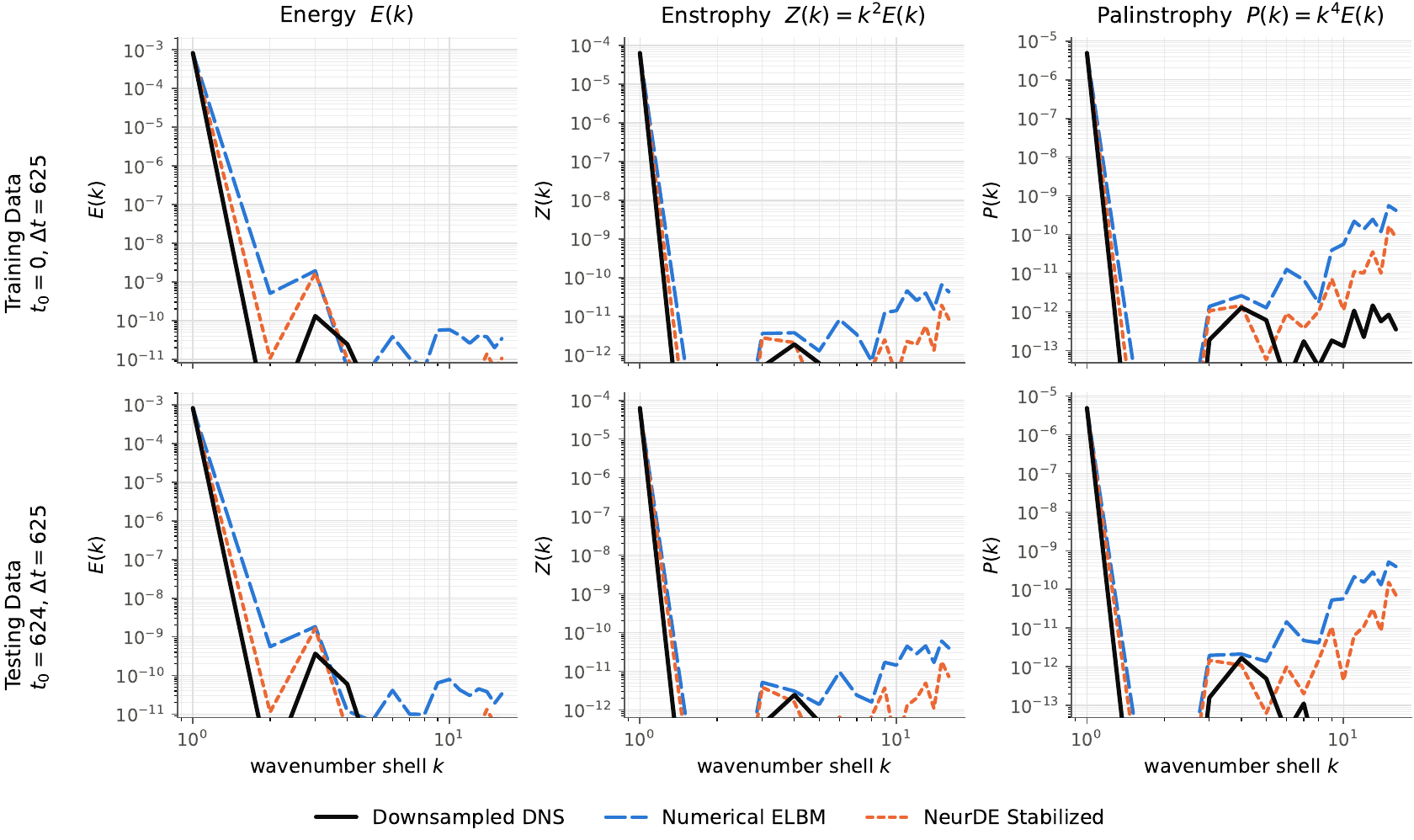}\hfill
  \appspectrapanel{d}{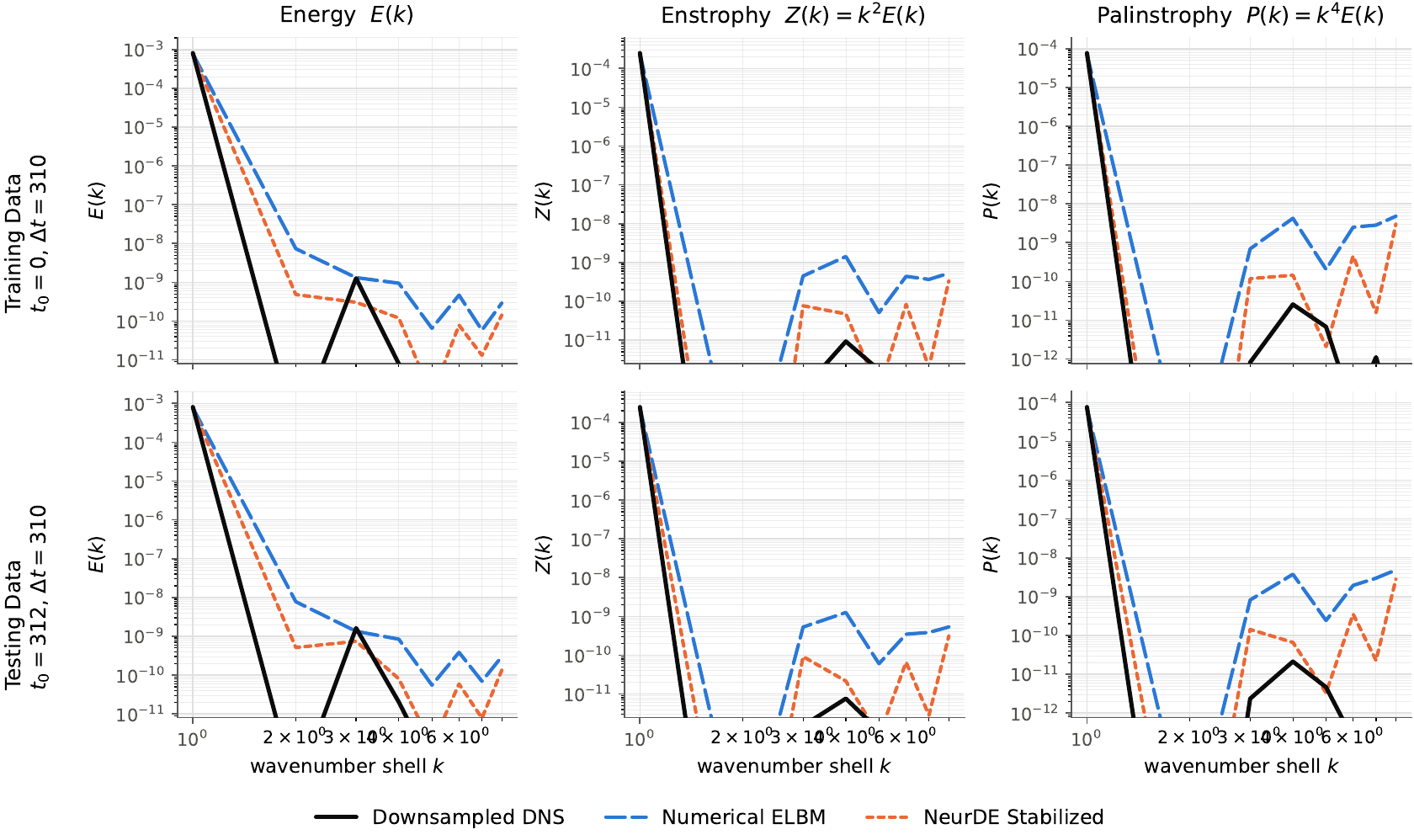}
  \caption{TGV at $\mathrm{Re}=50{,}000$ shell-averaged spectra at the final lead time on grids (a)~$128^2$, (b)~$64^2$, (c)~$32^2$, and (d)~$16^2$. Columns are the energy $E(k)$, enstrophy $Z(k)=k^2E(k)$, and palinstrophy $P(k)=k^4E(k)$ spectra, where $k$ is the wavenumber shell; rows are forecasts from the training (top) and testing (bottom) data. Curves compare the downsampled $512^2$ reference, numerical ELBM, and \emph{entropically-stabilized NeurDE}. The flow stays in a few low-$k$ shells, so the high-$k$ tail is at roundoff level.}
  \label{fig:app_spectra_tgv50k}
\end{figure}

\begin{figure}
    \centering
    \includegraphics[width=0.95\linewidth]{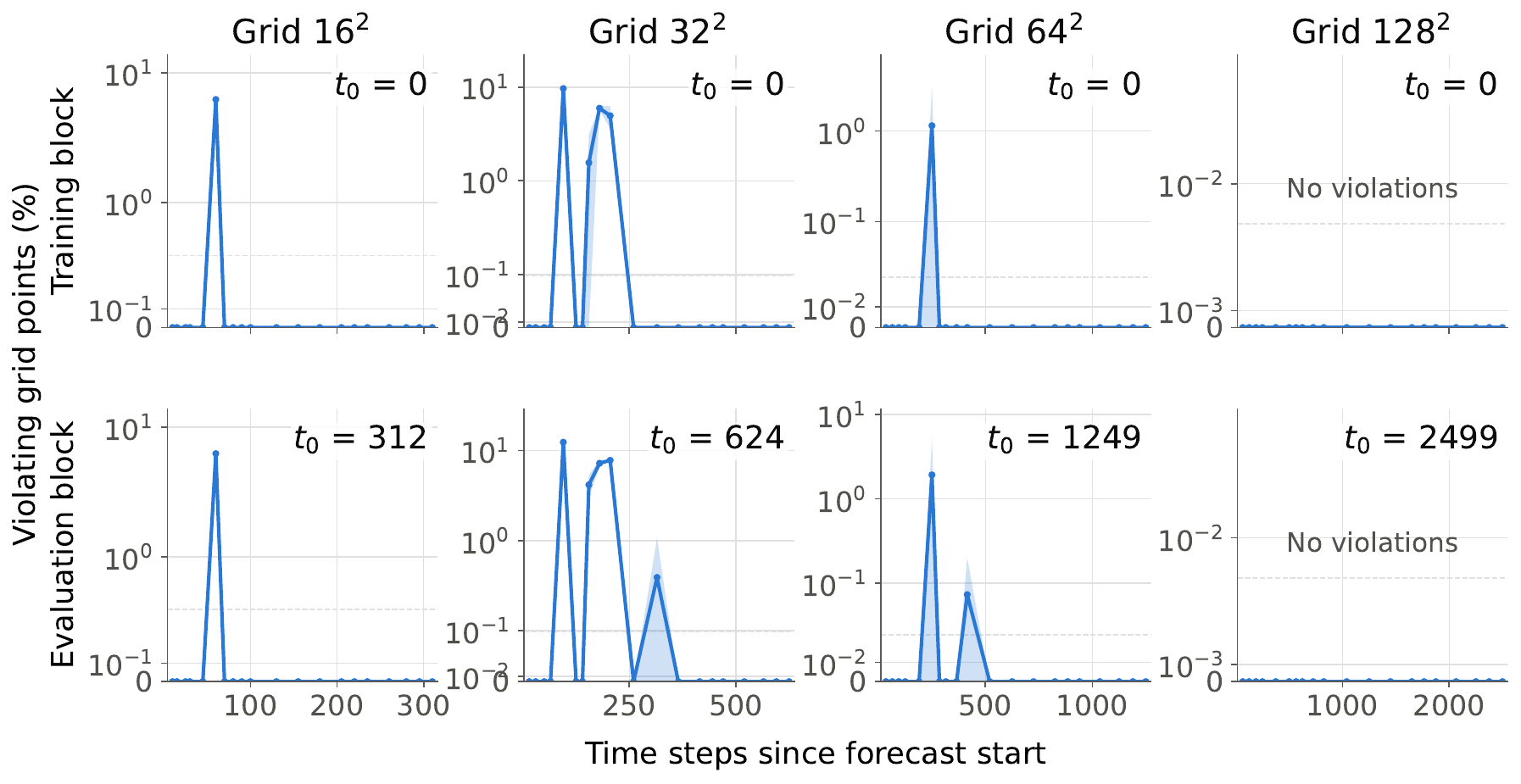}
    \caption{The percentage of pixels that violate the second law of thermodynamics over the \emph{NeurdE} emulation of the Taylor Green vortex experiment at $\mathrm{Re}=50{,}000$. We show the training and testing datasets for all grids. The mean and standard deviation are taken over all seeds.}
    \label{fig:app_entropy_tgv50k}
\end{figure}

\clearpage

\section{Fourier Neural Operator Baseline}\label{app:FNO_appx}
%

We use the Fourier Neural Operator (FNO)~\citet{li2021fourier} as a purely macroscopic neural-operator baseline for the doubly periodic shear layer with $\mathrm{Re}=50{,}000$ (DPSL-50K) forecasting task in Section~\ref{sec:experiments_main}.
This baseline does not evolve lattice populations and does not use the entropic collision mechanism.
Instead, it learns an autoregressive map directly between resolved macroscopic fields.
This makes the comparison useful: it separates the benefit of a high-capacity spectral neural operator from the kinetic and entropy-constrained structure used by the proposed method.

The DPSL geometry is doubly periodic, which is favorable for FNO because its spectral convolution layers are naturally matched to periodic domains.
However, the target dynamics contain sharp vorticity gradients and small-scale turbulent structures, which are known to be difficult for neural operators to resolve robustly at high Reynolds number~\citet{mcgreivy2024weak,sakarvadia2025false}.
We therefore evaluated relatively large Fourier bandwidths and lifting dimensions to give the baseline substantial representational capacity.

Because the DPSL-50K rollouts are long, treating time as an additional input dimension, as in some prior FNO turbulence experiments~\citet{li2021fourier}, is not practical at the resolutions and horizons considered here.
We therefore train FNO as a one-step map and roll it out autoregressively:
\[
    \mathcal{G}_{\theta}:
    \big(\rho,\Velocity,\omega\big)(t,\bx)
    \mapsto
    \big(\rho,\Velocity,\omega\big)(t+1,\bx).
\]
This protocol matches the forecasting use case in the main experiments, but it is also stricter than a short-window interpolation task because small one-step errors accumulate over thousands of steps.
Under this protocol, the FNO models we tested did not produce stable long-horizon DPSL forecasts.
We therefore keep the details in the appendix and focus the main experimental section on the more informative comparison between the proposed \emph{entropically-stabilized NeurDE} and the numerical ELBM baseline.

\paragraph{Computational Setup.}
The final model was trained on one NERSC Perlmutter node equipped with $4\times$ NVIDIA A100-SXM4 GPUs with $40$\,GB HBM2 memory (Table~\ref{tab:fno-cost}).
We implemented the models using the official Neural Operator library~\citet{kossaifi2025librarylearningneuraloperators} (version 0.3.0).

The final model uses the following configuration (Table~\ref{tab:fno-model}):
\begin{itemize}
    \item \textbf{Hidden Width ($d_v$):} $128$.
    \item \textbf{Fourier Modes ($\mathbf{k}_{\text{max}}$):} $64$.
    \item \textbf{Network Depth ($L$):} $4$ layers.
\end{itemize}

\begin{table}[ht]
  \centering
  \caption{FNO architecture.}
  \label{tab:fno-model}
  \begin{tabular}{ll}
    \toprule
    Hyperparameter & Value \\
    \midrule
    Spatial dimensions & 2 ($256 \times 256$ periodic grid) \\
    Input / output channels & 4 (density, $u_x$, $u_y$, vorticity) \\
    Fourier modes retained & 64 per dimension ($64 \times 33$ after the real FFT) \\
    Hidden channels (width) & 128 \\
    Fourier layers & 4 \\
    Lifting / projection channels & 256 / 256 \\
    Skip connection in Fourier layers & linear ($1 \times 1$ convolution) \\
    Activation & GELU \\
    Boundary padding & none (domain is periodic) \\
    Input normalization & per-variable batch statistics (running mean and std.) \\
    Precision & float32 \\
    Trainable parameters & 276{,}958{,}340 ($\approx 2.8 \times 10^{8}$) \\
    \bottomrule
  \end{tabular}
\end{table}

\paragraph{Hyperparameter Sweep.}
Before the final run, a grid search of $2 \times 3 \times 3 \times 2 = 36$ configurations varied the width, the batch size, and the learning-rate schedule (Table~\ref{tab:fno-sweep}).
All other settings were held at the values in Tables~\ref{tab:fno-model} and~\ref{tab:fno-train}: 64 Fourier modes, peak learning rate $10^{-4}$ and weight decay $10^{-5}$.
One difference from the final run is that the sweep runs supervised every step of the rollout, while the final run is supervised at most 32 steps (Table~\ref{tab:fno-train}).

\begin{table}[ht]
  \centering
  \caption{Sweep grid and the values used in the final run.}
  \label{tab:fno-sweep}
  \begin{tabular}{lll}
    \toprule
    Hyperparameter & Values swept & Final run \\
    \midrule
    Hidden channels & 128, 256 & 128 \\
    Global batch size & 16, 32, 64 & 16 \\
    OneCycle initial divisor (\texttt{div\_factor}) & 5, 25, 100 & 25 \\
    OneCycle final divisor (\texttt{final\_div\_factor}) & 100, 1000 & 100 \\
    \bottomrule
  \end{tabular}
\end{table}

\paragraph{Training Protocol.}
The model was trained on macroscopic channels only:
\[
(\rho, \Velocity, \omega)(t, \bx)
\;\mapsto\;
(\rho, \Velocity, \omega)(t+1, \bx),
\]
on a $256\times256$ periodic grid, where $\rho$ is density, $\Velocity$ is the bulk velocity, and $\omega=\nabla\times\Velocity$ is the scalar vorticity in two dimensions.
The FNO is therefore a direct field-to-field surrogate.
It is not given the lattice populations, the entropic parameter $\alpha$, or the discrete $H$-theorem residuals used by the proposed kinetic model.

The final model was trained with AdamW~\citet{loshchilov2019adamw} and a one-cycle learning-rate schedule~\citet{smith2019superconvergence} with cosine annealing and a peak learning rate of $10^{-4}$ (Table~\ref{tab:fno-train}).
The figures below show this final model.

\begin{table}[ht]
  \centering
  \caption{FNO training setup.}
  \label{tab:fno-train}
  \begin{tabular}{ll}
    \toprule
    Hyperparameter & Value \\
    \midrule
    \multicolumn{2}{l}{\emph{Data}} \\
    Simulation & D2Q9 weakly compressible isothermal LBM, $\mathrm{Re} = 5 \times 10^{4}$, $\mathrm{Ma} = 0.1$ \\
    Trajectory & one trajectory, $256 \times 256$ grid, $10^{4}$ saved steps \\
    Train / validation / test split & 50\,\% / 10\,\% / 40\,\%, contiguous in time \\
    Inputs and targets & density, velocity, vorticity \\
    \midrule
    \multicolumn{2}{l}{\emph{Loss and optimizer}} \\
    Loss & sum of per-variable MSE (density, velocity, vorticity) \\
    Optimizer & AdamW~\citet{loshchilov2019adamw} \\
    Weight decay & $10^{-5}$ \\
    Gradient clipping & global $L_2$ norm $\le 1.0$ \\
    Global batch size & 16 (4 per GPU) \\
    \midrule
    \multicolumn{2}{l}{\emph{Learning-rate schedule}} \\
    Schedule & OneCycle~\citet{smith2019superconvergence}, cosine annealing, stepped per batch \\
    Peak learning rate & $10^{-4}$ \\
    Warm-up fraction (\texttt{pct\_start}) & 0.3 \\
    Initial learning rate & $4 \times 10^{-6}$ (\texttt{div\_factor} $= 25$) \\
    Final learning rate & $4 \times 10^{-8}$ (\texttt{final\_div\_factor} $= 100$) \\
    Schedule length & 300 epochs \\
    \midrule
    \multicolumn{2}{l}{\emph{Rollout curriculum}} \\
    Rollout lengths & 1, 2, 3, 5, 10, 25, 50 steps \\
    Epoch budget per stage & 25, 25, 25, 50, 50, 50, 75 \\
    Early stage advance & after 5 epochs without validation improvement \\
    Supervised steps per rollout & at most 32, log-spaced (affects only the 50-step stage) \\
    \midrule
    \multicolumn{2}{l}{\emph{Run}} \\
    Epochs completed & 183 of 300 (job ended); best checkpoint at epoch 182 \\
    Parallelism & distributed data parallel (Table~\ref{tab:fno-cost}) \\
    Compilation & \texttt{torch.compile} \\
    Random seed & 2025 \\
    \bottomrule
  \end{tabular}
\end{table}

The curriculum advances to the next rollout length when a stage's epoch budget is used up, or earlier if the validation loss has not improved for 5 consecutive epochs.
In the final run, the stages occupied epochs 0--24, 25--49, 50--74, 75--96, 97--112, 113--152 and 153--182.
Training stopped during the 50-step stage, at which point the learning rate was $5.9 \times 10^{-5}$.

\paragraph{Training Cost.}
The final model was trained in a single 24-hour job on one NERSC Perlmutter GPU node (Table~\ref{tab:fno-cost}).
The job ran until the time limit.
The cost per epoch grows with the rollout length, because each training sample is unrolled for that many steps.
As a result, the 25- and 50-step stages account for 20.7 of the 23.7 training hours (Table~\ref{tab:fno-cost-stage}).
These figures cover only the final run, not the sweep or evaluation.

\begin{table}[ht]
  \centering
  \caption{Compute used to train the final FNO.}
  \label{tab:fno-cost}
  \begin{tabular}{ll}
    \toprule
    Quantity & Value \\
    \midrule
    System & NERSC Perlmutter, 1 GPU node \\
    GPU type & NVIDIA A100-SXM4, 40\,GB \\
    Number of GPUs & 4 \\
    CPU cores allocated & 64 physical (128 hardware threads) \\
    Job wall-clock time & 23.96\,h (24\,h limit) \\
    Training time (epoch 0 to epoch 182) & 23.67\,h \\
    GPU-hours (job) & 95.8 \\
    Node-hours (job) & 23.96 \\
    Energy (Slurm node counter) & $1.12 \times 10^{8}$\,J $\approx 31.2$\,kWh \\
    \bottomrule
  \end{tabular}
\end{table}

\begin{table}[ht]
  \centering
  \caption{Wall-clock time per curriculum stage (4 GPUs).}
  \label{tab:fno-cost-stage}
  \begin{tabular}{rrrr}
    \toprule
    Rollout length & Epochs & Wall-clock (h) & Minutes per epoch \\
    \midrule
     1 & 25 &  0.30 &  0.7 \\
     2 & 25 &  0.42 &  1.0 \\
     3 & 25 &  0.53 &  1.3 \\
     5 & 22 &  0.71 &  1.9 \\
    10 & 16 &  1.01 &  3.8 \\
    25 & 40 & 10.02 & 15.0 \\
    50 & 30 & 10.67 & 21.3 \\
    \midrule
    Total & 183 & 23.67 & \\
    \bottomrule
  \end{tabular}
\end{table}

\paragraph{Observed Behavior.}
Overall, this result of Fig.~\ref{fig:fno_forecast} should not be interpreted as a failure of spectral neural operators in general.
Rather, it indicates that a macroscopic one-step FNO, without kinetic variables or an entropy constraint, is not robust for this long-horizon underresolved DPSL setting.
The limitation is most visible during autoregressive rollout: the model may learn a reasonable short-term field update, but repeated application damps or distorts the vortical structures that control the later-time dynamics.
Given the substantial cost of scaling the space-time FNO formulation to the present horizons, we did not pursue further FNO-specific architectural modifications.

\section{Walrus Baseline}\label{app:walrus_appx}
Since Walrus is a large pretrained Transformer, which significantly differs from \emph{entropically-stabilized NeurDE}, we trained it within its own repository.
We used the default model parameters and started from the pretrained weights.
Unless otherwise specified, we used the default values within the repository.

\begin{figure}
    \centering
    \includegraphics[width=0.5\linewidth]{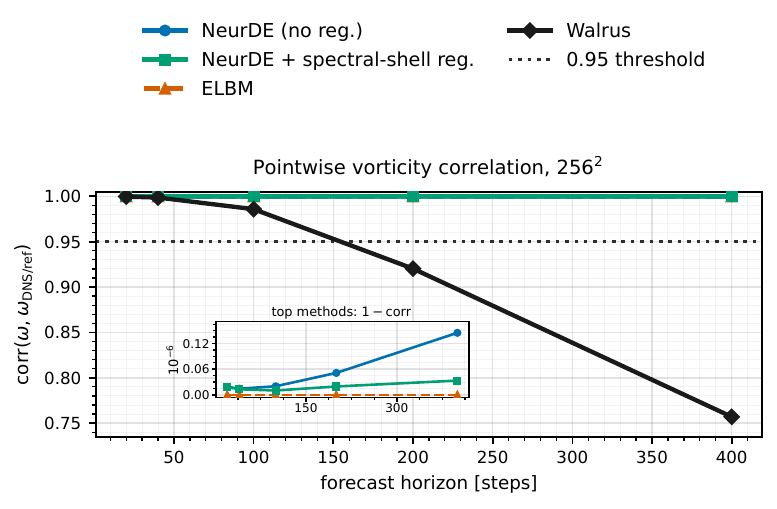}
    \caption{Diverging vorticity decay from DNS as forecast length increases.}
    \label{fig:walrus_corrs}
\end{figure}

\paragraph{Fine-tuning Setup.}
We used standard configuration variables found in the Walrus repository for the fine-tuning procedure: its trainer, the AdamW optimizer, and its inverse-square-root learning-rate schedule with linear warm-up and square-root cool-down.
We changed only what was needed for a short fine-tuning run on new data:
\begin{itemize}
  \item a learning rate of $10^{-4}$;
  \item warm-up and cool-down shortened to 6 and 3 epochs, to keep the proportions of the default schedule over 16 epochs;
  \item drop-path and input-field dropout turned off;
  \item no gradient accumulation;
  \item gradient clipping at 10;
  \item per-sample normalization in place of precomputed dataset statistics.
\end{itemize}
The pretrained Walrus model was fine-tuned on the same $256 \times 256$ DPSL-50K simulation used for the FNO baseline.
Each training sample gives the model 6 consecutive input states and asks it to predict the next state, with the time stride between states drawn from 1 to 5 saved steps.
Fine-tuning ran for 16 epochs of 2000 optimizer steps each, which is 32{,}000 steps in total.
It used fully sharded data parallelism (FSDP) across 8 GPUs, a batch of 2 per GPU (global batch 16), and float32 arithmetic.
Validation ran every second epoch.

\paragraph{Training Cost.}
Fine-tuning took 38\,h of wall-clock time on 8 GPUs, or about 305 GPU-hours (Table~\ref{tab:walrus-cost}).
These figures cover only fine-tuning, not pretraining or the forecast evaluation.

\begin{table}[h]
  \centering
  \caption{Compute used to fine-tune Walrus.}
  \label{tab:walrus-cost}
  \begin{tabular}{ll}
    \toprule
    Quantity & Value \\
    \midrule
    System & NERSC Perlmutter, 2 GPU nodes \\
    GPU type & NVIDIA A100-SXM4, 80\,GB \\
    Number of GPUs & 8 (4 per node) \\
    CPU cores allocated & 128 physical (256 hardware threads) \\
    Total wall-clock time & 38.11\,h \\
    GPU-hours & 304.8 \\
    Node-hours & 76.2 \\
    Energy (Slurm node counter) & $4.82 \times 10^{8}$\,J $\approx 134.0$\,kWh \\
    \bottomrule
  \end{tabular}
\end{table}

\end{document}